\documentclass[runningheads]{llncs}
 
\usepackage{eccv}

\usepackage{graphicx}
\usepackage{wrapfig}
\usepackage{booktabs}
\usepackage[accsupp]{axessibility}  

\usepackage[pagebackref,breaklinks,colorlinks,citecolor=eccvblue]{hyperref}

\usepackage{orcidlink}

\usepackage{multirow}
\usepackage[dvipsnames]{xcolor}
\usepackage{colortbl}
\usepackage{cancel}

\title{Evaluating the Adversarial Robustness of Semantic Segmentation: Trying Harder Pays Off\thanks{Accepted for ECCV 2024.}}
\titlerunning{Evaluating the Adversarial Robustness of Semantic Segmentation}

\author{%
Levente Halmosi\inst{1}\orcidlink{0009-0005-2375-2429} \and 
Bálint Mohos\inst{1}\orcidlink{0009-0000-0676-6566} \and
Márk Jelasity\inst{1,2}\orcidlink{0000-0001-9363-1482}}
\institute{University of Szeged, Hungary \and HUN-REN-SZTE Research Group on AI, Szeged, Hungary}

\begin{document}
\maketitle

\begin{abstract}
Machine learning models are vulnerable to tiny adversarial input perturbations
optimized to cause a very large output error.
To measure this vulnerability, we need reliable methods
that can find such adversarial perturbations.
For image classification models, evaluation methodologies have emerged that have stood the test of time.
However, we argue that in the area of semantic segmentation,
a good approximation of the sensitivity to adversarial perturbations requires
\emph{significantly more effort} than what is currently considered satisfactory.
To support this claim, we re-evaluate a number of well-known robust segmentation models
in an extensive empirical study.
We propose new attacks and combine them with the strongest attacks available in the literature.
We also analyze the sensitivity of the models in fine detail.
The results indicate that most of the state-of-the-art models have a \emph{dramatically larger
sensitivity} to adversarial perturbations than previously reported.
We also demonstrate a size-bias: small objects
are often more easily attacked, even if the large objects are robust,
a phenomenon not revealed by current evaluation metrics.
Our results also demonstrate that
a diverse set of strong attacks is necessary, because different models are often vulnerable to
different attacks.
Our implementation is available at \url{https://github.com/szegedai/Robust-Segmentation-Evaluation}.
\end{abstract}

\section{Introduction}
\label{sec:intro}

It has long been known that deep neural networks (and, in fact, most other machine learning
models as well) are sensitive to adversarial perturbation~\cite{Sturm2014a,szegedy13}.
In the case of image processing tasks, this means that---given a network and an input
image---an adversary can compute a specific input
perturbation that is invisible to the human eye yet changes the output arbitrarily.
This is not only a security problem but, more importantly, also a clue that the models
trained on image processing tasks have fundamental flaws regarding the feature representations
they evolve~\cite{adversarialbug19,engstrom19}.
In the context of image classification, this problem has received a lot of attention,
leading to a large number of attacks under various assumptions (just to mention a few,~\cite{carlini-wagner17,deepfool16,chen2020a,chen2020d,croce2022a}) and defenses (for example,~\cite{saddlepoint-iclr18,randsmoothing19}).

In image segmentation, the vulnerability to adversarial perturbation attacks has also been
demonstrated many times, for
example,~\cite{Cisse2017a,Arnab2018a,Xie2017a,Fischer2017a,Rony2022a,Sun2022a,Gupta2019a}.
However, interestingly, the problem of training models that are robust to adversarial perturbation
has not received a lot of attention until recently.
The first work that focuses on this problem in depth is by Xu et al.~\cite{Xu2021b} where the DDC-AT method was proposed,
followed by the improved SegPGD-AT method by Gu et al.~\cite{Gu2022a}.
More recently, Croce et al.~\cite{Croce2023a} also experimented with several configurations for adversarial training.

\begin{figure}[t]
\centering
\setlength{\tabcolsep}{10pt}
\begin{tabular}{cc}
\small clean samples & \small adversarial samples\\
\includegraphics[trim= 0mm 0mm 4mm 2mm,clip,width=0.45\columnwidth,height=35mm]{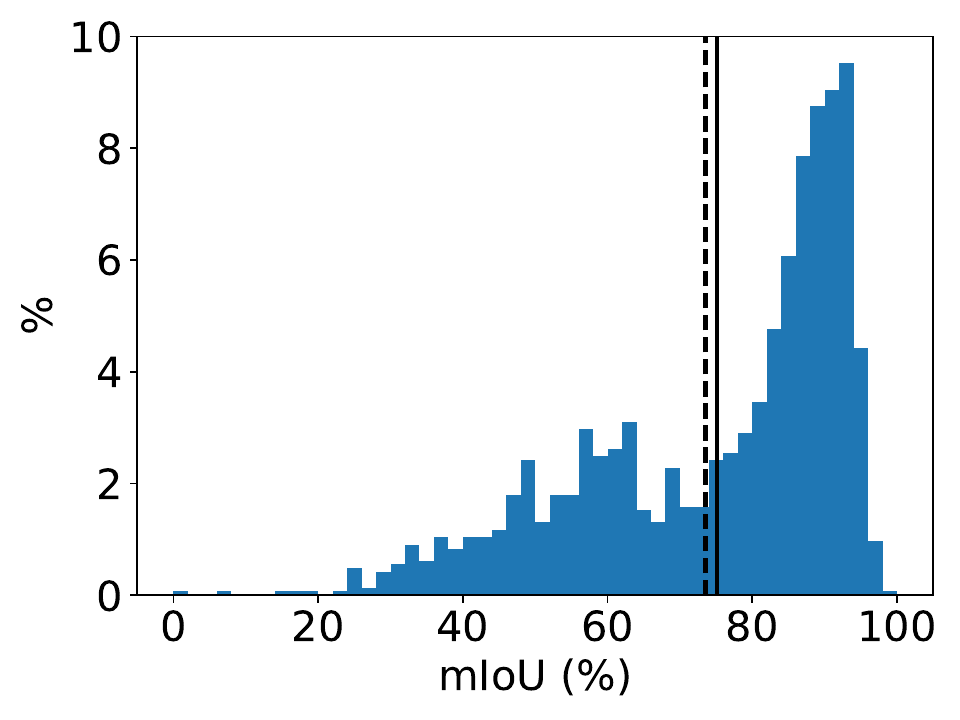} &
\includegraphics[trim= 0mm 0mm 4mm 2mm,clip,width=0.45\columnwidth,height=35mm]{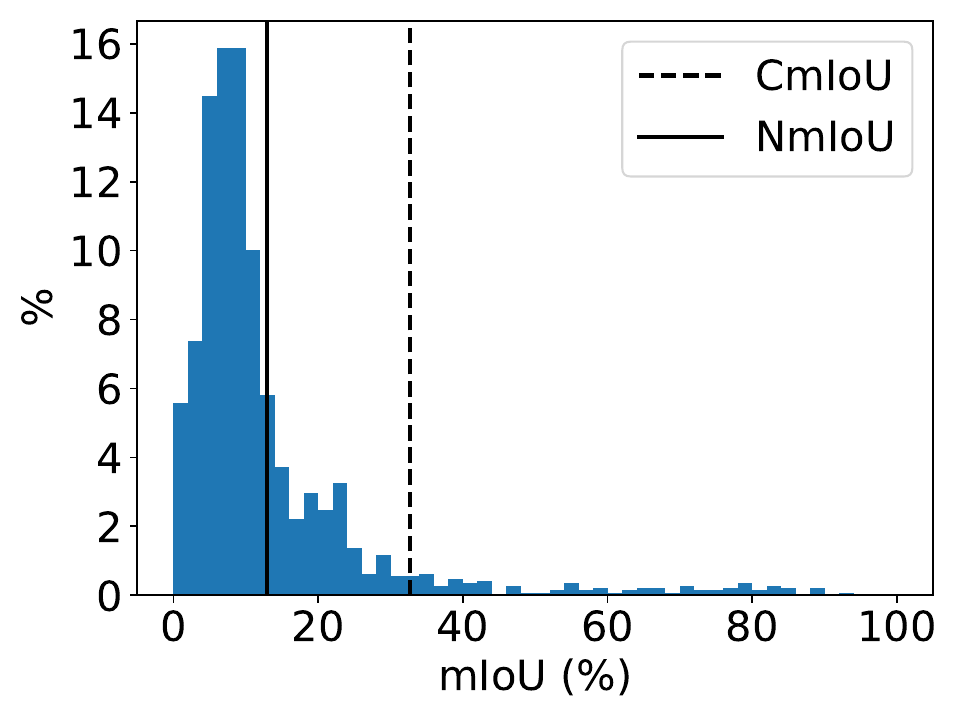}
\end{tabular}
\caption{Single image mIoU distributions of the `small' adversarially trained model of Croce et al.~\cite{Croce2023a} over the PASCAL VOC 2012 validation set.
The image-wise (NmIoU) and class-wise (CmIoU) aggregated mIoU metrics are shown using vertical lines (see \cref{sec:evaluation}).
The lack of robustness is apparent and the gap between the two aggregated mIoU metrics for adversarial input indicates a size-bias (see \cref{sec:evaluation}).}
\label{fig:teaser}
\end{figure}

These methods all use adversarial training~\cite{goodfellow15,saddlepoint-iclr18}, a method that has stood the test of time in image classification.
On clean input samples, all of the resulting models perform similarly to normally trained models.
At the same time, they are reported to have a non-trivial robustness.
This is rather surprising, because in image classification, it is well-known that there is a tradeoff between
accuracy and robustness~\cite{Zhang2019b}.
\emph{This motivates our hypothesis that these models are in fact not robust and
the current methodology for evaluating robustness is insufficient.}

To test this hypothesis, we perform a thorough empirical evaluation addressing two
shortcomings of current practice.
First, we apply the combination of the strongest attacks available in the literature,
along with our own attacks, to get state-of-the-art upper bounds on the robust performance.
The recently proposed attacks in our set include the ALMAProx attack~\cite{Rony2022a} and
the Segmentation Ensemble Attack (SEA)~\cite{Croce2023a}.
We apply the attack set in an ensemble fashion, attacking each input with all the attacks.
We then select the most successful attack for each input, according to a given metric.

Second, we demonstrate that the usual practice of using only pixel accuracy and class-wise aggregated mIoU over the test set hides an important
robustness problem, because these metrics are relatively insensitive to the misclassification of small objects.
We therefore propose to examine the image-wise average mIoU as well when evaluating robust models, because
it balances between smaller and larger objects better, especially over datasets where some classes have
only a few instances in most images.

\subsection{Contributions}

In stark contrast with the results reported previously,
using a thorough evaluation methodology we demonstrate that the best-known models proposed in the
literature \emph{cannot be considered significantly robust}. 
We focus on the state-of-the-art adversarial training methods including DDC-AT~\cite{Xu2021b} and SegPGD-AT~\cite{Gu2022a}, as well as
the approach of Croce et al.~\cite{Croce2023a}.
We show that

\begin{itemize}
\item using 50\% adversarial and 50\% clean samples during training---the setup used
by DDC-AT and SegPGD-AT---results in
a \emph{complete lack of robustness} regardless of which performance metric is considered
\item using 100\% adversarial samples might result in some robustness
according to some of the metrics, but \emph{in terms of the image-wise average mIoU metric these models are also vulnerable}
\item the models of Croce et al.---that
have the highest robust pixel accuracy among the models we examine---are vulnerable in terms of image-wise average mIoU, which
suggests that these models sacrifice the small objects and focus on protecting the large ones (see \cref{fig:teaser})
\item the \emph{diversity of our attack set is essential}, because different scenarios and
models might require different attacks, no attack dominates all the others
\end{itemize}

Our results indicate that, in the case of semantic segmentation models, a very thorough robustness
evaluation is necessary using a diverse set of attacks, and the
distribution of the mIoU values over the dataset should also be examined.

\subsection{Related work}
\label{sec:related}

Here, we overview related work specifically in the area of adversarial
robustness in semantic segmentation.

\textbf{Attacks.}
Adaptations of the gradient-based adversarial attacks using the segmentation
loss function have been proposed relatively
early~\cite{Fischer2017a,Xie2017a,Arnab2018a}.
The Houdini attack by Cissé et al.\ uses a novel surrogate function more
tailored to adversarial example generation~\cite{Cisse2017a}.
Agnihotri et al.\ propose the cosine similarity as a surrogate~\cite{Agnihotri2023a}.
The ALMAProx attack proposed by Rony et al.~\cite{Rony2022a}
defines a constrained optimization problem to find the minimum perturbation to
change the prediction over a given proportion of pixels.
This is a fairly expensive, yet accurate baseline to evaluate defenses.
The dynamic divide-and-conquer (DDC) attack by Xu et al.~\cite{Xu2021b} is based
on grouping pixels dynamically during the attack.
The segmentation PGD (SegPGD) attack by
by Gu et al.~\cite{Gu2022a} is an efficient adaptation of the PGD algorithm for the segmentation task.
Finally, the Segmentation Ensemble Attack (SEA) by Croce et al~\cite{Croce2023a}
is a collection of four different adaptive attacks with the goal of serving as a reliable
ensemble for evaluating robust models.

\textbf{Special purpose attacks.}
Some works introduce different versions of the adversarial perturbation problem
with specific practical applications in mind.
Metzen et al. study universal (input independent) perturbations~\cite{Metzen2017a}.
Cai et al.~\cite{Cai2022a} study semantically consistent (context sensitive) attacks
that can fool defenses that are based on verifying the semantic consistency of the predicted scene.
Chen et al \cite{Chen2022b} also consider semantic attacks where
the predicted scene is still meaningful, only some elements are deleted, for example.

\textbf{Detection as a defense.} Klingner et al. proposed an approach to detect adversarial inputs based on the consistency on
different tasks~\cite{Klingner2022a}.
Another detection approach was suggested by Bär et al.~\cite{Baer2019a,Baer2020a}
where a specifically designed ensemble of models is applied that involves
a dynamic student network that explicitly attempts to become different from
another model of the ensemble.
While detection approaches
are useful in practice, they are not hard defenses because adversarial inputs
can be constructed to mislead multiple tasks or multiple models
as well simultaneously, as the authors also note.

\textbf{Multi-tasking as a defense.}
Another approach involves multi-task networks, with the underlying idea that
if a network is trained on several different tasks then it will naturally become
more robust~\cite{Mao2020a,Klingner2020a}.
While this is certainly a very promising direction for defense, many uncertainties
are involved such as generalizability to different tasks and datasets, and the critical
number of tasks.

\textbf{Adversarial training.}
In image classification,
the most successful approach is adversarial training~\cite{saddlepoint-iclr18}.
In semantic segmentation, a notable application of adversarial training
is the DDC-AT algorithm by Xu et al.~\cite{Xu2021b}.
More recently, the SegPGD-AT algorithm was proposed by Gu et al.~\cite{Gu2022a}, where
SegPGD was used to implement adversarial training.
Croce et al.~\cite{Croce2023a} have also proposed techniques for adversarial training
such as using a robust backbone.
These approaches are the subject of our study.

\section{Background}
\label{sec:back}

Here, we briefly summarize the basic notions of adversarial attacks and adversarial training.
We focus on the white-box setting, and we assume that the model to be attacked is differentiable and
deterministic.
Let the pre-trained model be
$f_\theta: \mathcal X\mapsto \mathcal Y$ and the
loss function $\mathcal L(\theta, x, y)\in\mathbb R$ that characterizes the error of the
prediction $f_\theta(x)$ given the ground truth output $y\in \mathcal Y$.

\subsection{Adversarial Attacks}
\label{sec:back-attacks}

Intuitively, the goal of an adversarial attack is to find a very small perturbation of a
given input $x$ in such a way that the prediction of the model $f_\theta(x)$ is completely
wrong.
Clearly, for an input $x\in\mathcal X$ and a perturbation $\delta$ we require that
$x+\delta\in\mathcal X$ as well.
For simplicity, we will omit this constraint from the discussion below.

The most common type of attack is based on solving the constrained maximization problem
\begin{equation}
\delta^*=\arg\max_{\delta\in\Delta} \mathcal L(\theta, x+\delta, y),
\label{eq:untargeted-obj}
\end{equation}
which gives us the perturbed input $x+\delta^*$ that \emph{causes the most damage} in terms
of the loss function \emph{within a small domain $\Delta$}.
Here, the set $\Delta$ captures the idea of ``very small perturbation''.
Throughout the paper, we adopt the widely used definition $\Delta_\epsilon=\{\delta: \|\delta\|_\infty\leq\epsilon\}$
that defines the neighborhood of an input $x$ in terms of the maximum absolute difference
in any tensor value.

Another possible type of attack is defined by the constrained minimization problem
\begin{equation}
\delta^*=\arg\min \|\delta\|_\infty\ \ \ \mbox{s.t.}
\ \ \mathcal C(f_\theta(x+\delta), f_\theta(x))>0,
\label{eq:minpert-obj}
\end{equation}
where the function $\mathcal C$ expresses the amount of damage.
This allows the definition of constraints that, for example, require
that the predicted label is wrong (in classification) or that 99\% of the predicted
pixel labels are wrong (in segmentation).
Many early approaches adopted this formalism, for example,~\cite{szegedy13,carlini-wagner17,deepfool16}.

Note that \cref{eq:minpert-obj} does not guarantee the perturbation to stay within a
certain small domain $\Delta$.
Since here, we are interested in perturbations from a given $\Delta_\epsilon$, we will use the clipped perturbation
$\max(\delta^*,\epsilon\frac{\delta^*}{\|\delta^*\|})$ in the case of the attacks that solve the minimization problem in \cref{eq:minpert-obj}.

\subsection{Adversarial Training}

Adversarial training has proven to be a reliable heuristic solution to achieve
robustness~\cite{goodfellow15,saddlepoint-iclr18}.
The idea behind adversarial training is to use adversarial examples during
training as a form of augmentation.
The adversarial examples are always created based on the current model in the given
update step.
Formally, we wish to solve the following learning task:
\begin{equation}
\label{eq:at-task}
    \theta^*=\arg\min_\theta \mathbb{E}_{p(x,y)}[
    \max_{\delta \in \Delta} \mathcal{L}(\theta, x + \delta, y)],
\end{equation}
where $p(x,y)$ is the distribution of the data and we assumed the more usual
maximum damage attack formalism.

In the outer minimization (learning) task one can use an arbitrary learning method, and
in the inner maximization task one can select any suitable attack to
perturb the samples used by the learning algorithm.

\section{Our Battery of Segmentation Attacks}

Every attack is constrained to the perturbation set
$\Delta_\epsilon=\{\delta: \|\delta\|_\infty\leq\epsilon\}$ with
$\epsilon=8/255$, in line with general practice.

We build on the notions in \cref{sec:back}, noting 
that in semantic segmentation, $\mathcal X=[0,1]^{H\times W\times 3}$, that is,
the inputs are 3-channel color images of width $W$ and height $H$, with all the
values normalized into the interval $[0,1]$.
The output space $\mathcal Y$ is $[0,1]^{H\times W\times C}$, where
$C$ is the number of possible categories for each pixel.
Note that $\mathcal Y$ is the softmax output, which defines a probability distribution for
each pixel over the categories that can be used to compute the
final segmentation mask by taking the maximum probability category.

\subsection{PAdam attacks}
We include two attacks of our own: PAdam-CE and PAdam-Cos.
PAdam stands for Projected Adam.
It is an algorithm similar to projected gradient descent (PGD)~\cite{Kurakin2017a,saddlepoint-iclr18},
but it uses the Adam optimizer~\cite{adam15} instead of the vanilla gradient descent
used by PGD.
PAdam can be thought of as an alternative to APGD~\cite{Croce2020a} for adaptive step-size control.
Our two PAdam attacks both use the AMSGrad variant~\cite{Reddi2018a} of Adam with
200 iterations and a step size of $2/255$,
projected onto the feasible solution set $\Delta$ after each update like in PGD.

PAdam-CE solves the problem in \cref{eq:untargeted-obj} using PAdam assuming
the cross entropy loss
\begin{equation}
\mathcal L_{CE}(\theta, x, y) = 
\frac{1}{H W}\sum_{h=1}^H\sum_{w=1}^W\sum_{c=1}^C -y_{h,w,c}\log f_\theta(x)_{h,w,c},
\end{equation}
while
PAdam-Cos solves the problem
\begin{equation}
\delta^*=\arg\min_{\delta\in\Delta} \mathrm{CosSim}(\mathrm{OneHot}(y),F_{\theta}(x+\delta))
\end{equation}
using PAdam, where $F_\theta$ computes the logit layer of $f_\theta$,
$\mathrm{CosSim}(x,y) = \frac{x\cdot y}{\|x\|_2\|y\|_2}$,
and OneHot$(y)$ computes the one-hot encoded ground truth label so that the dimensions of
the label match the logit tensor dimensions.

PAdam-Cos is not to be confused with CosPGD~\cite{Agnihotri2023a} where cosine
similarity is also used but in a different way, combined with cross entropy.
Croce et al.~\cite{Croce2023a} found that SEA dominates CosPGD.
However, PAdam-Cos is clearly not dominated as we demonstrate later.

\subsection{The SEA attack set}
We include the four attacks in SEA introduced by
Croce et al.~\cite{Croce2023a}.
The authors propose an improved version of APGD~\cite{Croce2020a} that uses
progressive radius reduction and apply it for 300 iterations.
They include four attacks defined by four loss functions: \emph{balanced cross-entropy (SEA-BCE)} as used in SegPGD,
\emph{masked cross-entropy (SEA-MCE)}, where pixels that are incorrectly classified are excluded, \emph{Jensen-Shannon divergence (SEA-JSD)}
between the softmax output and the one-hot encoded label, and \emph{masked spherical loss (SEA-MSL)}, 
where the logit of the correct class is minimized, but first the logit is projected on the unit sphere.
The parameter settings we used were identical to those in~\cite{Croce2023a}.

\subsection{Clipped Minimum Perturbation Attacks}
\label{sec:minpertattacks}

We include a number of attacks that solve the problem in \cref{eq:minpert-obj}, also clipping
the result into the bounded perturbation space $\Delta$,
as described in \cref{sec:back-attacks}.
The attacks we include are ALMAProx~\cite{Rony2022a}, DAG~\cite{Xie2017a}, and PDPGD~\cite{Matyasko2021a}.

Proposed by Rony et al.~\cite{Rony2022a}, the ALMAProx attack is a 
proximal gradient method for solving the perturbation
minimization problem in \cref{eq:minpert-obj}, where the constraint requires a
99\% pixel label error.
The problem is transformed into an unconstrained problem by moving the constraints
into the objective using Lagrangian penalty functions.
The parameter settings we used were identical to those in~\cite{Rony2022a}.

Xie et al.\ proposed the dense adversary generation (DAG)
algorithm~\cite{Xie2017a}, which attempts to attack each pixel using a
gradient method that takes into consideration only the correctly predicted
pixels in each gradient step.
The algorithm terminates when reaching the maximum iteration number, or
when all the targeted pixels are successfully attacked.
We can use this algorithm in our evaluation framework by recording the
perturbation size at termination, along with the ratio of the successfully attacked
pixels.
The maximum iteration number was set to 200, and we used two step-sizes:
0.001 and 0.003.

Matyasko and Chau proposed the primal-dual proximal gradient descent adversarial attack
(PDPGD)~\cite{Matyasko2021a} that,
like ALMAProx, also uses proximal splitting but uses a different
optimization method.
PDPGD was adapted to the semantic segmentation task in~\cite{Rony2022a} via
introducing a constraint on each pixel.
We applied the same parameter settings as~\cite{Rony2022a}.

\subsection{Aggregating the Attacks}

We use our set of ten different attacks by running each attack on each input and taking the most successful
result.
Depending on the metric in question, the most successful attack is the one with the smallest pixel accuracy, or the smallest mean IoU,
respectively.
Note that the aggregated performance of the set of attacks can in principle be much better than any of
the individual attacks.

\section{Investigated Models}
\label{sec:models}

We investigate the best known state-of-the-art robust models~\cite{Xu2021b,Gu2022a,Croce2023a}.
We also include a number of our own models to illustrate the effect of some design
choices.
Our model-set includes 22 robust models as described below.

\paragraph{DDC-AT.} Xu et al.~\cite{Xu2021b} propose the DDC attack and design an
adversarial training algorithm based on DDC.
We include their model checkpoints in our set.
These checkpoints are trained over the Cityscapes dataset~\cite{Cordts2016a} and
the PASCAL VOC 2012~\cite{Everingham2010a} dataset
with the PSPNet~\cite{Zhao2017a} and DeepLabv3~\cite{Chen2017a}
architectures, using a ResNet-50 backbone pretrained on ImageNet.
Thus, we include four DDC-AT model instances altogether.
DDC-AT uses a sophisticated method to create the training batches with
50\% adversarial samples and
we do not include any modified versions of the published method.

\paragraph{PGD-AT.} Xu et al.~\cite{Xu2021b} include a simple baseline adversarial
training algorithm that uses the 3-step PGD attack.
The implementation uses training batches with 50\% adversarial
and 50\% clean samples.
Here, we also include the checkpoints the authors shared in all the four settings
similar to those of DDC-AT.
In addition, we train our own robust models as well in all the four settings with
an identical configuration, except using 100\% adversarial batches during training.
Thus, we include eight PGD-AT models altogether.

\paragraph{SegPGD-AT.} 
Gu et al.~\cite{Gu2022a} improve DDC-AT with the help of the SegPGD attack.
They publish measurements with models trained using the 3-step and 7-step SegPGD,
using batches with 50\% adversarial and 50\% clean samples.
Since the checkpoints used in the paper are not available publicly, we trained
our own models using the implementation provided by the authors. 
We created eight models altogether using configurations that are identical
to those of the PGD-AT models.

\paragraph{SEA-AT.}
Croce et al.~\cite{Croce2023a} test their SEA ensemble attack using their own
adversarially trained models.
Note that instead of the SEA attack, they use PGD for adversarial training.
The dataset is an extended PASCAL VOC 2011 dataset, and the architecture 
of the model is UPerNet~\cite{Xiao2018a} with a ConvNeXt~\cite{Liu2022a} pretrained ImageNet backbone.
All the training batches are 100\% adversarial.
We include in our set two checkpoints provided by the authors
that were both trained for 50 epochs with a 5-step PGD and a robust backbone initialization.
The two models use the tiny and the small version of the ConvNeXt architecture, respectively.

\paragraph{Normal.} We also include models trained on clean samples for all the
possible combinations of databases and architectures mentioned above.
The normal PSPNet and DeepLabv3 models are identical to the ones used in~\cite{Xu2021b}.
The normal UPerNet models were trained by us using the implementation of ~\cite{Croce2023a}
for 50 epochs.

\subsection{General Notes on Training}

The internal attacks applied in adversarial training used the $\ell_\infty$-norm neighborhood
$\Delta=\{\delta: \|\delta\|_\infty\leq\epsilon\}$ with $\epsilon=0.03\approx 8/255$, the same
perturbation set the attacks use.
The input channels are scaled to the range $[0,1]$.

We note that on the PASCAL VOC dataset adversarial training was implemented assuming that
the background class is a regular class, both in terms of training and attack.
However, the Cityscapes models were all trained with masking the `void' class out.
We adopt this slightly inconsistent methodology from related work in order to obtain comparable results.
For more details on the training methodology, please refer to the supplementary material (\cref{sec:supp-train}
and the implementation).

\begin{table*}[tbh]
\caption{Clean / robust metrics (\%). PN: PSPNet, DL: DeepLabv3, P: PASCAL VOC 2012, CS: Cityscapes, \cancel{B}: pixels with background ground truth label ignored.}
\label{tab:eval}
\centering
\resizebox{\textwidth}{!}{%
\begin{tabular}{cl}
\, \\
\multirow{6}{*}{\rotatebox[origin=c]{90}{Accuracy}}
&\,\\
&\,\\
&\,\\
&\,\\
&\,\\
&\,\\
\multirow{6}{*}{\rotatebox[origin=c]{90}{CmIoU}}
&\,\\
&\,\\
&\,\\
&\,\\
&\,\\
&\,\\
\multirow{6}{*}{\rotatebox[origin=c]{90}{NmIoU}}
&\,\\
&\,\\
&\,\\
&\,\\
&\,\\
&\,\\
\end{tabular}%
\begin{tabular}{l|c|c|c|c|c|c}
             & Normal       & DDC-AT       & PGD-AT       & SegPGD-AT    & PGD-AT-100    & SegPGD-AT-100 \\\hline
PN+P     & 91.85 / 0.00 & 91.42 / 0.00 & 91.23 / 0.51 & 88.19 / 0.19 & 79.26 / 39.25 & 79.12 / 46.15 \\
DL+P    & 91.81 / 0.00 & 91.45 / 0.01 & 89.81 / 0.00 & 88.23 / 0.01 & 80.08 / 46.16 & 79.37 / 48.04 \\
\rowcolor[gray]{0.85}
PN+P \cancel{B}  & 87.63 / 0.00 & 86.98 / 0.00 & 86.71 / 0.04 & 67.34 / 0.00 & 40.59 / \;\;8.15  & 38.46 / \;\;9.06 \\
\rowcolor[gray]{0.85}
DL+P \cancel{B} & 87.84 / 0.00 & 86.71 / 0.00 & 85.58 / 0.00 & 67.87 / 0.00 & 39.81 / \;\;8.20  & 39.68 / \;\;9.46 \\
PN+CS    & 93.09 / 0.00 & 92.80 / 0.01 & 92.52 / 0.03 & 91.77 / 0.01 & 83.83 / 51.45 & 83.32 / 69.84 \\
DL+CS   & 93.08 / 0.00 & 92.78 / 0.04 & 92.62 / 0.01 & 91.83 / 0.00 & 84.69 / 48.38 & 84.04 / 68.76 \\\hline
PN+P     & 68.87 / 0.00 & 67.53 / 0.00 & 66.73 / 0.06 & 51.59 / 0.02 & 24.31 / \;\;5.37  & 23.27 / \;\;5.88 \\ 
DL+P    & 68.80 / 0.00 & 67.30 / 0.00 & 63.24 / 0.00 & 52.50 / 0.01 & 24.70 / \;\;5.70  & 24.15 / \;\;6.20 \\
\rowcolor[gray]{0.85}
PN+P \cancel{B}  & 78.84 / 0.00 & 78.12 / 0.00 & 77.28 / 0.02 & 53.82 / 0.00 & 25.10 / \;\;4.50  & 23.59 / \;\;4.98 \\
\rowcolor[gray]{0.85}
DL+P \cancel{B} & 79.08 / 0.00 & 77.46 / 0.00 & 75.28 / 0.00 & 55.26 / 0.00 & 24.70 / \;\;4.60  & 24.51 / \;\;5.02 \\
PN+CS    & 66.28 / 0.00 & 63.90 / 0.01 & 61.85 / 0.04 & 59.71 / 0.01 & 32.74 / 16.37 & 31.19 / 20.29 \\
DL+CS   & 66.96 / 0.00 & 64.01 / 0.04 & 63.29 / 0.01 & 59.87 / 0.00 & 34.79 / 15.73 & 33.32 / 20.72 \\\hline
PN+P     & 70.70 / 0.00 & 69.32 / 0.00 & 68.46 / 0.12 & 55.76 / 0.05 & 37.01 / 11.35 & 36.10 / 13.31 \\
DL+P    & 69.00 / 0.00 & 67.92 / 0.00 & 64.01 / 0.00 & 54.94 / 0.01 & 35.23 / 11.61 & 33.85 / 12.21 \\
\rowcolor[gray]{0.85}
PN+P \cancel{B}  & 43.41 / 0.00 & 42.25 / 0.00 & 41.86 / 0.02 & 28.00 / 0.01 & 14.50 / \;\;3.14  & 13.55 / \;\;3.59 \\
\rowcolor[gray]{0.85}
DL+P \cancel{B} & 42.89 / 0.00 & 41.59 / 0.00 & 39.26 / 0.00 & 27.89 / 0.01 & 13.73 / \;\;3.06  & 13.41 / \;\;3.52 \\
PN+CS    & 52.57 / 0.00 & 50.97 / 0.01 & 49.14 / 0.02 & 47.27 / 0.01 & 33.13 / 17.64 & 32.41 / 22.67 \\
DL+CS   & 51.18 / 0.00 & 49.63 / 0.02 & 48.28 / 0.01 & 45.91 / 0.00 & 33.37 / 15.86 & 32.68 / 21.82 \\
\end{tabular}}
\end{table*}

\begin{table*}[tbh]
\caption{Clean / robust metrics of SEA-AT (\%). \cancel{B}: pixels with background ground truth label ignored.}
\label{tab:seaat}
\centering
\begin{tabular}{l|c|c|c|c}
             & Normal-Tiny  & SEA-AT-Tiny   & Normal-Small & SEA-AT-Small \\\hline
Accuracy    & 93.10 / 0.00  & 92.75 / 71.84 & 93.78 / 0.00 & 93.10 / 70.57 \\
Accuracy \cancel{B} & 88.21 / 0.00  & 86.39 / 52.99 & 90.12 / 0.00 & 88.53 / 55.50 \\
\rowcolor[gray]{0.85}
CmIoU       & 72.49 / 0.00  & 72.09 / 32.79 & 75.40 / 0.00 & 73.53 / 32.66 \\
\rowcolor[gray]{0.85}
CmIoU \cancel{B}    & 80.55 / 0.00  & 79.68 / 43.14 & 83.20 / 0.00 & 81.76 / 43.76 \\
NmIoU       & 73.33 / 0.00  & 74.07 / 13.20 & 77.08 / 0.00 & 75.11 / 12.89 \\
NmIoU \cancel{B}    & 43.10 / 0.00  & 42.29 / \;\;7.31  & 45.54 / 0.00 & 43.48 / \;\;7.51  \\
\end{tabular}
\end{table*}

\section{Evaluation}
\label{sec:evaluation}

Our models, datasets and attacks allow for 280 possible combinations.
We evaluated all these combinations.
Here, we present a representative sample of our results to support our main
findings, the complete set of results is presented in the supplementary material.

We apply the same evaluation methodology for every model we study, as we discuss below.
This methodology differs from the ones used in the original evaluations of these models.
We present the details of the original methodologies in the supplementary material in \cref{sec:supp-method},
noting here that some of these details are not documented and had to be learned directly from the implementation.

\textbf{The evaluation procedure.}
Following general practice, we evaluate over the validation sets of the PASCAL VOC 2012 and Cityscapes
datasets.
On PASCAL VOC, before evaluation, the longer dimension of each image is scaled to 512 pixels, but no further
rescaling or cropping is applied.
On Cityscapes, the images are scaled down to 1024x512.
After computing the prediction mask, no augmentation method is applied to improve the mask.

Here, we deviate from related work in that Croce et al.~\cite{Croce2023a} resize to 512x512 and then
crop to 473x473, and Gu et al.\ and Xu et al.~\cite{Gu2022a,Xu2021b} use a tiled evaluation with
overlapping tiles, and in addition, over the clean inputs further augmentations are applied based on
mirrored inputs.
Due to these differences, our measurements are not identical to those published in the original works.
However, when applying these techniques, we are able to reproduce the published values.

\textbf{How to aggregate IoU?}
We use \emph{Pixel Accuracy} (or simply accuracy) and \emph{mean intersection over union (mIoU)} as
our performance metrics.
Regarding mIoU, the aggregation procedure to compute the global mIoU value based on the
images in the evaluation set is often overlooked.
We will demonstrate that in the case of evaluating robustness, this aggregation procedure plays an important role.
The usual approach, also used in~\cite{Gu2022a,Xu2021b,Croce2023a}, performs an aggregation over
the images to compute the global IoU of each class, and then computes the average:
\begin{equation}
\mbox{CmIoU} = \frac{1}{C}\sum_{c=1}^C \frac{\sum_{n=1}^N \mbox{TP}_{cn}}
{\sum_{n=1}^N \mbox{FP}_{cn}+\mbox{TP}_{cn}+\mbox{FN}_{cn}},
\end{equation}
where $N$ is the number of images and $C$ is the number of classes.

However, we will also apply a different aggregation to emphasize the errors made
on individual images, which is the main goal of adversarial attacks.
Here, we simply average the image-wise IoU values:
\begin{equation}
\mbox{NmIoU} = \frac{1}{NC}\sum_{n=1}^N \sum_{c=1}^C \frac{\mbox{TP}_{cn}}
{\mbox{FP}_{cn}+\mbox{TP}_{cn}+\mbox{FN}_{cn}}.
\end{equation}
This metric captures the image-wise performance much better,
especially over datasets where there are only a few objects in most images,
some of which are large and some are small.
This is the case, for example, in the PASCAL VOC dataset.

\subsection{Results}

\cref{tab:eval} contains our results with the DDC-AT and SegPGD-AT models, along with
baselines (normal model and PGD-AT models).
As described in \cref{sec:models},
the two models that use 100\% adversarial batches during training (PGD-AT-100 and SegPGD-AT-100) were trained by us,
just like SegPGD-AT.
The remaining models are checkpoints from~\cite{Xu2021b}.

\cref{tab:seaat} shows our results with the SEA-AT models.
These models are all checkpoints taken from~\cite{Croce2023a}.
The normal models were trained by ourselves.

The tables contain clean and robust metrics.
The robust metrics were computed over the adversarially perturbed inputs
that were generated using the aggregated attack of our set of ten attacks.

\textbf{No robustness with 50\% adversarial batches.}
The most striking observation is that our aggregated attack achieves 
a value of near zero according to all the three metrics for all the
models that use only 50\% adversarial samples. 
This means that the models published in~\cite{Xu2021b,Gu2022a} show
\emph{no sign of robustness}. (See \cref{sec:supp-minpert} in the supplementary material for further evidence.)

\textbf{Using 100\% adversarial samples is not sufficient.}
While we can achieve non-trivial robustness in terms of pixel accuracy,
our attacks significantly reduce both the CmIoU and NmIoU metrics
in the case of PGD-AT-100 and SegPGD-AT-100, indicating a very low robustness.
Also, the performance of these models over the clean samples is much worse than that of the normal model.

In the case of the SEA-AT models (\cref{tab:seaat})
the clean performance is very close to that of the normal model according to
all the metrics.
However, robust NmIoU drops to a fraction of the clean NmIoU, again, indicating
a very low level of robustness.

\begin{figure}[t]
\centering
\setlength{\tabcolsep}{3pt}
\footnotesize
\begin{tabular}{cccc}
& Normal, Clean & SegPGD-AT-100, clean & SegPGD-AT-100, robust \\
\rotatebox[origin=l]{90}{\hspace{8mm}Cityscapes} & 
\includegraphics[trim= 2mm 0mm 4mm 2mm,clip,width=.32\textwidth]{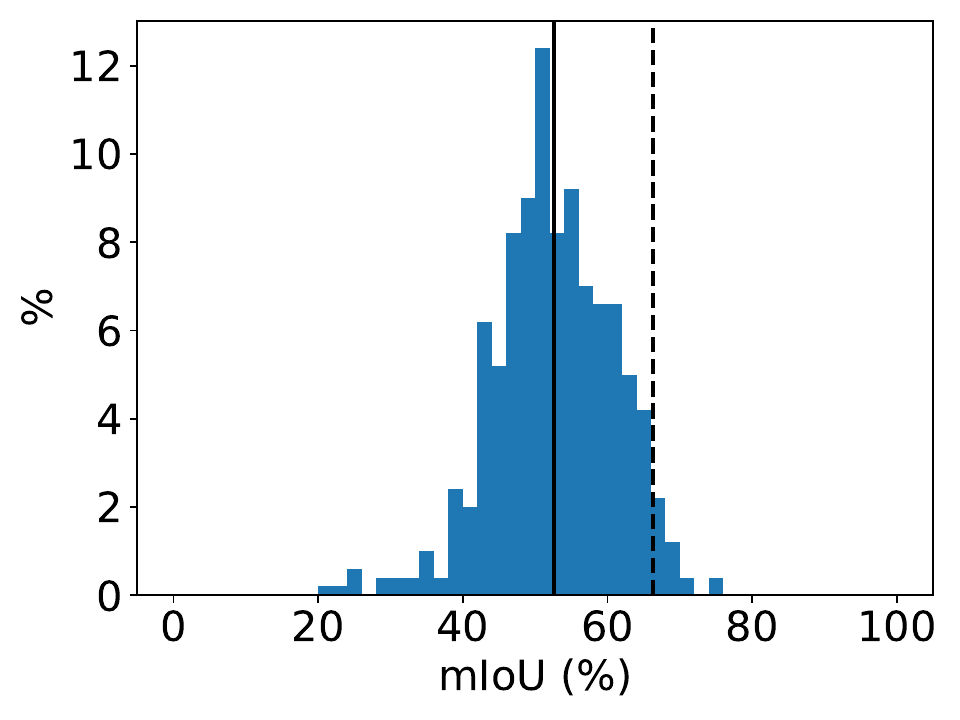} &
\includegraphics[trim= 13mm 0mm 4mm 2mm,clip,width=.3\textwidth]{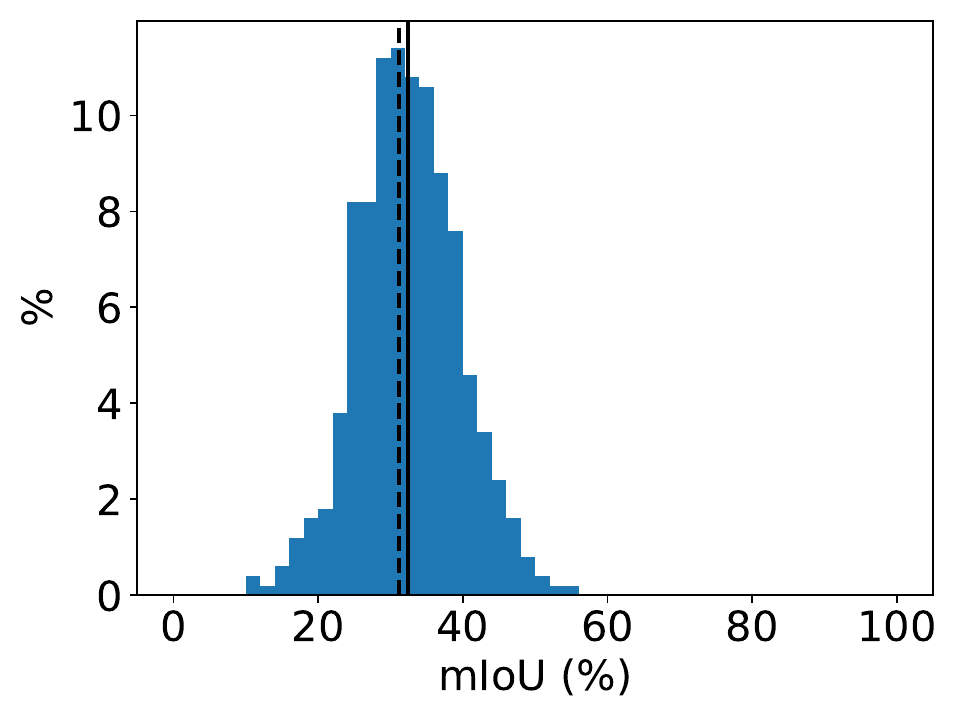} &
\includegraphics[trim= 13mm 0mm 4mm 2mm,clip,width=.3\textwidth]{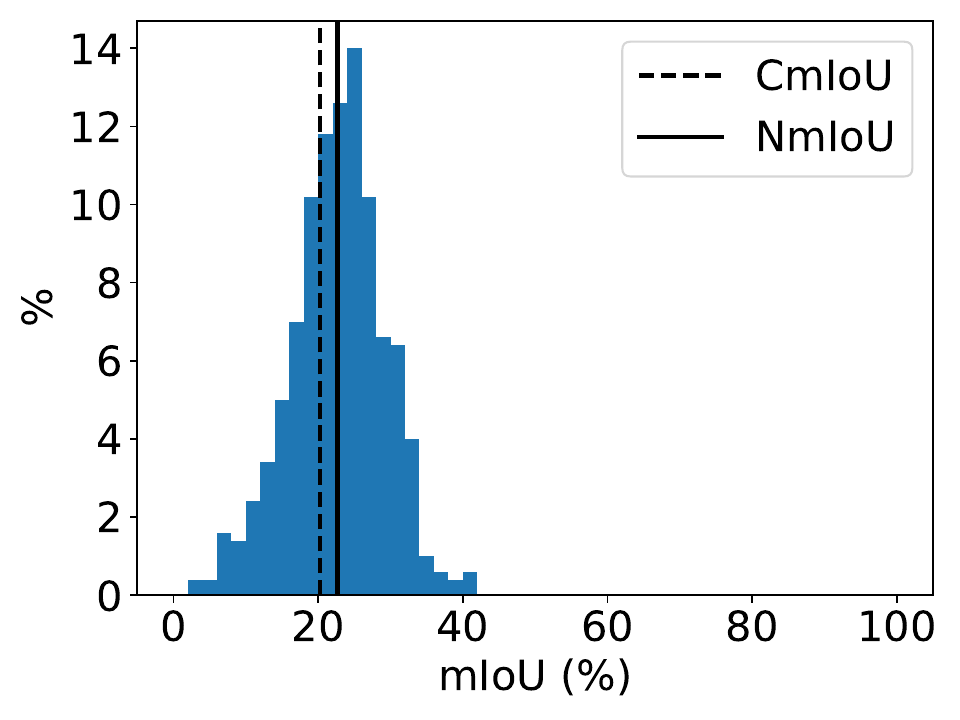} \\
\rotatebox[origin=l]{90}{\hspace{1mm}PASCAL VOC 2012} & 
\includegraphics[trim= 2mm 0mm 4mm 2mm,clip,width=.32\textwidth]{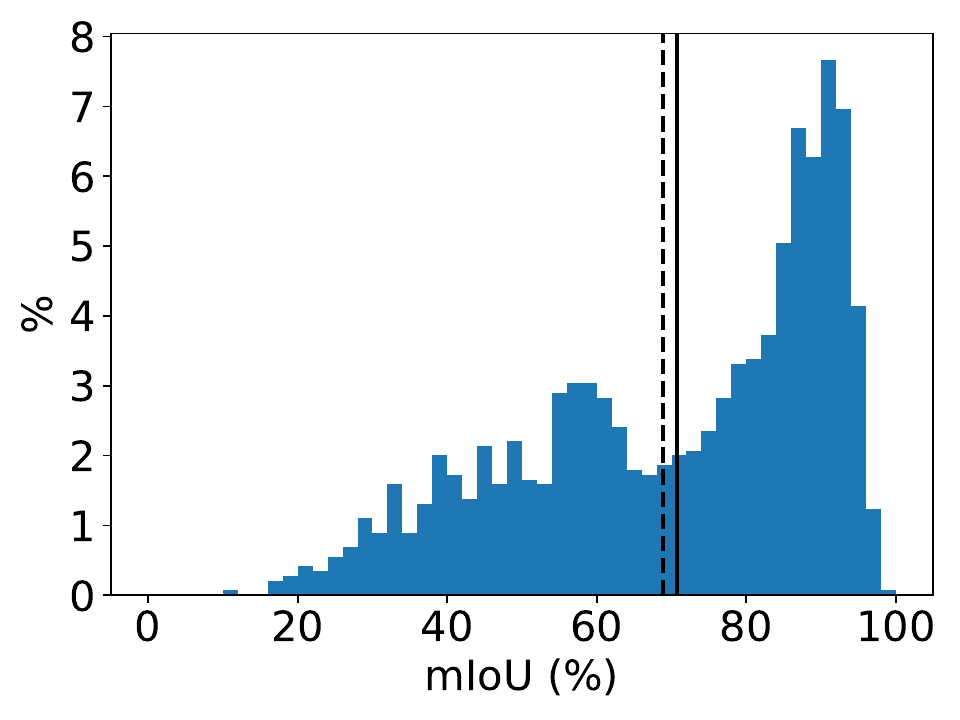} &
\includegraphics[trim= 12mm 0mm 4mm 2mm,clip,width=.3\textwidth]{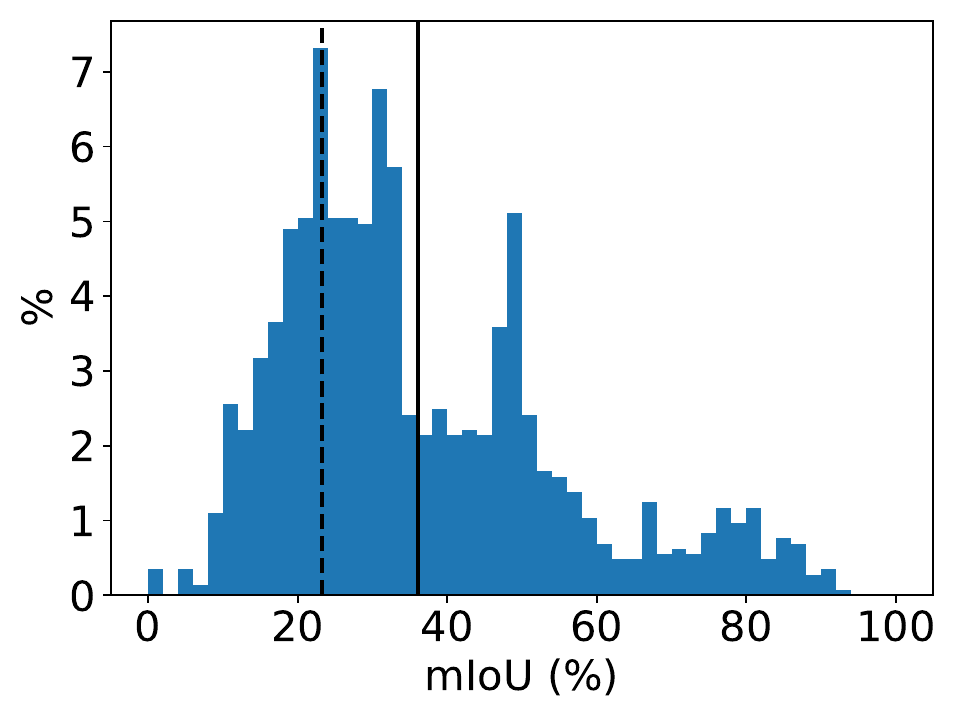} &
\includegraphics[trim= 13mm 0mm 4mm 2mm,clip,width=.3\textwidth]{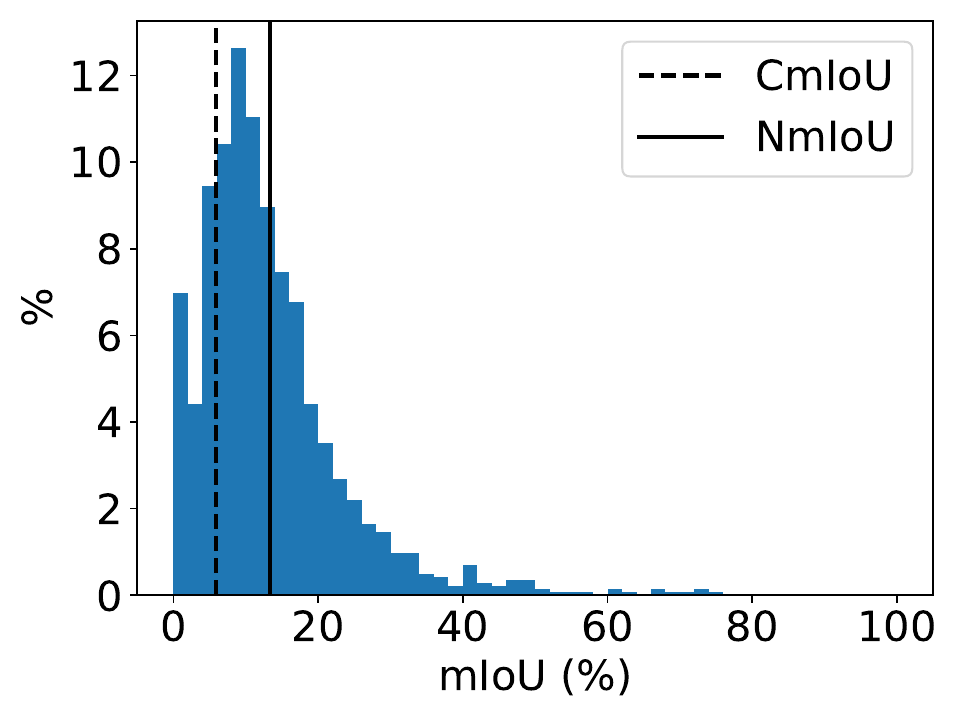} \\
\end{tabular}
\caption{Distributions of single image mIoU with PSPNET for the Normal model and for SegPGD-AT-100, for clean
and adversarial inputs.
CmIoU and NmIoU are shown using vertical lines.
}
\label{fig:miouhistograms}
\end{figure}

\textbf{Size-bias in the SEA-AT models.}
In the case of the SEA-AT models the robust CmIoU is significantly higher than the robust NmIoU,
while the rest of the models show the opposite behavior.
To understand this better, we take a closer look at the distribution of the single image mIoU over the evaluation set.
\cref{fig:miouhistograms,fig:teaser} show a representative sample of such distributions, with
CmIoU and NmIoU indicated. (For more, please refer to \cref{sec:supp-miouhistograms}).

\setlength{\intextsep}{3pt plus 1.0pt minus 2.0pt}
\begin{wrapfigure}[17]{r}{0.6\textwidth}
\centering
\setlength{\tabcolsep}{0.2pt}
\scriptsize
\begin{tabular}{ccc}
Label & SEA-Tiny & SegPGD-100\\
\includegraphics[height=16mm,width=0.18\columnwidth]{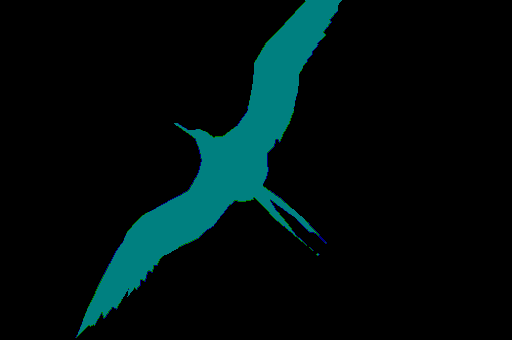}&
\includegraphics[height=16mm,width=0.18\columnwidth]{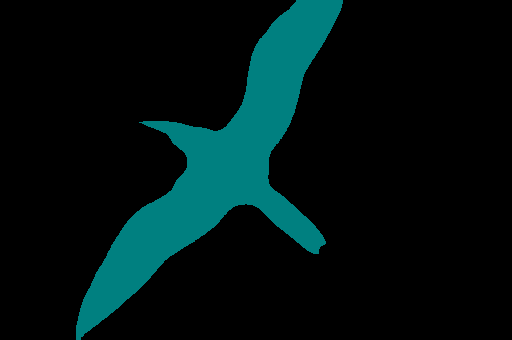}&
\includegraphics[height=16mm,width=0.18\columnwidth]{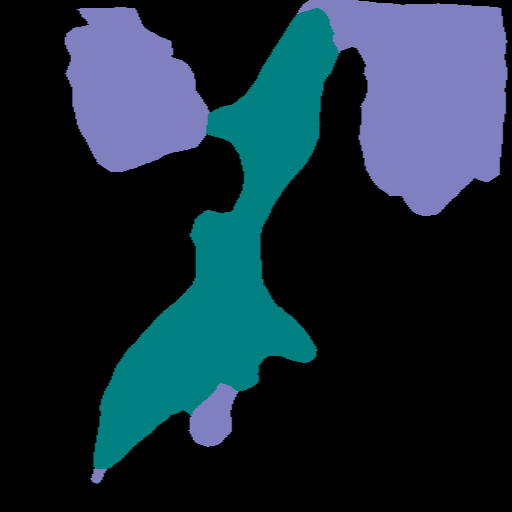}\\[-0.5mm]
\includegraphics[height=16mm,width=0.18\columnwidth]{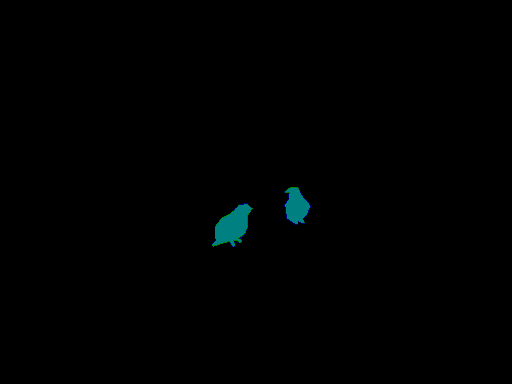}&
\includegraphics[height=16mm,width=0.18\columnwidth]{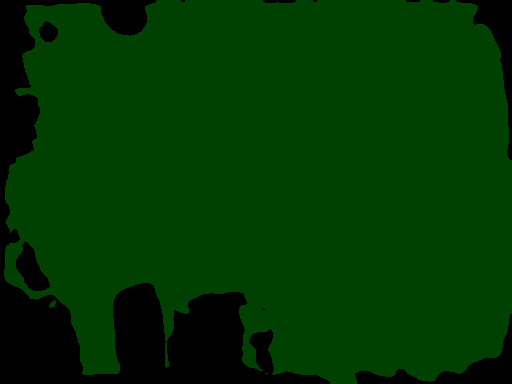}&
\includegraphics[height=16mm,width=0.18\columnwidth]{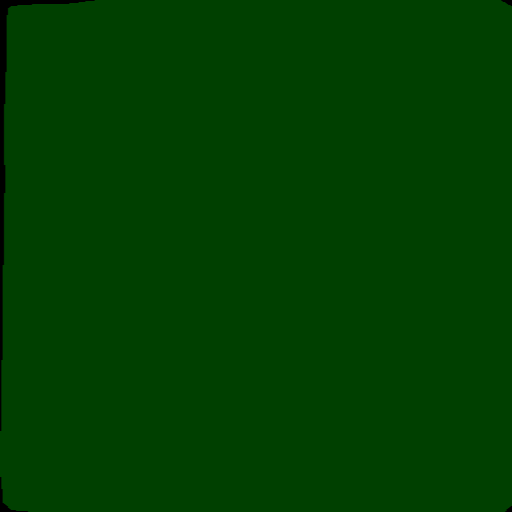}
\end{tabular}
\caption{Illustration of the size-bias on two bird samples.
The predicted masks on adversarial input are shown for two models.
The small birds are deleted completely by our adversarial attacks,
while the large bird is better preserved, especially by SEA-AT-Tiny.
The NmIoU metric captures this important problem better than CmIoU or accuracy do.}
\label{fig:smallremove}
\end{wrapfigure}

It is striking that in the case of the PASCAL VOC dataset the robust single image mIoU distributions of
SegPGD-AT-100 (\cref{fig:miouhistograms}) and SEA-AT-S (\cref{fig:teaser}) are very similar but
the aggregated mIoU measures dramatically differ.
Since the NmIoU metric is more sensitive to errors in the segmentation of small objects, a
possible explanation is that the SEA-AT model learns to protect the largest objects while 
it sacrifices the small ones.
In other words, the attacks
essentially remove smaller objects from the image, causing very low mIoU values on many images.
This problem is not captured well by the CmIoU or the accuracy metrics.
Looking at the predicted masks confirms this hypothesis.
\cref{fig:smallremove} shows an example for this phenomenon.

\textbf{The foreground is more vulnerable.}
In the PASCAL VOC dataset, the images contain lots of pixels in the
background class that might dominate our measurements.
In fact, the background pixels form 74\% of all the pixels in the validation set.

To examine the effect of the background class, in all the tables, we also show the measurement results with the background pixels removed.
That is, the metrics are computed only on the foreground pixels, and the IoU of the background class
is not taken into account.

We can see that the accuracy and the NmIoU over the foreground are much worse than on the complete image.
This problem is more severe in the case of SegPGD-AT-100 and PGD-AT-100.
These  models essentially focus on predicting and protecting the background,
which is not the intended behavior.
The SEA-AT models also show this effect to some extent.

\textbf{Robustness-accuracy tradeoff.}
Our results confirm that the apparent lack of robustness-accuracy tradeoff can be
explained by the insufficient evaluation methodology both in terms of
using only weak attacks, or not using the right performance metrics.
Indeed, those models that have a good clean performance, close to that of the normally
trained models, turn out not to be very robust, when measured appropriately.
Clearly, the best models we examined are the SEA-AT models, but even those models show
a weak performance in the NmIoU metric due to their mIoU distribution (\cref{fig:teaser}).

\begin{figure*}[t]
\centering
\setlength{\tabcolsep}{1pt}
\scriptsize
\begin{tabular}{ccccc}
& Normal & DDC-AT & PGD-AT & SegPGD-AT-100\\
\rotatebox[origin=l]{90}{\ PN+Cityscapes} &
\includegraphics[trim= 6mm 0mm 16mm 13mm,clip,width=.255\textwidth]{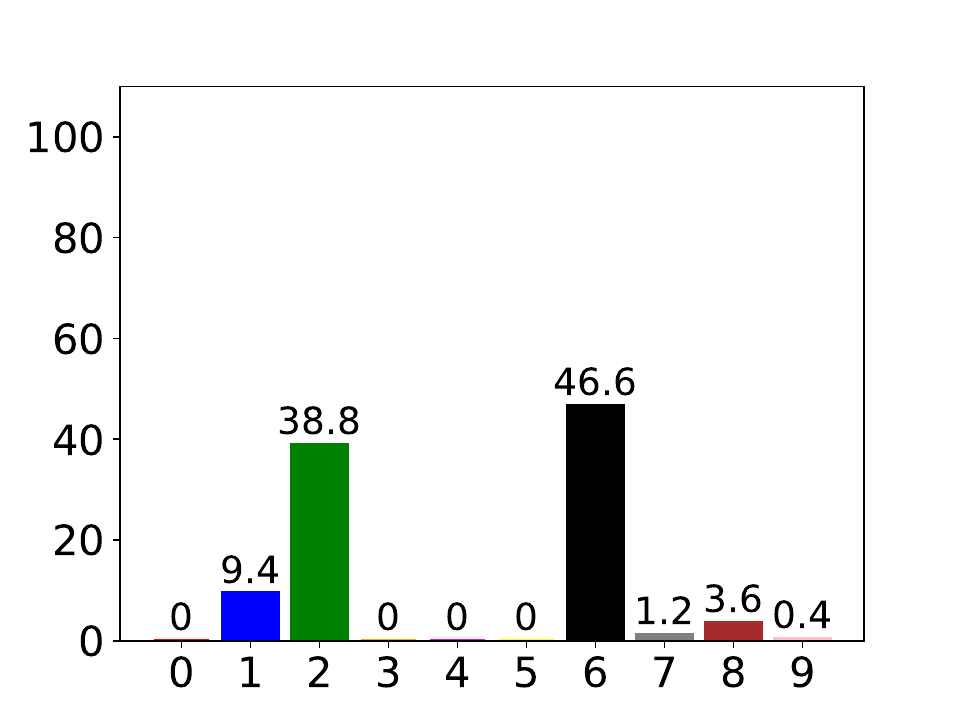} &
\includegraphics[trim= 20mm 0mm 16mm 13mm,clip,width=.23\textwidth]{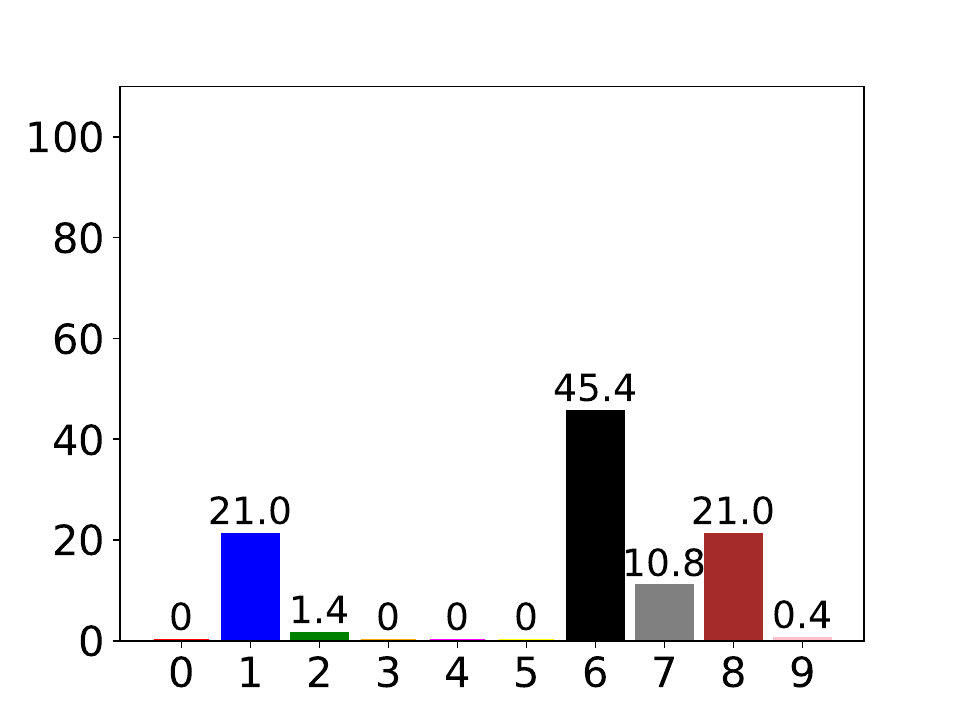} &
\includegraphics[trim= 20mm 0mm 16mm 13mm,clip,width=.23\textwidth]{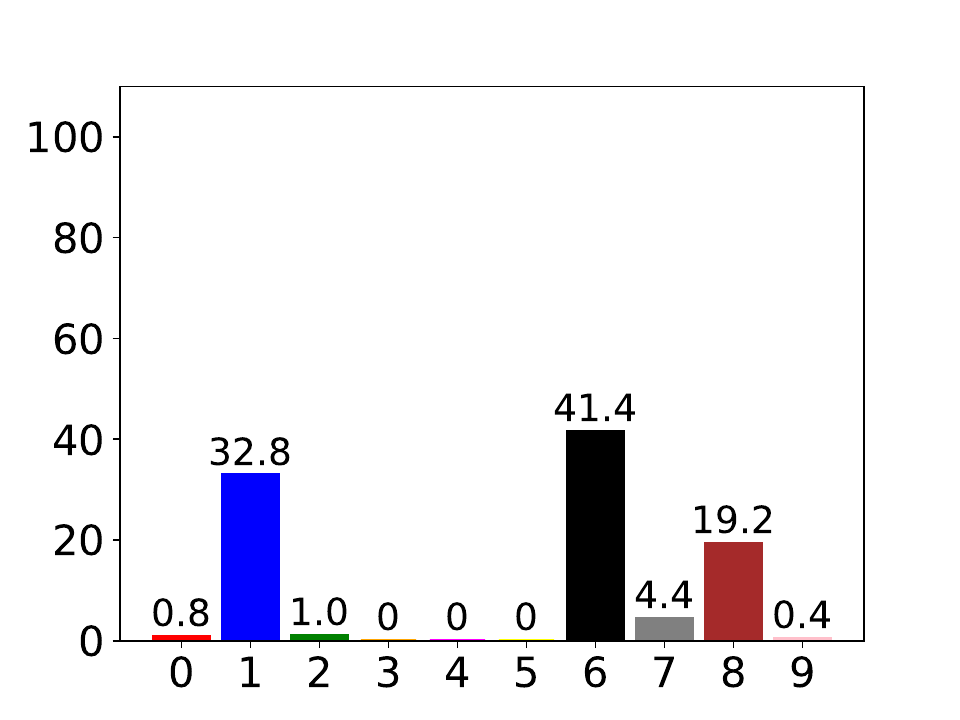} &
\includegraphics[trim= 20mm 0mm 16mm 13mm,clip,width=.23\textwidth]{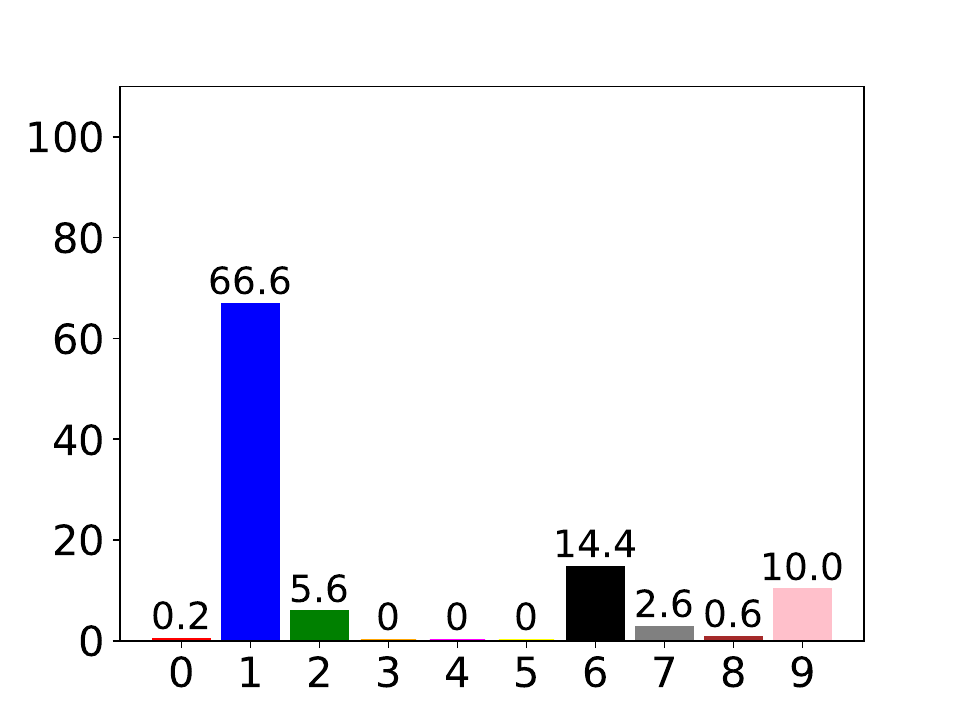}\\
\rotatebox[origin=l]{90}{\ \ PN+PASCAL} &
\includegraphics[trim= 6mm 0mm 16mm 13mm,clip,width=.255\textwidth]{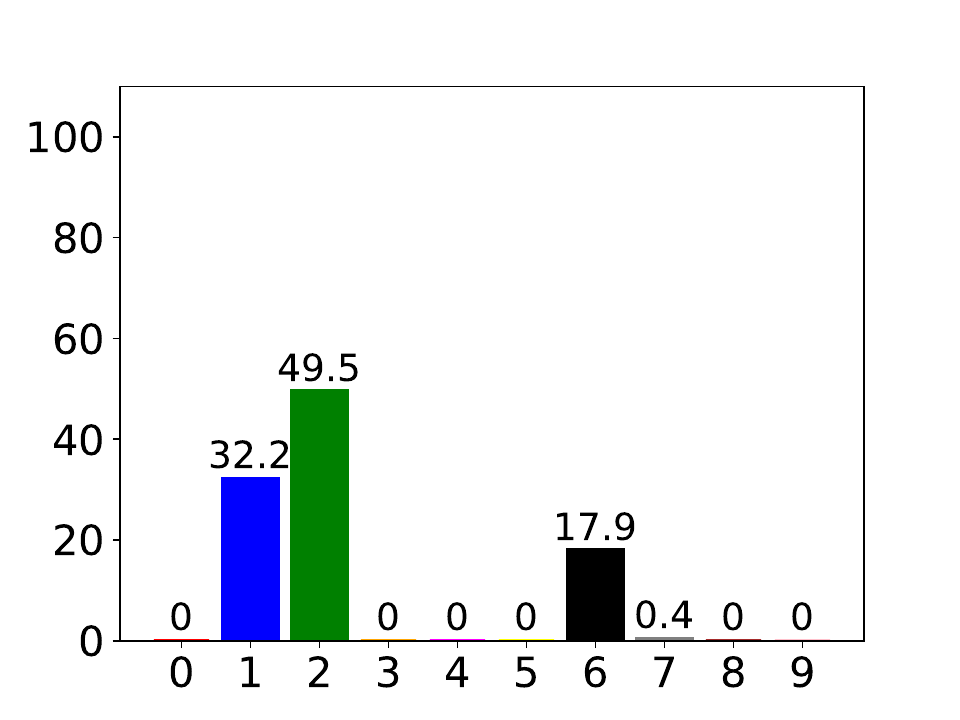} &
\includegraphics[trim= 20mm 0mm 16mm 13mm,clip,width=.23\textwidth]{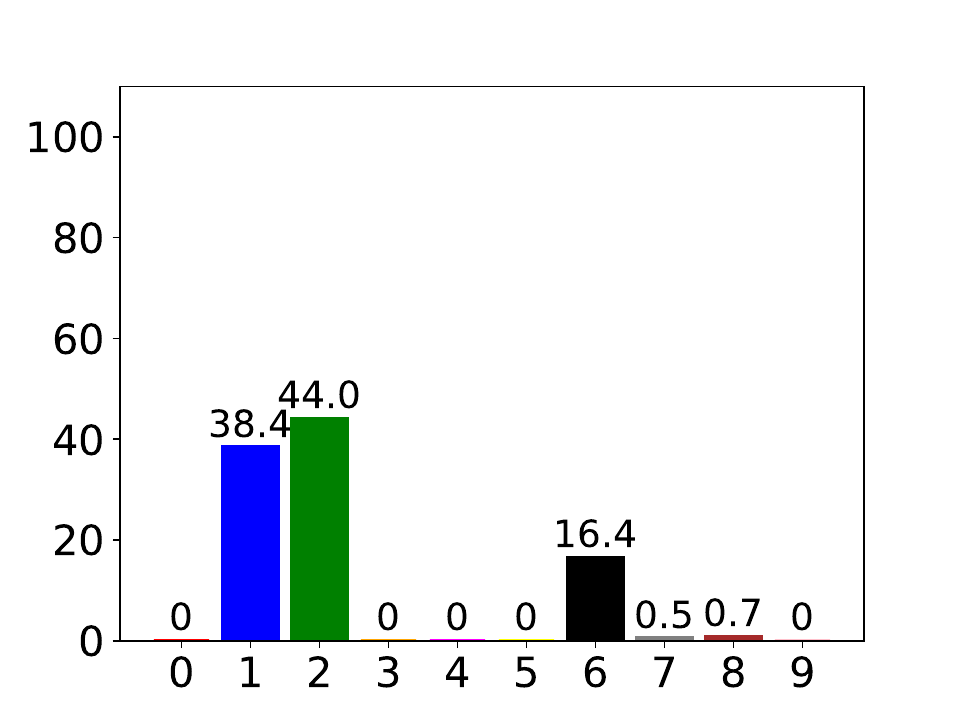} &
\includegraphics[trim= 20mm 0mm 16mm 13mm,clip,width=.23\textwidth]{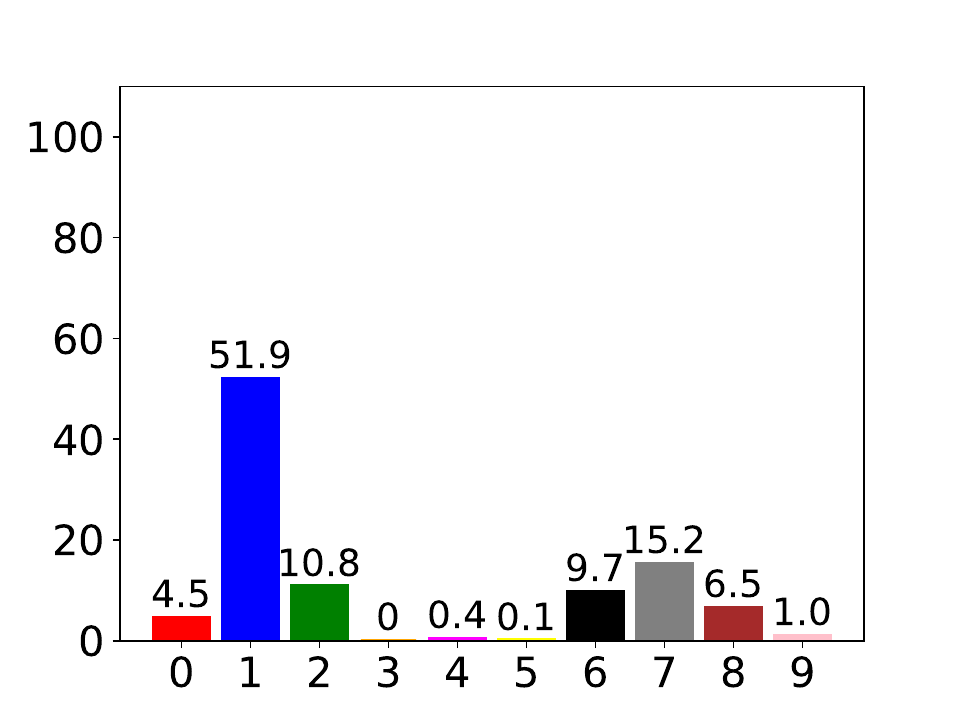} &
\includegraphics[trim= 20mm 0mm 16mm 13mm,clip,width=.23\textwidth]{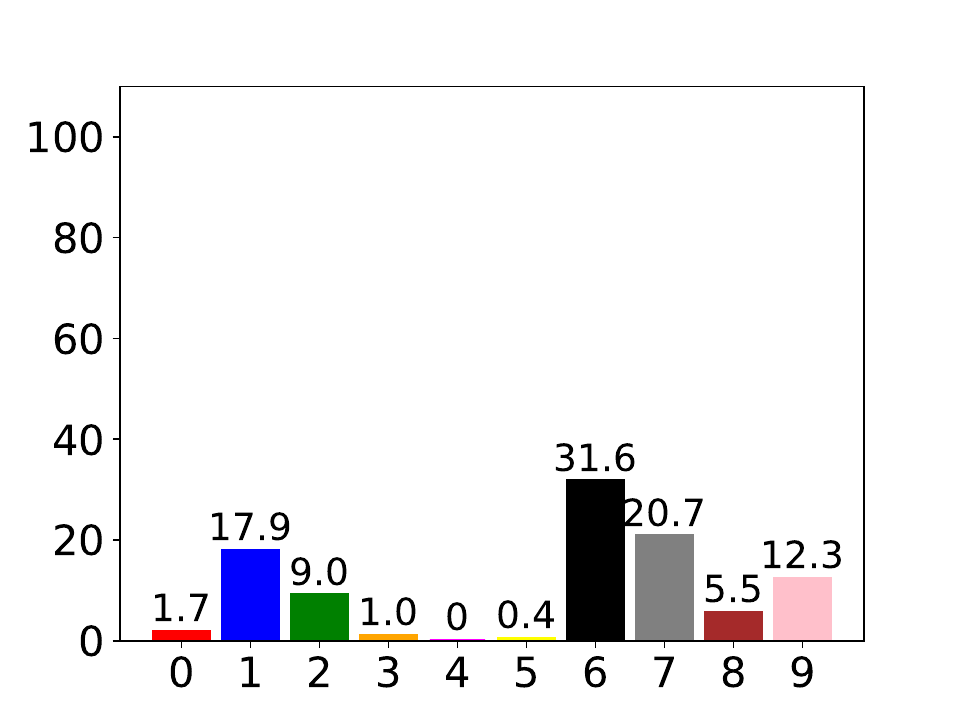}\\
& Normal-Tiny & Tiny & Normal-Small & Small\\
\rotatebox[origin=l]{90}{\ SEA-AT family} &
\includegraphics[trim= 6mm 0mm 16mm 13mm,clip,width=.255\textwidth]{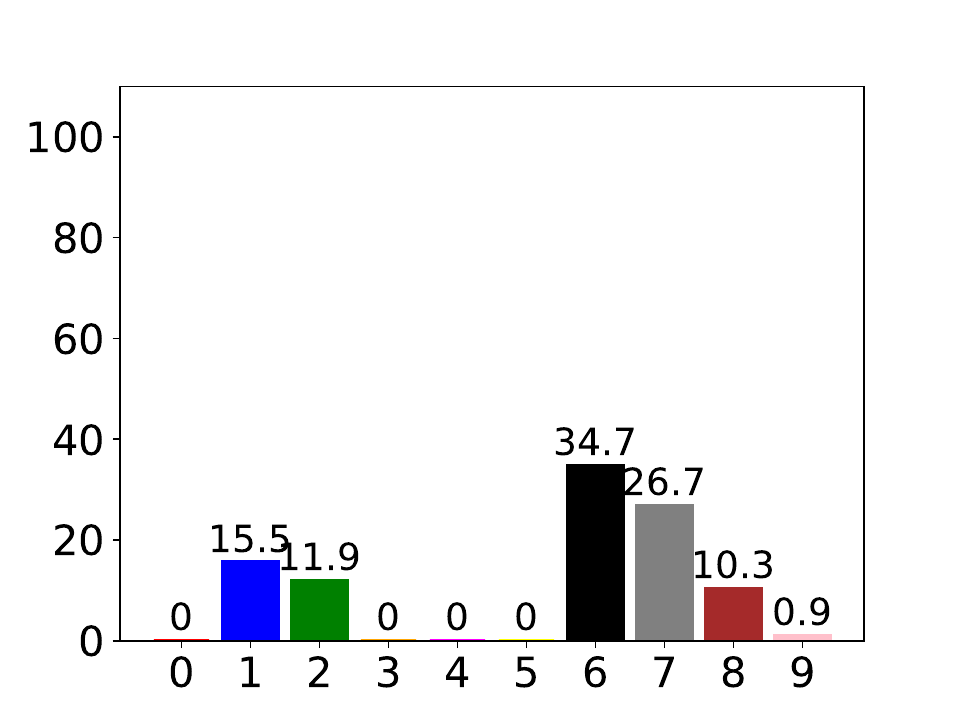} &
\includegraphics[trim= 20mm 0mm 16mm 13mm,clip,width=.23\textwidth]{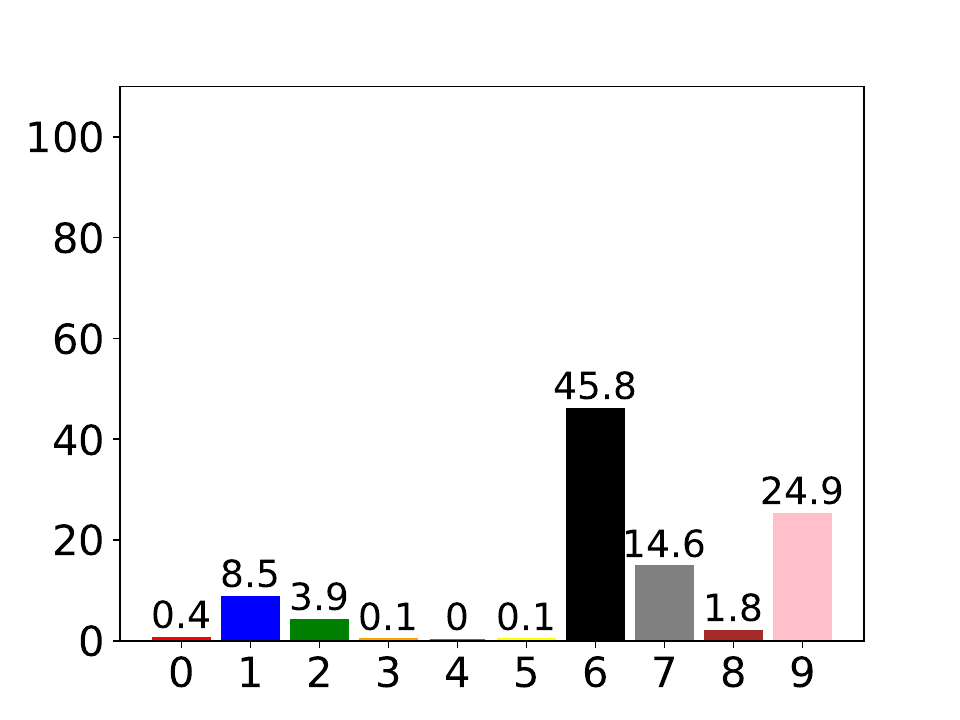} &
\includegraphics[trim= 20mm 0mm 16mm 13mm,clip,width=.23\textwidth]{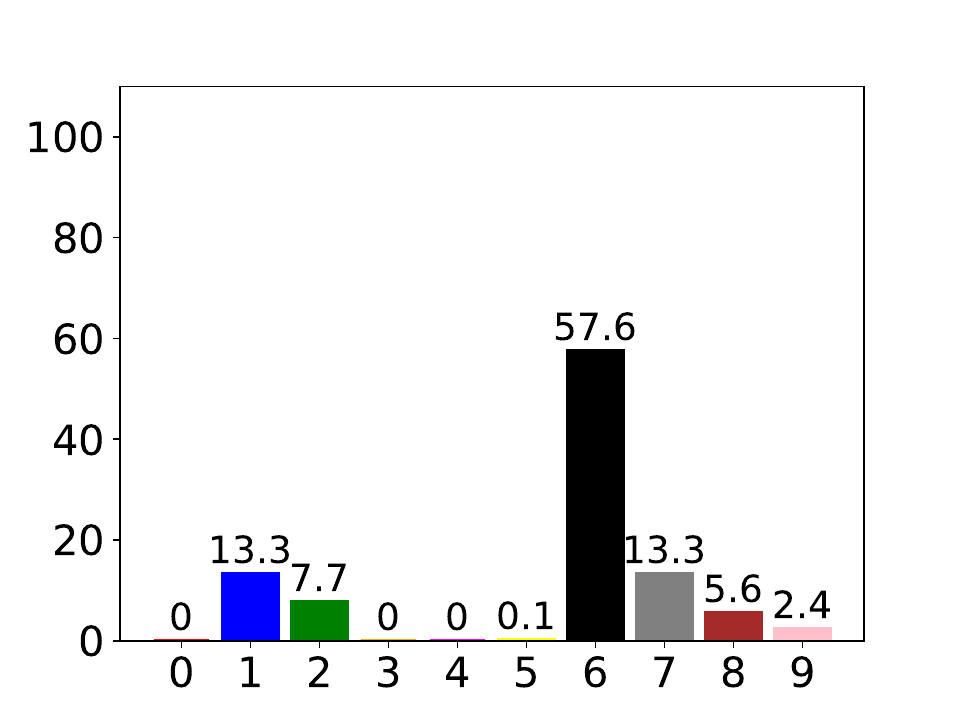} & 
\includegraphics[trim= 20mm 0mm 16mm 13mm,clip,width=.23\textwidth]{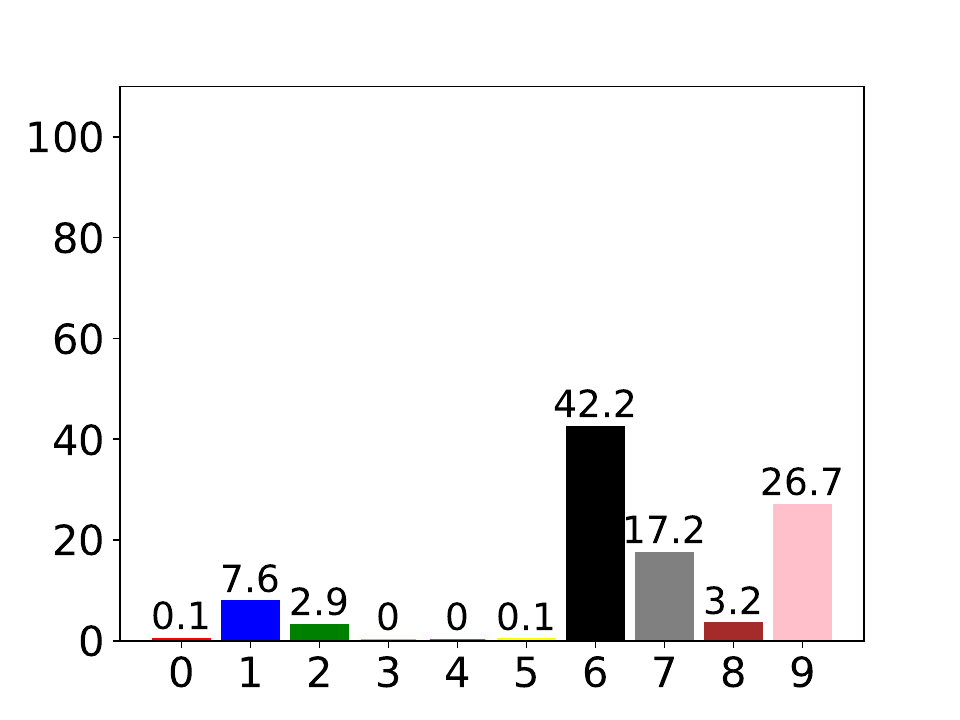} \\
\end{tabular}
\caption{Best attack distribution over the validation set according to mIoU (PN: PSPNet).
The attacks are 0: ALMAProx, 1: PAdam-CE, 2: PAdam-Cos, 3: DAG-0.001, 4: DAG-0.003, 5: PDPGD, 6: SEA-JSD, 7: SEA-MCE, 8: SEA-MSL, 9: SEA-BCE.}
\label{fig:histograms}
\end{figure*}

\begin{figure*}[t]
\centering
\setlength{\tabcolsep}{.2pt}
\scriptsize
\begin{tabular}{ccccccc}
No Attack & PAdam-CE & PAdam-Cos & SEA-JSD & SEA-MCE & SEA-MSL & SEA-BCE \\
\includegraphics[width=.14\textwidth]{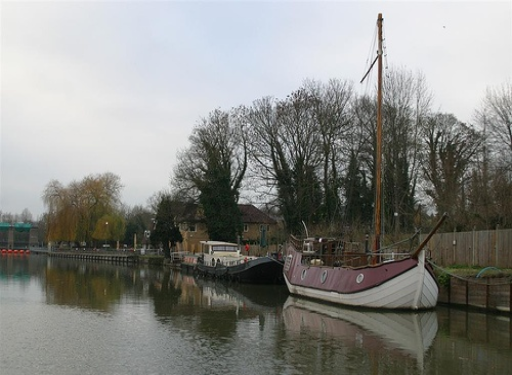} &
\includegraphics[width=.14\textwidth]{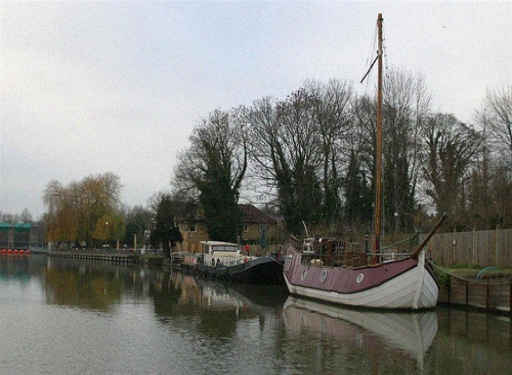} &
\includegraphics[width=.14\textwidth]{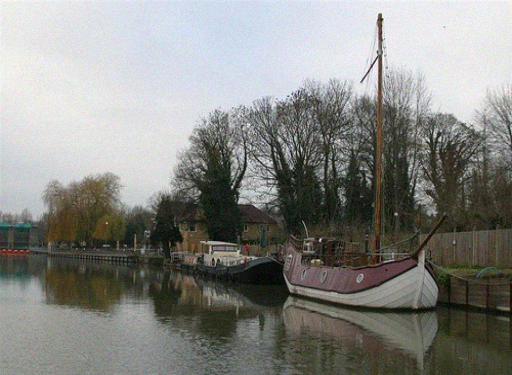} &
\includegraphics[width=.14\textwidth]{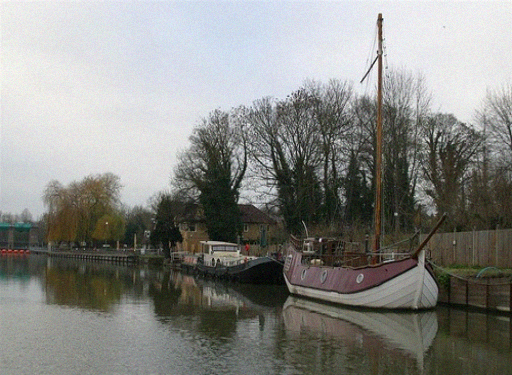} &
\includegraphics[width=.14\textwidth]{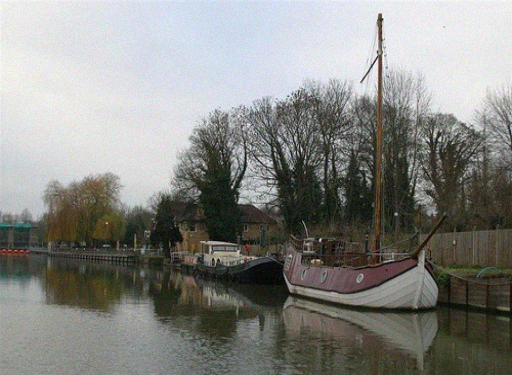} &
\includegraphics[width=.14\textwidth]{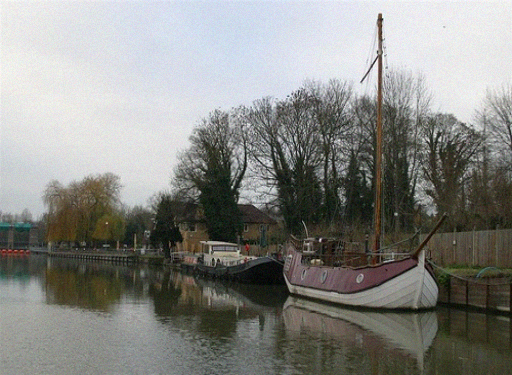} &
\includegraphics[width=.14\textwidth]{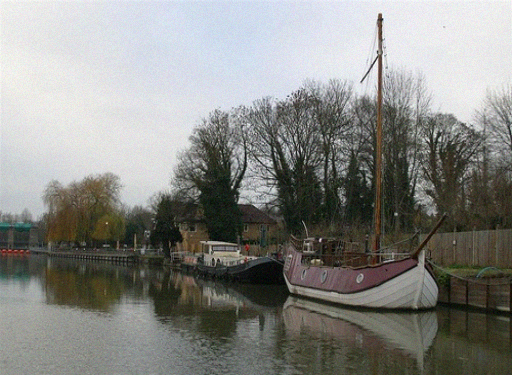}\\[-.5mm]
\includegraphics[width=.14\textwidth]{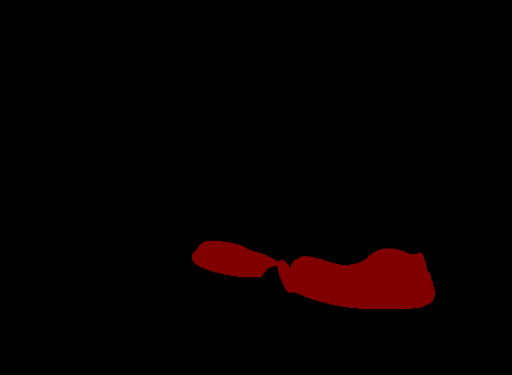} &
\includegraphics[width=.14\textwidth]{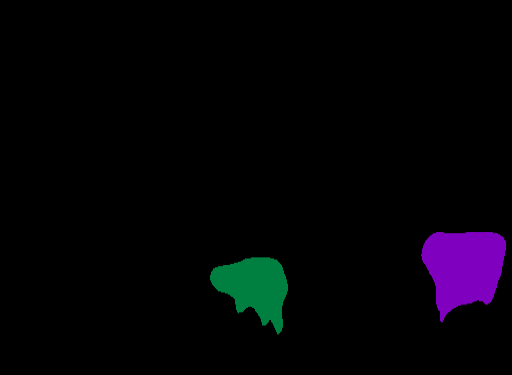} &
\includegraphics[width=.14\textwidth]{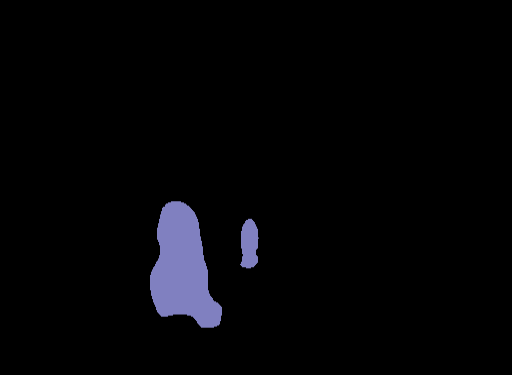} &
\includegraphics[width=.14\textwidth]{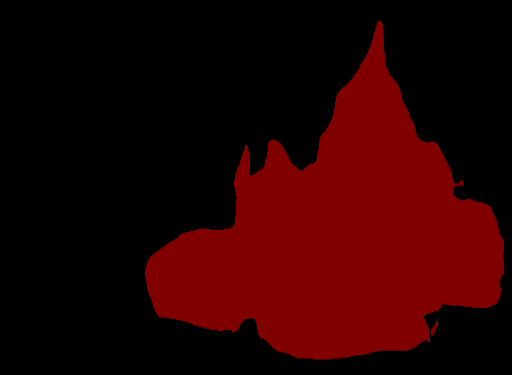} &
\includegraphics[width=.14\textwidth]{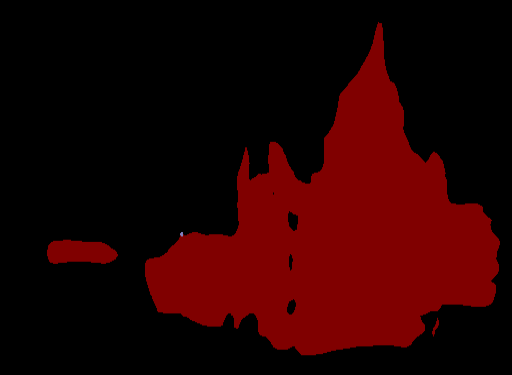} &
\includegraphics[width=.14\textwidth]{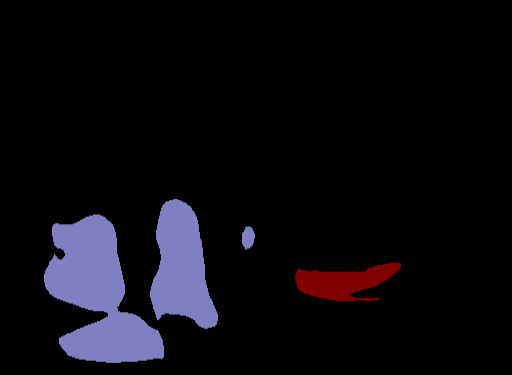} &
\includegraphics[width=.14\textwidth]{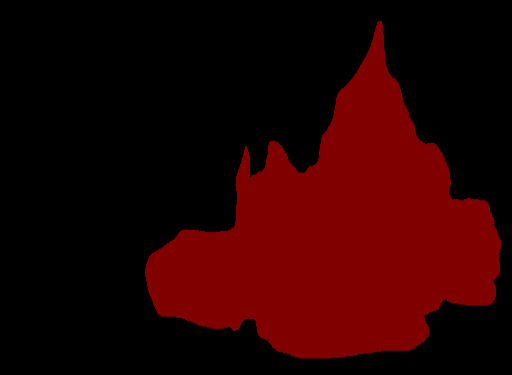} \\
\end{tabular}
\caption{Result of some attacks on an example PASCAL-VOC image on SEA-AT-Tiny.
Top row: perturbed images; bottom row: predicted mask on the perturbed image.}
\label{fig:examples-attacks}
\end{figure*}

\begin{figure*}[t]
\centering
\setlength{\tabcolsep}{.2pt}
\scriptsize
\begin{tabular}{ccccccc}
Normal & DDC-AT & PGD-AT & SegPGD-AT & PGD-AT-100 & SegPGD-100 & SEA-Small\\
\includegraphics[width=.14\textwidth]{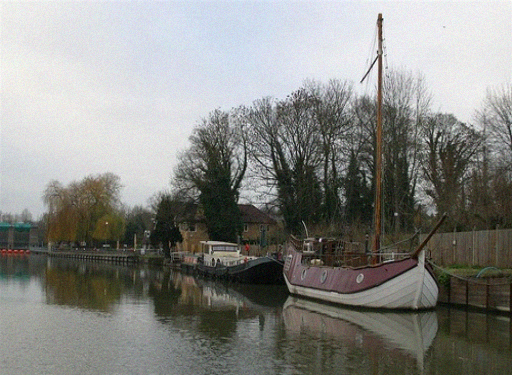} &
\includegraphics[width=.14\textwidth]{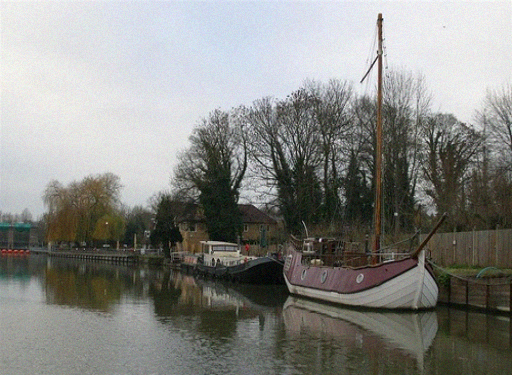} &
\includegraphics[width=.14\textwidth]{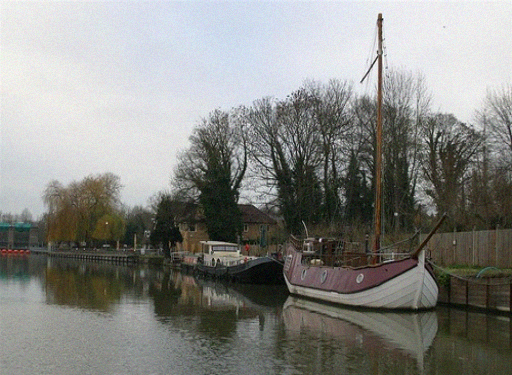} &
\includegraphics[width=.14\textwidth]{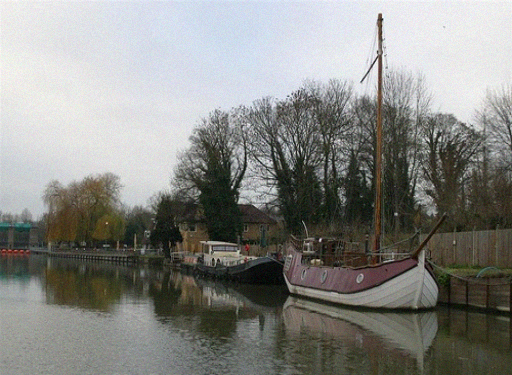} &
\includegraphics[width=.14\textwidth]{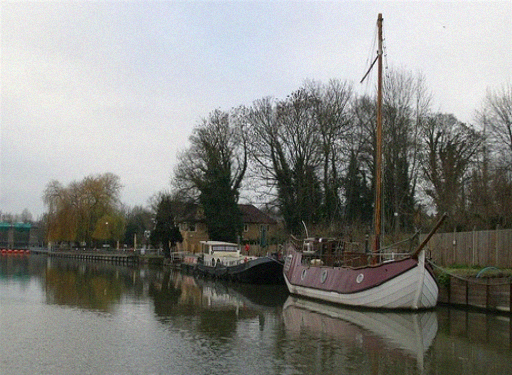} &
\includegraphics[width=.14\textwidth]{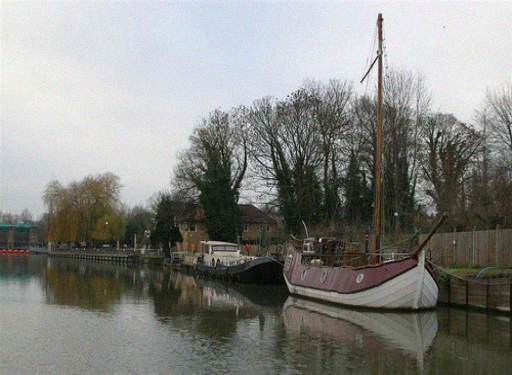} &
\includegraphics[width=.14\textwidth]{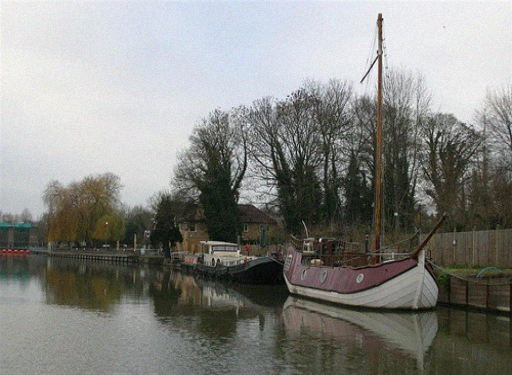} \\[-.5mm]
\includegraphics[width=.14\textwidth]{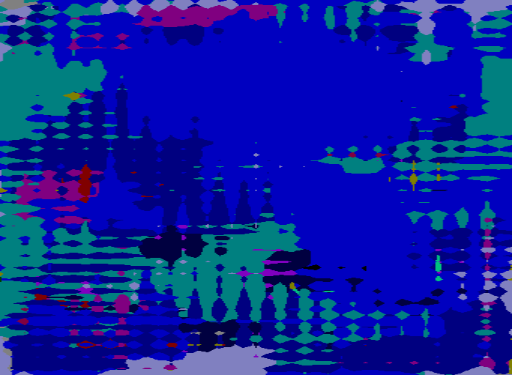} &
\includegraphics[width=.14\textwidth]{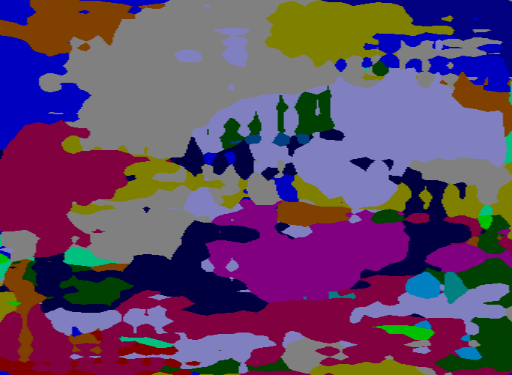} &
\includegraphics[width=.14\textwidth]{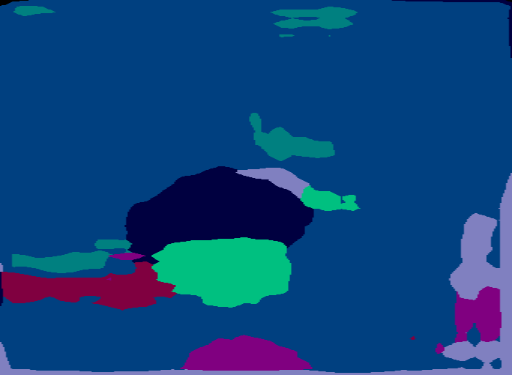} &
\includegraphics[width=.14\textwidth]{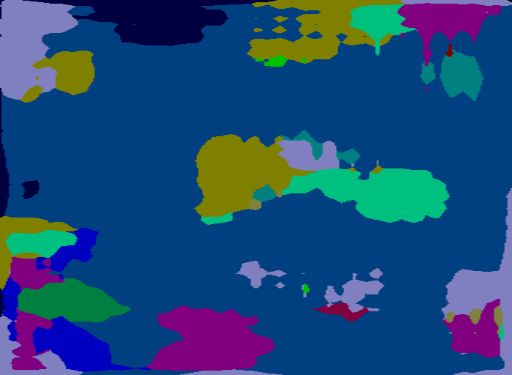} &
\includegraphics[width=.14\textwidth]{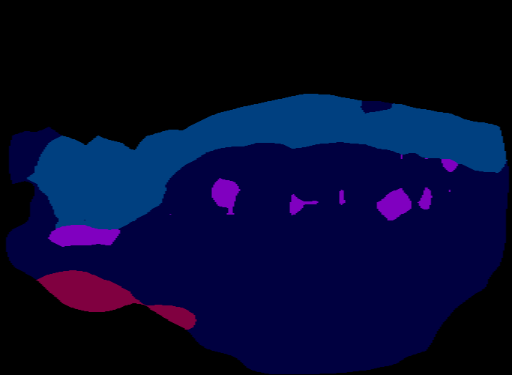} &
\includegraphics[width=.14\textwidth]{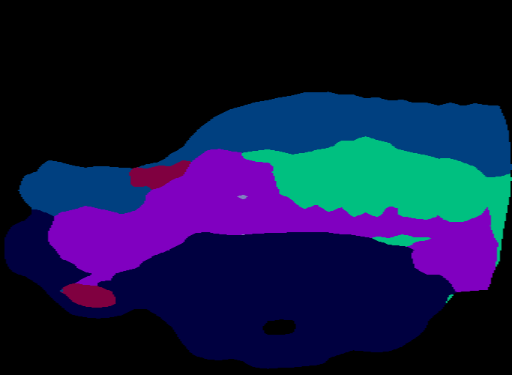} &
\includegraphics[width=.14\textwidth]{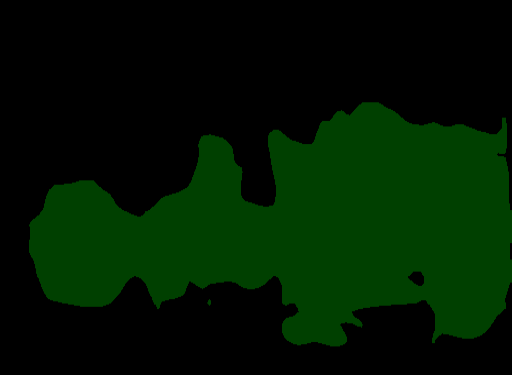} \\
\end{tabular}
\caption{Result of PAdam-Cos attack on an example PASCAL-VOC image for various PSPNet models and SEA-AT-Small.
Top row: perturbed images; bottom row: predicted mask on the perturbed image.}
\label{fig:examples-models}
\end{figure*}

\subsection{A Closer Look at the Individual Attacks}

Let us now examine how the individual attacks in our attack-set contribute to the aggregated
robust metrics.
\cref{tab:attacks} shows the results of every attack separately in one scenario
(every scenario is similar in this regard, please refer to the supplementary material in \cref{sec:supp-indattacks}).
We can observe that the aggregated attack is often much stronger than any of the
individual attacks.
Also, different attacks are effective against different kinds of models.
For example, for the normal models PAdam-Cos tends to be the best individual attack,
while for the more robust models the SEA attacks perform better.

\setlength{\intextsep}{3pt plus 1.0pt minus 2.0pt}
\begin{wraptable}[17]{r}{0.6\textwidth}
\caption{Attacks on Cityscapes with DeepLabv3 (NmIoU \%).}
\label{tab:attacks}
\small
\centering
\setlength{\tabcolsep}{5pt}
\resizebox{0.6\textwidth}{!}{%
\begin{tabular}{lrrrrrrr}
	  & \rotatebox[origin=r]{90}{\footnotesize Normal} 
	  & \rotatebox[origin=r]{90}{\footnotesize \ DDC-AT} 
	  & \rotatebox[origin=r]{90}{\footnotesize PGD-AT} 
	  & \rotatebox[origin=r]{90}{\footnotesize \ SegPGD} 
	   \rotatebox[origin=r]{90}{\footnotesize -AT} 
	  & \rotatebox[origin=r]{90}{\footnotesize \ PGD-AT} 
	   \rotatebox[origin=r]{90}{\footnotesize -100} 
	  & \rotatebox[origin=r]{90}{\footnotesize \ SegPGD} 
	   \rotatebox[origin=r]{90}{\footnotesize -AT-100} \\\hline

\rowcolor[gray]{0.85}
clean     & 51.18 & 49.63 & 48.28 & 45.91 & 33.37 & 32.68\\
PAdam-CE & 1.39   & 1.09  & 0.31 & 0.85 & 20.66 & 23.11\\
PAdam-Cos & 0.00  & 0.64  & 0.30 & 0.43 & 22.58 & 25.40\\
\rowcolor[gray]{0.85}
SEA-JSD   & 0.67  & 0.74  & 0.58 & 1.12 & 17.15 & 25.14\\
\rowcolor[gray]{0.85}
SEA-MCE   & 1.51  & 1.03  & 0.67 & 0.79 & 18.13 & 26.23\\
\rowcolor[gray]{0.85}
SEA-MSL   & 1.53  & 0.73  & 0.62 & 0.83 & 20.95 & 28.41\\
\rowcolor[gray]{0.85}
SEA-BCE   & 1.73  & 2.81  & 1.20 & 2.43 & 17.74 & 25.48\\
ALMAProx  & 1.87  & 1.35  & 1.73  & 1.37 & 30.86 & 30.53\\
DAG-0.001 & 5.48  & 45.29 & 35.25 & 27.67 & 33.51 & 32.80\\
DAG-0.003 & 1.04  & 20.80 & 13.24 & 9.31 & 33.51 & 32.80\\
PDPGD     & 1.48  & 11.22 & 9.88 & 1.75 & 32.34 & 31.87\\
\rowcolor[gray]{0.85}
aggregated & 0.00 & 0.02  & 0.01 & 0.00 & 15.86 & 21.82\\
\end{tabular}}
\end{wraptable}

\cref{fig:histograms} shows the distribution of the most successful attack over the
validation set in a number of scenarios.
In other words, for each input example, we determine which attack was the most successful and
show the resulting distribution.
The supplementary material covers all the scenarios in \cref{sec:supp-indattacks}.

It is striking how different these distributions are in the various scenarios.
The two attacks proposed in this work, PAdam-CE and PAdam-Cos, dominate
many scenarios including normal and some robust models as well.
Interestingly, the SEA-AT family of models is most sensitive to the SEA attack set.
Nevertheless, it is clear that every attack has its contribution, and different models
require different attacks.

\cref{fig:examples-attacks,fig:examples-models} illustrate the diversity of the attacks
and the models through an example image.
The perturbations remain invisible in all the cases.
For this particular image, PAdam-Cos can completely alter the output of all the models.
Further examples are shown in the supplementary material in \cref{sec:supp-examples}.

\section{Conclusions and Limitations}

Our contribution was of a methodological nature.
We empirically proved that using a strong set of attacks can dramatically reduce
the known upper bounds of the robustness metrics of the state-of-the-art models,
in many cases completely diminishing them.
We also pointed out that the choice of mIoU aggregation method matters, because
robust models tend to have a strong size-bias that is not revealed by class-wise
aggregation, only by image-wise aggregation (NmIoU).

We demonstrated our methodology mostly
on public model checkpoints from related work and we did only a limited exploration of
variants.
It would be very informative to also study, for example,
the effect of the proportion of adversarial samples in the batches,
or combinations of backbone models, architectures and training hyperparameters.
These limitations are mostly due to the prohibitive cost of such a study.

We do believe that our thorough analysis of the highly cited models we selected
still provides a useful contribution to the community in terms of how to move forward with
the analysis of robust semantic segmentation models.

\section*{Acknowledgements}

This research was supported by the AI Competence Centre of the Cluster of Science and Mathematics of the 
Centre of Excellence for Interdisciplinary Research, Development and Innovation of the University of Szeged. 
We also thank the support by the the European Union project RRF-2.3.1-21-2022-00004
within the framework of the Artificial Intelligence National Laboratory, and by
project TKP2021-NVA-09 that has been implemented with the support provided  by the Ministry of Culture
and Innovation of Hungary from the National Research, Development and Innovation Fund, financed under
the TKP2021-NVA funding scheme.
We thank András Balogh for his comments on earlier versions that helped us improve the presentation.

\bibliographystyle{splncs04}
\bibliography{references}

\clearpage
\begin{center}
\large{\bf Evaluating the Adversarial Robustness of Semantic Segmentation: Trying Harder Pays Off\\[.5em] Supplementary Material}
\ 
\end{center}

\renewcommand{\thesection}{S.\arabic{section}}

\section{Details on Training Methodology}
\label{sec:supp-train}

As mentioned in \cref{sec:models}, apart from using checkpoints from related work, we also
trained a number of models, in particular, we trained PGD-AT-100 and SegPGD-AT-100 in all the
scenarios, as well as the normally trained models for the SEA-AT architectures.

The internal attacks applied in adversarial training used the $\ell_\infty$-norm neighborhood
$\Delta=\{\delta: \|\delta\|_\infty\leq\epsilon\}$ with $\epsilon=0.03\approx 8/255$,
in line with the general practice.
Note that this assumes that the input channels are scaled to the range $[0,1]$.

In each case, over the PASCAL VOC dataset the background class was used as a regular class, and it was included
in the cross-entropy training loss.
Our main references also follow this practice~\cite{Xu2021b,Gu2022a,Croce2023a}.
Although we evaluated all the models with and without taking the background class into account,
no models were trained without the background.
Over Cityscapes, the void class was not used during training, like in~\cite{Xu2021b,Gu2022a}.

\subsection{PGD-AT-100 and SegPGD-AT-100}

In our training setup, we followed Xu et al.~\cite{Xu2021b}, except for the fact that
we used 100\% adversarial batches for training, as opposed to 50\%.
The training of every model in our investigation shared the
hyperparameters that are recommended in the semseg package~\cite{semseg2019}.
These settings were also used by Xu et al.~\cite{Xu2021b}.
In more detail, the batch size was 16, we used SGD with momentum as the learning algorithm
with a dynamic learning rate (starting at $0.01$),
a weight decay coefficient of $10^{-4}$.

We also used a number of augmentations including rotation, scaling and cropping.
For the PASCAL VOC models we used random crops of size 473x473.
The training lasted for 50 epochs, without early stopping.
For Cityscapes, the models were trained for 400 epochs without early stopping, and random crops of 449x449 were
used after scaling down the images to half-size.

\subsection{Normal Models}

We trained normal (non-robust) models for the two SEA-AT models, as we could not find checkpoints
for these baselines.

Our training setup was identical to that of Croce et al.~\cite{Croce2023a}, and
we used their implementation without any changes.
The architecture was identical to the corresponding SEA-AT models, and we used a non-robust
pre-trained backbone for initialization.

The optimization algorithm used for training was AdamW~\cite{Loshchilov2019a}.
A linear learning rate decay was applied with a warmup of 10 epochs,
with a base learning rate of 0.001 and a weight decay of 0.01.
The training was run for 50 epochs without early stopping.
The batch size was 32.

As for augmentations, horizontal flip and random scaling was used, and then a random 473x473 crop (with zero padding) was
selected, and finally Gaussian blur was used.
For all the details, please consult Croce et al.~\cite{Croce2023a}.

\section{Details on Evaluation Methodology}
\label{sec:supp-method}

\subsection{Running the Attacks}

The attacks were run following the methodology of 
Rony et al.~\cite{Rony2022a}.
There, the PASCAL VOC images were scaled to have 512 as their smaller dimension,
while keeping the aspect ratio, and the full image was attacked.
The Cityscapes images were scaled down to half-size (1024x512), and the full image was attacked.

In the case of the minimum perturbation attacks the background class was masked out
in every dataset (like in~\cite{Rony2022a}), while in the case of the maximum damage attacks, the background (void) class
was masked out only for Cityscapes images (like in~\cite{Xu2021b}).

Like in training, the maximum damage attacks were run assuming an $\ell_\infty$-norm neighborhood
$\Delta=\{\delta: \|\delta\|_\infty\leq\epsilon\}$ with $\epsilon=0.03\approx 8/255$.

\subsection{Computing the Metrics}

The image-wise performance metrics---pixel accuracy and mIoU---were computed on the
mask that was predicted on the attacked image.
No post-processing of the mask was performed.

The aggregated attack based on our attack set was computed by selecting the most
successful attack for each image separately, based on the metric under investigation.
That is, when the target metric was accuracy, then the lowest robust accuracy wins
over a given picture, and when the target metric is CmIoU or NmIoU, then the
lowest robust mIoU wins.

\subsection{Differences from Related Work}

The evaluation methodologies of Xu et al.~\cite{Xu2021b} and Gu et al.~\cite{Gu2022a} are similar to each other.
They both report only CmIoU. 
Most importantly, they perform the attacks on overlapping crops of the input organized in
overlapping tiles.
The masks predicted over these overlapping tiles are then averaged in the overlapping
regions and this aggregated mask is used to predict the classes.
This method results in an increased quality because of the ensemble effect in the overlapping
regions.

Interestingly, the authors use an extra post-processing step when evaluating clean inputs.
In this step, the input is mirrored, and the mask of the mirror image is then mirrored again and
averaged with the original mask. 
This further improves performance metrics over clean inputs rather significantly.

Croce et al.~\cite{Croce2023a} follow yet another methodology for evaluation.
First, they scale the PASCAL VOC images to 512x512, then they take a central crop of size 473x473, and then
they attack this cropped image.
The metrics are computed over the result of this attack.

\section{Distribution of the Image-wise mIoU}
\label{sec:supp-miouhistograms}

\begin{figure*}
\setlength{\tabcolsep}{1pt}
\centering
\tiny
\begin{tabular}{ccccc}
& clean, DeepLabv3 & robust, DeepLabv3 & clean, PSPNet & robust, PSPNet\\
\rotatebox[origin=l]{90}{\hspace{8mm}Normal} & 
\includegraphics[width=0.24\textwidth]{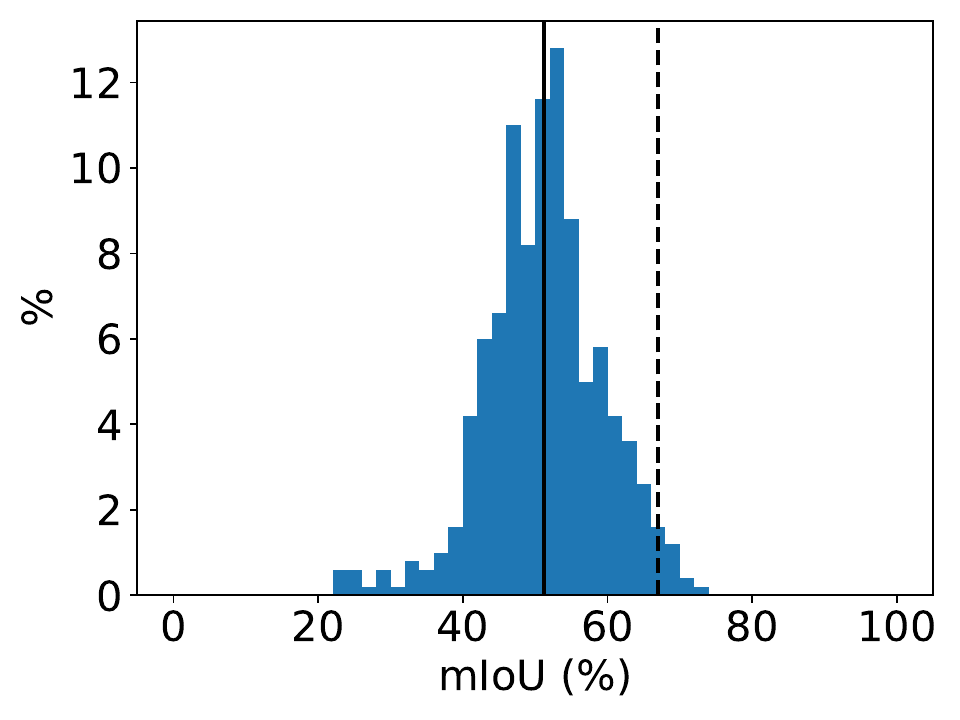} &
\includegraphics[width=0.24\textwidth]{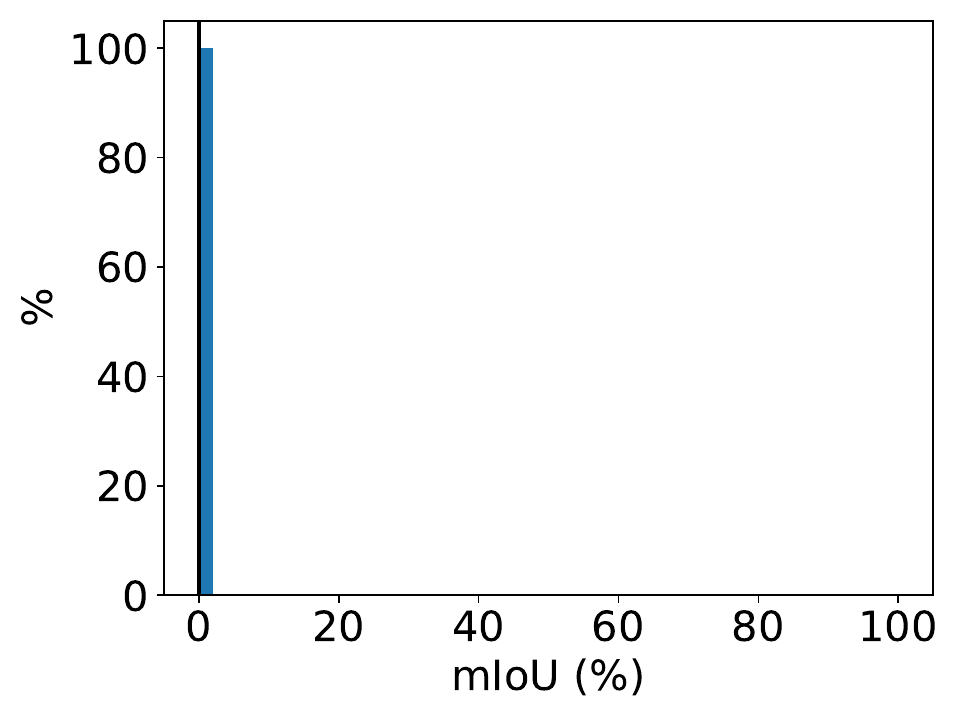} &
\includegraphics[width=0.24\textwidth]{gdrive/data_final/cityscapes/normal_pspnet/miou_normal_dist_with_bg} &
\includegraphics[width=0.24\textwidth]{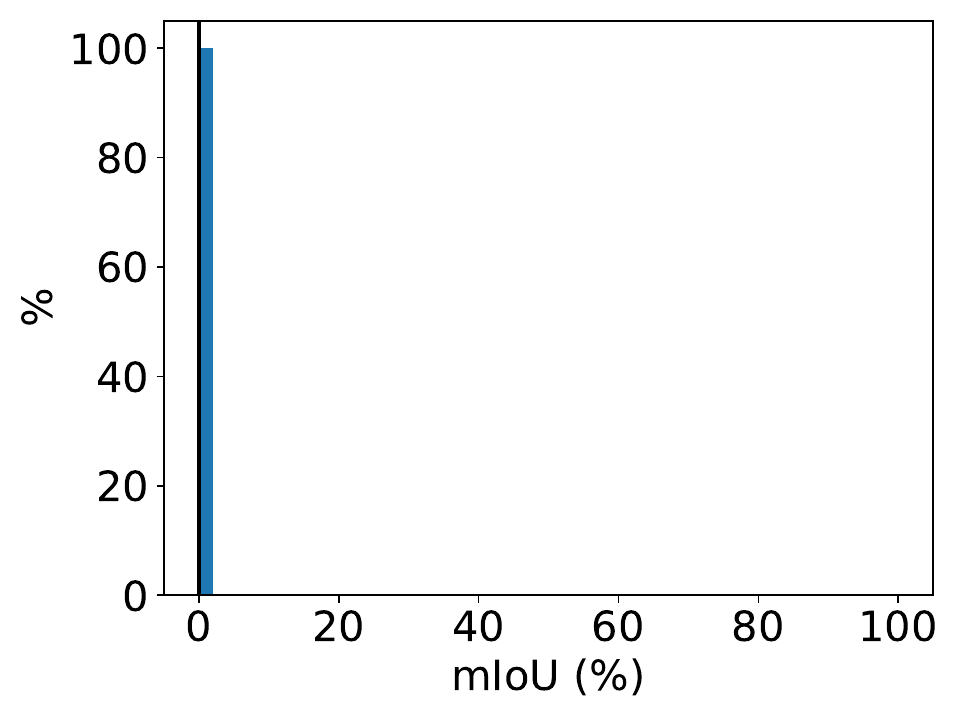} \\
\rotatebox[origin=l]{90}{\hspace{8mm}DDC-AT} & 
\includegraphics[width=0.24\textwidth]{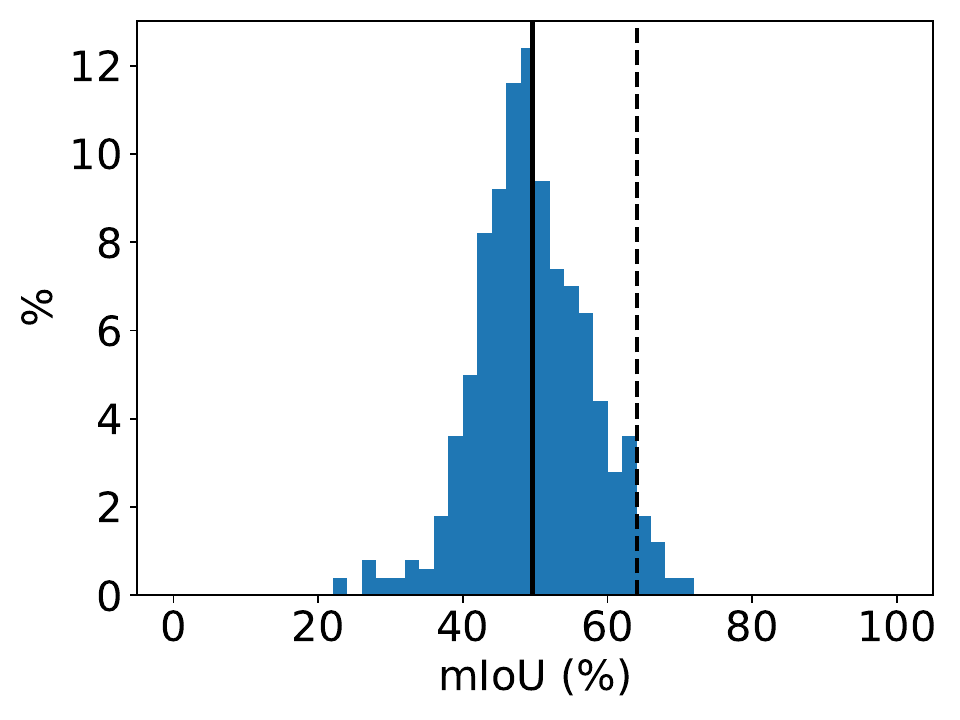} &
\includegraphics[width=0.24\textwidth]{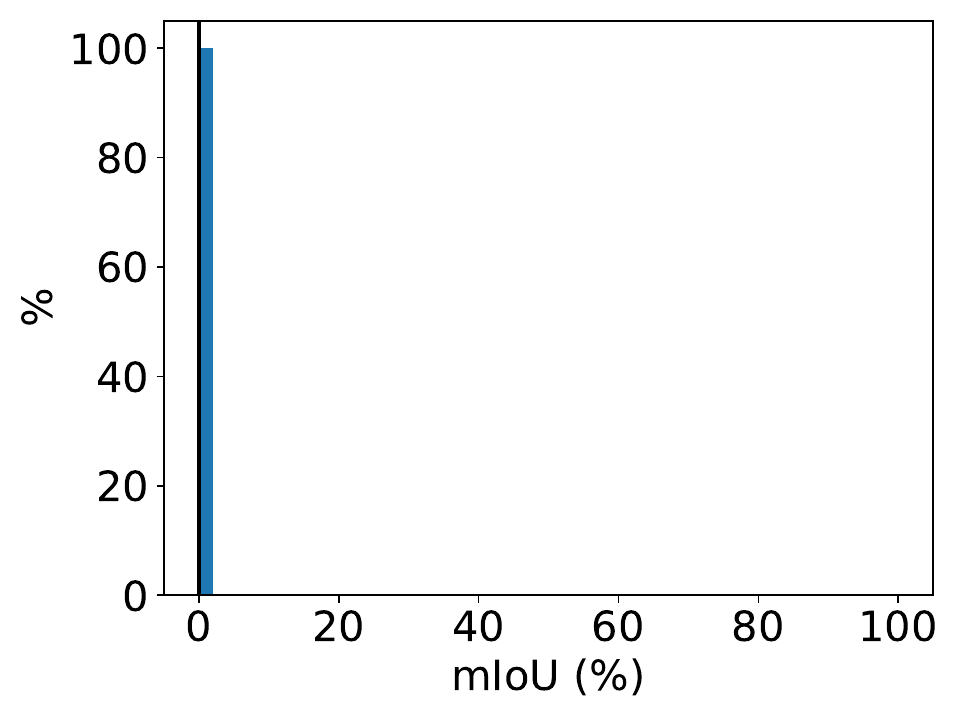} &
\includegraphics[width=0.24\textwidth]{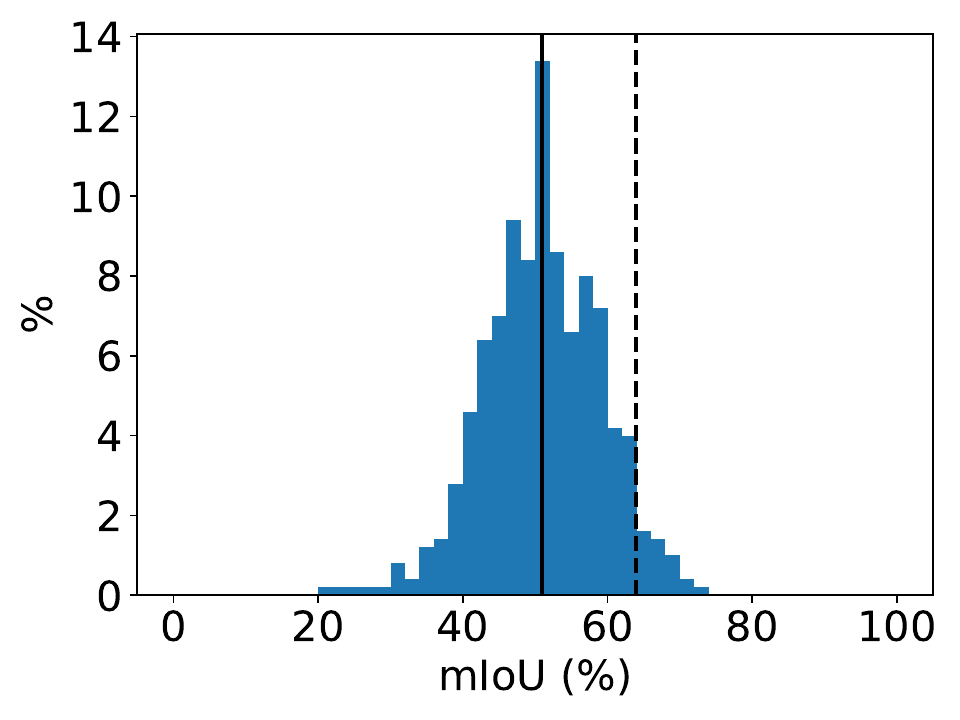} &
\includegraphics[width=0.24\textwidth]{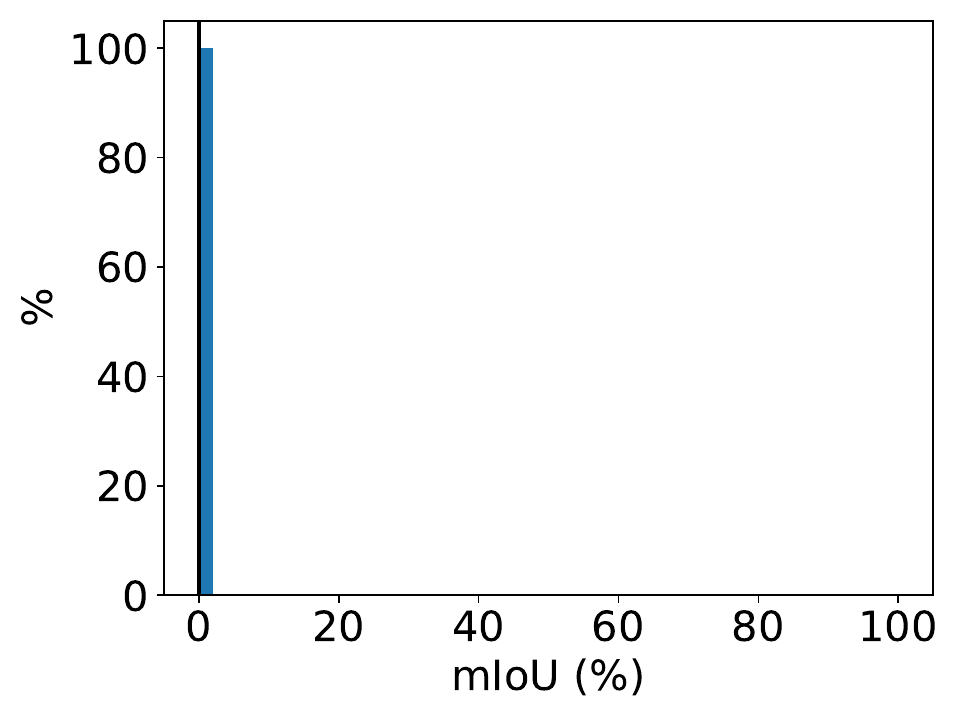} \\
\rotatebox[origin=l]{90}{\hspace{8mm}PGD-AT} & 
\includegraphics[width=0.24\textwidth]{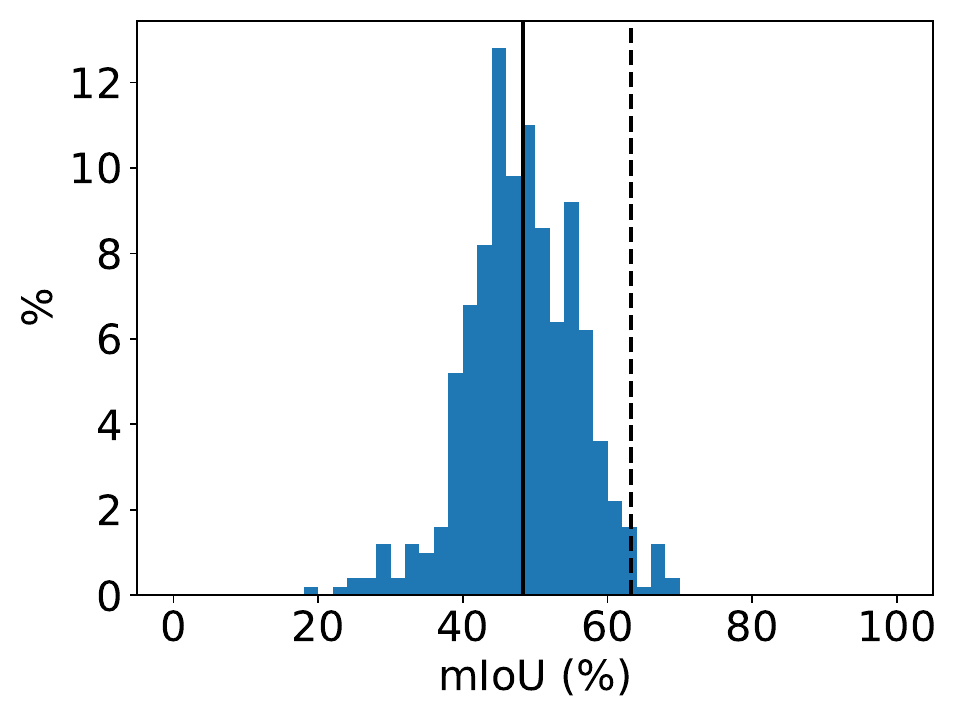} &
\includegraphics[width=0.24\textwidth]{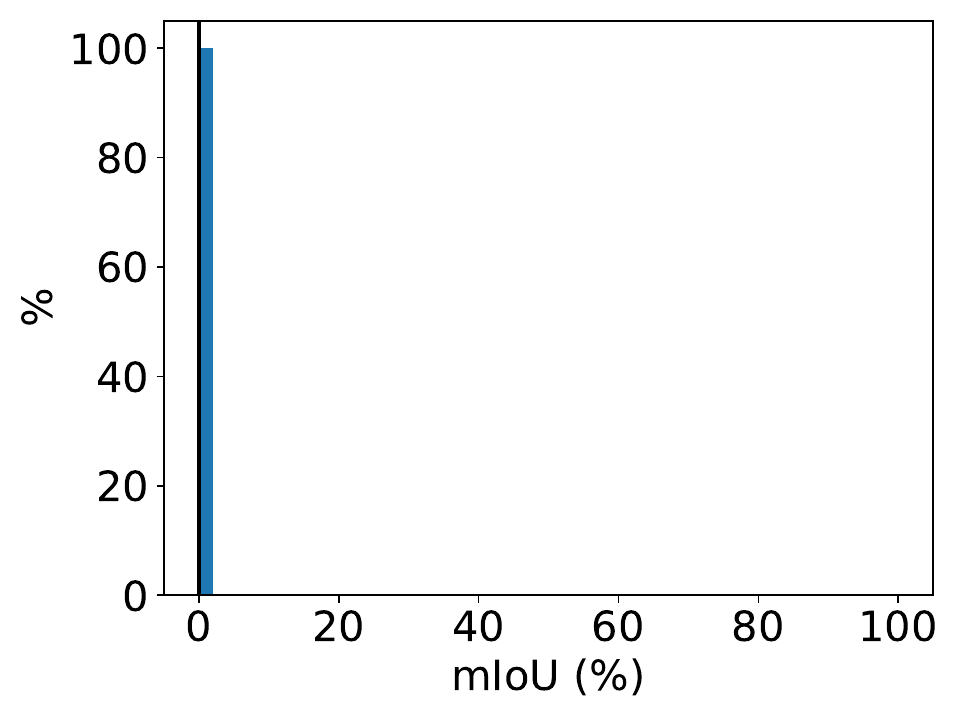} &
\includegraphics[width=0.24\textwidth]{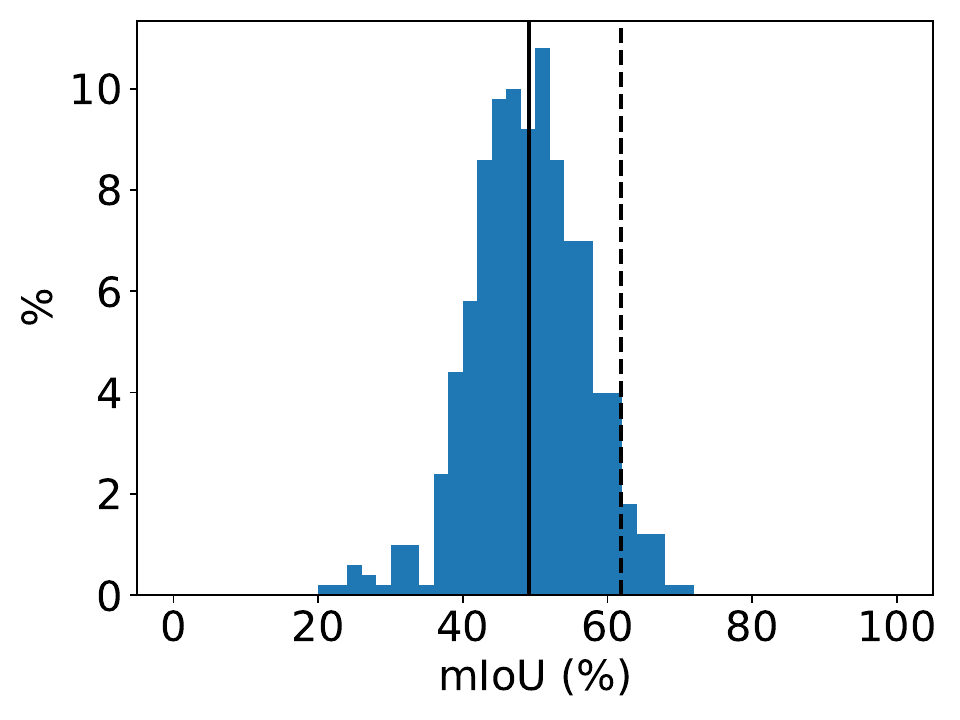} &
\includegraphics[width=0.24\textwidth]{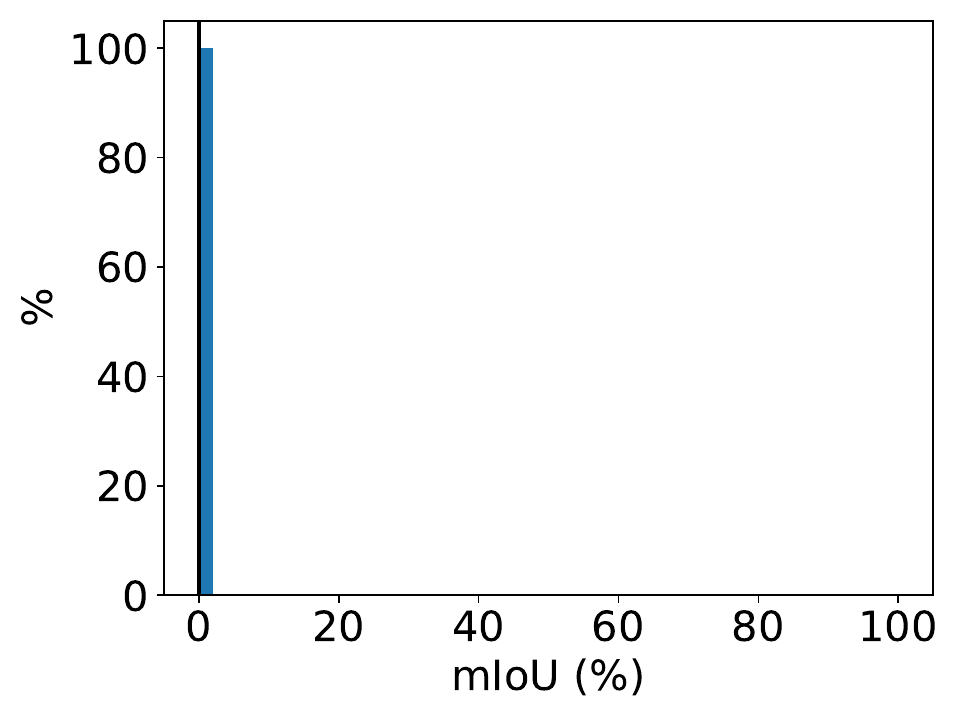} \\
\rotatebox[origin=l]{90}{\hspace{4mm}SegPGD-AT} & 
\includegraphics[width=0.24\textwidth]{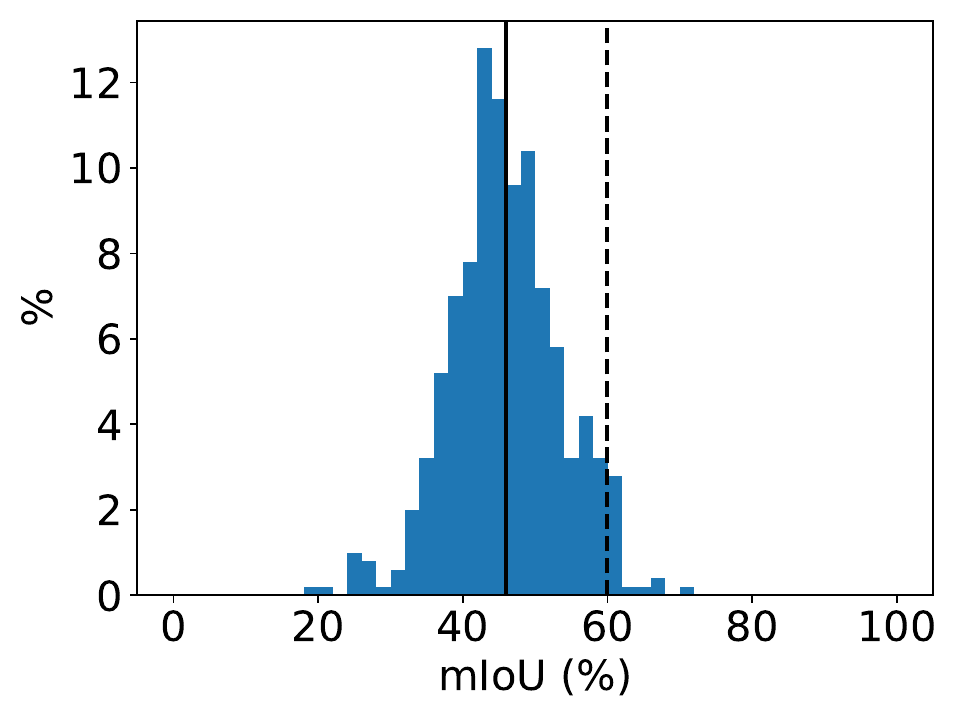} &
\includegraphics[width=0.24\textwidth]{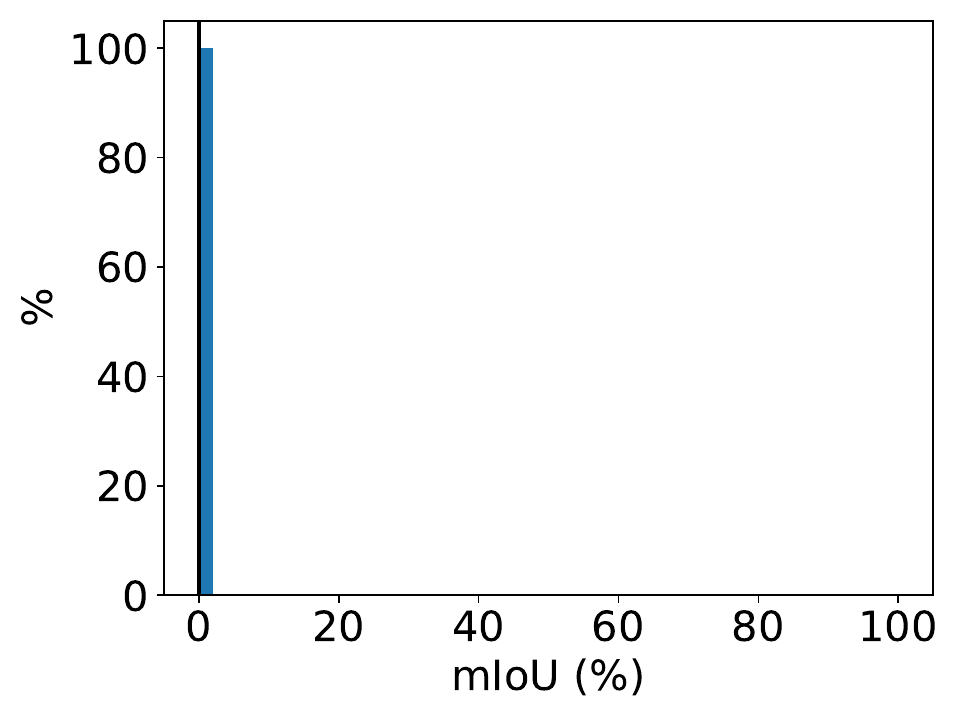} &
\includegraphics[width=0.24\textwidth]{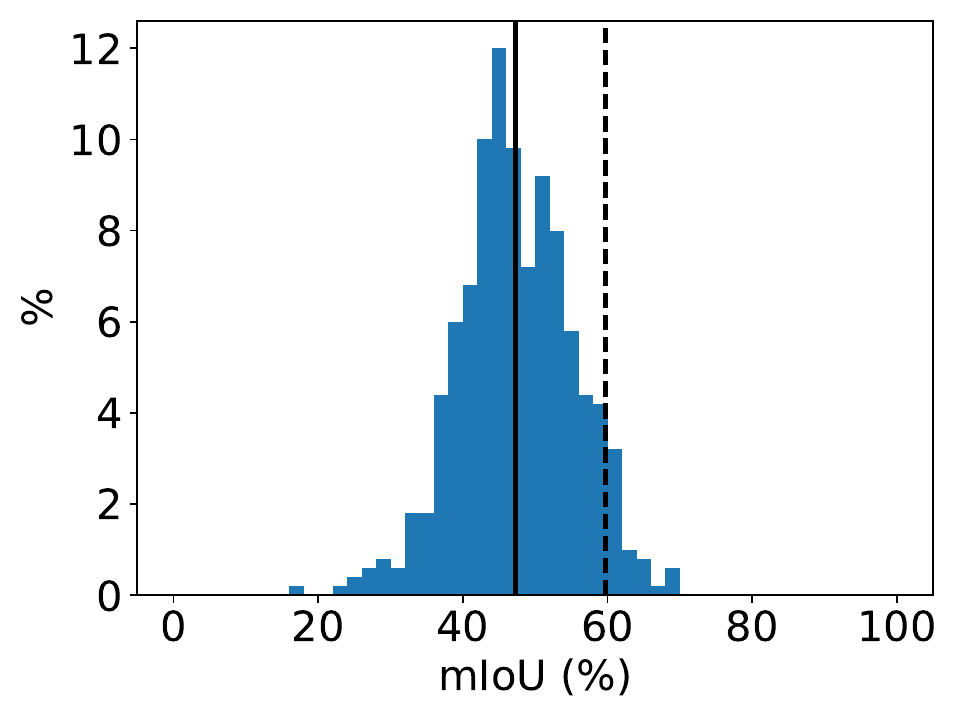} &
\includegraphics[width=0.24\textwidth]{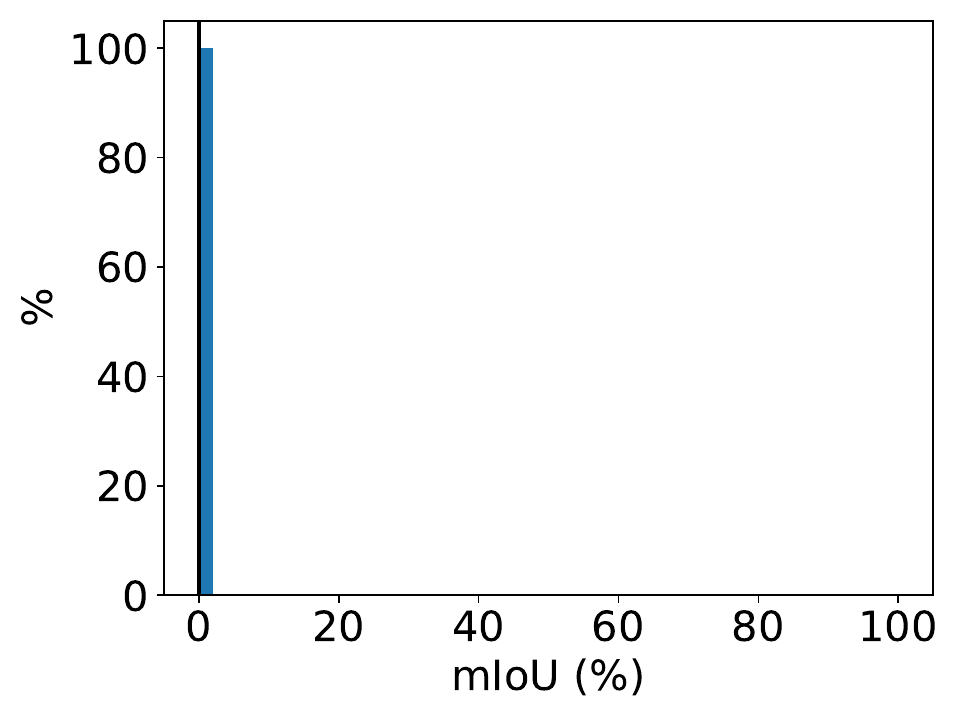} \\
\rotatebox[origin=l]{90}{\hspace{4mm}PGD-AT-100} & 
\includegraphics[width=0.24\textwidth]{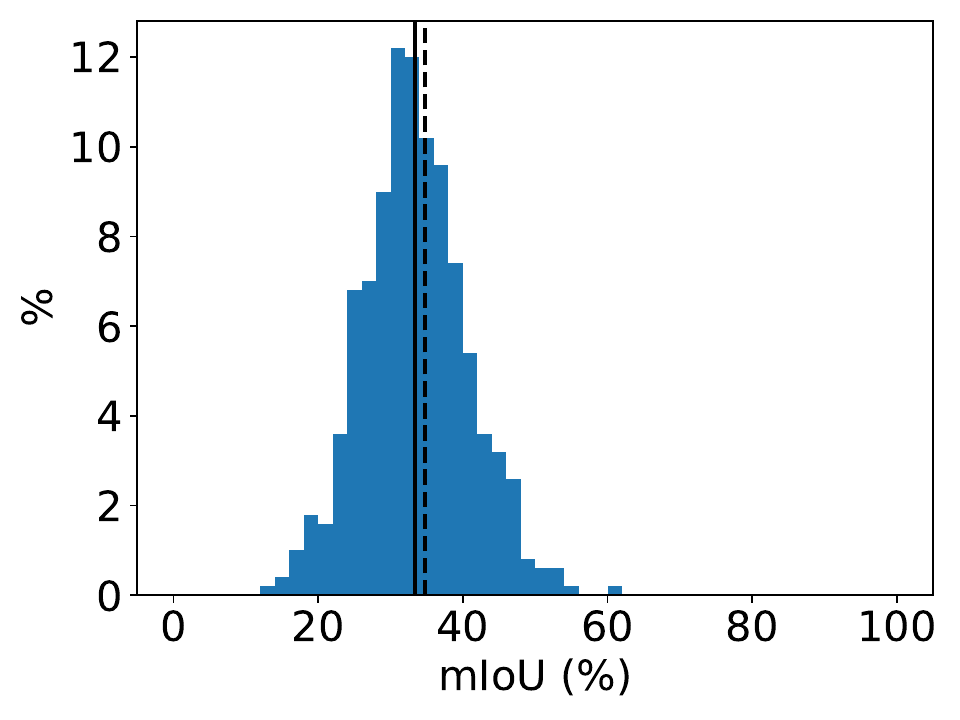} &
\includegraphics[width=0.24\textwidth]{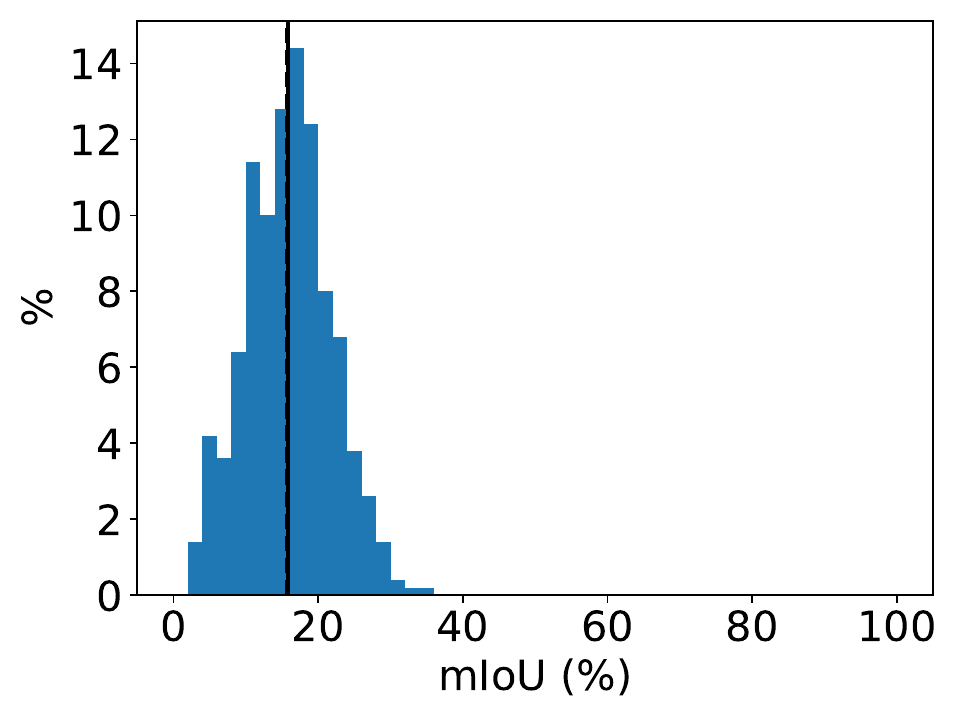} &
\includegraphics[width=0.24\textwidth]{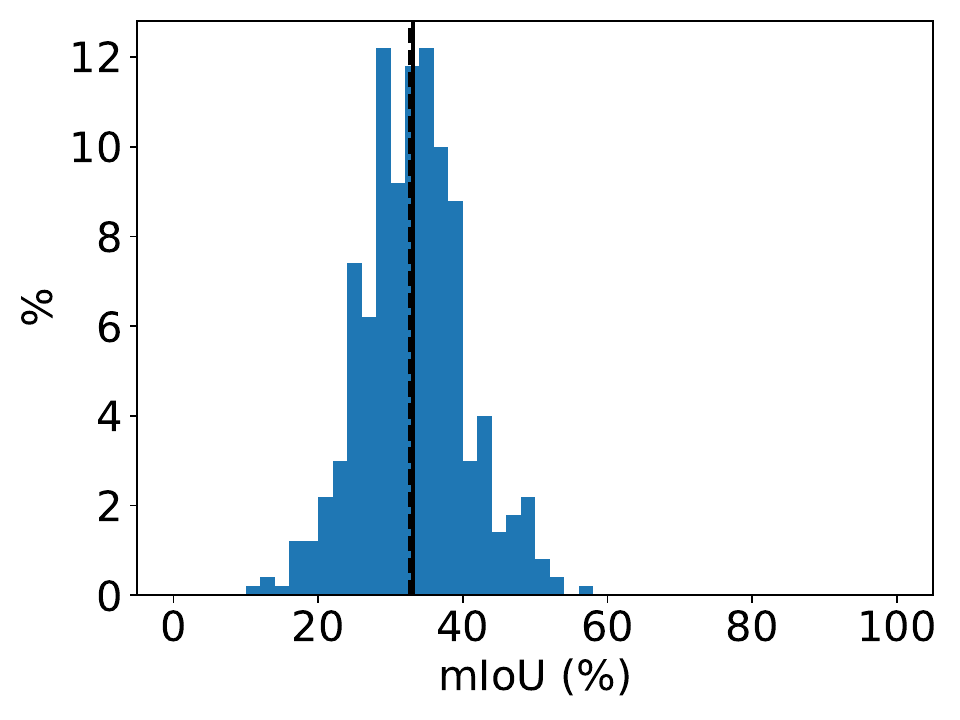} &
\includegraphics[width=0.24\textwidth]{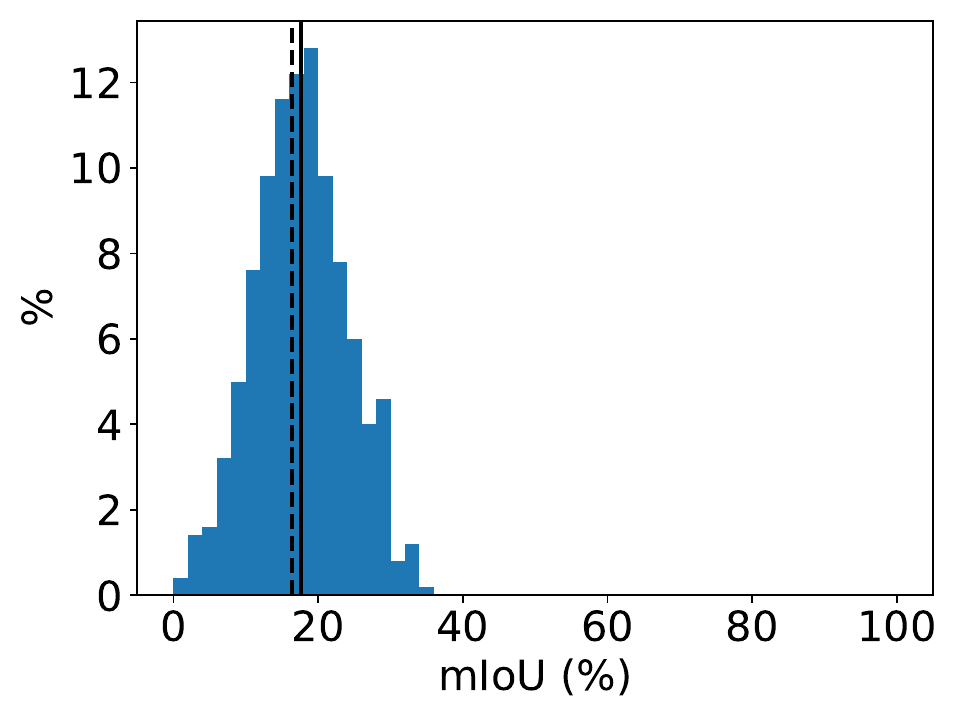} \\
\rotatebox[origin=l]{90}{\hspace{2mm}SegPGD-AT-100} & 
\includegraphics[width=0.24\textwidth]{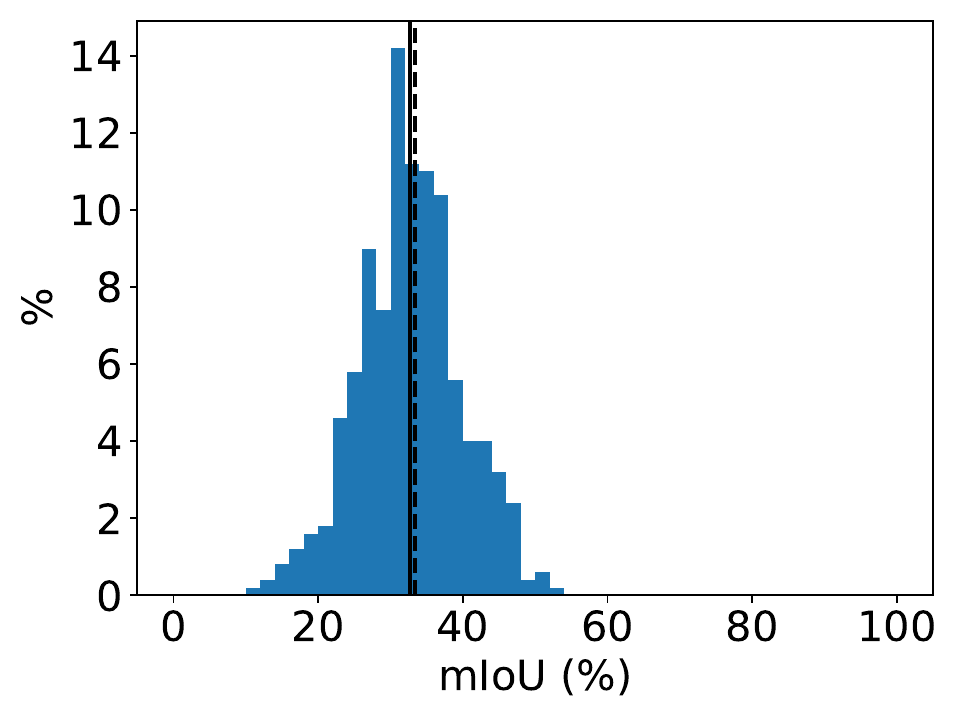} &
\includegraphics[width=0.24\textwidth]{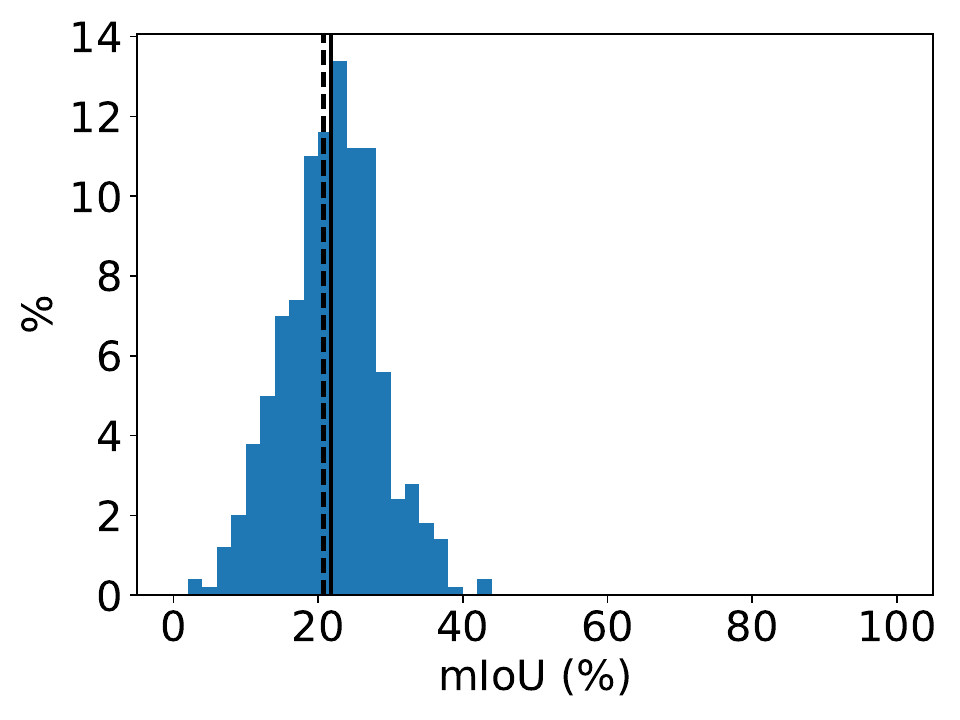} &
\includegraphics[width=0.24\textwidth]{gdrive/data_final/cityscapes/seg_pgd-100_pspnet/miou_normal_dist_with_bg} &
\includegraphics[width=0.24\textwidth]{gdrive/data_final/cityscapes/seg_pgd-100_pspnet/miou_min_dist_with_bg} \\
\end{tabular}
\caption{Distributions of single image mIoU on Cityscapes.
CmIoU and NmIoU are shown using vertical lines.}
\label{fig:supp-cs-miouhistograms}
\end{figure*}

Here, we include the histograms of the distributions of image-wise mIoU for all
the scenarios that we studied in
\cref{fig:supp-cs-miouhistograms,fig:supp-p-miouhistograms,fig:supp-pnb-miouhistograms,fig:supp-p-sea-miouhistograms,fig:supp-pnb-sea-miouhistograms}.
These results further confirm the conclusions in \cref{sec:evaluation} and allow us to
compare and understand the different methods more thoroughly.

\subsection{DDC-AT and SegPGD-AT}

The DDC-AT, PGD-AT and SegPGD-AT models show now robustness, and even their clean performance
looks identical to that of the normal model in all the scenarios, except for the SegPGD-AT model
over the PASCAL VOC dataset, where its clean performance is slightly worse than that of the normal model.
This is especially clear when looking at \cref{fig:supp-pnb-miouhistograms}, where SegPGD-AT
has a clearly worse clean performance over non-background pixels, which is better reflected by the NmIoU score.

The 100\% adversarially trained variants are also similar to each other, which further confirms that
the most decisive factor is the percentage of adversarial samples during training.

We can also conclude that the choice between PSPNet and DeepLabv3 make no significant difference.

\subsection{SEA-AT}

\begin{figure*}
\setlength{\tabcolsep}{1pt}
\centering
\tiny
\begin{tabular}{ccccc}
& clean, DeepLabv3 & robust, DeepLabv3 & clean, PSPNet & robust, PSPNet\\
\rotatebox[origin=l]{90}{\hspace{8mm}Normal} & 
\includegraphics[width=0.24\textwidth]{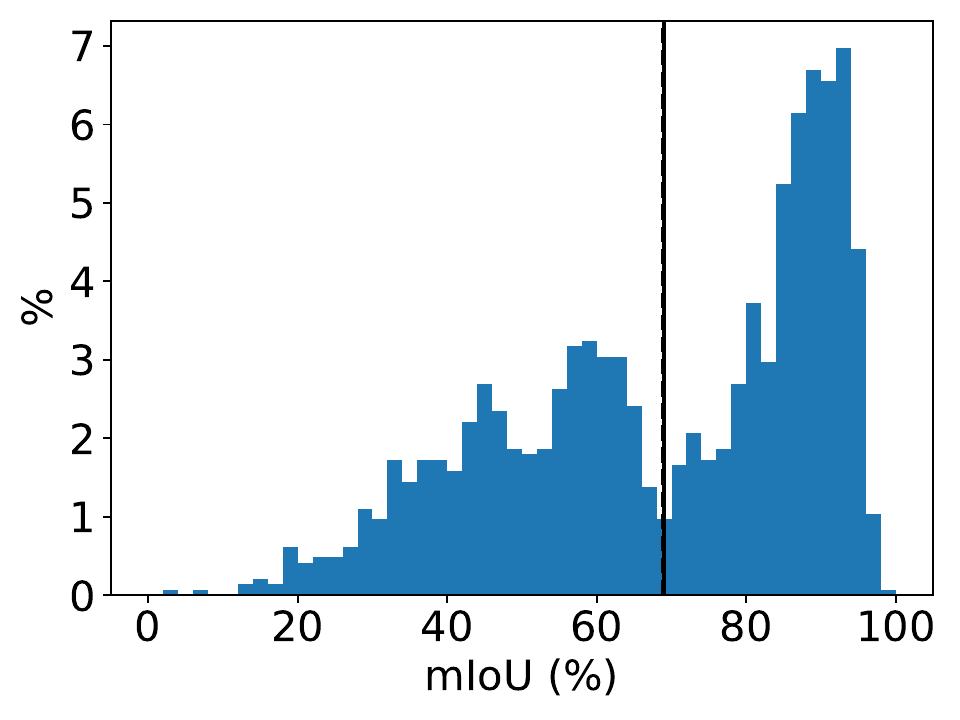} &
\includegraphics[width=0.24\textwidth]{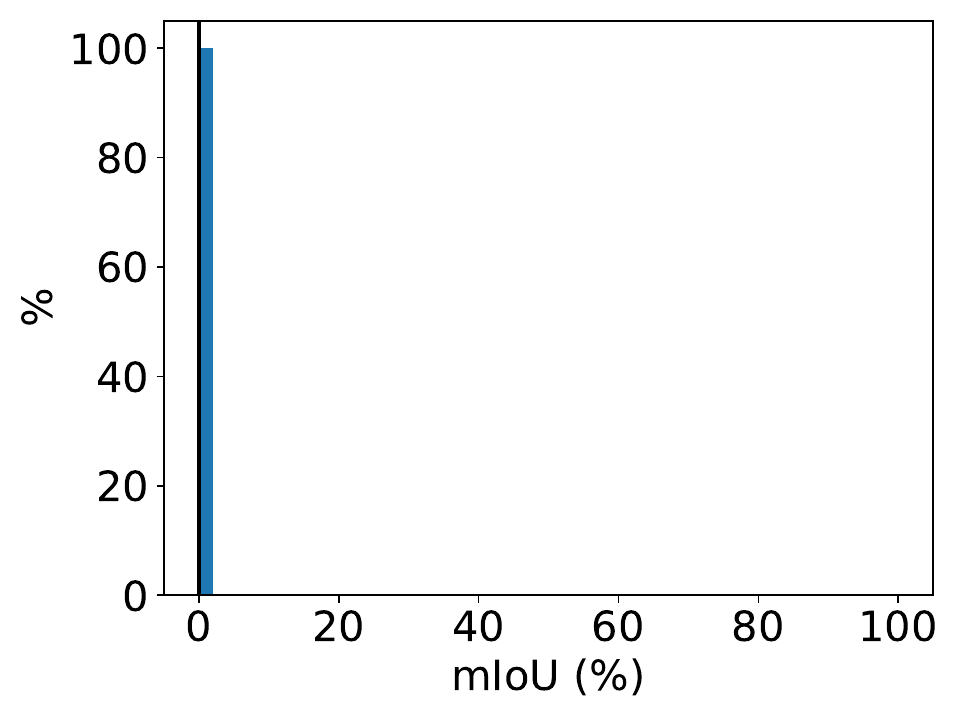} &
\includegraphics[width=0.24\textwidth]{gdrive/data_final/pascal/normal_pspnet/miou_normal_dist_with_bg} &
\includegraphics[width=0.24\textwidth]{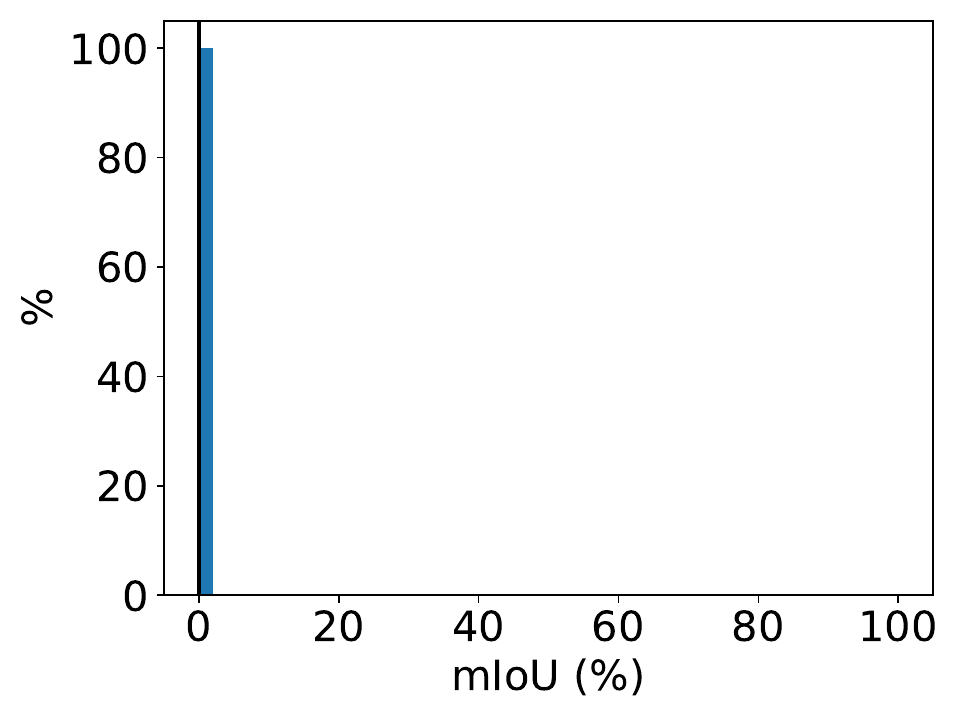} \\
\rotatebox[origin=l]{90}{\hspace{8mm}DDC-AT} & 
\includegraphics[width=0.24\textwidth]{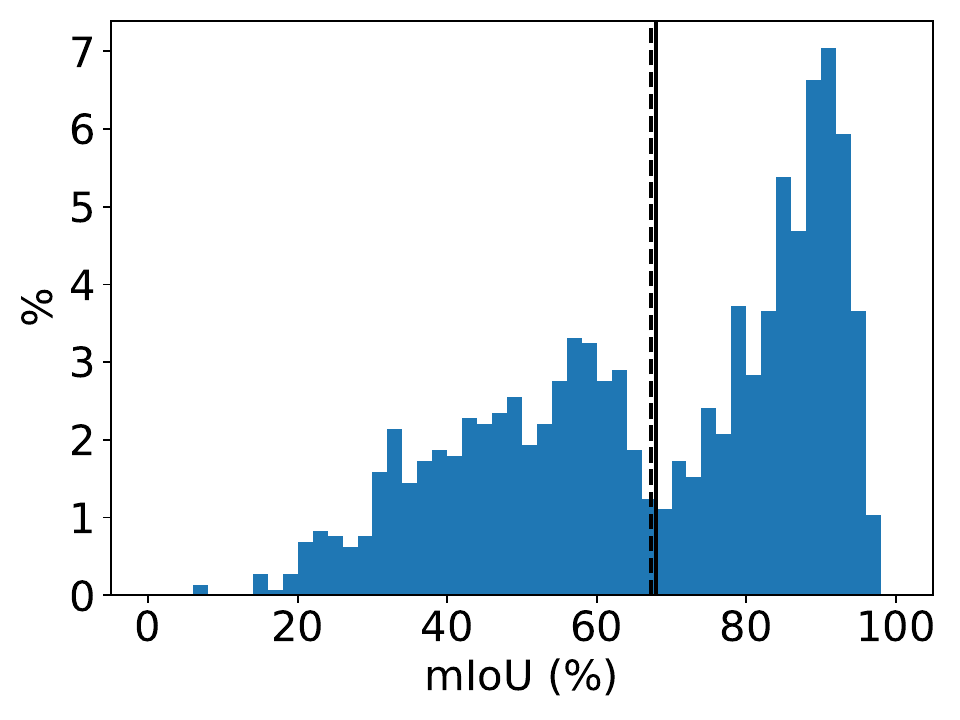} &
\includegraphics[width=0.24\textwidth]{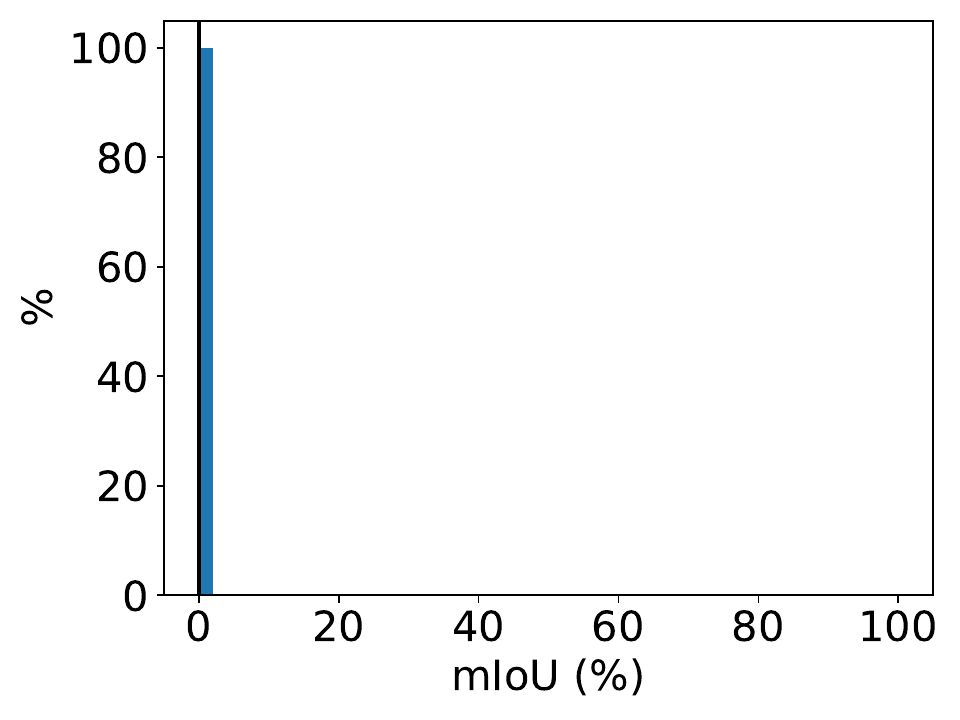} &
\includegraphics[width=0.24\textwidth]{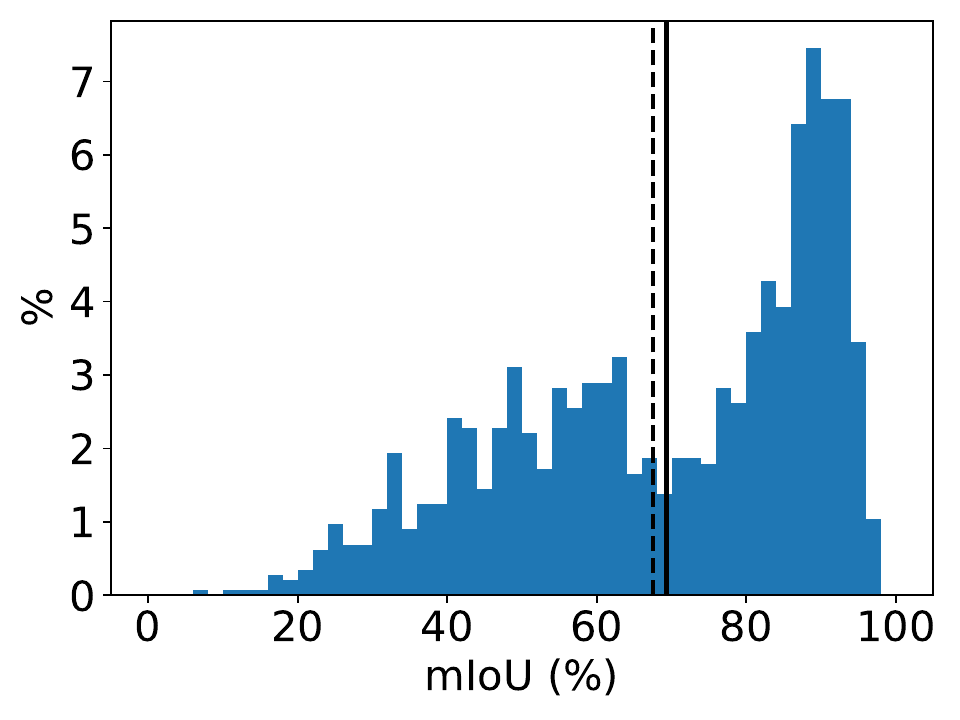} &
\includegraphics[width=0.24\textwidth]{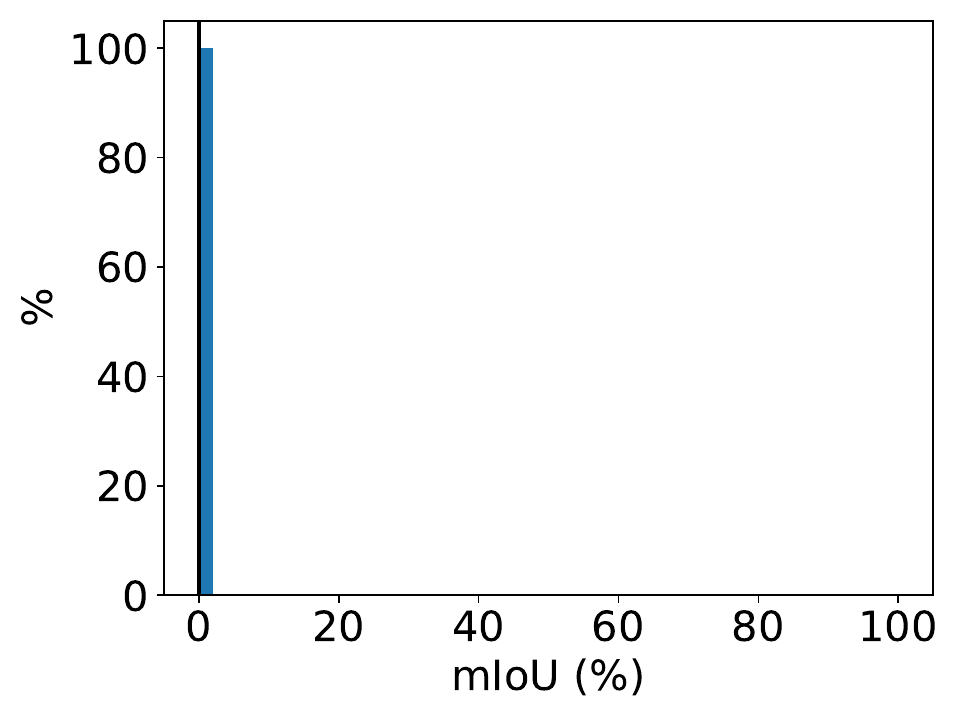} \\
\rotatebox[origin=l]{90}{\hspace{8mm}PGD-AT} & 
\includegraphics[width=0.24\textwidth]{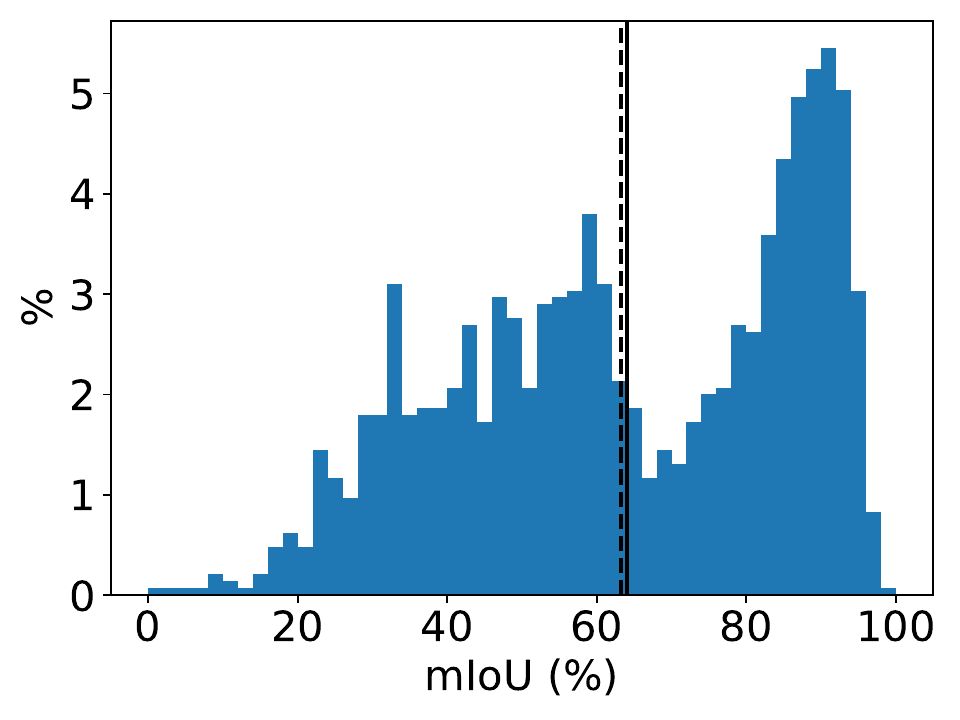} &
\includegraphics[width=0.24\textwidth]{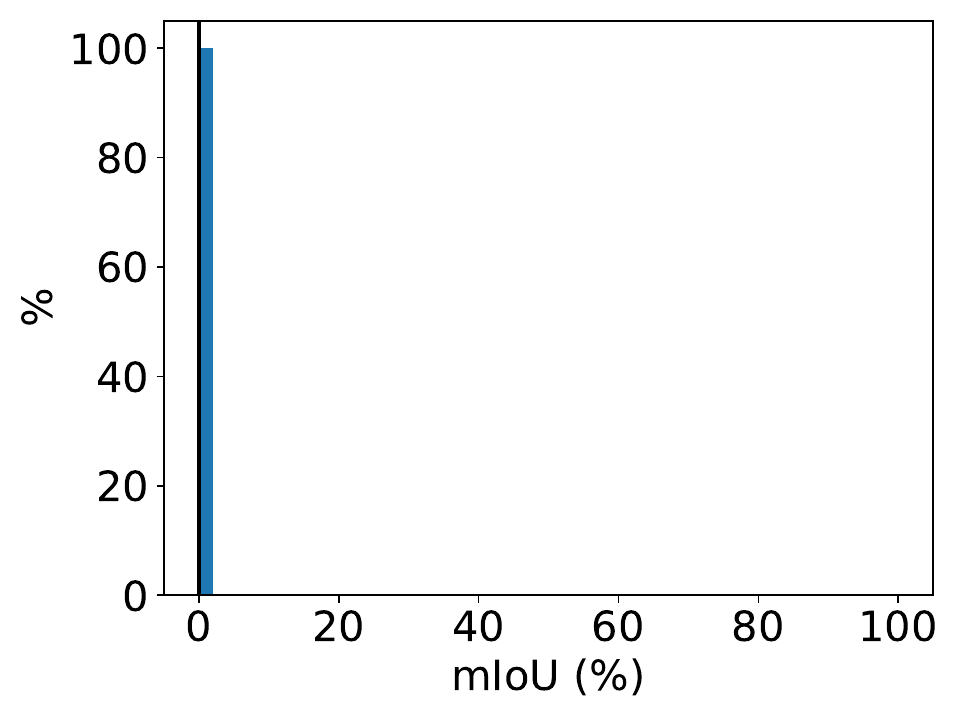} &
\includegraphics[width=0.24\textwidth]{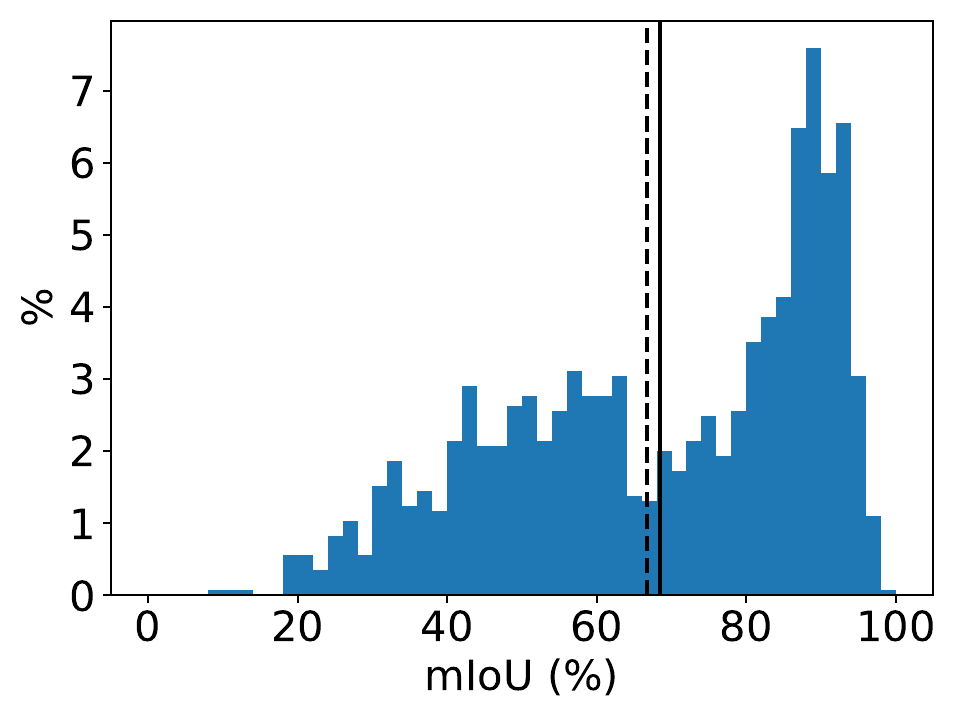} &
\includegraphics[width=0.24\textwidth]{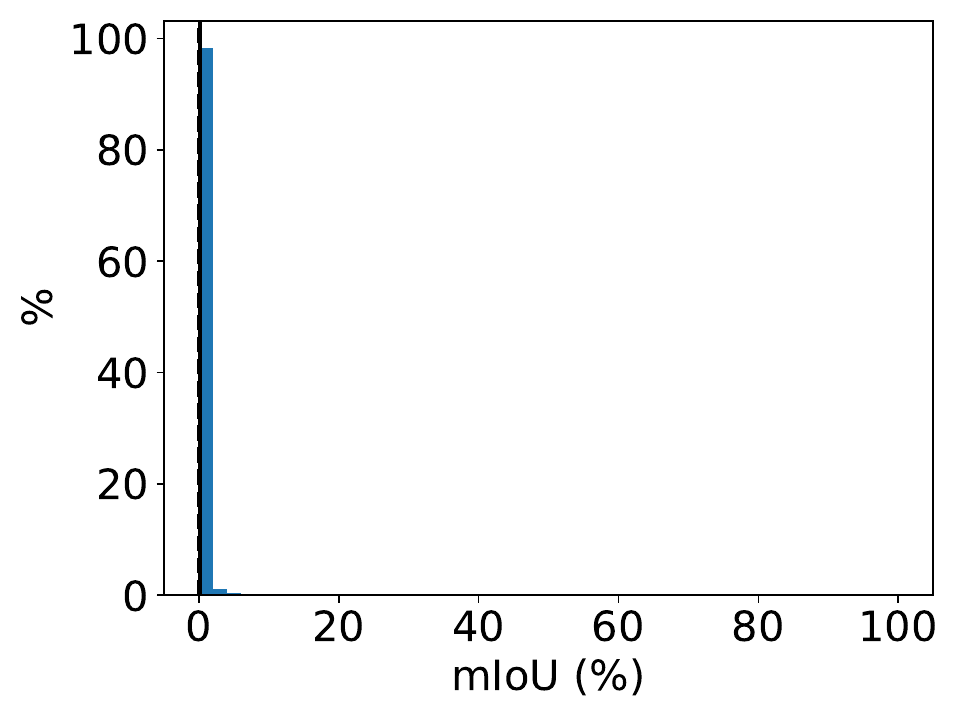} \\
\rotatebox[origin=l]{90}{\hspace{4mm}SegPGD-AT} & 
\includegraphics[width=0.24\textwidth]{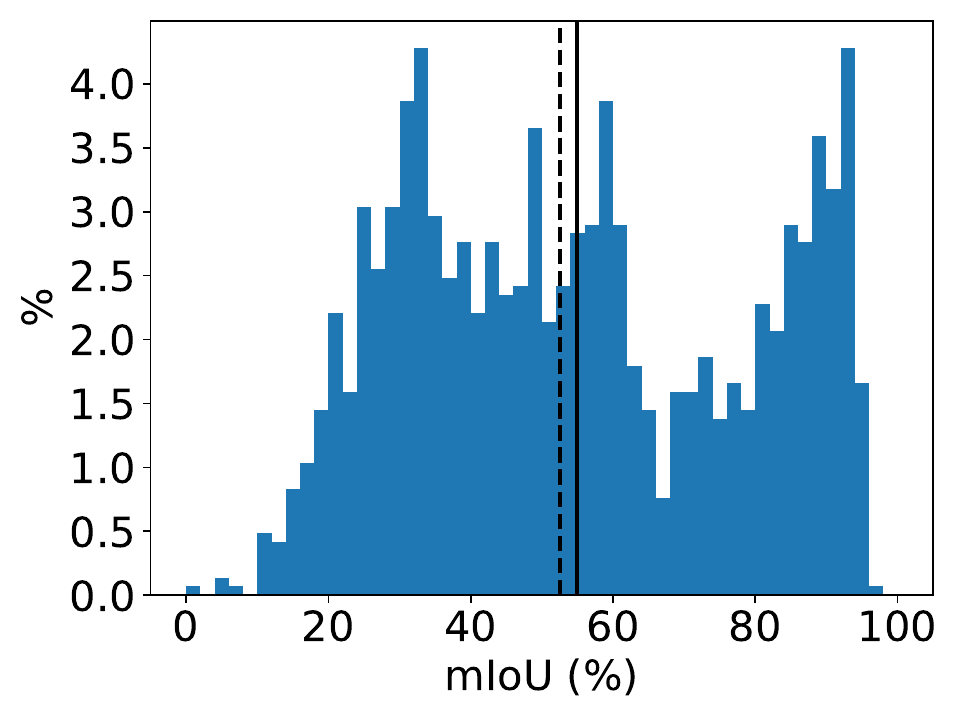} &
\includegraphics[width=0.24\textwidth]{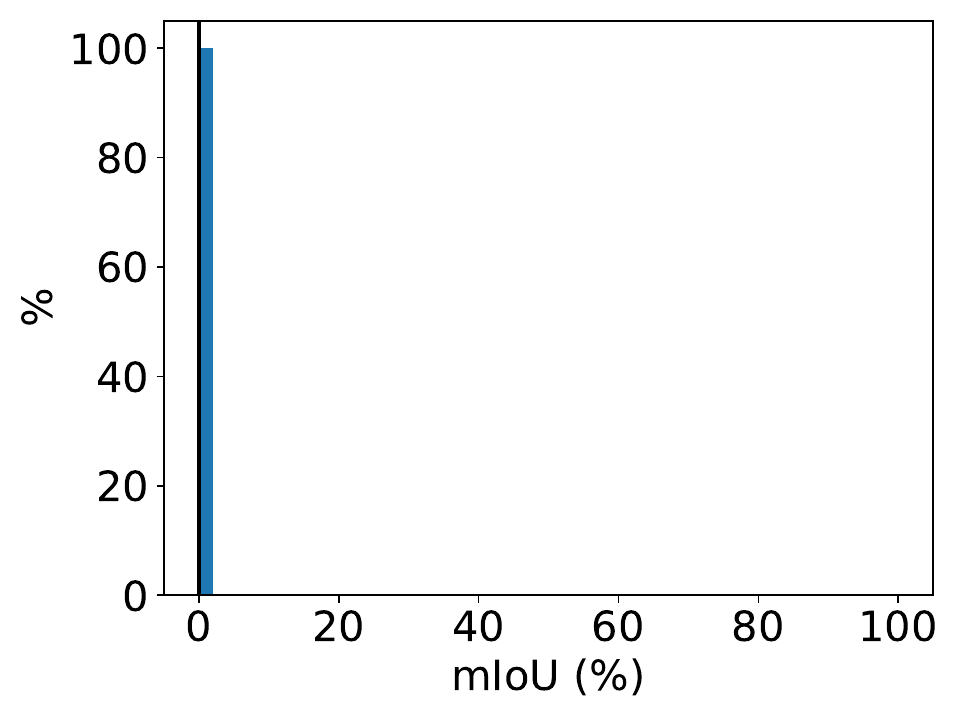} &
\includegraphics[width=0.24\textwidth]{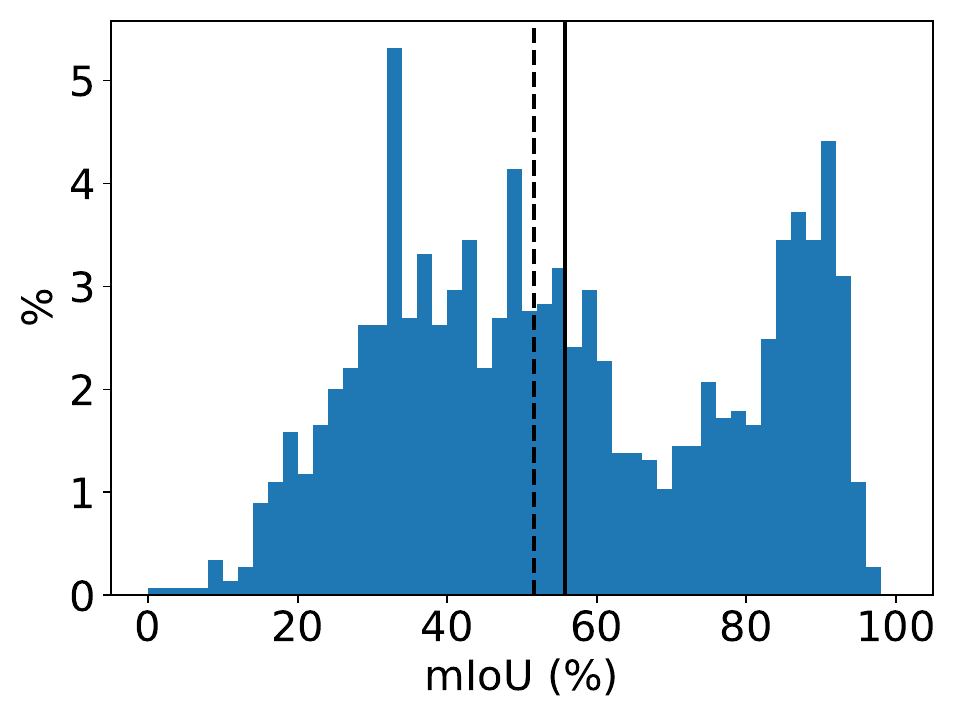} &
\includegraphics[width=0.24\textwidth]{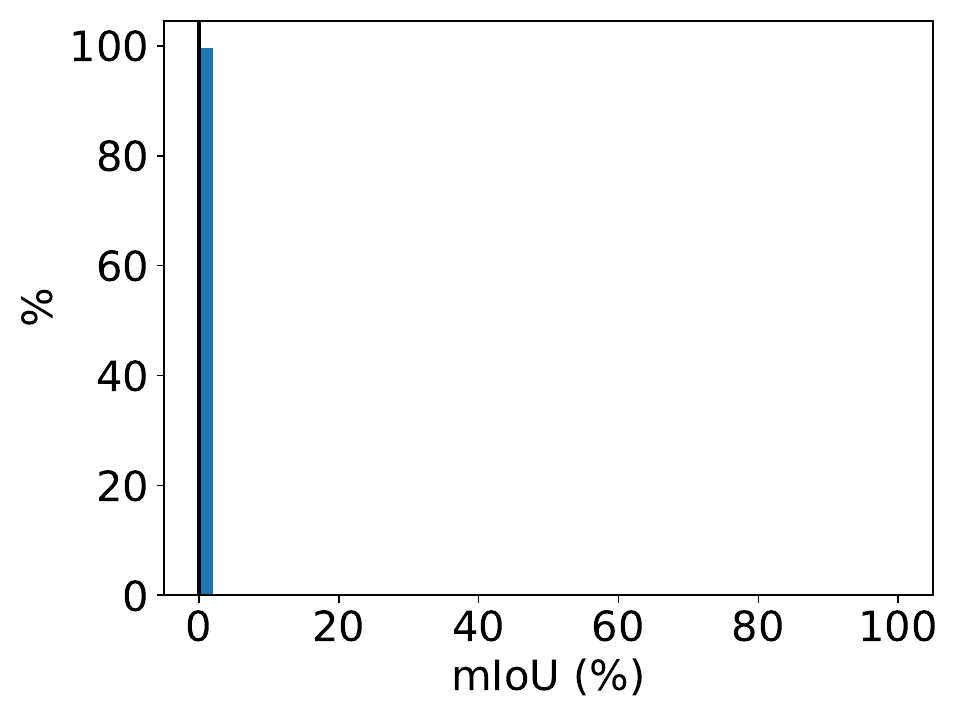} \\
\rotatebox[origin=l]{90}{\hspace{4mm}PGD-AT-100} & 
\includegraphics[width=0.24\textwidth]{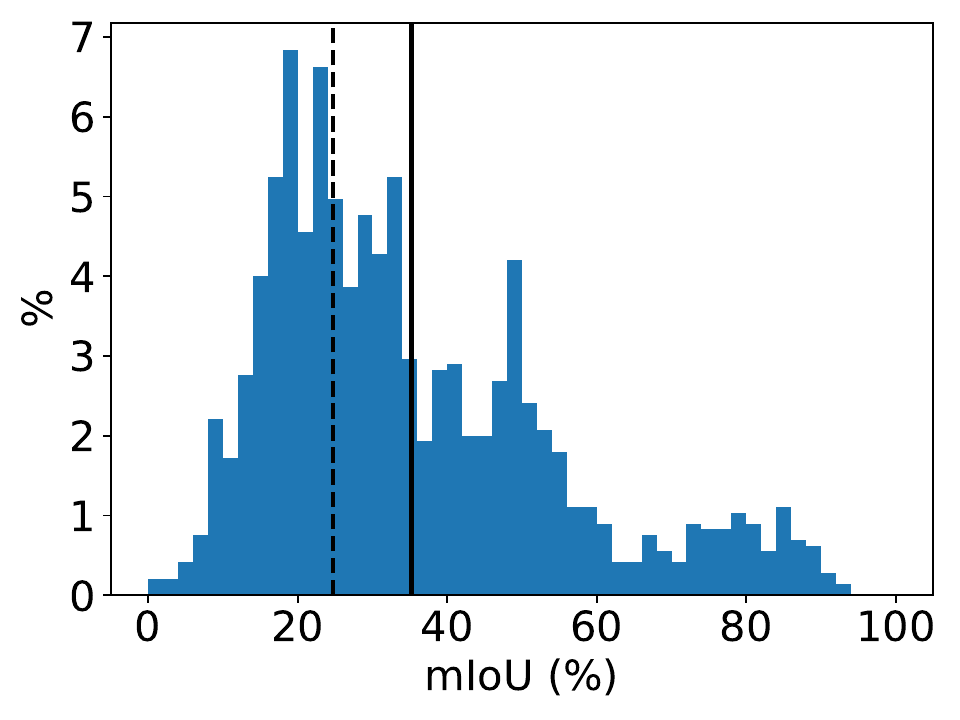} &
\includegraphics[width=0.24\textwidth]{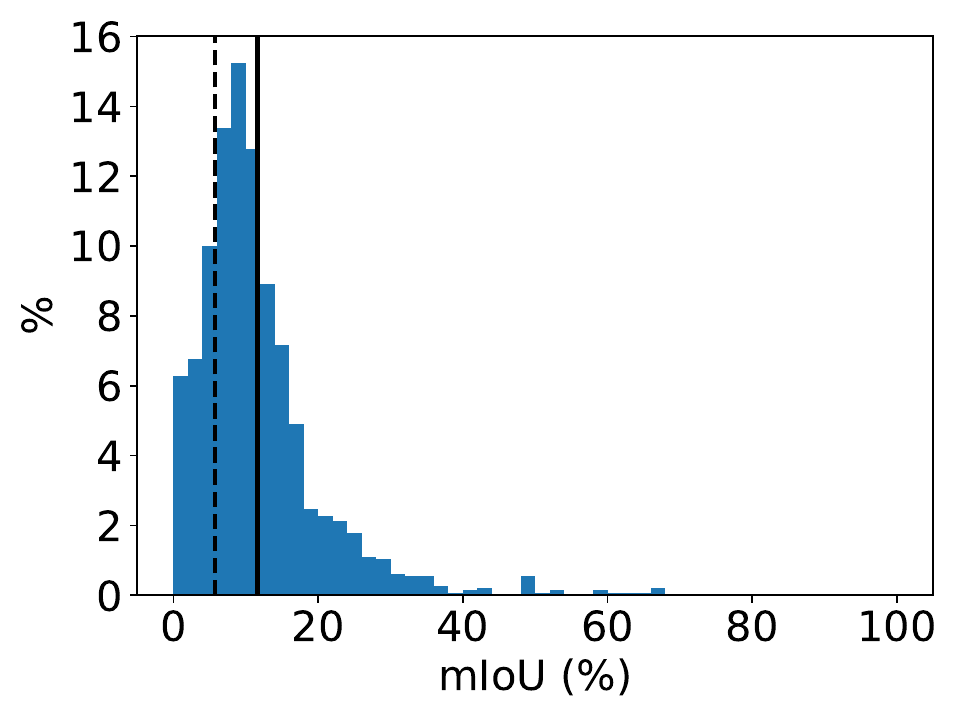} &
\includegraphics[width=0.24\textwidth]{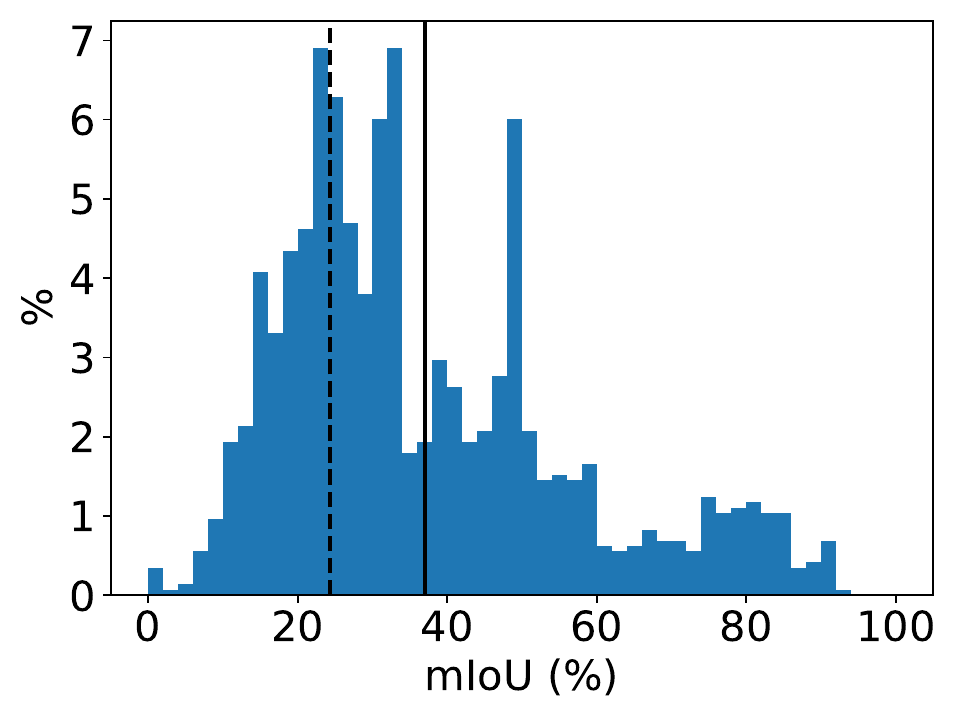} &
\includegraphics[width=0.24\textwidth]{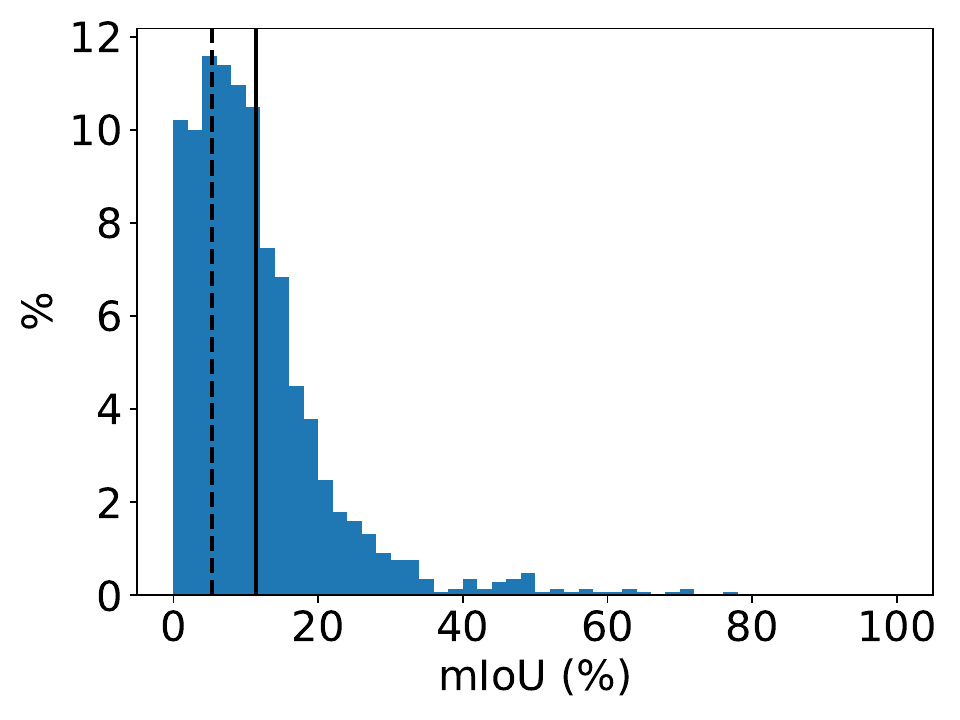} \\
\rotatebox[origin=l]{90}{\hspace{2mm}SegPGD-AT-100} & 
\includegraphics[width=0.24\textwidth]{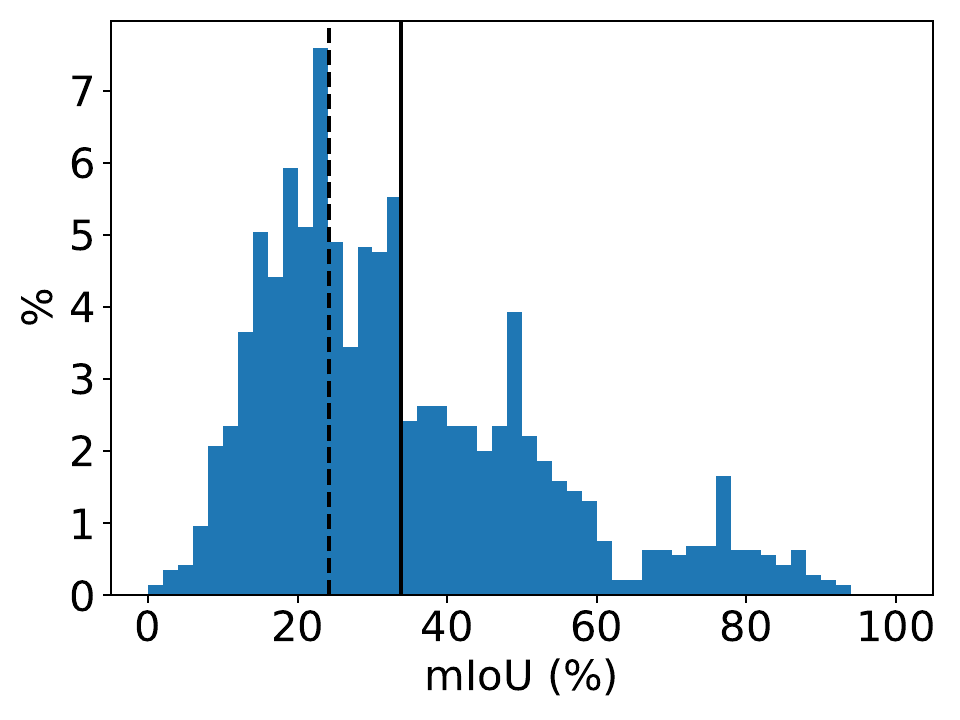} &
\includegraphics[width=0.24\textwidth]{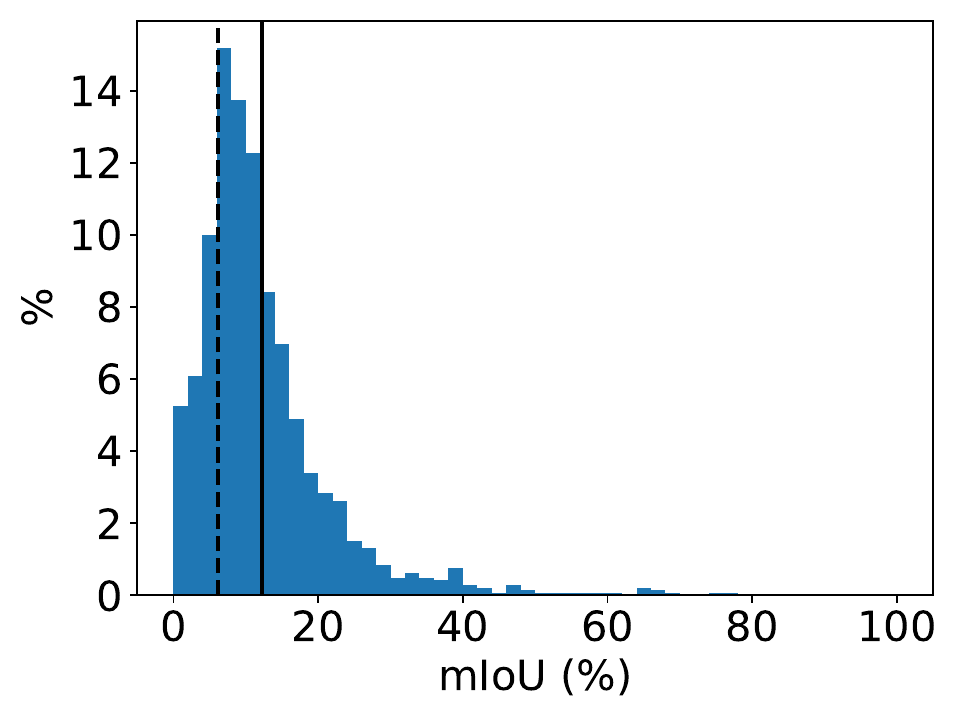} &
\includegraphics[width=0.24\textwidth]{gdrive/data_final/pascal/seg_pgd-100_pspnet/miou_normal_dist_with_bg} &
\includegraphics[width=0.24\textwidth]{gdrive/data_final/pascal/seg_pgd-100_pspnet/miou_min_dist_with_bg} \\
\end{tabular}
\caption{Distributions of single image mIoU on PASCAL VOC 2012.
CmIoU and NmIoU are shown using vertical lines.}
\label{fig:supp-p-miouhistograms}
\end{figure*}

\begin{figure*}
\setlength{\tabcolsep}{1pt}
\centering
\tiny
\begin{tabular}{ccccc}
& clean, DeepLabv3 & robust, DeepLabv3 & clean, PSPNet & robust, PSPNet\\
\rotatebox[origin=l]{90}{\hspace{8mm}Normal} & 
\includegraphics[width=0.24\textwidth]{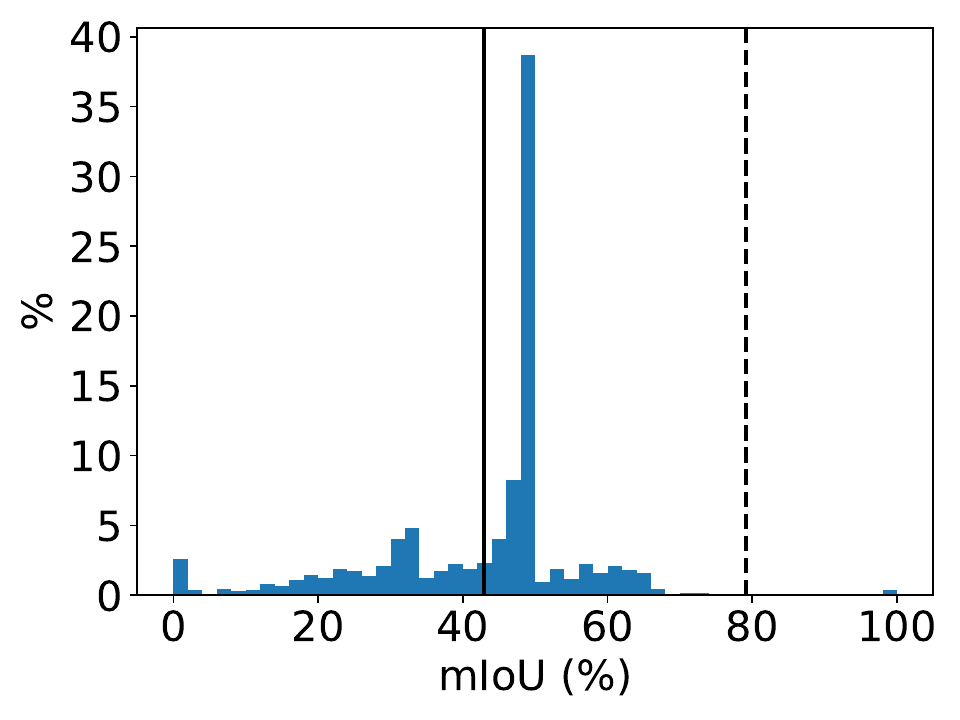} &
\includegraphics[width=0.24\textwidth]{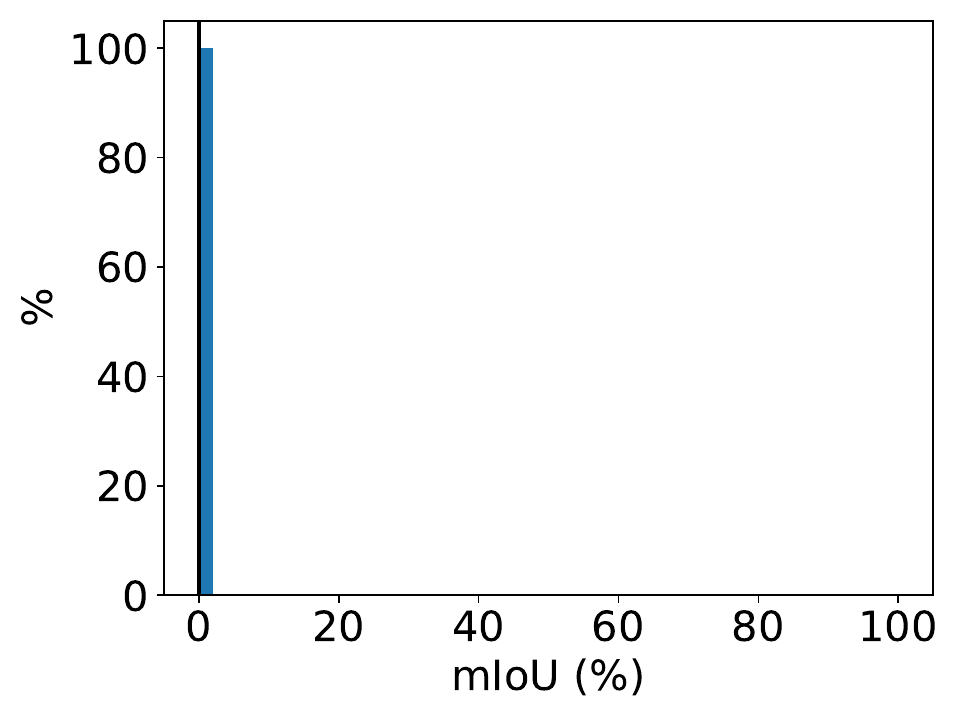} &
\includegraphics[width=0.24\textwidth]{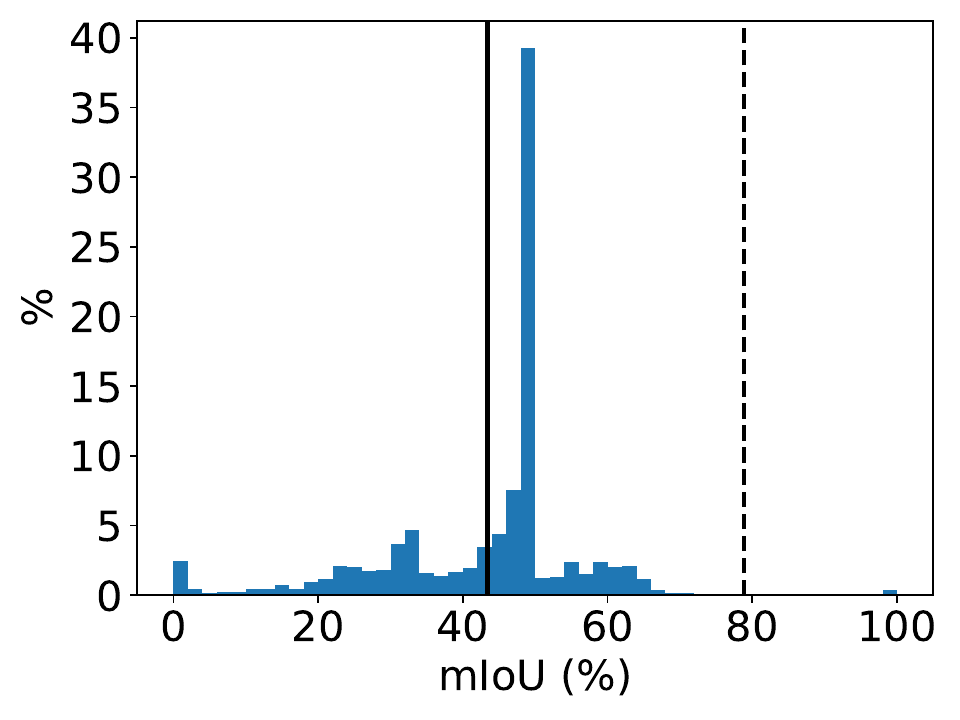} &
\includegraphics[width=0.24\textwidth]{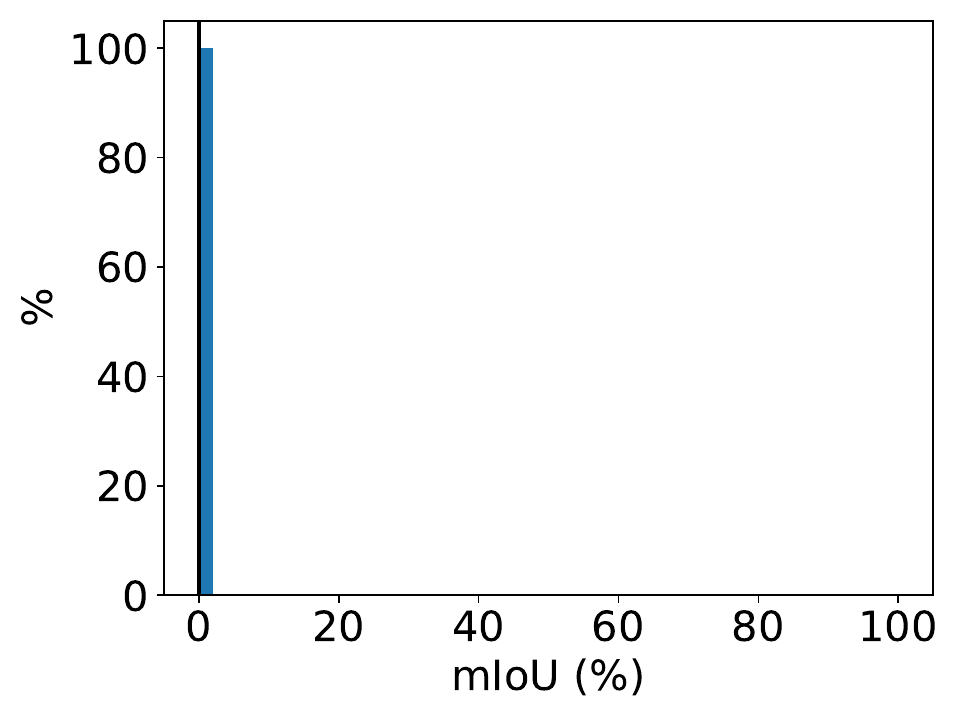} \\
\rotatebox[origin=l]{90}{\hspace{8mm}DDC-AT} & 
\includegraphics[width=0.24\textwidth]{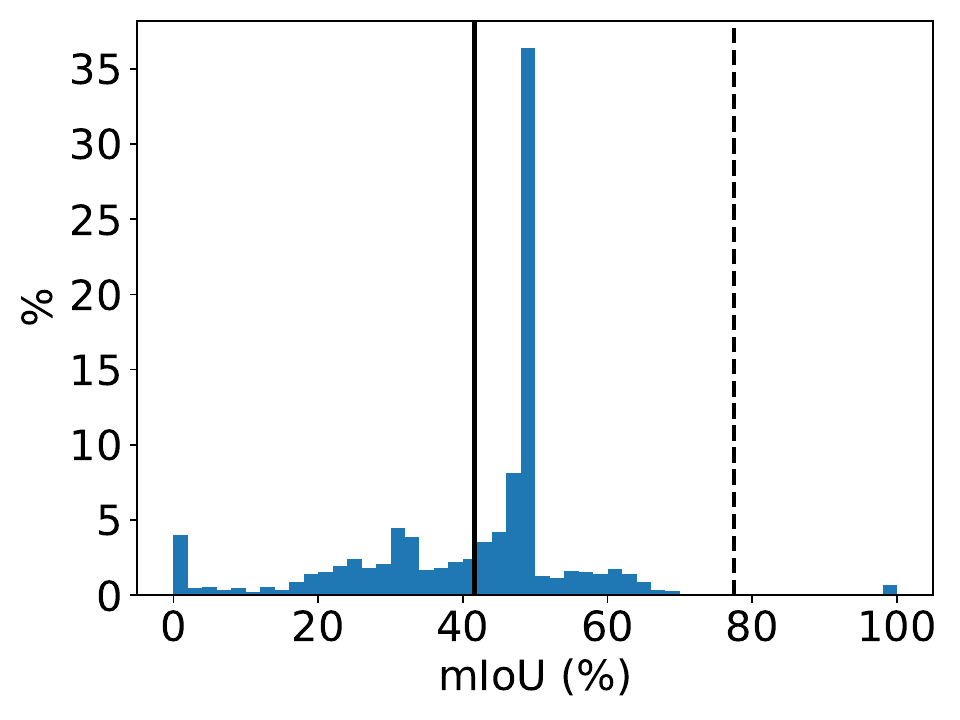} &
\includegraphics[width=0.24\textwidth]{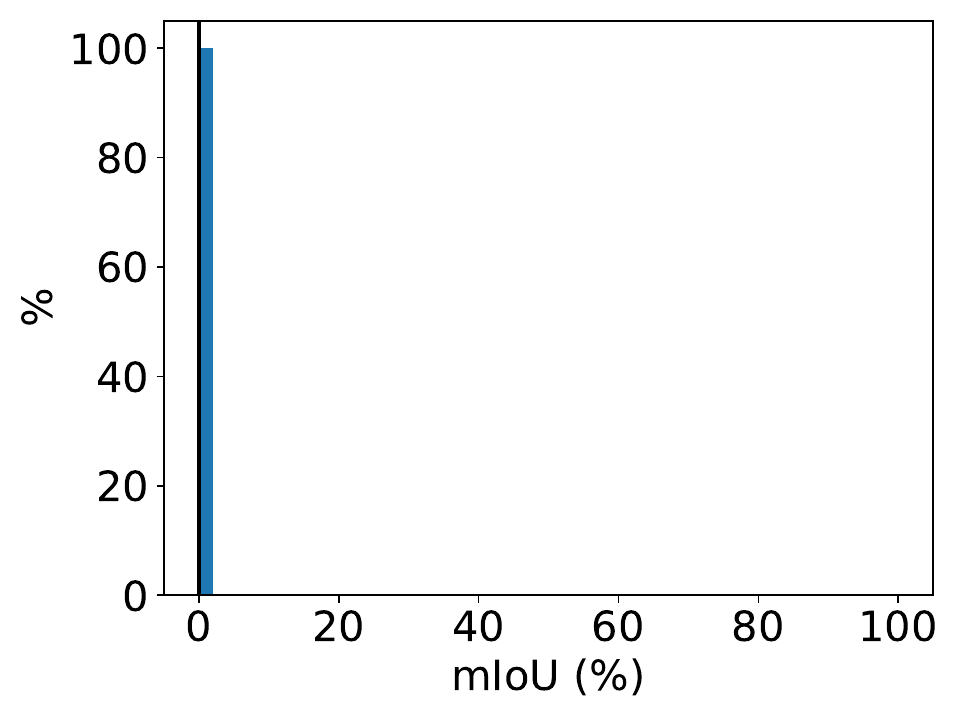} &
\includegraphics[width=0.24\textwidth]{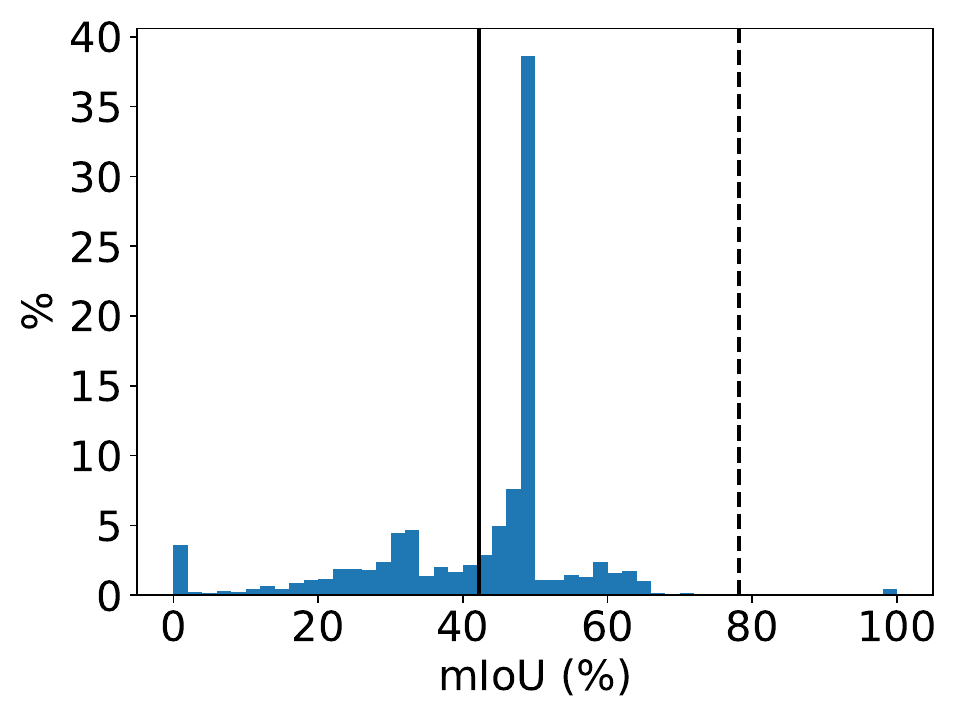} &
\includegraphics[width=0.24\textwidth]{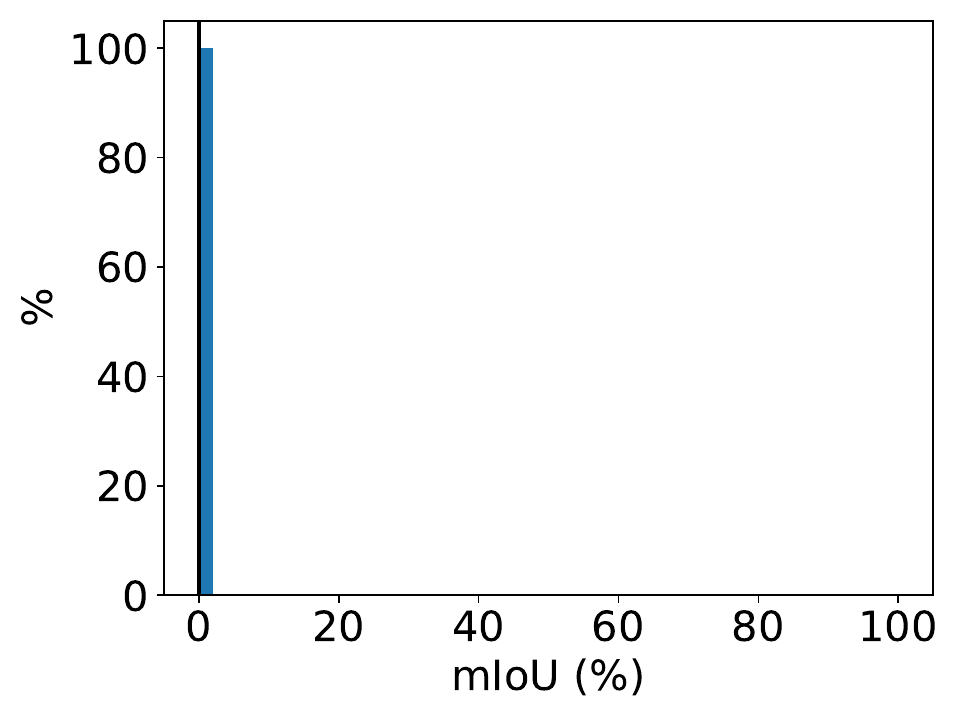} \\
\rotatebox[origin=l]{90}{\hspace{8mm}PGD-AT} & 
\includegraphics[width=0.24\textwidth]{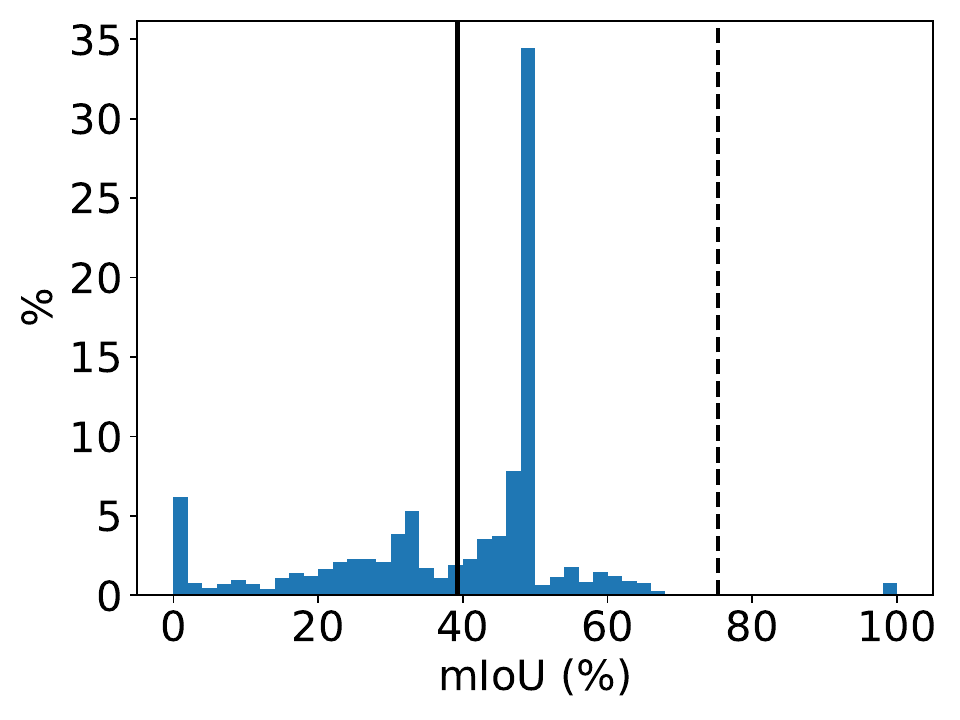} &
\includegraphics[width=0.24\textwidth]{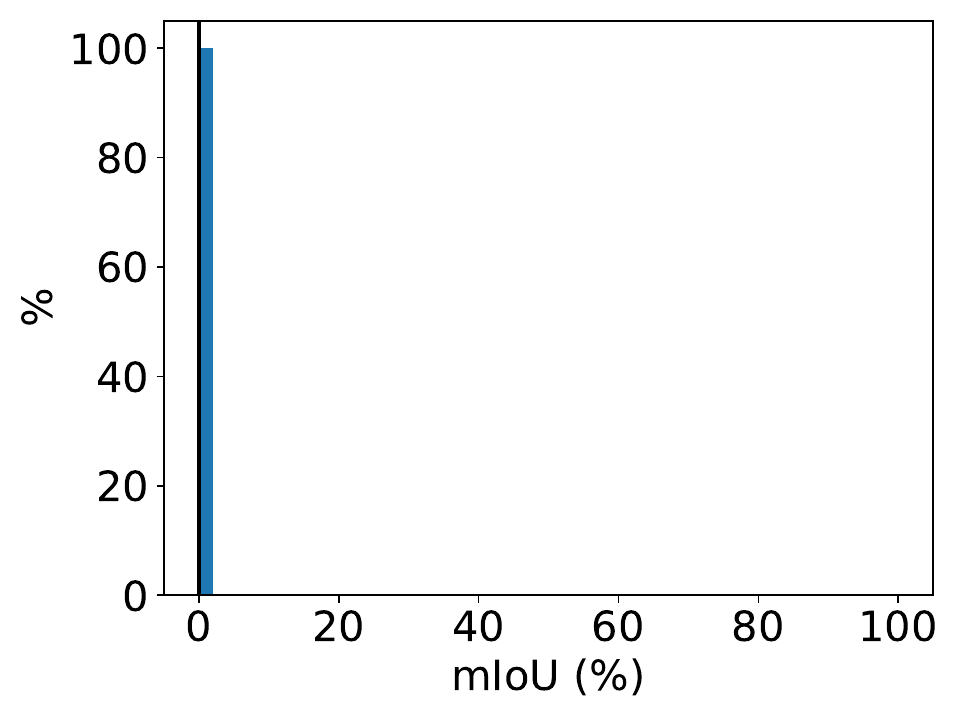} &
\includegraphics[width=0.24\textwidth]{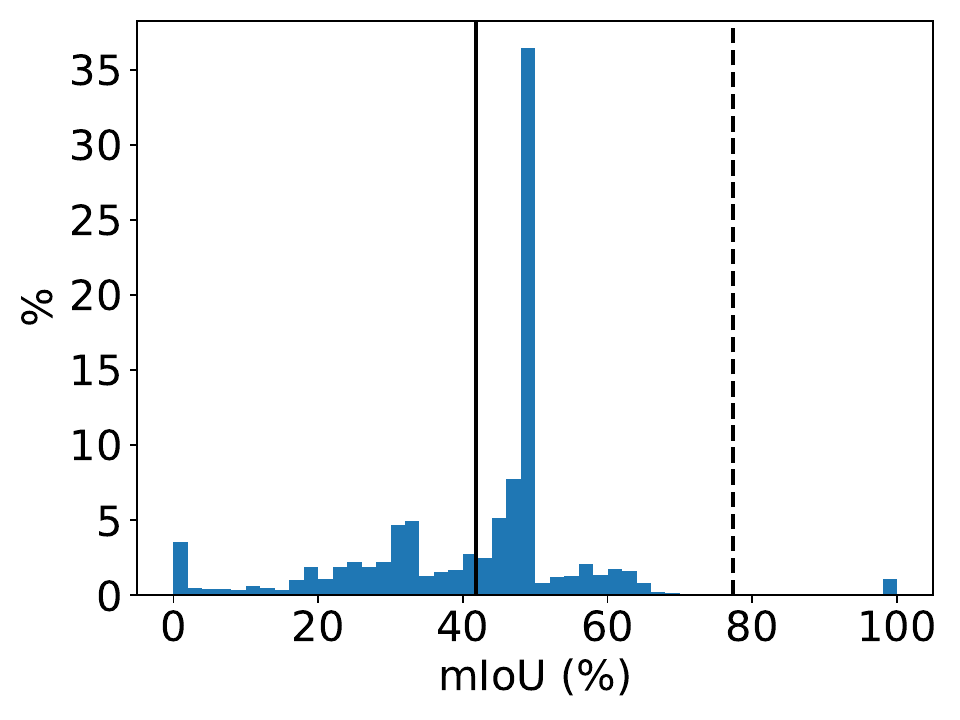} &
\includegraphics[width=0.24\textwidth]{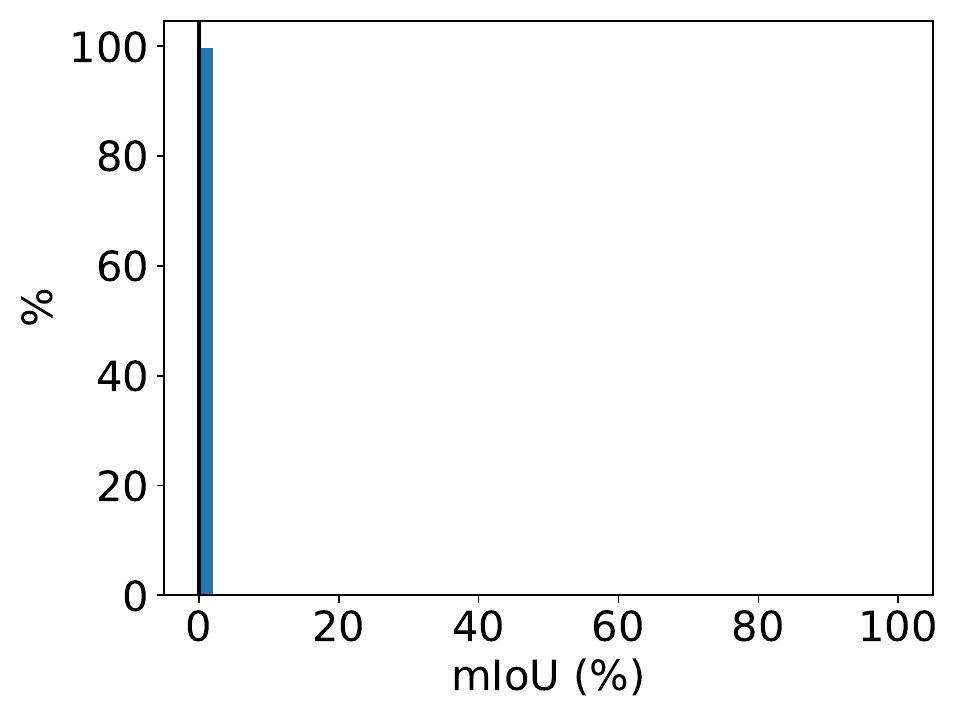} \\
\rotatebox[origin=l]{90}{\hspace{4mm}SegPGD-AT} & 
\includegraphics[width=0.24\textwidth]{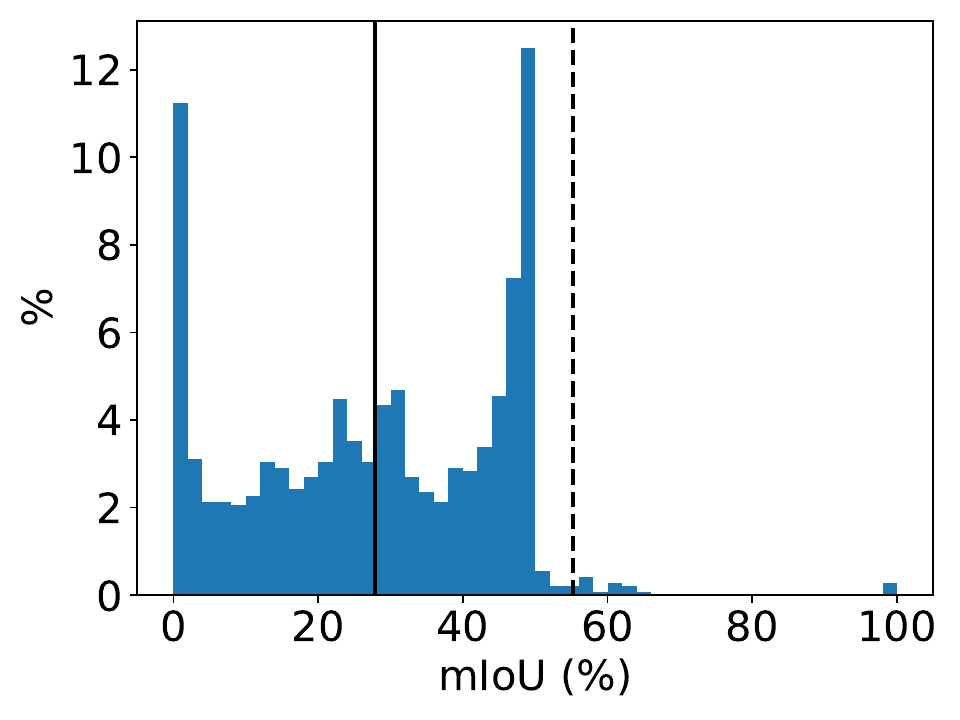} &
\includegraphics[width=0.24\textwidth]{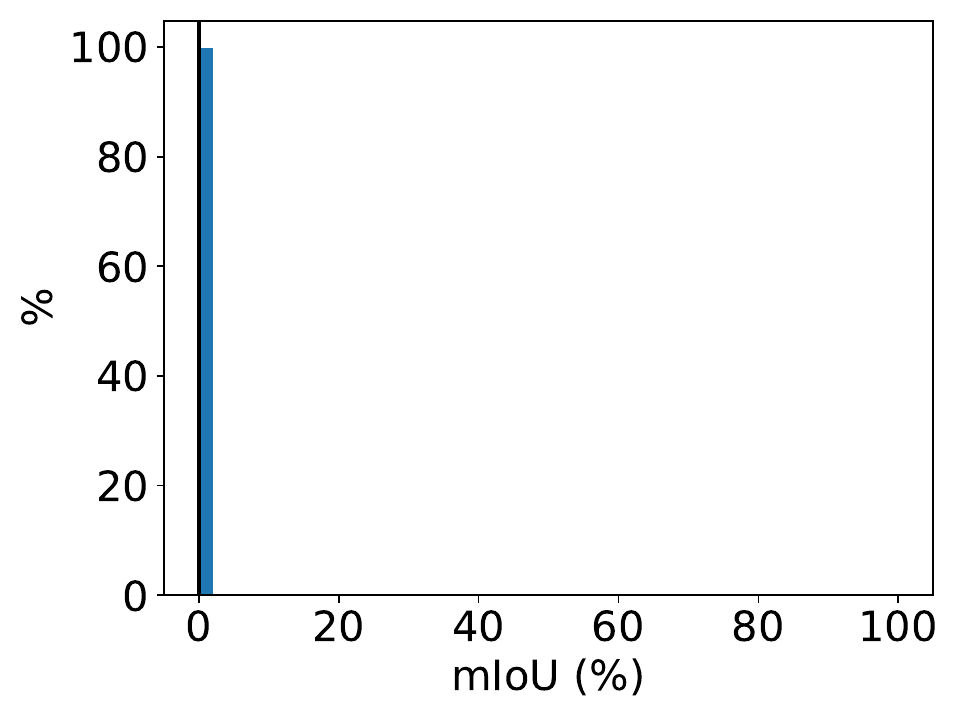} &
\includegraphics[width=0.24\textwidth]{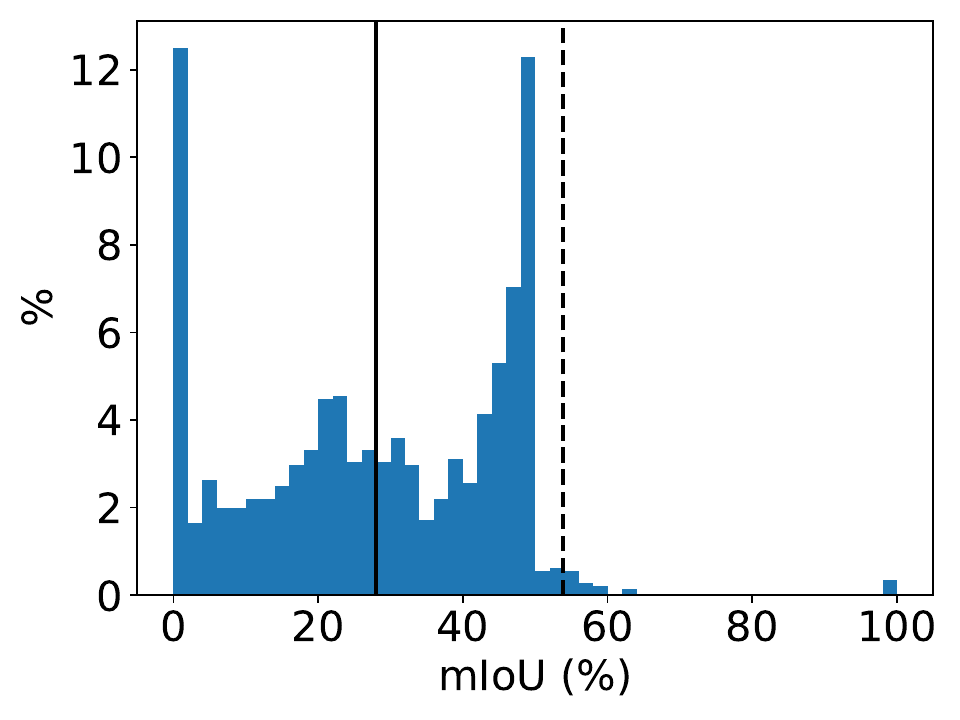} &
\includegraphics[width=0.24\textwidth]{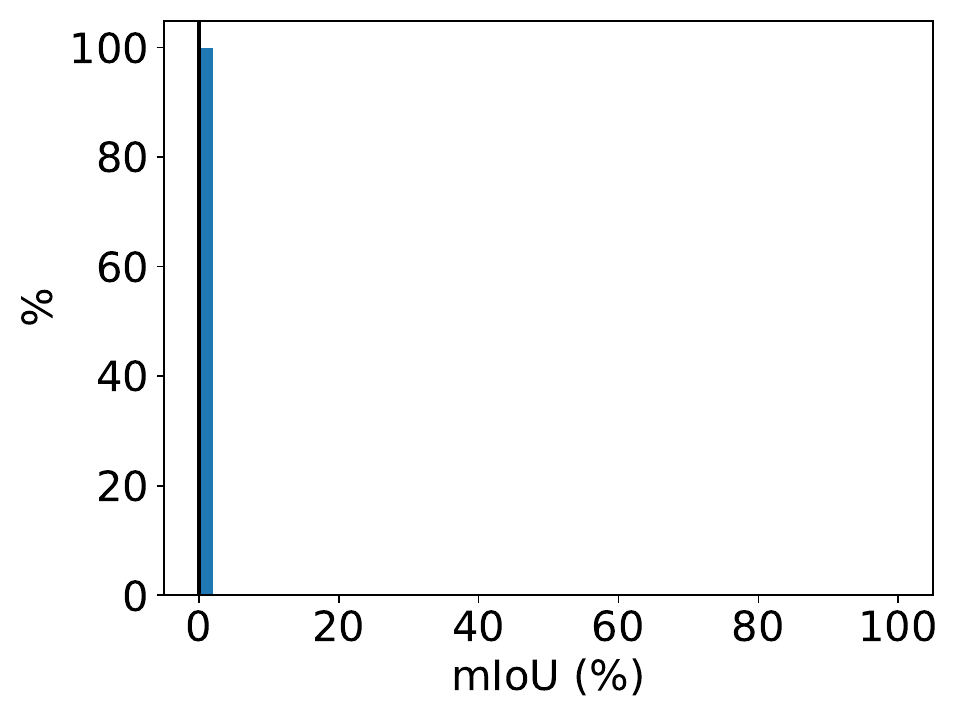} \\
\rotatebox[origin=l]{90}{\hspace{4mm}PGD-AT-100} & 
\includegraphics[width=0.24\textwidth]{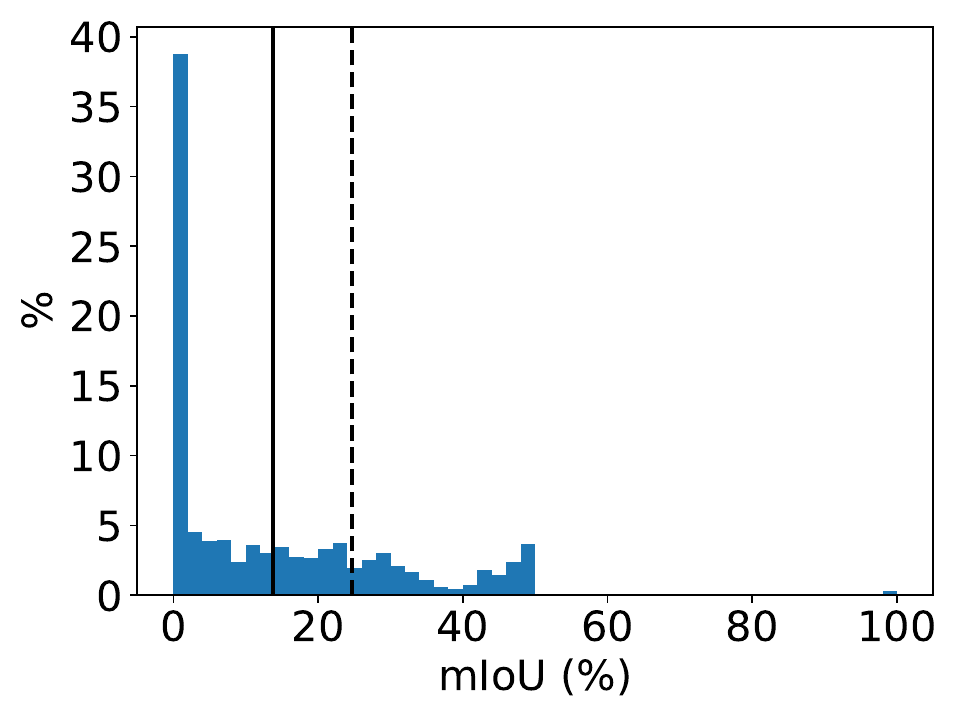} &
\includegraphics[width=0.24\textwidth]{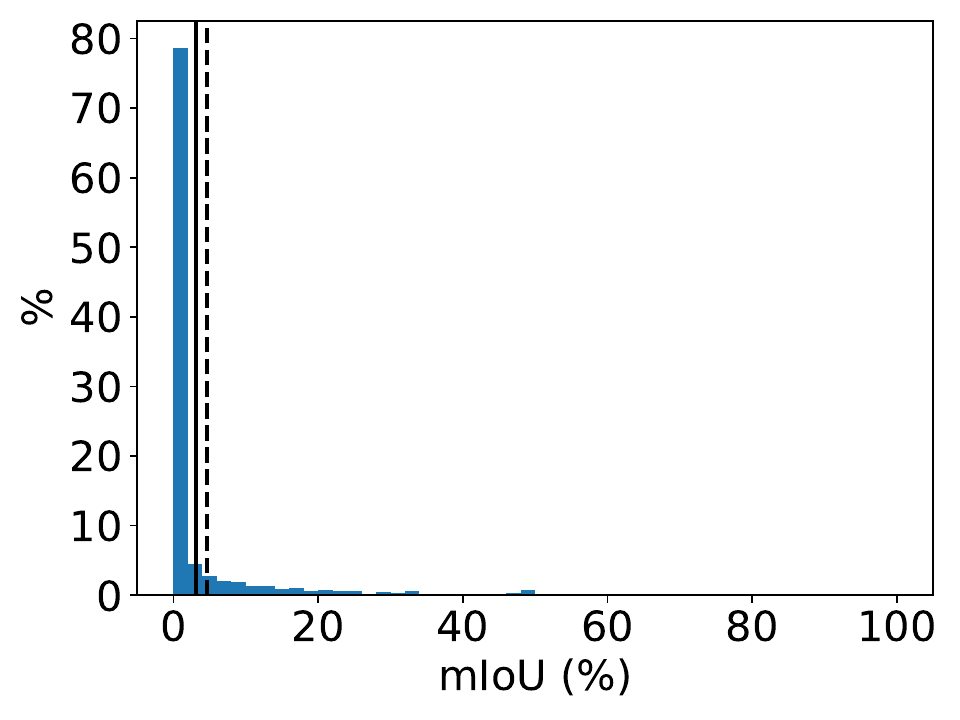} &
\includegraphics[width=0.24\textwidth]{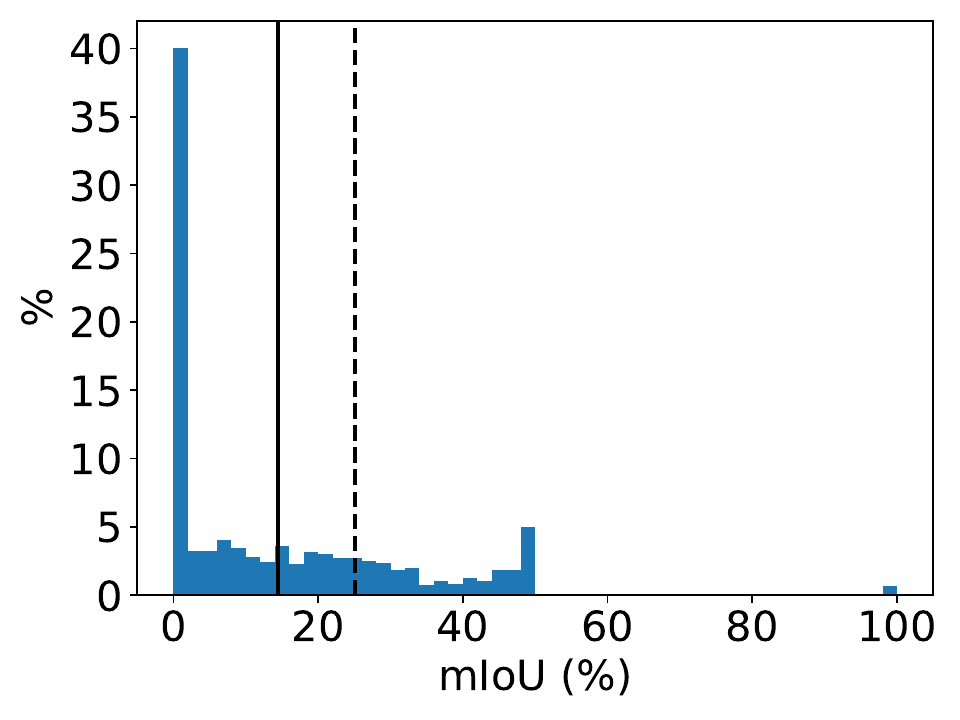} &
\includegraphics[width=0.24\textwidth]{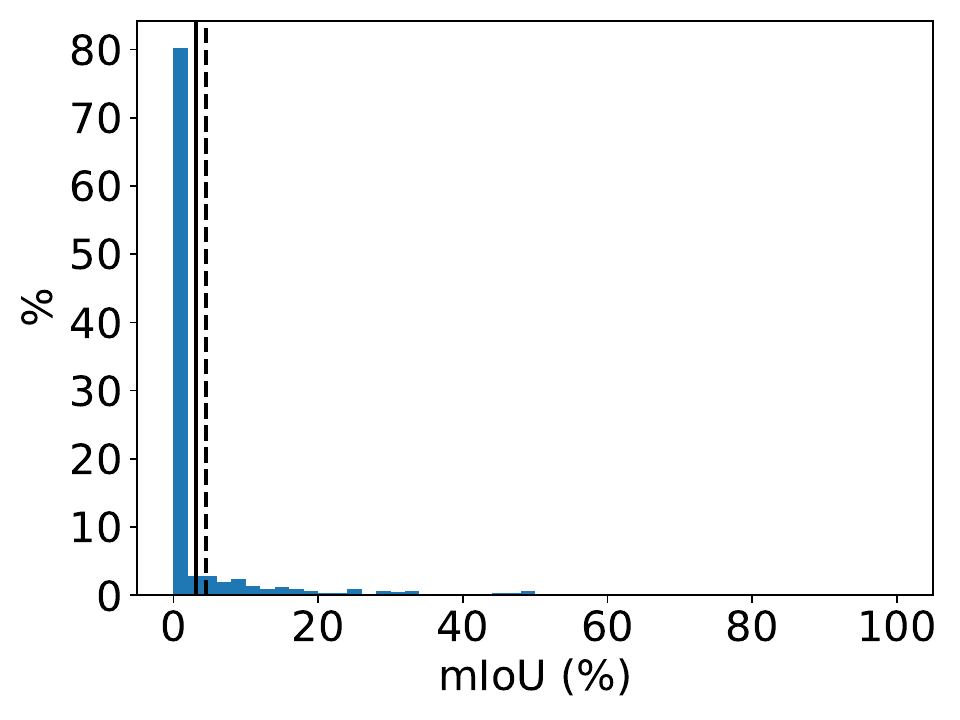} \\
\rotatebox[origin=l]{90}{\hspace{2mm}SegPGD-AT-100} & 
\includegraphics[width=0.24\textwidth]{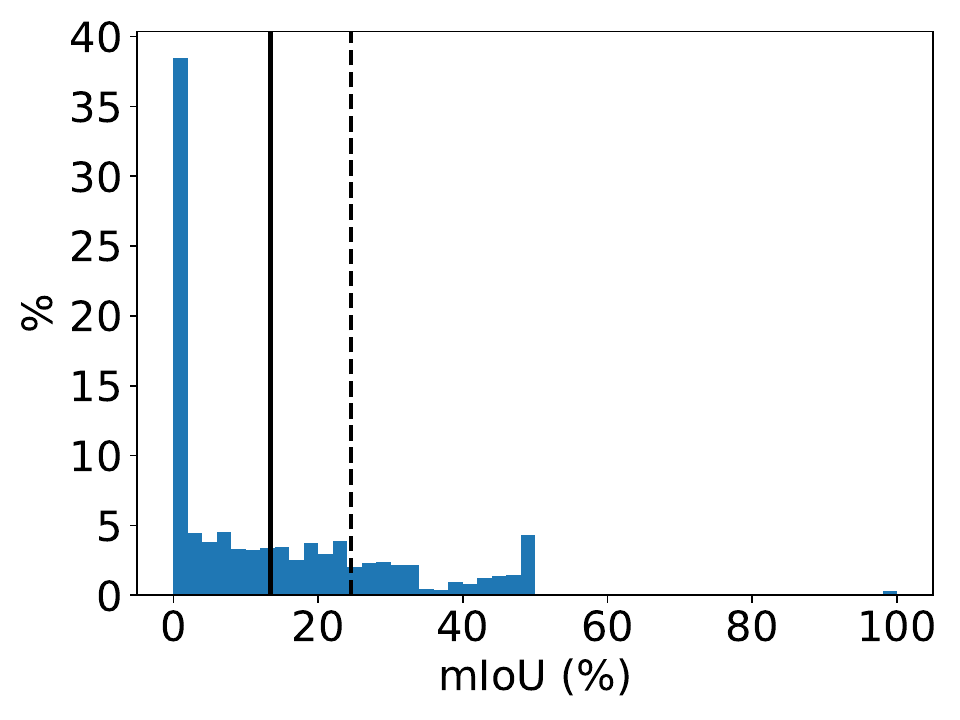} &
\includegraphics[width=0.24\textwidth]{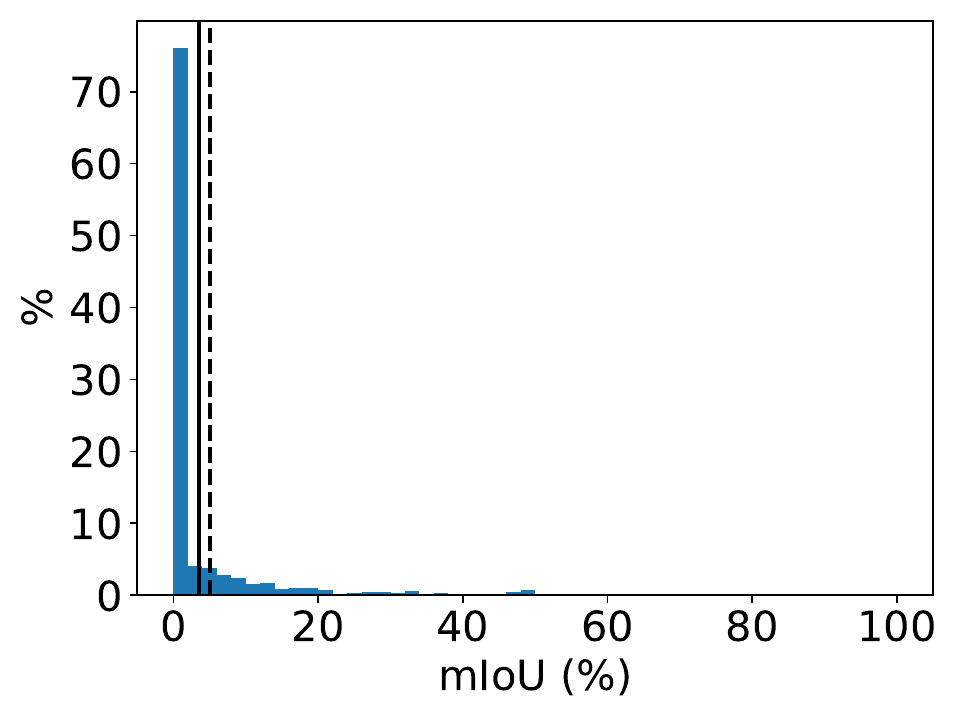} &
\includegraphics[width=0.24\textwidth]{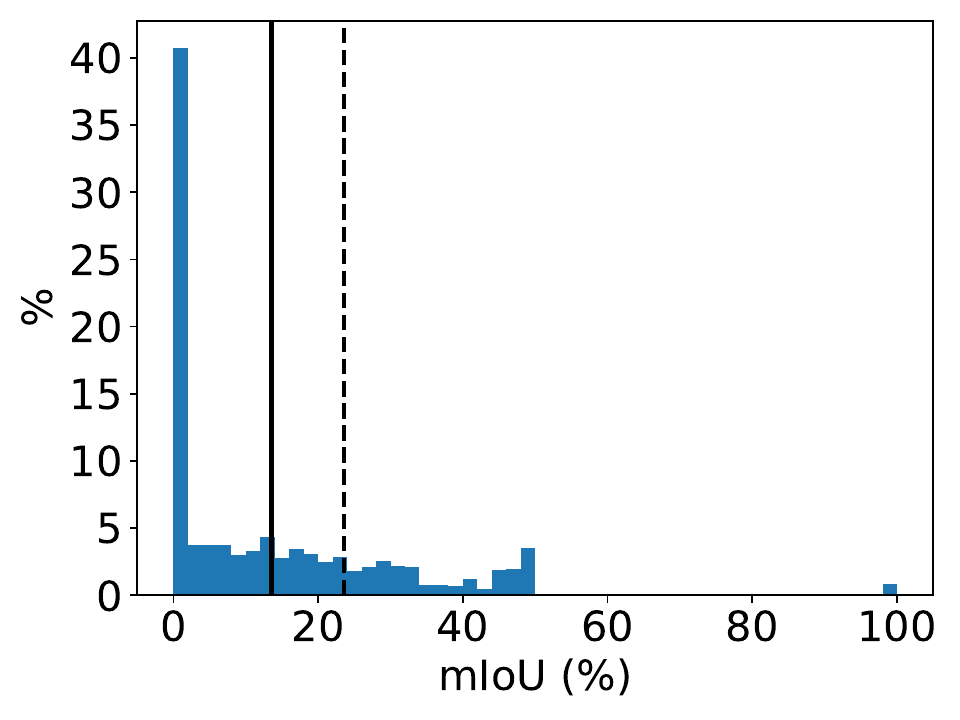} &
\includegraphics[width=0.24\textwidth]{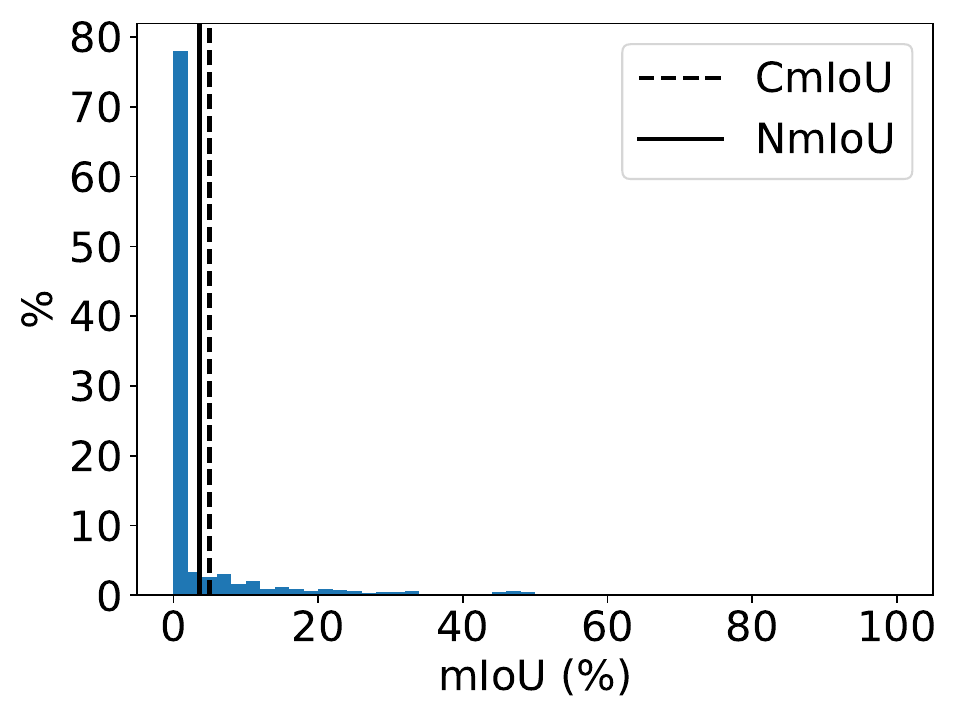} \\
\end{tabular}
\caption{Distributions of single image mIoU on PASCAL VOC 2012 with pixels with the background ground truth label ignored.
CmIoU and NmIoU are shown using vertical lines.}
\label{fig:supp-pnb-miouhistograms}
\end{figure*}

\begin{figure*}
\setlength{\tabcolsep}{1pt}
\centering
\tiny
\begin{tabular}{ccccc}
& clean, Tiny & robust, Tiny & clean, Small & robust, Small\\
\rotatebox[origin=l]{90}{\hspace{8mm}Normal} & 
\includegraphics[width=0.24\textwidth]{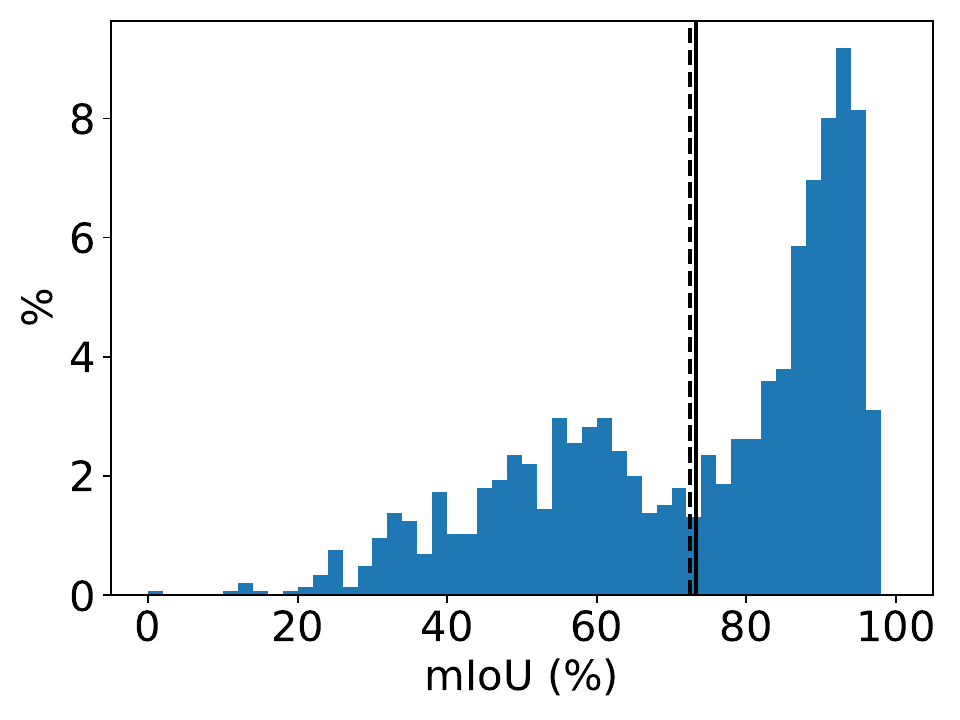} &
\includegraphics[width=0.24\textwidth]{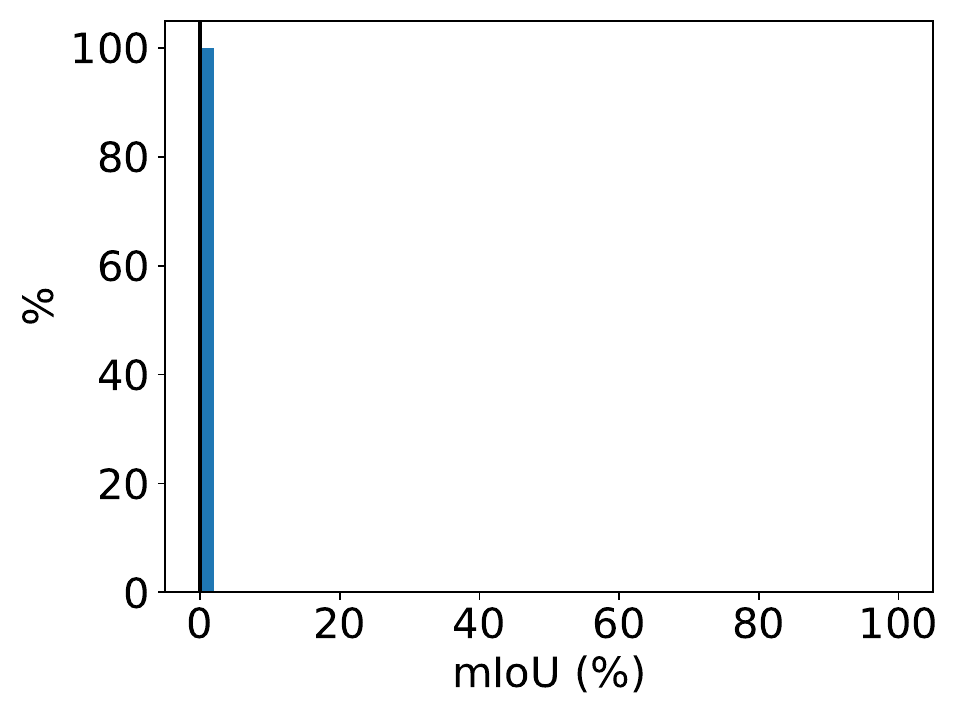} &
\includegraphics[width=0.24\textwidth]{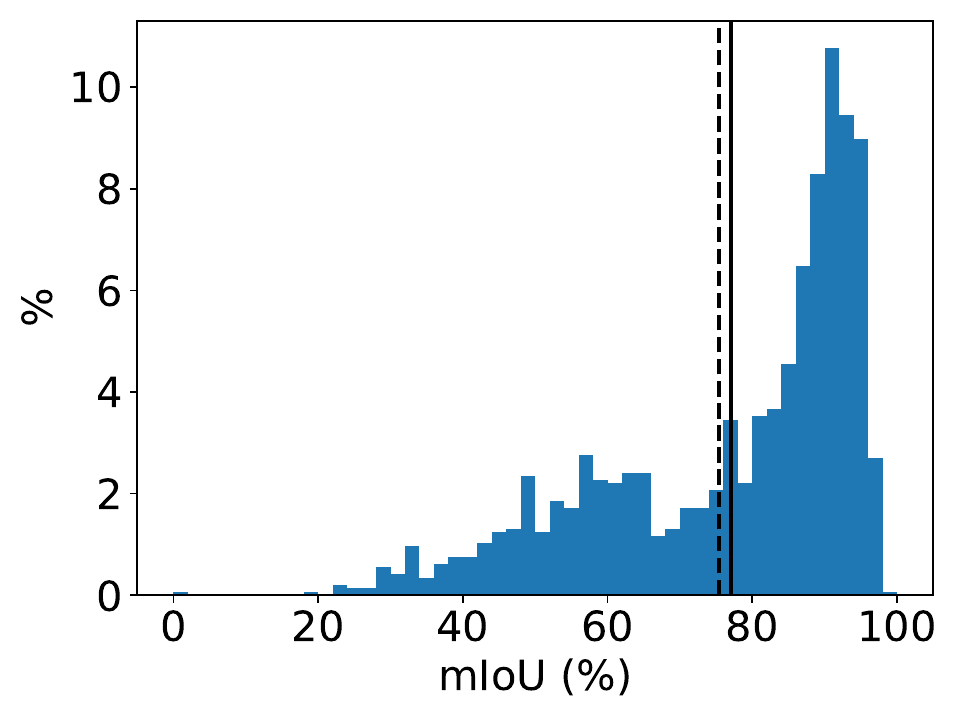} &
\includegraphics[width=0.24\textwidth]{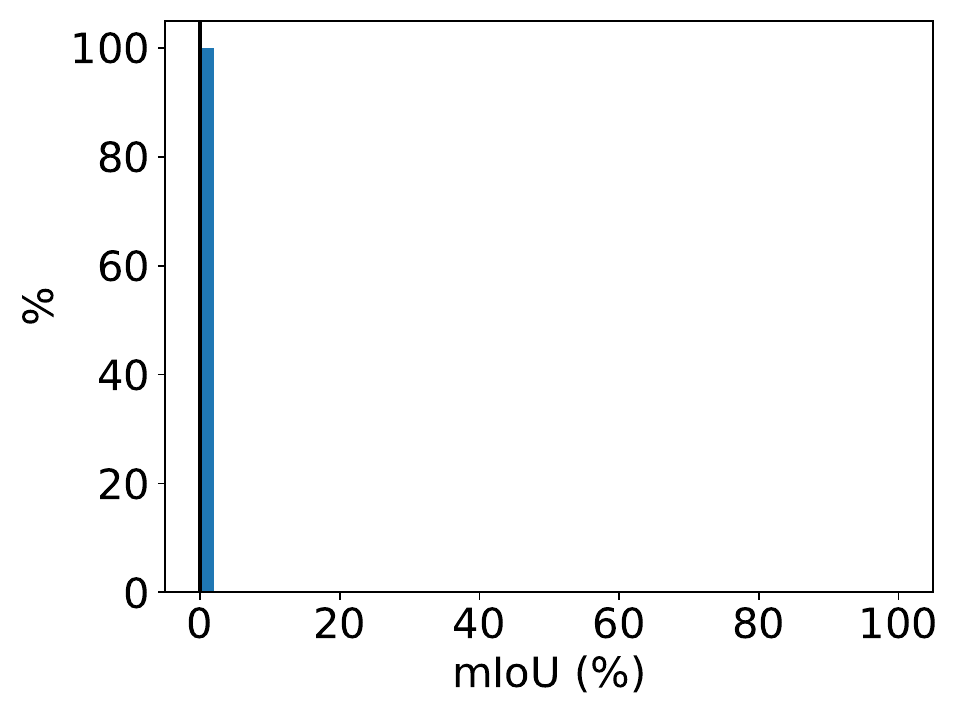} \\
\rotatebox[origin=l]{90}{\hspace{8mm}SEA-AT} & 
\includegraphics[width=0.24\textwidth]{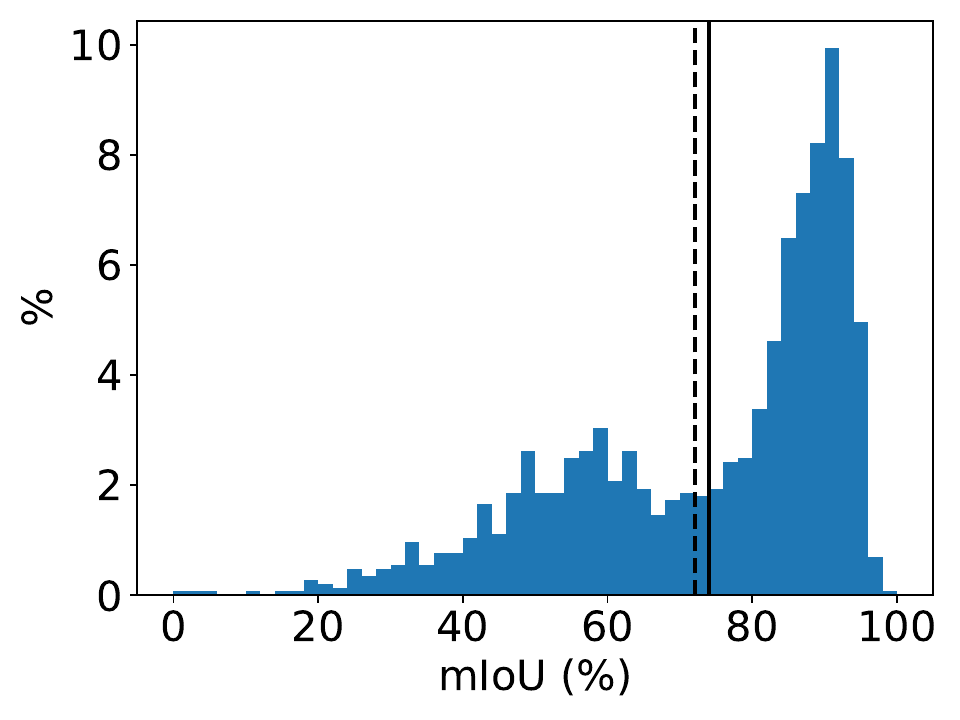} &
\includegraphics[width=0.24\textwidth]{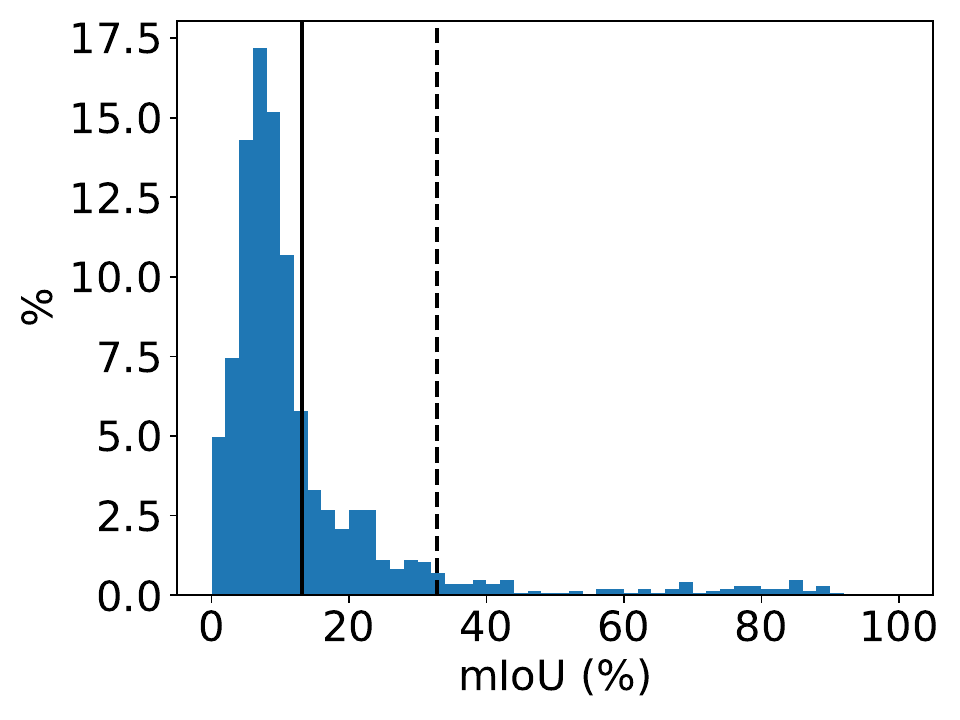} &
\includegraphics[width=0.24\textwidth]{gdrive/data_final/pascal/ConvNeXt-S_CVST_ROB/miou_normal_dist_with_bg} &
\includegraphics[width=0.24\textwidth]{gdrive/data_final/pascal/ConvNeXt-S_CVST_ROB/miou_min_dist_with_bg} \\
\end{tabular}
\caption{Distributions of single image mIoU for the SEA-AT model family on PASCAL VOC 2012.
CmIoU and NmIoU are shown using vertical lines.}
\label{fig:supp-p-sea-miouhistograms}
\end{figure*}

\begin{figure*}
\setlength{\tabcolsep}{1pt}
\centering
\tiny
\begin{tabular}{ccccc}
& clean, Tiny & robust, Tiny & clean, Small & robust, Small\\
\rotatebox[origin=l]{90}{\hspace{8mm}Normal} & 
\includegraphics[width=0.24\textwidth]{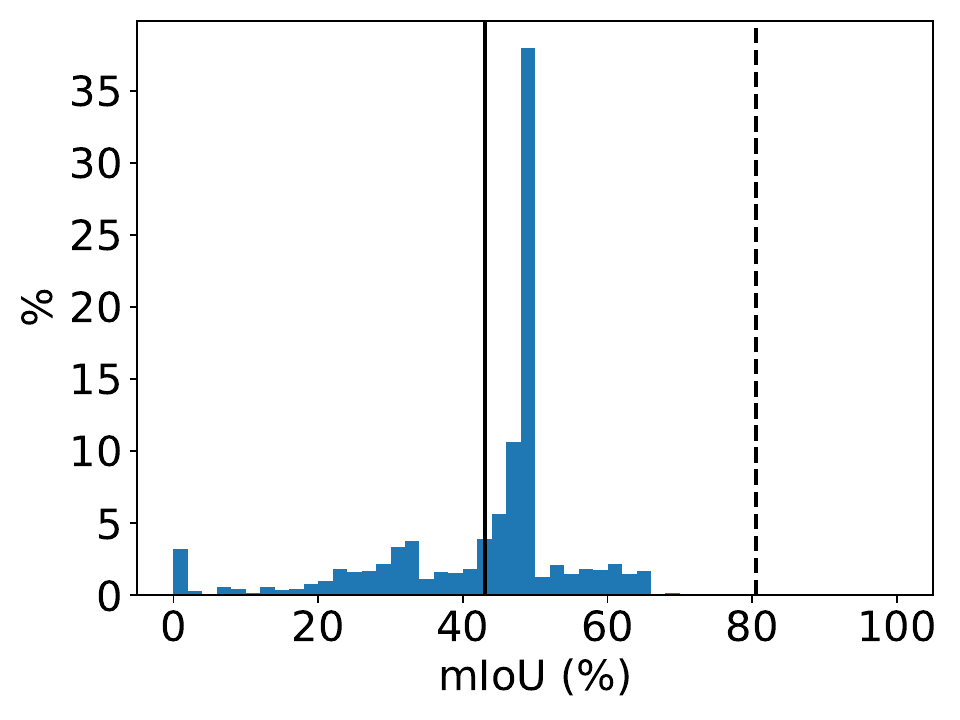} &
\includegraphics[width=0.24\textwidth]{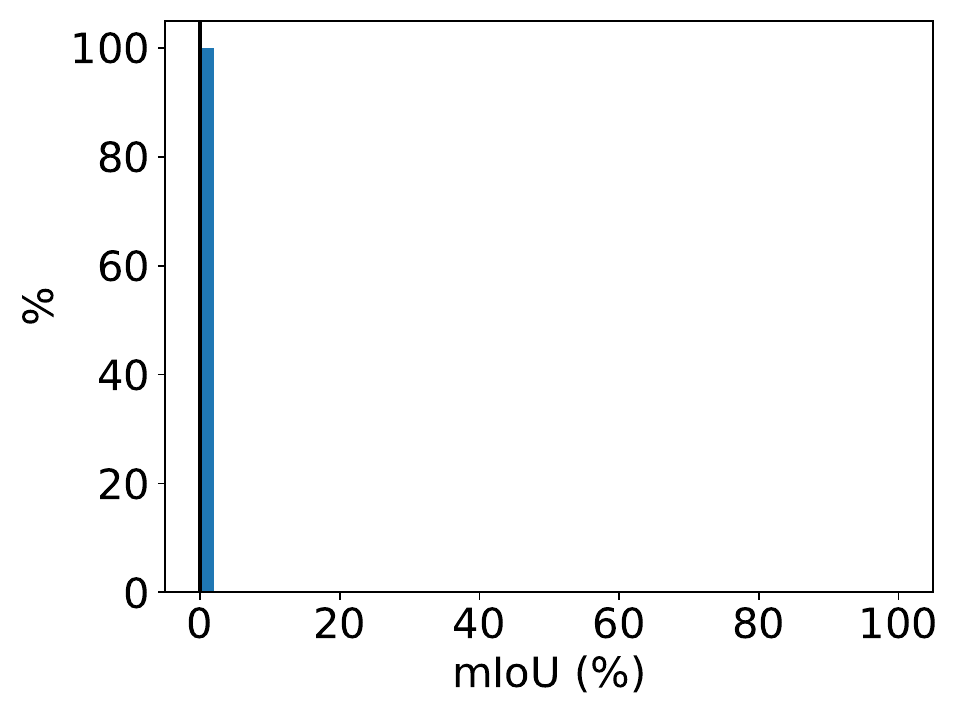} &
\includegraphics[width=0.24\textwidth]{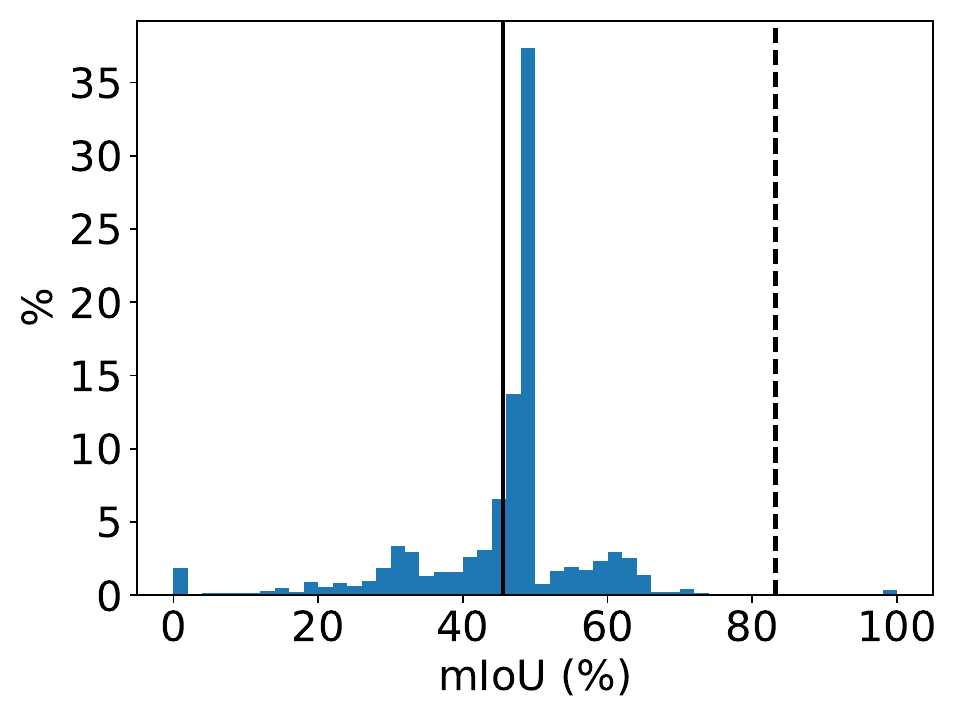} &
\includegraphics[width=0.24\textwidth]{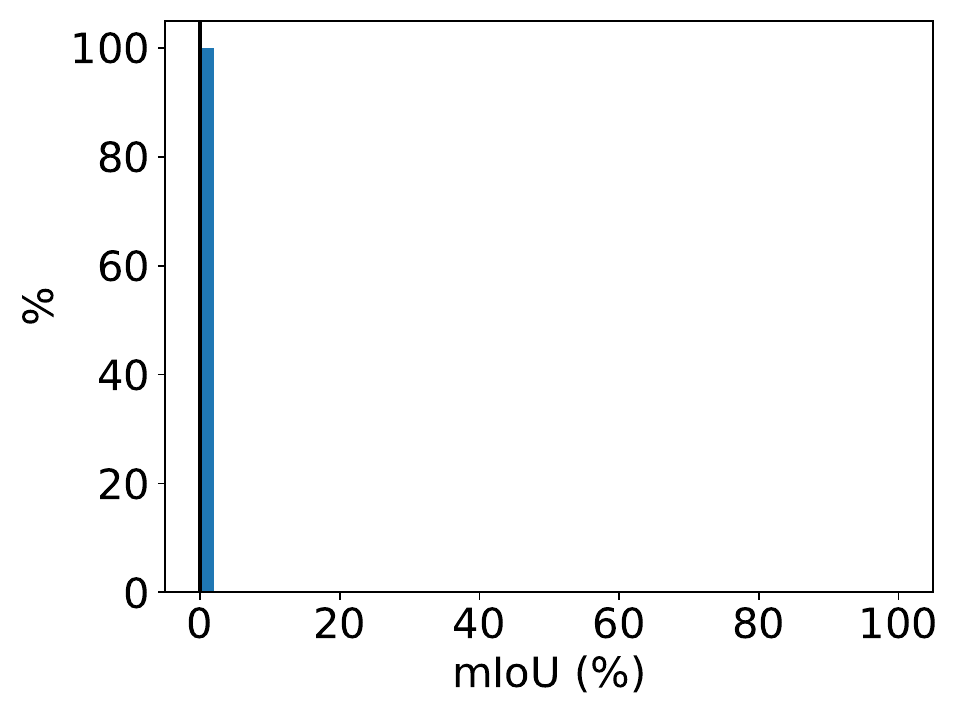} \\
\rotatebox[origin=l]{90}{\hspace{8mm}SEA-AT} & 
\includegraphics[width=0.24\textwidth]{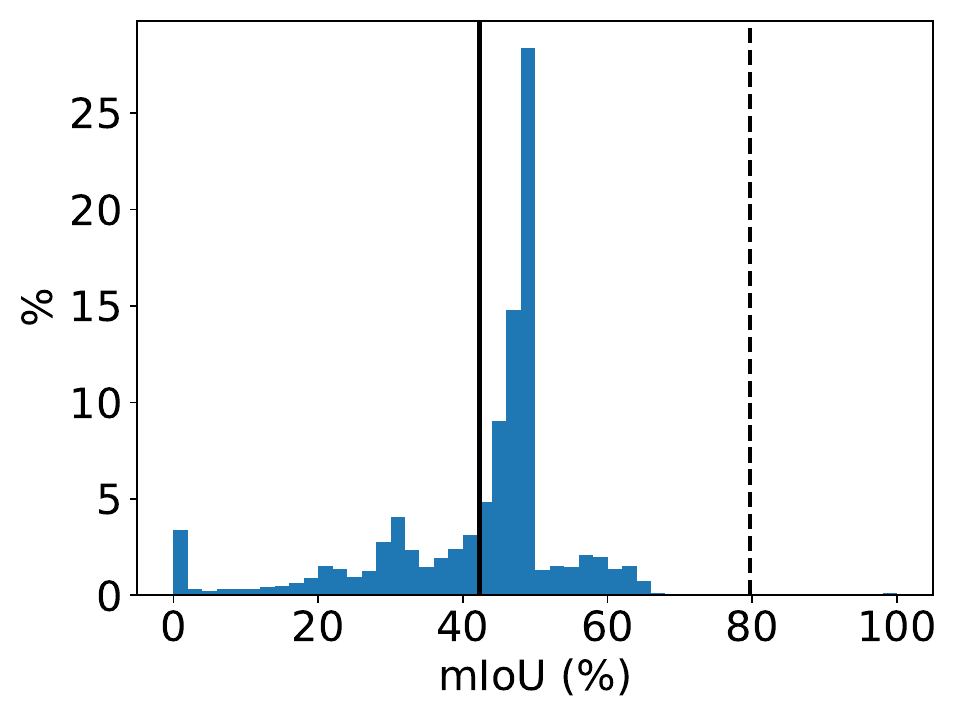} &
\includegraphics[width=0.24\textwidth]{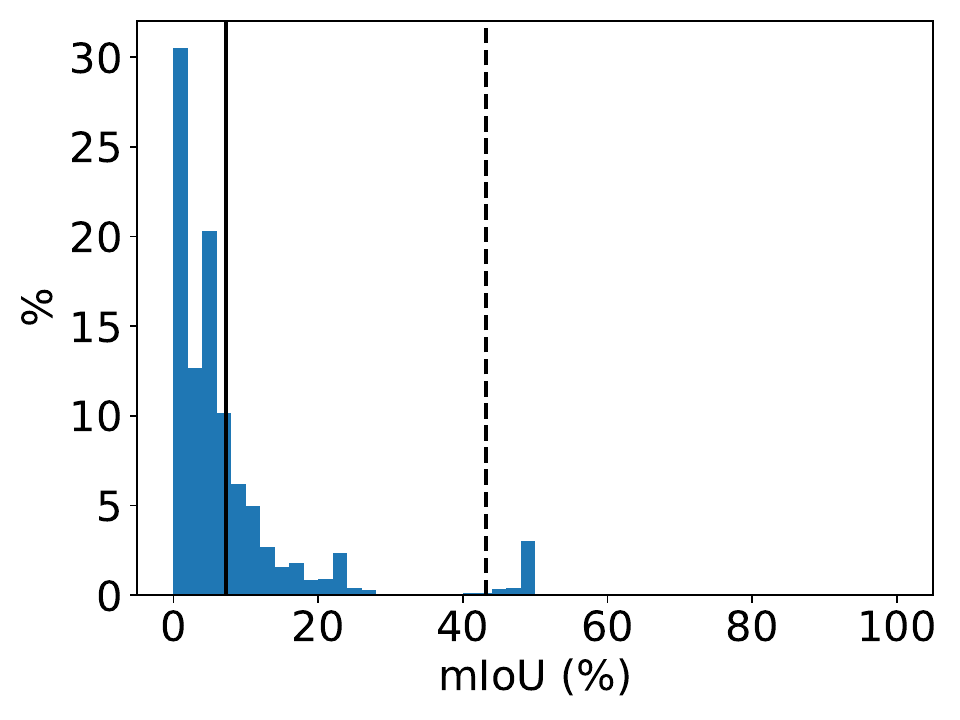} &
\includegraphics[width=0.24\textwidth]{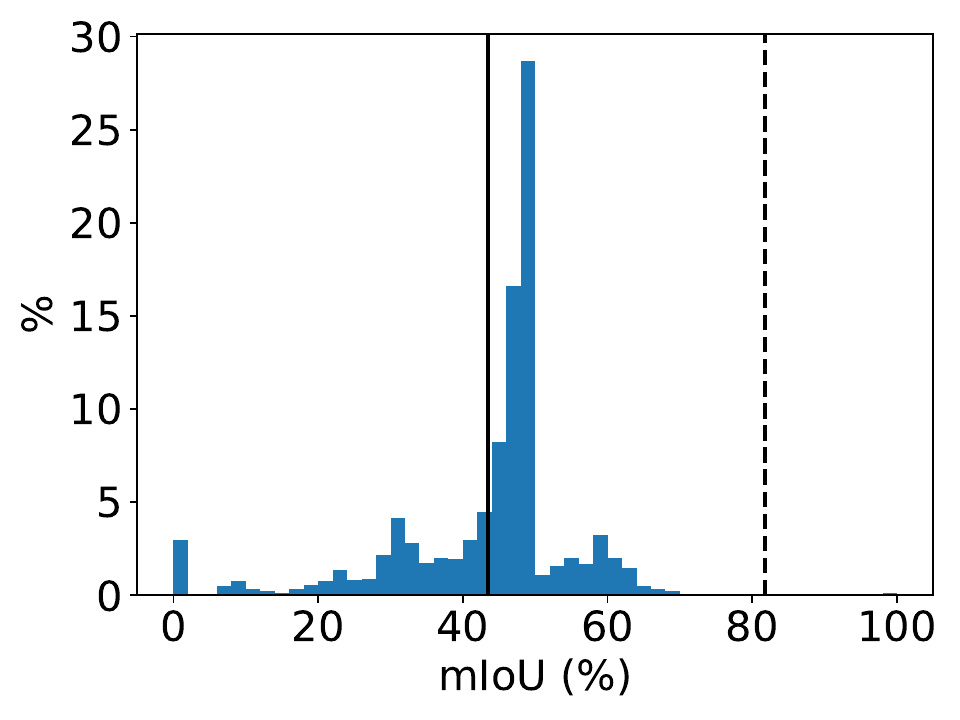} &
\includegraphics[width=0.24\textwidth]{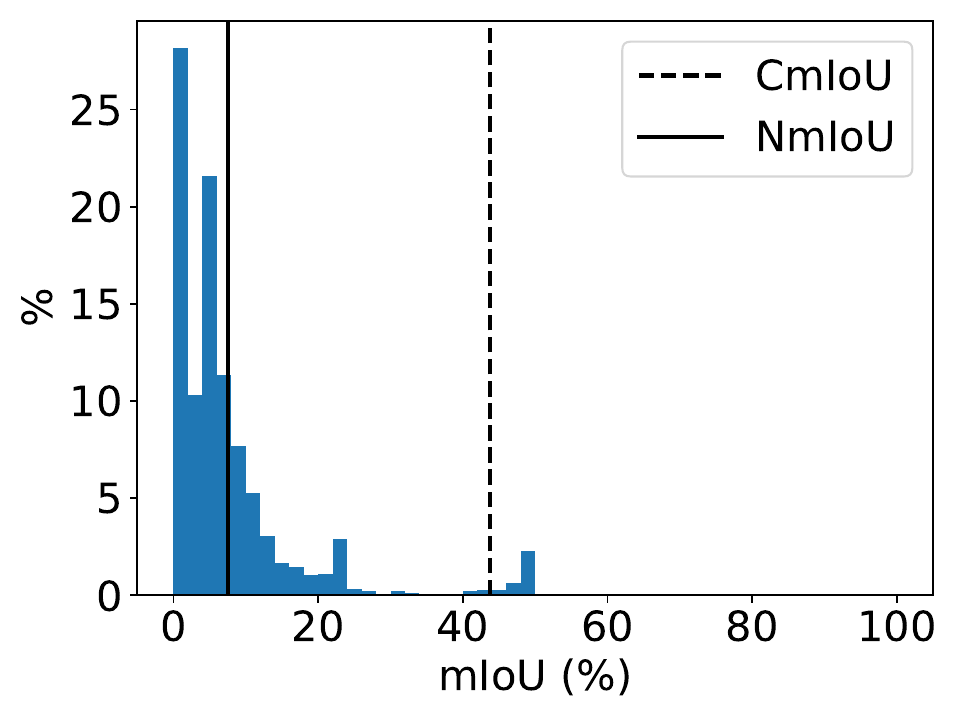} \\
\end{tabular}
\caption{Distributions of single image mIoU for the SEA-AT model family on PASCAL VOC 2012 with pixels with the background ground truth label ignored.
CmIoU and NmIoU are shown using vertical lines.}
\label{fig:supp-pnb-sea-miouhistograms}
\end{figure*}

As we can see in \cref{fig:supp-p-sea-miouhistograms,fig:supp-pnb-sea-miouhistograms}, the clean performance
of the SEA-AT models is also very similar to that of the normal models, like in the case of DDC-AT and
SegPGD-AT.
Also, the robust mIoU distribution is very similar visually to that of PGD-AT-100 and SegPGD-AT-100, as can be seen
comparing \cref{fig:supp-p-miouhistograms} and \cref{fig:supp-p-sea-miouhistograms}, except for the CmIoU values, which
dramatically differ.
This indicates the strong size-bias of the SEA-AT models, that is, large objects are protected more.
This is supported even more clearly if we consider also \cref{fig:supp-pnb-miouhistograms,fig:supp-pnb-sea-miouhistograms}.
In these figures, the difference between the CmIoU values is even larger, so the size-bias is related to the
foreground objects.

\section{A Closer Look at Individual Attacks}

Here, we present our complete set of measurement data about all the scenarios we studied for both
maximum damage attacks and minimum perturbation attacks.

\subsection{Maximum Damage Attacks}
\label{sec:supp-indattacks}

The performance of all the attacks in all the scenarios is included in
\cref{tab:supp-cs,tab:supp-p,tab:supp-pnb,tab:supp-sea}.
We also include the distribution of the most successful attack in all the
scenarios in
\cref{fig:supp-cs-attackhistograms,fig:supp-p-attackhistograms,fig:supp-p-sea-attackhistograms}.
Again, the conclusions of \cref{sec:evaluation} are further confirmed.

PAdam-Cos, our own attack, seems to be particularly successful in
the case of the Cityscapes dataset, and on non-robust models over the PASCAL VOC dataset,
while the SEA attacks are more useful in the case of the most robust models, and in particular,
in the case of the SEA-AT models.

This clearly confirms that the choice of dataset and training methodology has a significant
effect on which attack performs best.

\begin{table*}[tb]
\caption{Attacks on Cityscapes DeepLabv3 (left) and PSPNet (right) models.}
\label{tab:supp-cs}
\small
\centering
\setlength{\tabcolsep}{5pt}
\resizebox{.49\textwidth}{!}{%
\begin{tabular}{lrrrrrrr}
	  & \rotatebox[origin=r]{90}{\footnotesize Normal} 
	  & \rotatebox[origin=r]{90}{\footnotesize DDC-AT} 
	  & \rotatebox[origin=r]{90}{\footnotesize PGD-AT} 
	  & \rotatebox[origin=r]{90}{\footnotesize SegPGD-AT} 
	  & \rotatebox[origin=r]{90}{\footnotesize PGD-AT-100} 
	  & \rotatebox[origin=r]{90}{\footnotesize SegPGD-AT-100} \\

Accuracy (\%)\\\hline
\rowcolor[gray]{0.85}
clean   & 93.08 & 92.78 & 92.62 & 91.83 & 84.69 & 84.04\\
ALMAProx   & 0.83 & 0.87 & 0.96 & 0.87 & 82.38 & 82.05\\
DAG-0.001   & 15.69 & 6.18 & 1.80 & 7.58 & 65.86 & 72.12\\
\rowcolor[gray]{0.85}
DAG-0.003   & 0.00 & 0.63 & 0.30 & 0.53 & 67.91 & 75.46\\
\rowcolor[gray]{0.85}
PDPGD   & 8.97 & 83.37 & 65.56 & 53.66 & 84.72 & 84.07\\
SEA-JSD   & 1.06 & 37.19 & 23.56 & 17.30 & 84.72 & 84.07\\
\end{tabular}
}~%
\resizebox{.49\textwidth}{!}{%
\begin{tabular}{lrrrrrrr}
	  & \rotatebox[origin=r]{90}{\footnotesize Normal} 
	  & \rotatebox[origin=r]{90}{\footnotesize DDC-AT} 
	  & \rotatebox[origin=r]{90}{\footnotesize PGD-AT} 
	  & \rotatebox[origin=r]{90}{\footnotesize SegPGD-AT} 
	  & \rotatebox[origin=r]{90}{\footnotesize PGD-AT-100} 
	  & \rotatebox[origin=r]{90}{\footnotesize SegPGD-AT-100} \\

Accuracy (\%)\\\hline
\rowcolor[gray]{0.85}
clean   & 93.09 & 92.80 & 92.52 & 91.77 & 83.83 & 83.32\\
ALMAProx   & 0.85 & 0.95 & 1.18 & 0.89 & 81.72 & 81.38\\
DAG-0.001   & 15.00 & 7.48 & 5.94 & 10.85 & 68.47 & 72.40\\
\rowcolor[gray]{0.85}
DAG-0.003   & 0.00 & 1.04 & 1.25 & 2.52 & 70.81 & 76.22\\
\rowcolor[gray]{0.85}
PDPGD   & 10.31 & 71.53 & 78.36 & 50.51 & 83.86 & 83.35\\
SEA-JSD   & 1.41 & 23.02 & 38.96 & 22.15 & 83.86 & 83.35\\
\end{tabular}
}
\end{table*}

\begin{table*}[tb]
\caption{Attacks on PASCAL VOC 2012 DeepLabv3 (left) and PSPNet (right) models.}
\label{tab:supp-p}
\small
\centering
\setlength{\tabcolsep}{5pt}
\resizebox{.49\textwidth}{!}{%
\begin{tabular}{lrrrrrrr}
	  & \rotatebox[origin=r]{90}{\footnotesize Normal} 
	  & \rotatebox[origin=r]{90}{\footnotesize DDC-AT} 
	  & \rotatebox[origin=r]{90}{\footnotesize PGD-AT} 
	  & \rotatebox[origin=r]{90}{\footnotesize SegPGD-AT} 
	  & \rotatebox[origin=r]{90}{\footnotesize PGD-AT-100} 
	  & \rotatebox[origin=r]{90}{\footnotesize SegPGD-AT-100} \\

Accuracy (\%)\\\hline
\rowcolor[gray]{0.85}
clean   & 91.81 & 91.45 & 89.81 & 88.23 & 80.08 & 79.37\\
ALMAProx   & 0.90 & 1.10 & 1.19 & 0.86 & 66.66 & 65.32\\
DAG-0.001   & 13.25 & 16.56 & 16.42 & 6.96 & 51.94 & 52.71\\
\rowcolor[gray]{0.85}
DAG-0.003   & 0.00 & 0.09 & 0.10 & 2.47 & 53.12 & 54.40\\
\rowcolor[gray]{0.85}
PDPGD   & 10.57 & 30.93 & 32.64 & 2.74 & 77.65 & 77.04\\
SEA-JSD   & 3.67 & 14.05 & 15.37 & 1.32 & 76.10 & 75.91\\
\end{tabular}
}~%
\resizebox{.49\textwidth}{!}{%
\begin{tabular}{lrrrrrrr}
	  & \rotatebox[origin=r]{90}{\footnotesize Normal} 
	  & \rotatebox[origin=r]{90}{\footnotesize DDC-AT} 
	  & \rotatebox[origin=r]{90}{\footnotesize PGD-AT} 
	  & \rotatebox[origin=r]{90}{\footnotesize SegPGD-AT} 
	  & \rotatebox[origin=r]{90}{\footnotesize PGD-AT-100} 
	  & \rotatebox[origin=r]{90}{\footnotesize SegPGD-AT-100} \\

Accuracy (\%)\\\hline
\rowcolor[gray]{0.85}
clean   & 91.85 & 91.42 & 91.23 & 88.19 & 79.26 & 79.12\\
ALMAProx   & 0.91 & 0.91 & 7.35 & 3.50 & 60.47 & 64.44\\
DAG-0.001   & 34.44 & 14.13 & 11.16 & 8.31 & 45.43 & 50.52\\
\rowcolor[gray]{0.85}
DAG-0.003   & 0.01 & 0.05 & 2.75 & 1.50 & 47.78 & 53.35\\
\rowcolor[gray]{0.85}
PDPGD   & 16.44 & 24.89 & 72.60 & 62.00 & 77.09 & 76.71\\
SEA-JSD   & 6.36 & 9.82 & 63.53 & 45.96 & 75.38 & 75.38\\
\end{tabular}
}
\end{table*}

\begin{table*}[tbh]
\caption{Attacks on PASCAL VOC 2012 DeepLabv3 (left) and PSPNet (right). Metrics computed ignoring the background pixels.}
\label{tab:supp-pnb}
\small
\centering
\setlength{\tabcolsep}{5pt}
\resizebox{.49\textwidth}{!}{%
\begin{tabular}{lrrrrrrr}
	  & \rotatebox[origin=r]{90}{\footnotesize Normal} 
	  & \rotatebox[origin=r]{90}{\footnotesize DDC-AT} 
	  & \rotatebox[origin=r]{90}{\footnotesize PGD-AT} 
	  & \rotatebox[origin=r]{90}{\footnotesize SegPGD-AT} 
	  & \rotatebox[origin=r]{90}{\footnotesize PGD-AT-100} 
	  & \rotatebox[origin=r]{90}{\footnotesize SegPGD-AT-100} \\

Accuracy (\%)\\\hline
\rowcolor[gray]{0.85}
clean   & 87.84 & 86.71 & 85.58 & 67.87 & 39.81 & 39.68\\
ALMAProx   & 0.77 & 0.77 & 0.89 & 3.41 & 25.51 & 24.88\\
DAG-0.001   & 17.10 & 25.53 & 24.75 & 2.04 & 11.42 & 12.16\\
\rowcolor[gray]{0.85}
DAG-0.003   & 0.00 & 0.08 & 0.26 & 0.34 & 17.10 & 17.83\\
\rowcolor[gray]{0.85}
PDPGD   & 12.56 & 28.84 & 28.38 & 10.85 & 30.52 & 30.00\\
SEA-JSD   & 4.74 & 14.24 & 13.90 & 5.21 & 30.02 & 29.67\\
\end{tabular}
}~%
\resizebox{.49\textwidth}{!}{%
\begin{tabular}{lrrrrrrr}
	  & \rotatebox[origin=r]{90}{\footnotesize Normal} 
	  & \rotatebox[origin=r]{90}{\footnotesize DDC-AT} 
	  & \rotatebox[origin=r]{90}{\footnotesize PGD-AT} 
	  & \rotatebox[origin=r]{90}{\footnotesize SegPGD-AT} 
	  & \rotatebox[origin=r]{90}{\footnotesize PGD-AT-100} 
	  & \rotatebox[origin=r]{90}{\footnotesize SegPGD-AT-100} \\

Accuracy (\%)\\\hline
\rowcolor[gray]{0.85}
clean   & 87.63 & 86.98 & 86.71 & 67.34 & 40.59 & 38.46\\
ALMAProx   & 0.45 & 0.68 & 9.19 & 3.26 & 25.01 & 23.93\\
DAG-0.001   & 5.17 & 24.77 & 5.37 & 3.83 & 11.76 & 12.16\\
\rowcolor[gray]{0.85}
DAG-0.003   & 0.00 & 0.11 & 0.78 & 0.69 & 17.12 & 17.76\\
\rowcolor[gray]{0.85}
PDPGD   & 14.00 & 25.73 & 46.39 & 39.52 & 31.94 & 29.20\\
SEA-JSD   & 5.16 & 11.58 & 40.73 & 27.95 & 31.50 & 28.73\\
\end{tabular}
}
\end{table*}

\begin{table*}[tbh]
\caption{Attacks on the SEA-AT model family on PASCAL VOC 2012 evaluated with (left) and without (right) the background pixels.}
\label{tab:supp-sea}
\centering
\setlength{\tabcolsep}{5pt}
\resizebox{.49\textwidth}{!}{%
\begin{tabular}{lrrrrrr}
	  & \rotatebox[origin=r]{90}{\footnotesize Normal-Tiny} 
	  & \rotatebox[origin=r]{90}{\footnotesize SEA-AT-Tiny} 
	  & \rotatebox[origin=r]{90}{\footnotesize Normal-Small} 
	  & \rotatebox[origin=r]{90}{\footnotesize SEA-AT-Small} \\
Accuracy (\%)\\\hline
\rowcolor[gray]{0.85}
clean   & 93.10 & 92.75 & 93.78 & 93.10 \\
ALMAProx   & 0.93 & 90.63 & 0.93 & 90.66 \\
DAG-0.001   & 8.87 & 80.82 & 7.99 & 81.30 \\
\rowcolor[gray]{0.85}
DAG-0.003   & 0.10 & 80.76 & 0.10 & 79.99 \\
\rowcolor[gray]{0.85}
PDPGD   & 41.49 & 92.37 & 50.90 & 92.70 \\
SEA-JSD   & 15.73 & 92.37 & 19.76 & 92.68 \\
\end{tabular}
}
\resizebox{.49\textwidth}{!}{%
\begin{tabular}{lrrrrrr}
	  & \rotatebox[origin=r]{90}{\footnotesize Normal-Tiny} 
	  & \rotatebox[origin=r]{90}{\footnotesize SEA-AT-Tiny} 
	  & \rotatebox[origin=r]{90}{\footnotesize Normal-Small} 
	  & \rotatebox[origin=r]{90}{\footnotesize SEA-AT-Small} \\
Accuracy (\%)\\\hline
\rowcolor[gray]{0.85}
clean   & 88.21 & 86.39 & 90.12 & 88.53 \\
ALMAProx   & 0.65 & 81.73 & 0.77 & 83.46 \\
DAG-0.001   & 14.89 & 68.39 & 8.16 & 71.84 \\
\rowcolor[gray]{0.85}
DAG-0.003   & 0.23 & 63.91 & 0.24 & 66.82 \\
\rowcolor[gray]{0.85}
PDPGD   & 38.21 & 83.70 & 47.21 & 85.74 \\
SEA-JSD   & 15.34 & 83.69 & 20.79 & 85.64 \\
\end{tabular}
}
\end{table*}

\begin{figure*}
\setlength{\tabcolsep}{1pt}
\centering
\tiny
\begin{tabular}{ccccc}
& Accuracy, DeepLabv3 & mIoU, DeepLabv3 & Accuracy, PSPNet & mIoU, PSPNet\\
\rotatebox[origin=l]{90}{\hspace{8mm}Normal} & 
\includegraphics[width=0.24\textwidth]{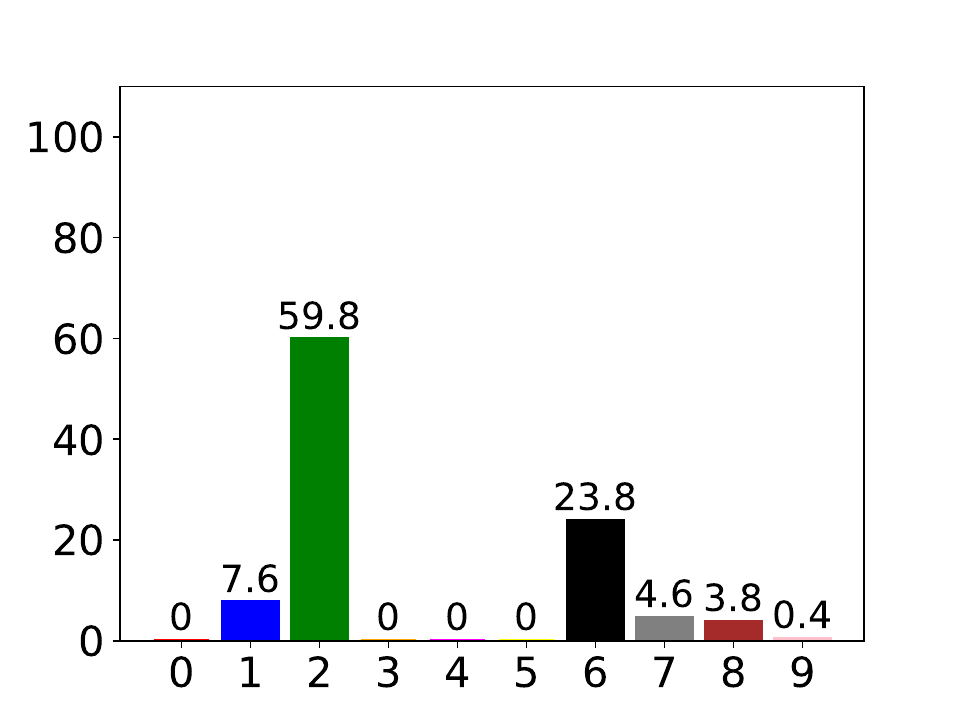} &
\includegraphics[width=0.24\textwidth]{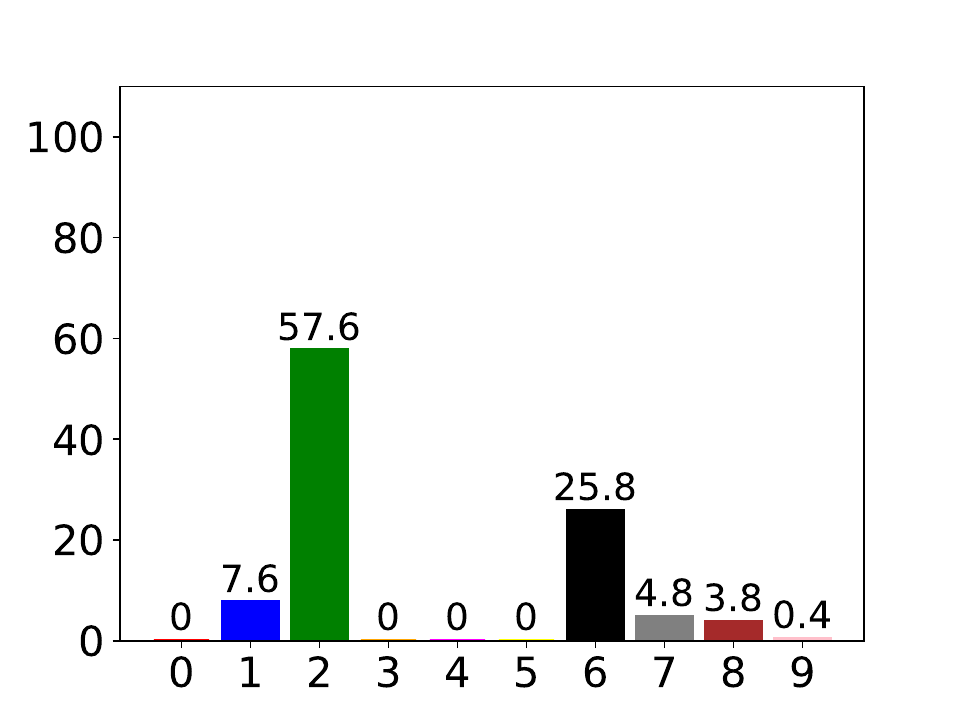} &
\includegraphics[width=0.24\textwidth]{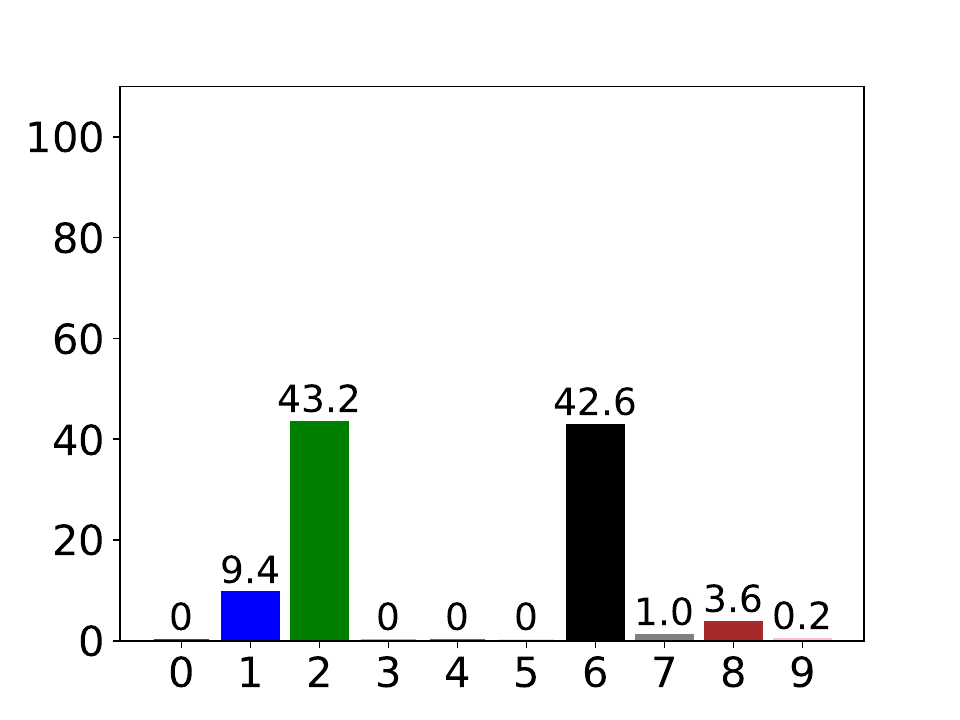} &
\includegraphics[width=0.24\textwidth]{gdrive/data_final/distribution_attacks_fix/results_city/normal_pspnet_iou1s} \\
\rotatebox[origin=l]{90}{\hspace{8mm}DDC-AT} & 
\includegraphics[width=0.24\textwidth]{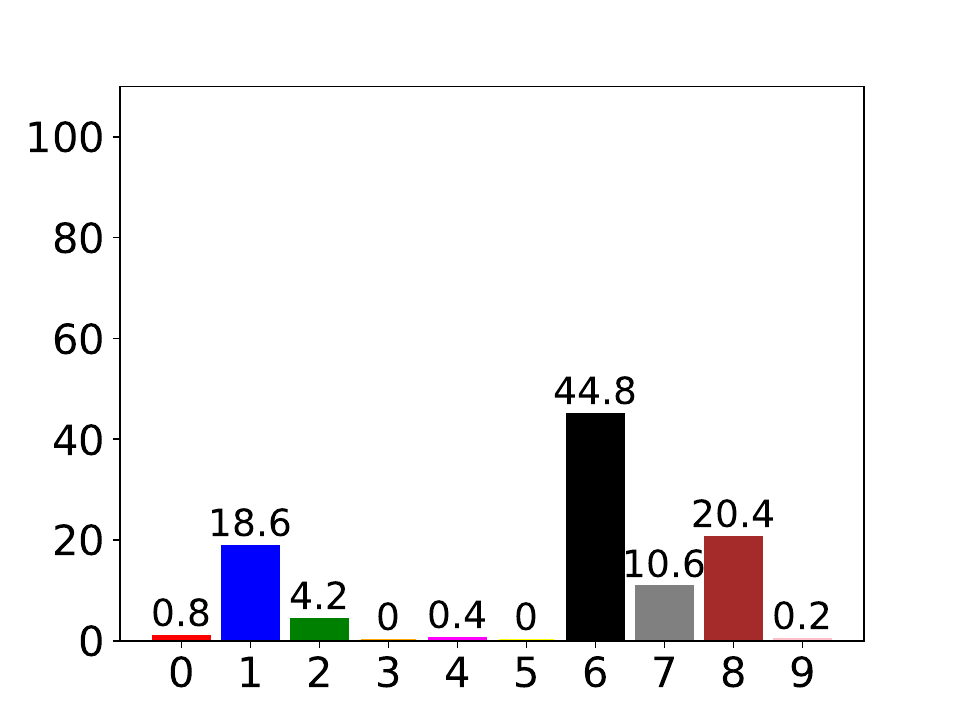} &
\includegraphics[width=0.24\textwidth]{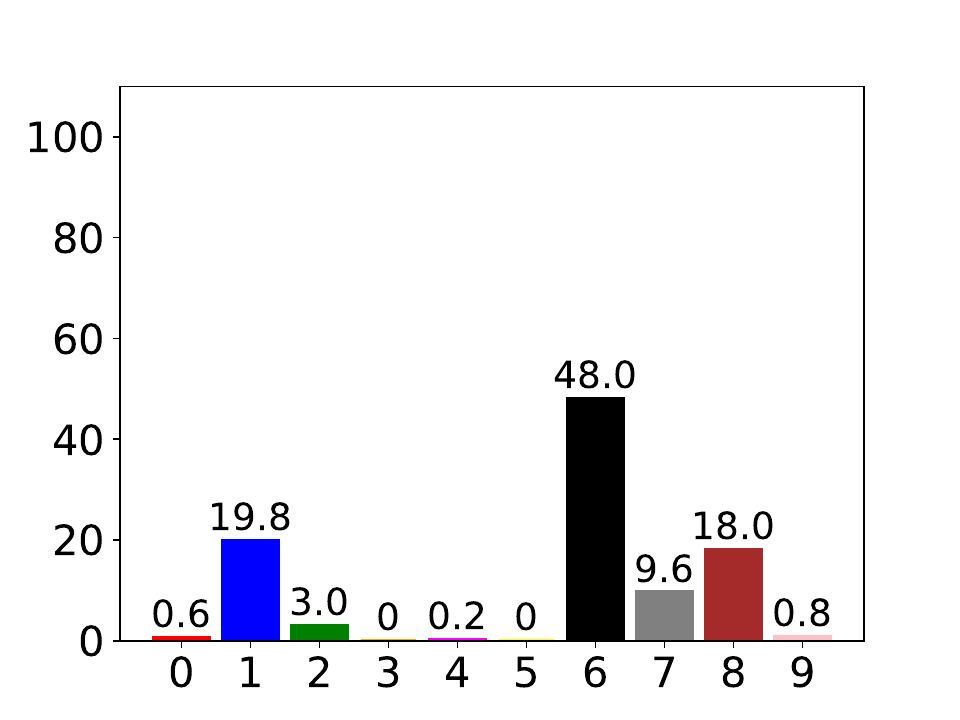} &
\includegraphics[width=0.24\textwidth]{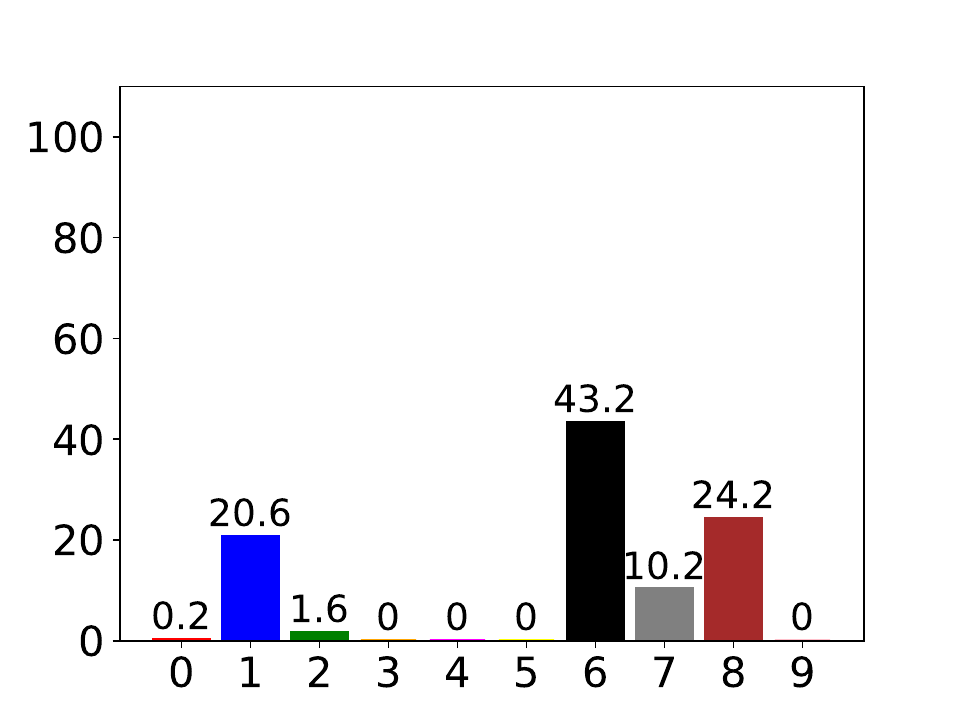} &
\includegraphics[width=0.24\textwidth]{gdrive/data_final/distribution_attacks_fix/results_city/ddcat_pspnet_iou1s} \\
\rotatebox[origin=l]{90}{\hspace{8mm}PGD-AT} & 
\includegraphics[width=0.24\textwidth]{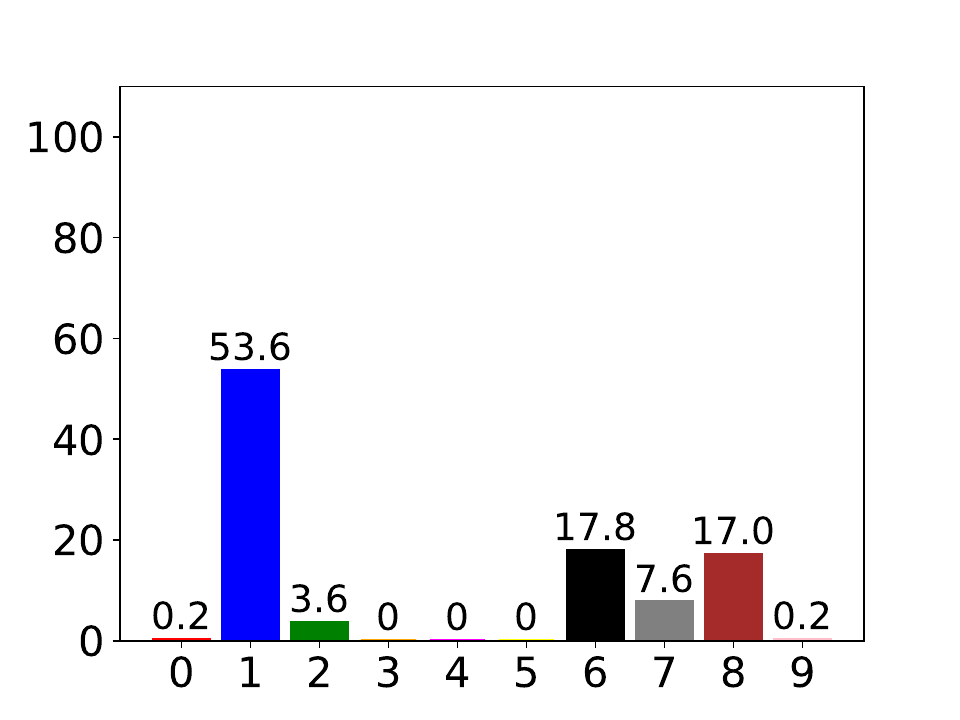} &
\includegraphics[width=0.24\textwidth]{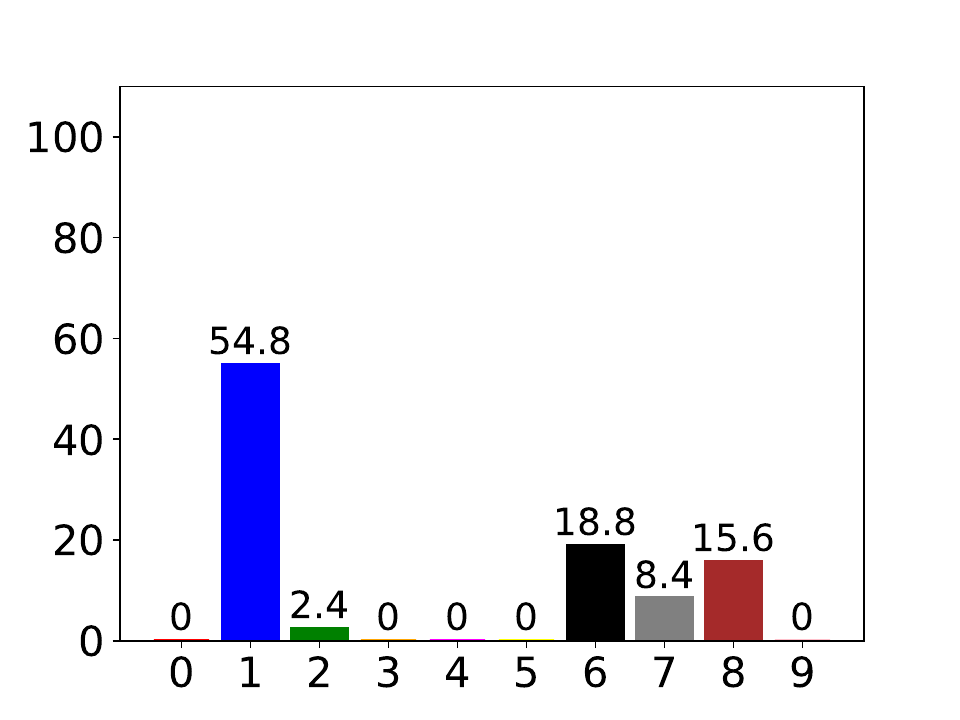} &
\includegraphics[width=0.24\textwidth]{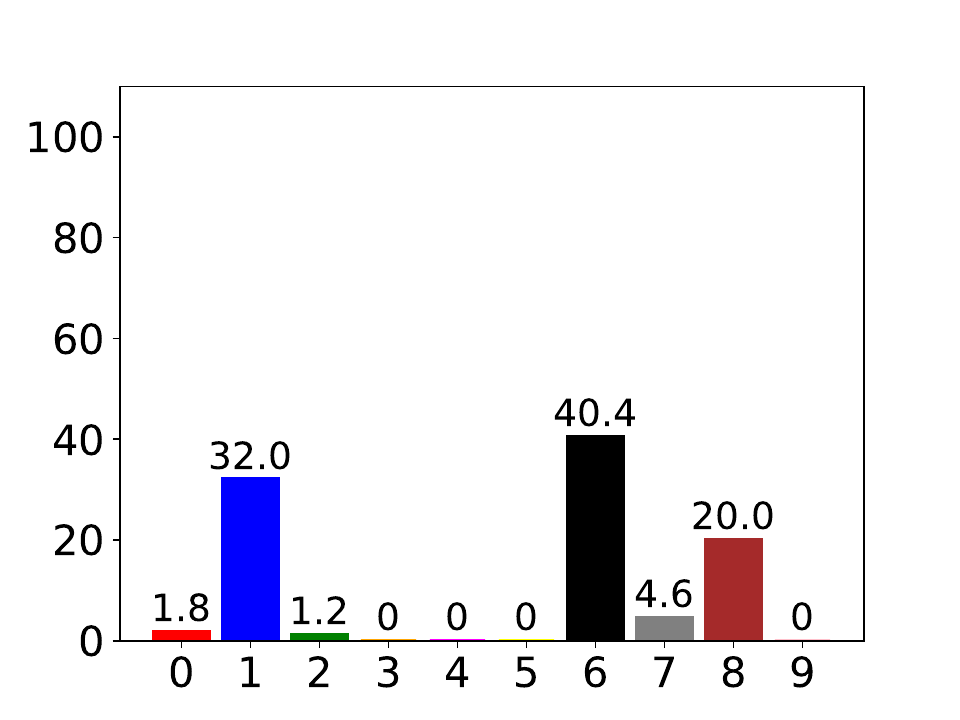} &
\includegraphics[width=0.24\textwidth]{gdrive/data_final/distribution_attacks_fix/results_city/sat_pspnet_iou1s} \\
\end{tabular}
\caption{Distributions of `best attack' selected based on Accuracy and mIoU, on Cityscapes.
The attacks are 0: ALMAProx, 1: PAdam-CE, 2: PAdam-Cos, 3: DAG-0.001, 4: DAG-0.003, 5: PDPGD, 6: SEA-JSD, 7: SEA-MCE, 8: SEA-MSL, 9: SEA-BCE.}
\label{fig:supp-cs-attackhistograms}
\end{figure*}

\begin{figure*}
\setlength{\tabcolsep}{1pt}
\centering
\tiny
\begin{tabular}{ccccc}
& Accuracy, DeepLabv3 & mIoU, DeepLabv3 & Accuracy, PSPNet & mIoU, PSPNet\\
\rotatebox[origin=l]{90}{\hspace{8mm}Normal} & 
\includegraphics[width=0.24\textwidth]{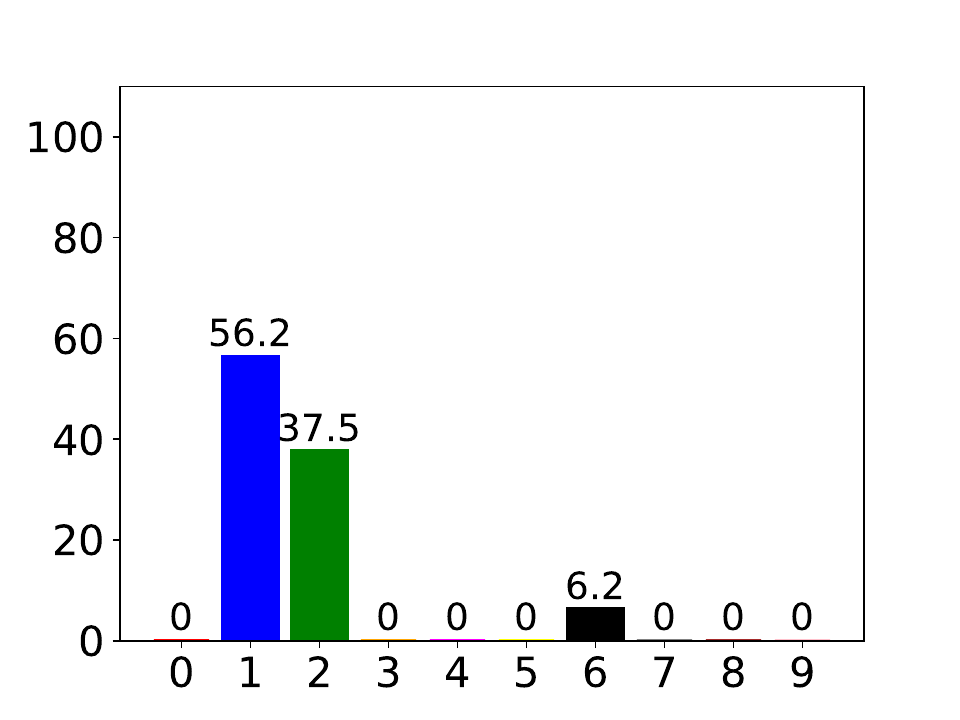} &
\includegraphics[width=0.24\textwidth]{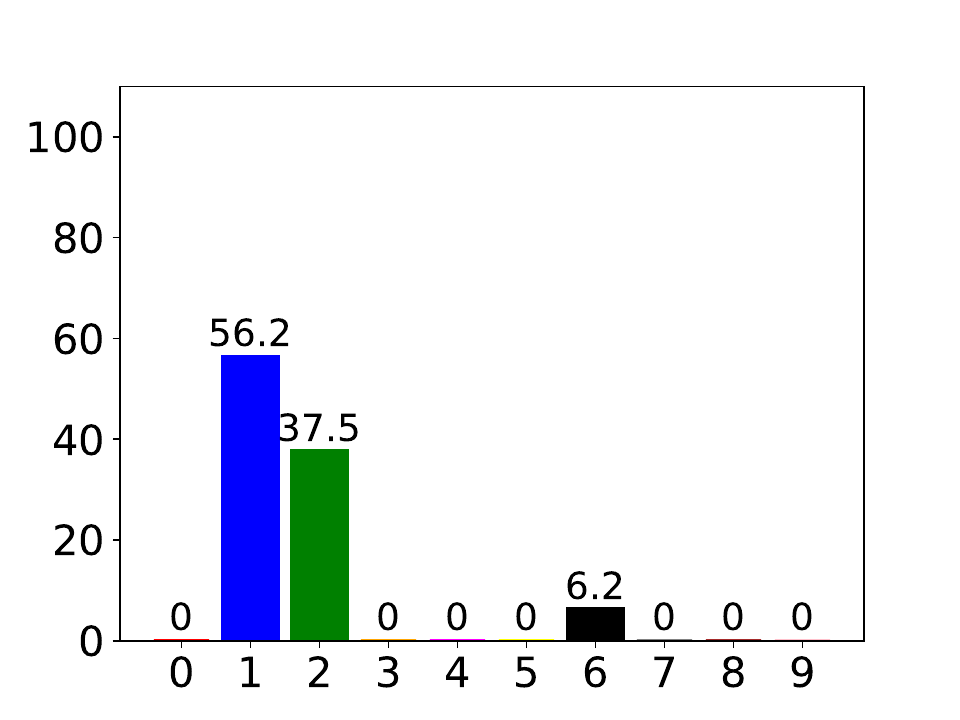} &
\includegraphics[width=0.24\textwidth]{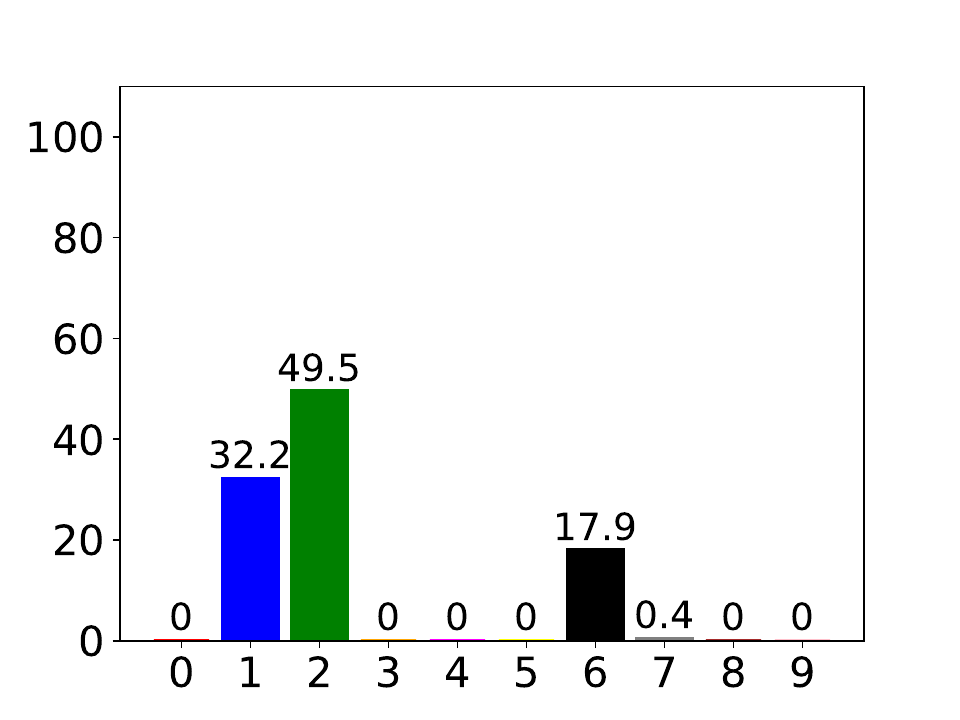} &
\includegraphics[width=0.24\textwidth]{gdrive/data_final/distribution_attacks_fix/results_pascal/normal_pspnet_iou1s} \\
\rotatebox[origin=l]{90}{\hspace{8mm}DDC-AT} & 
\includegraphics[width=0.24\textwidth]{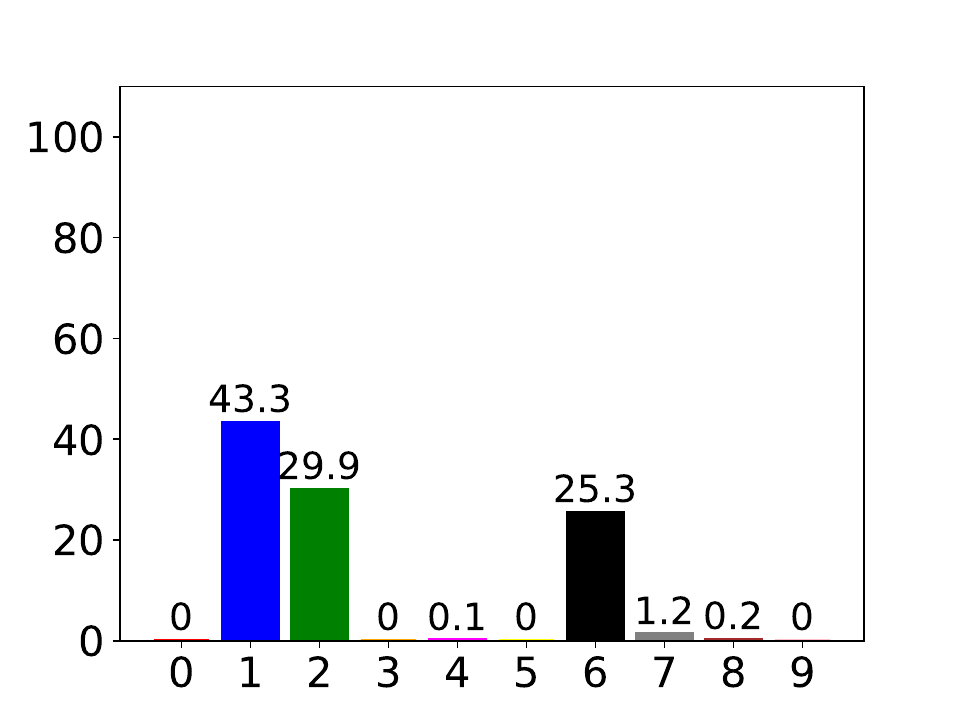} &
\includegraphics[width=0.24\textwidth]{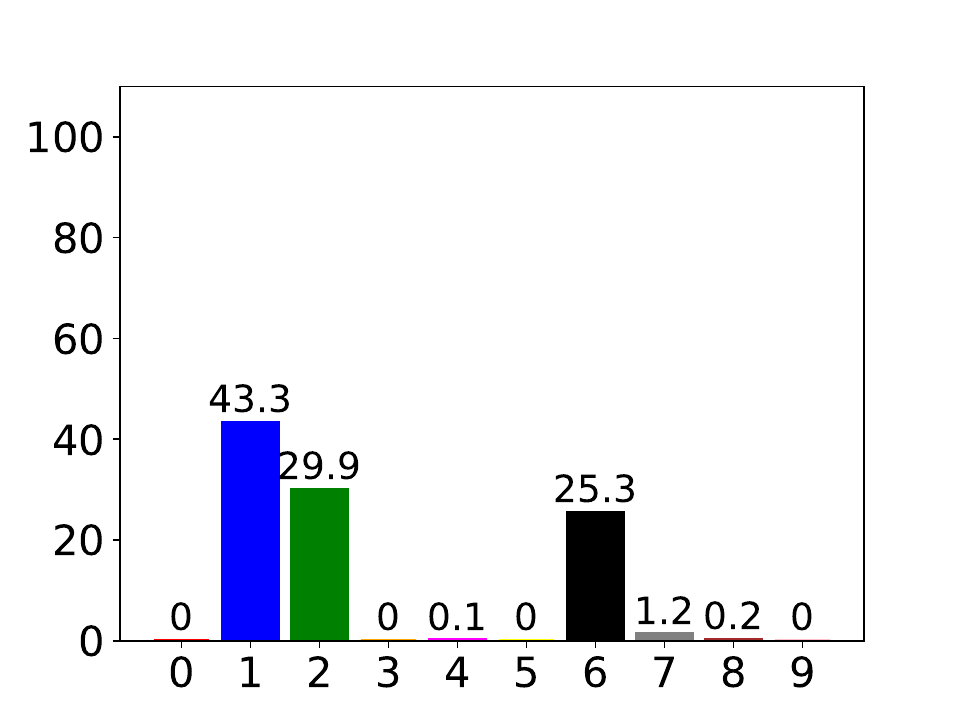} &
\includegraphics[width=0.24\textwidth]{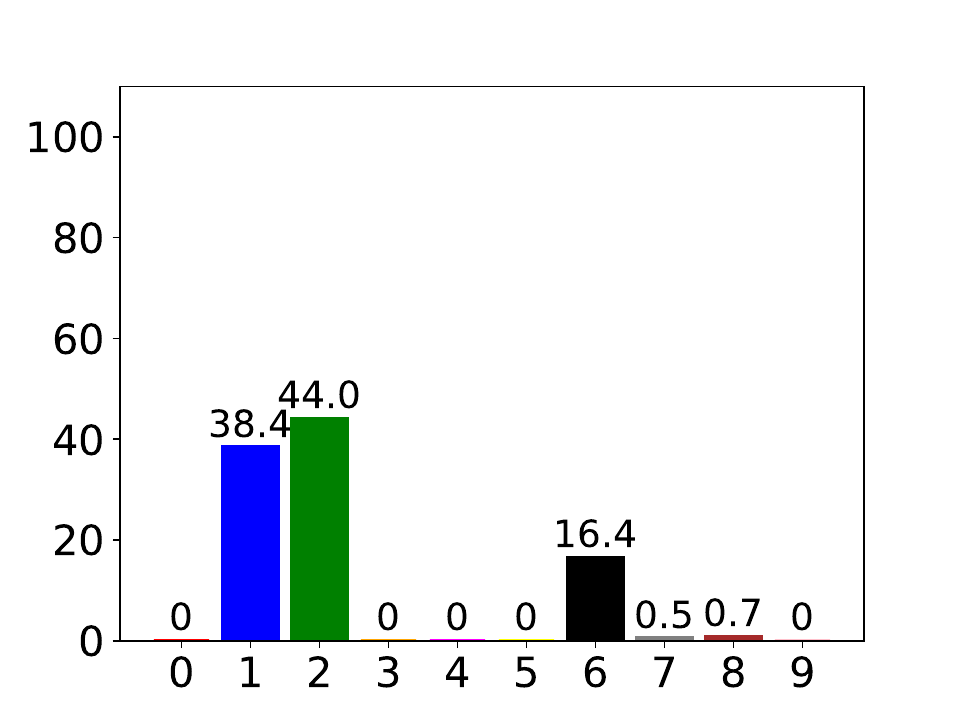} &
\includegraphics[width=0.24\textwidth]{gdrive/data_final/distribution_attacks_fix/results_pascal/ddcat_pspnet_iou1s} \\
\rotatebox[origin=l]{90}{\hspace{8mm}PGD-AT} & 
\includegraphics[width=0.24\textwidth]{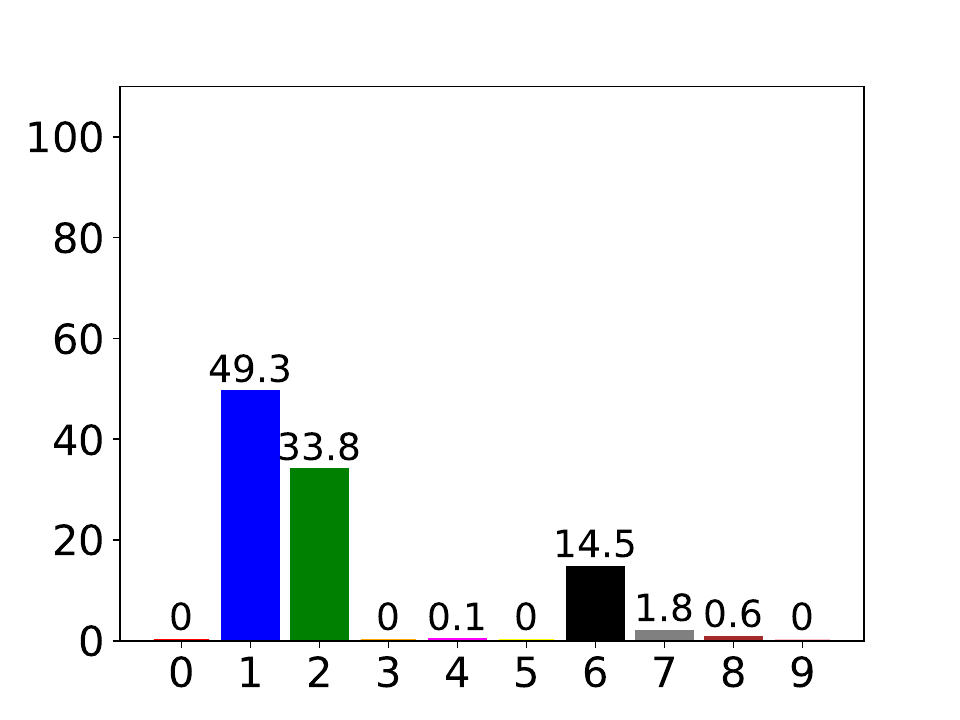} &
\includegraphics[width=0.24\textwidth]{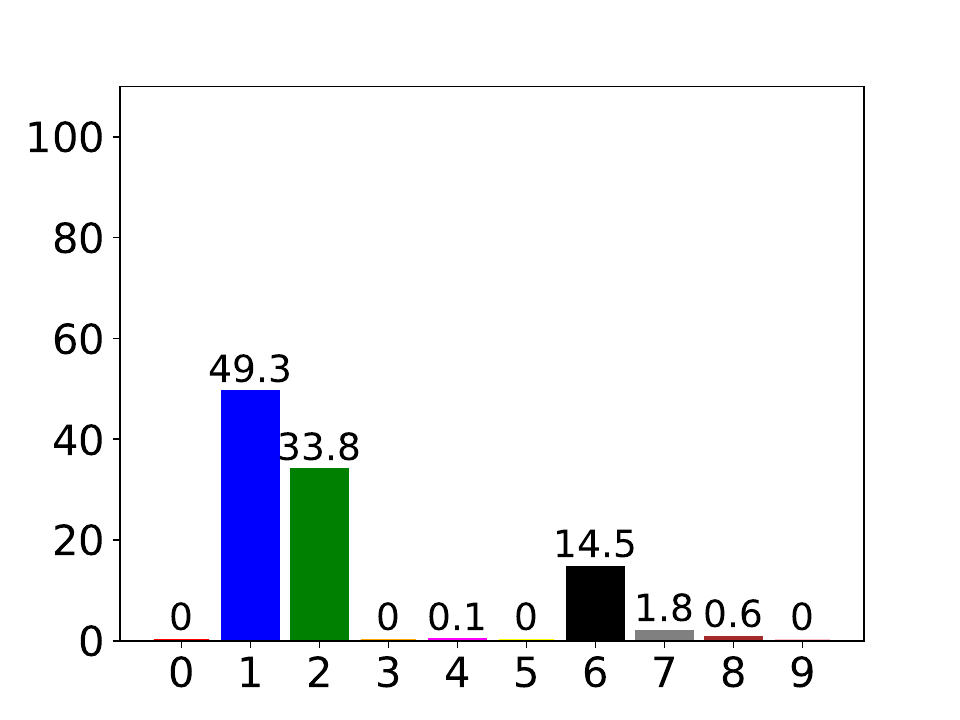} &
\includegraphics[width=0.24\textwidth]{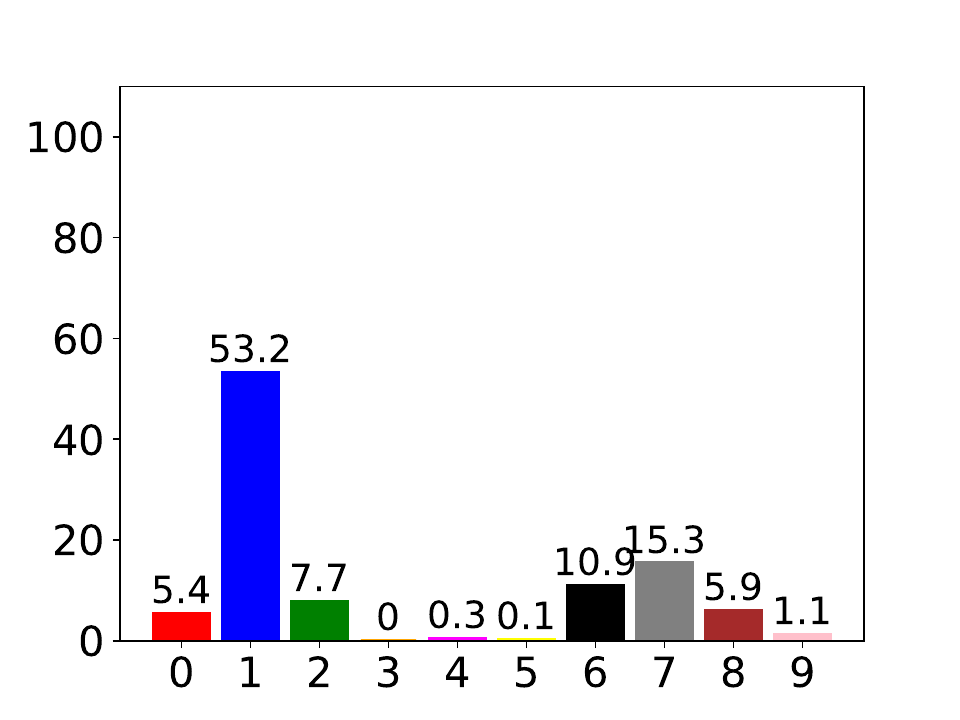} &
\includegraphics[width=0.24\textwidth]{gdrive/data_final/distribution_attacks_fix/results_pascal/sat_pspnet_iou1s} \\
\end{tabular}
\caption{Distributions of `best attack' selected based on Accuracy and mIoU, on PASCAL VOC 2012.
The attacks are 0: ALMAProx, 1: PAdam-CE, 2: PAdam-Cos, 3: DAG-0.001, 4: DAG-0.003, 5: PDPGD, 6: SEA-JSD, 7: SEA-MCE, 8: SEA-MSL, 9: SEA-BCE.}
\label{fig:supp-p-attackhistograms}
\end{figure*}

\begin{figure*}
\setlength{\tabcolsep}{1pt}
\centering
\tiny
\begin{tabular}{ccccc}
& Accuracy, Tiny & mIoU, Tiny & Accuracy, Small & mIoU, Small\\
\rotatebox[origin=l]{90}{\hspace{8mm}Normal} & 
\includegraphics[width=0.24\textwidth]{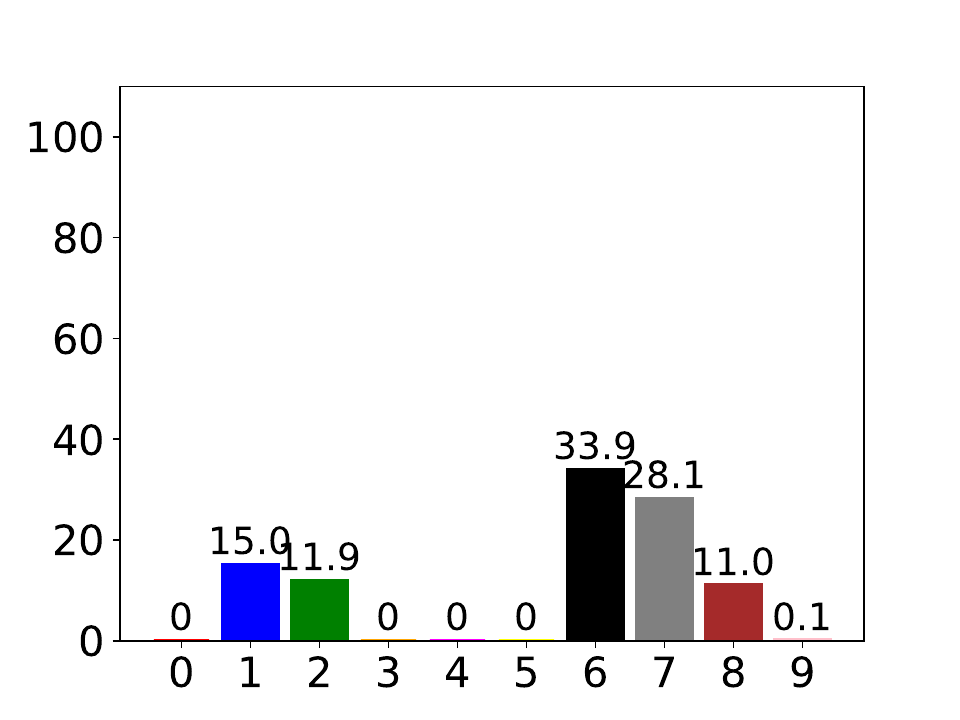} &
\includegraphics[width=0.24\textwidth]{gdrive/data_final/distribution_attacks_fix/results_pascal/ConvNeXt-T_CVST_NORMAL_iou1s} &
\includegraphics[width=0.24\textwidth]{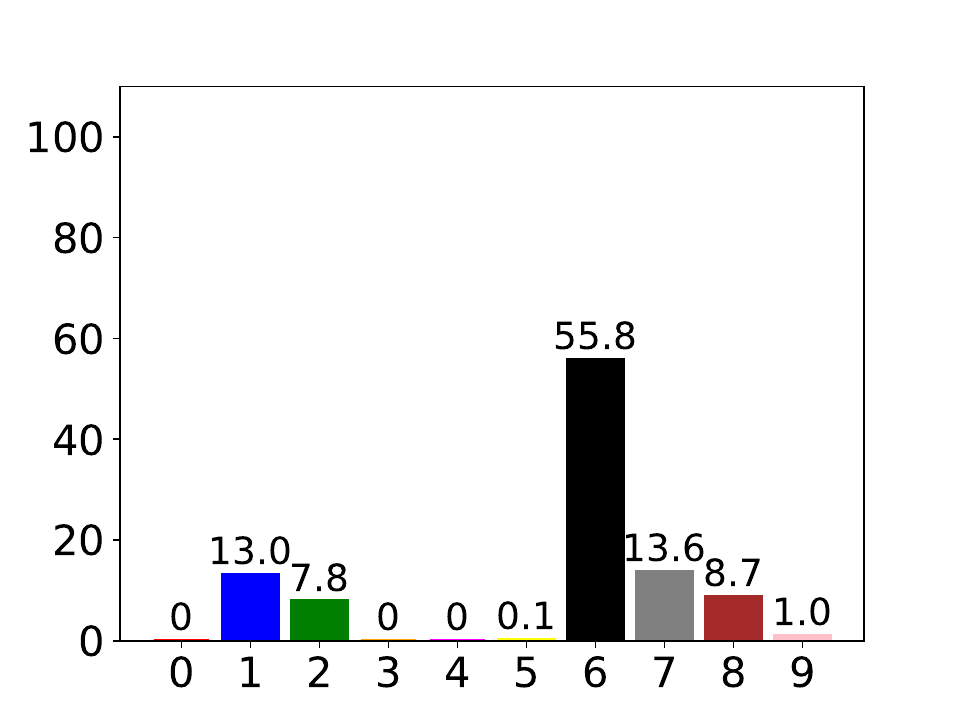} &
\includegraphics[width=0.24\textwidth]{gdrive/data_final/distribution_attacks_fix/results_pascal/ConvNeXt-S_CVST_NORMAL_iou1s} \\
\rotatebox[origin=l]{90}{\hspace{8mm}SEA-AT} & 
\includegraphics[width=0.24\textwidth]{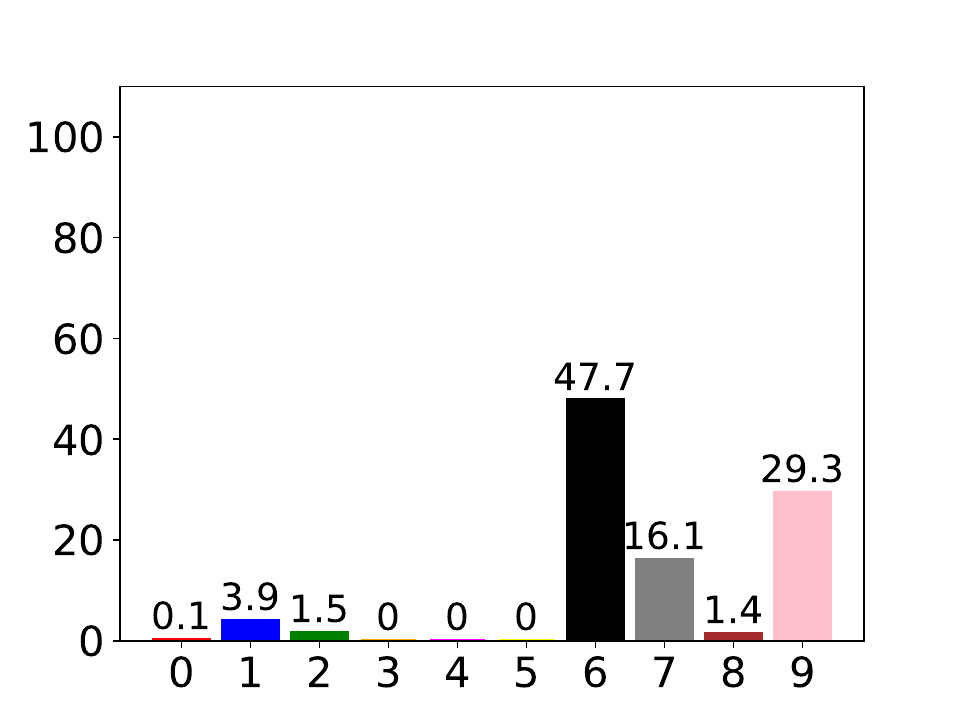} &
\includegraphics[width=0.24\textwidth]{gdrive/data_final/distribution_attacks_fix/results_pascal/ConvNeXt-T_CVST_ROB_iou1s} &
\includegraphics[width=0.24\textwidth]{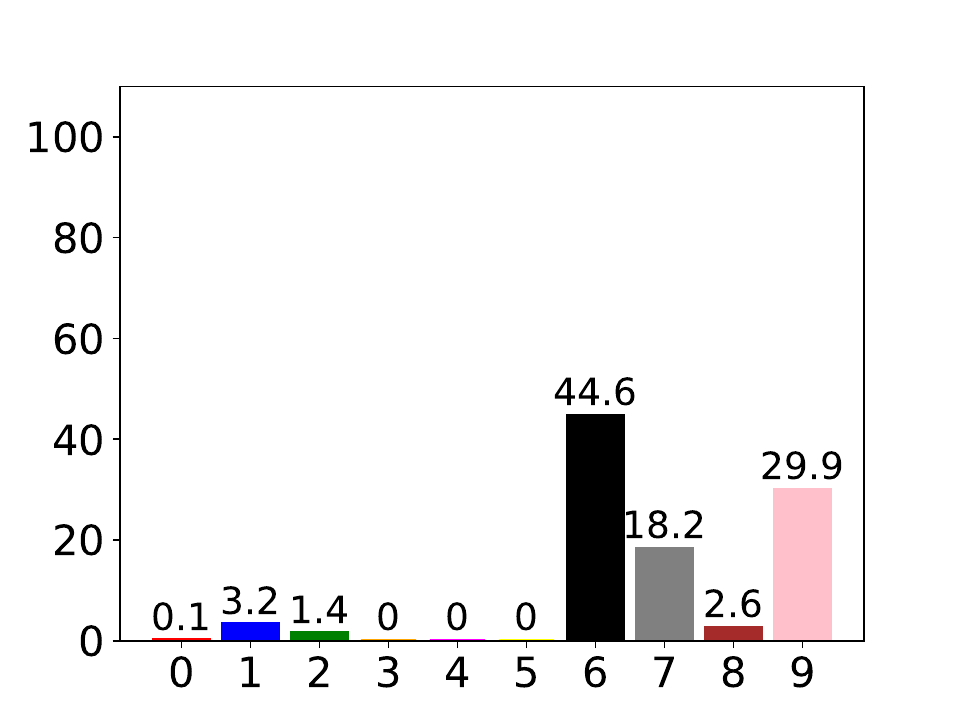} &
\includegraphics[width=0.24\textwidth]{gdrive/data_final/distribution_attacks_fix/results_pascal/ConvNeXt-S_CVST_ROB_iou1s} \\
\end{tabular}
\caption{Distributions of `best attack' selected based on Accuracy and mIoU, on PASCAL VOC 2012 for the SEA-AT model family.
The attacks are 0: ALMAProx, 1: PAdam-CE, 2: PAdam-Cos, 3: DAG-0.001, 4: DAG-0.003, 5: PDPGD, 6: SEA-JSD, 7: SEA-MCE, 8: SEA-MSL, 9: SEA-BCE.}
\label{fig:supp-p-sea-attackhistograms}
\end{figure*}

\subsection{Minimum Perturbation Attacks}
\label{sec:supp-minpert}

\begin{figure*}
\setlength{\tabcolsep}{0pt}
\centering
\tiny
\begin{tabular}{cccccc}
& ALMAProx & DAG-0.001 & DAG-0.003 & PDPGD & Aggregated\\
\rotatebox[origin=l]{90}{\hspace{0mm}DeepLabv3, 99\%} & 
\includegraphics[width=0.195\textwidth]{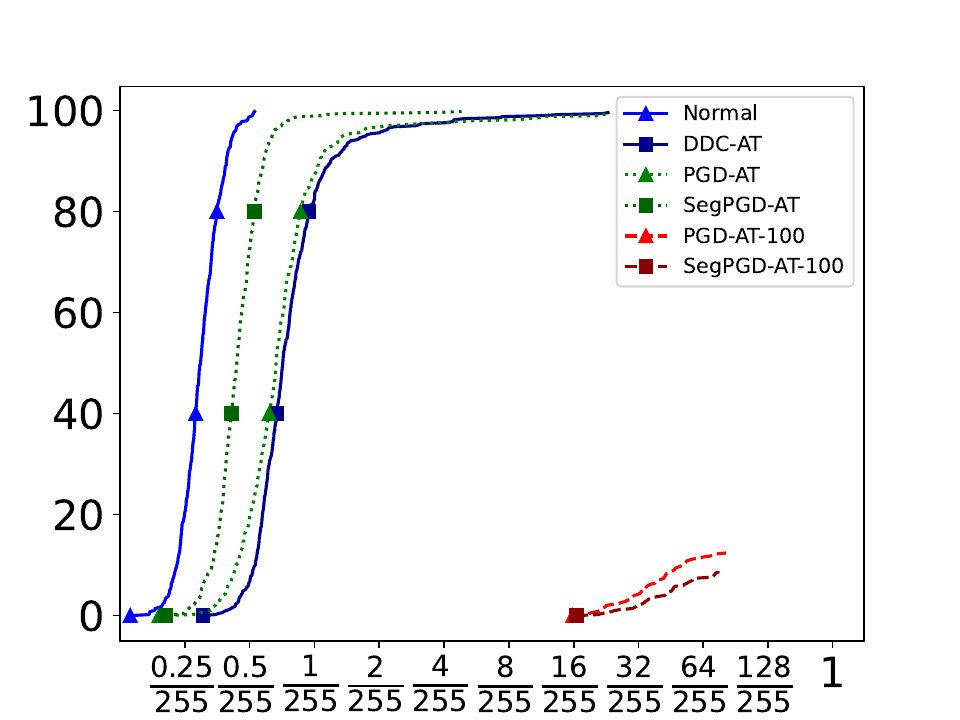} &
\includegraphics[width=0.195\textwidth]{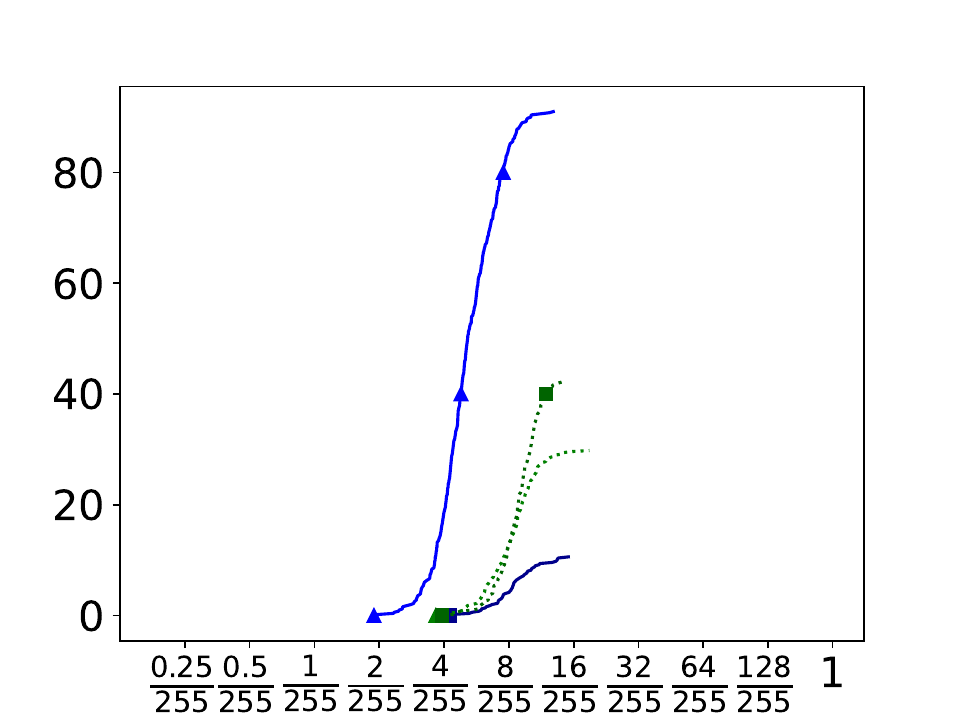} &
\includegraphics[width=0.195\textwidth]{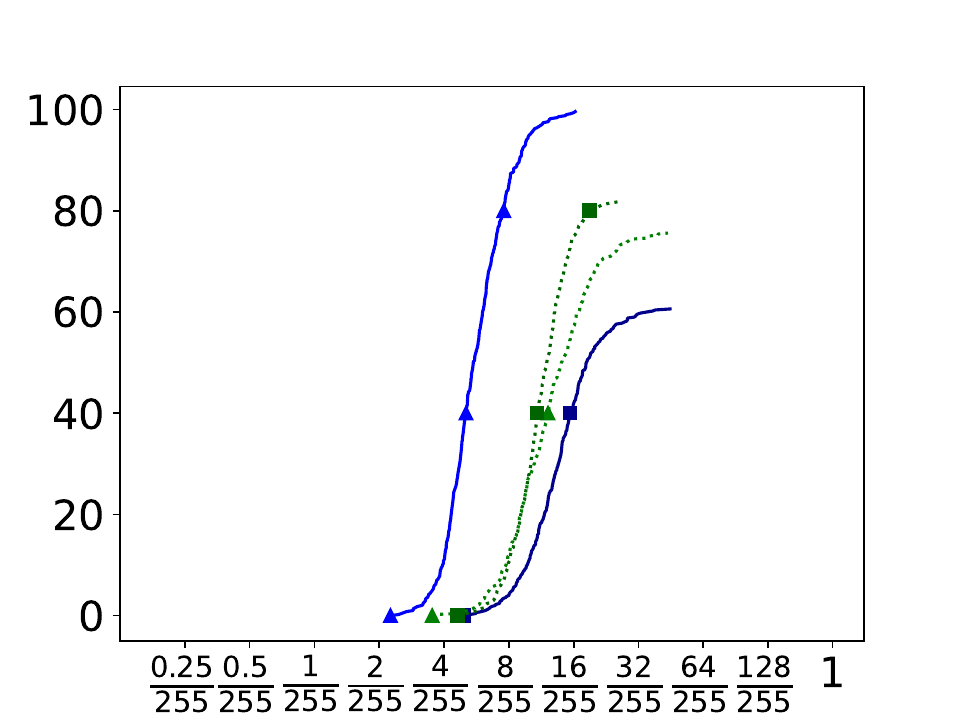} &
\includegraphics[width=0.195\textwidth]{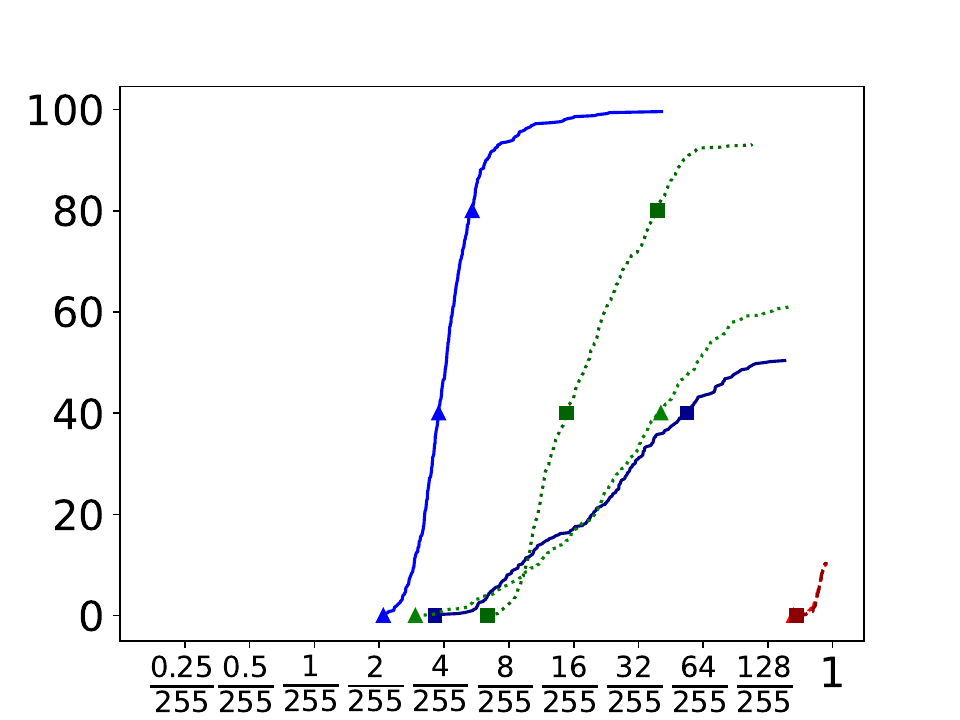} &
\includegraphics[width=0.195\textwidth]{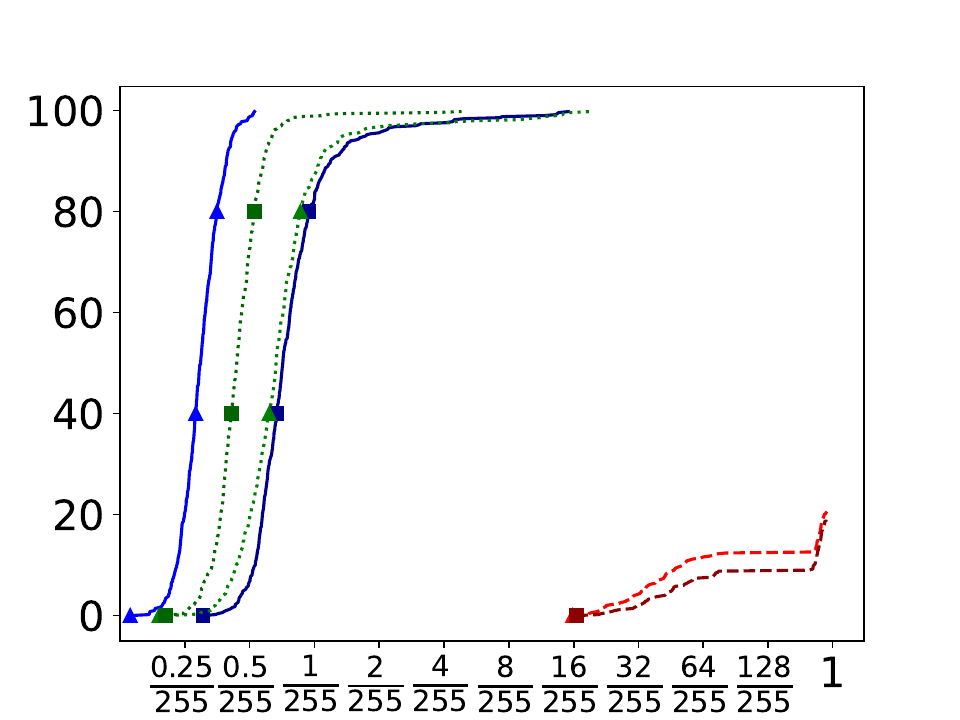} \\
\rotatebox[origin=l]{90}{\hspace{2mm}PSPNet, 99\%} & 
\includegraphics[width=0.195\textwidth]{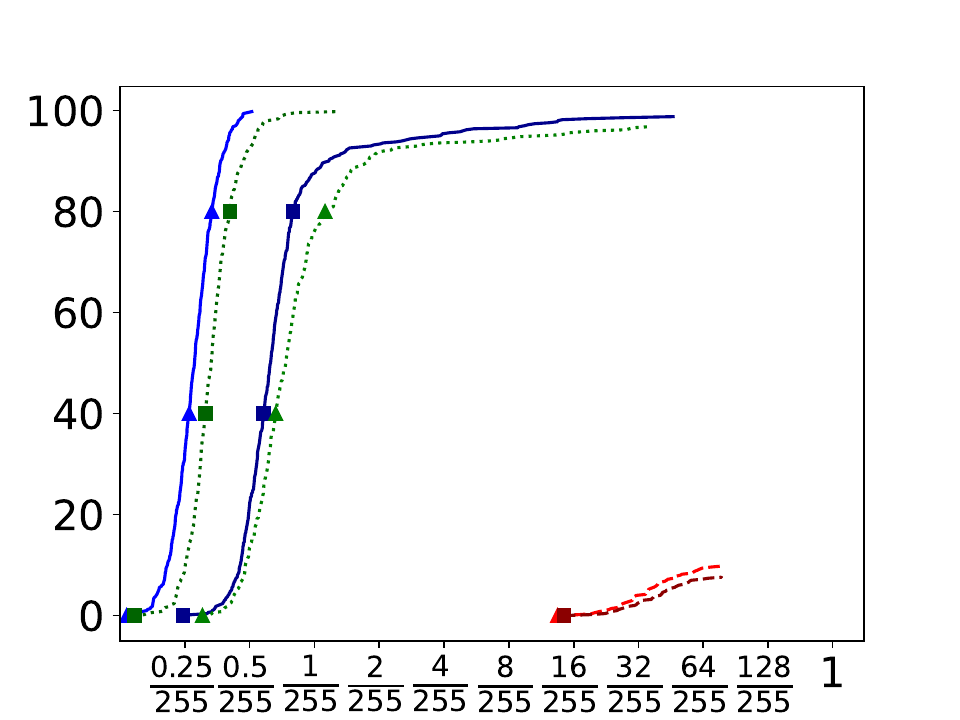} &
\includegraphics[width=0.195\textwidth]{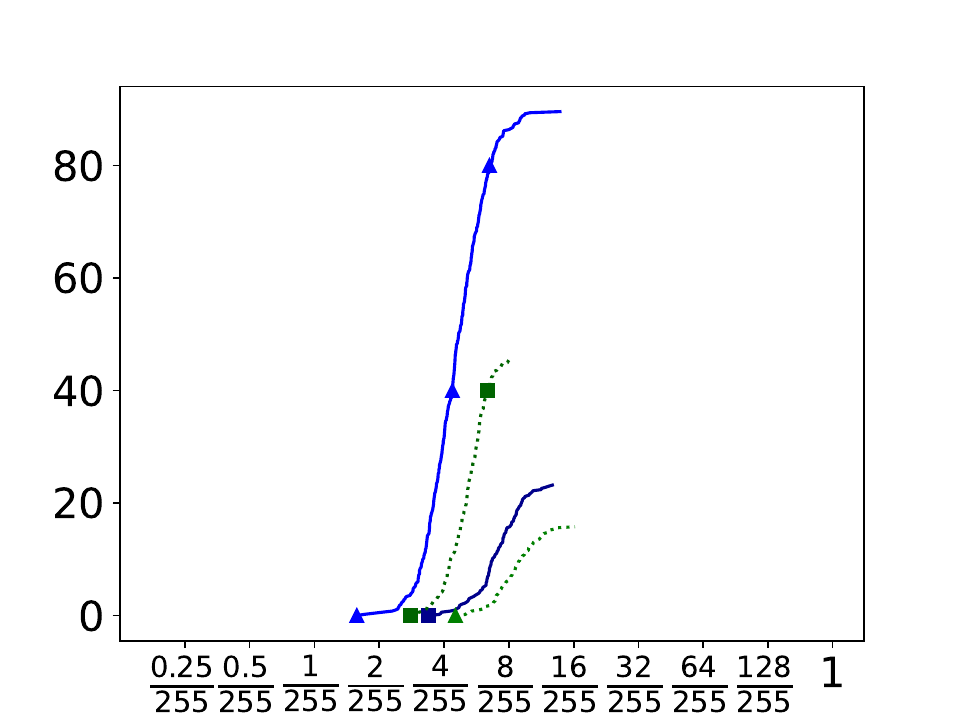} &
\includegraphics[width=0.195\textwidth]{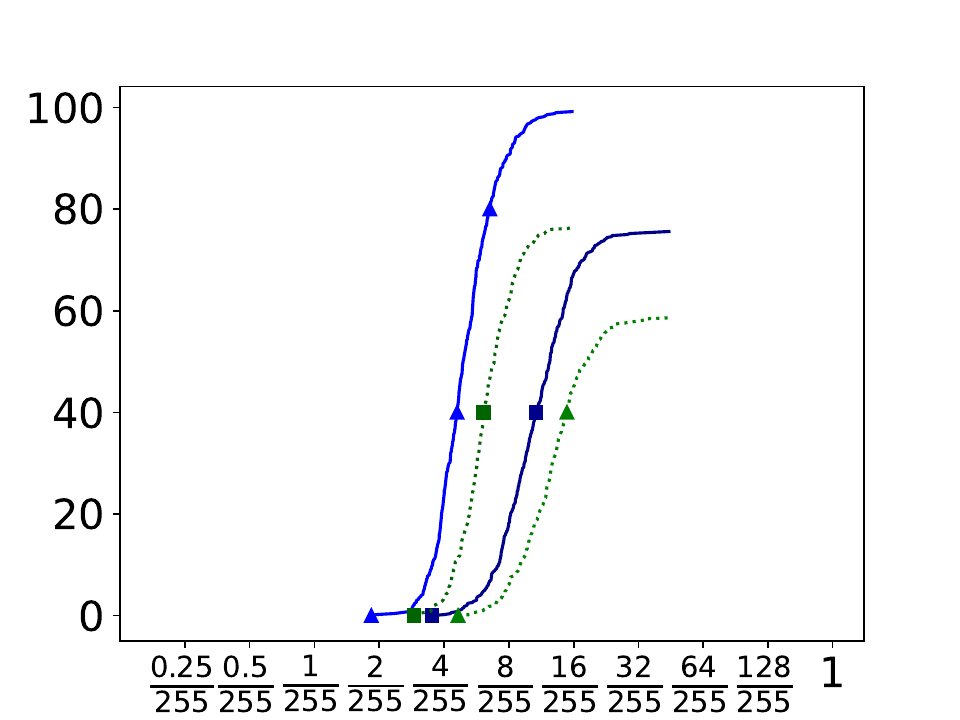} &
\includegraphics[width=0.195\textwidth]{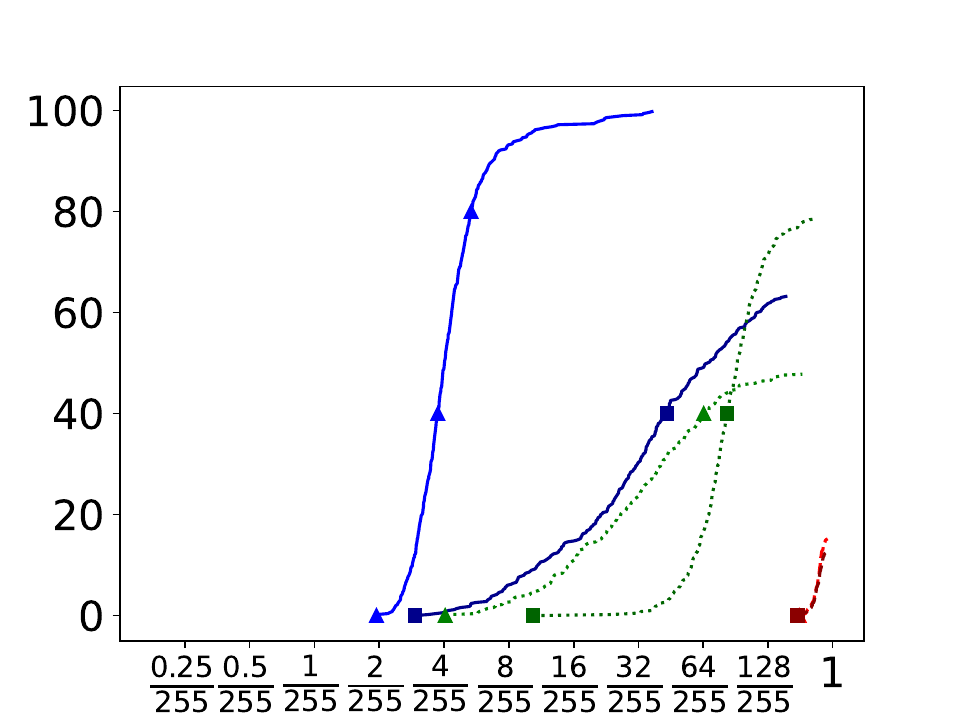} &
\includegraphics[width=0.195\textwidth]{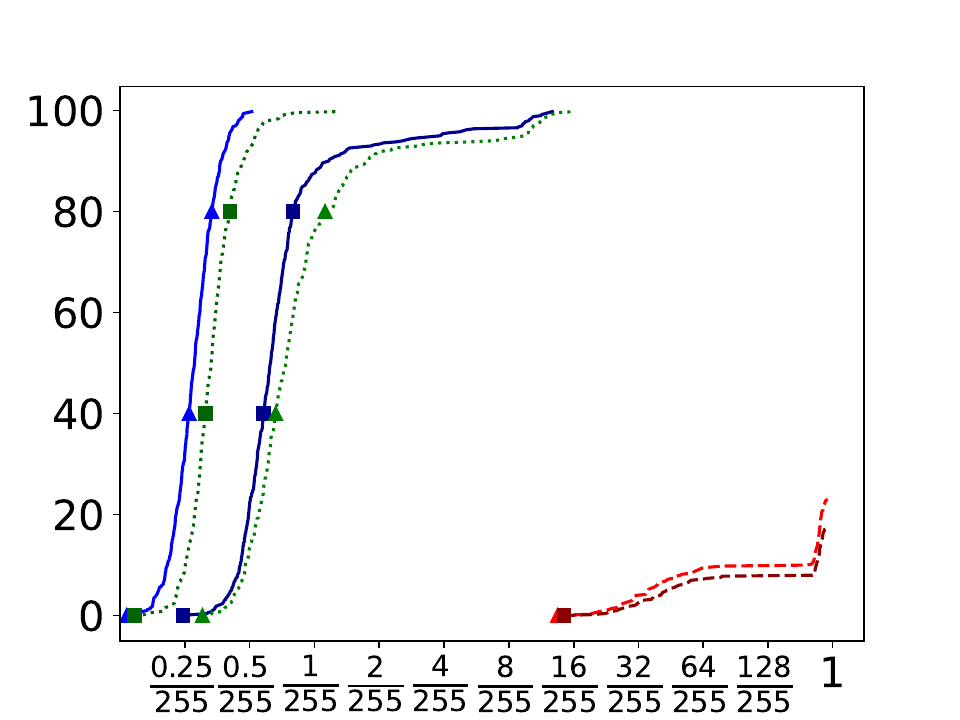} \\
\rotatebox[origin=l]{90}{\hspace{0mm}DeepLabv3, 90\%} & 
\includegraphics[width=0.195\textwidth]{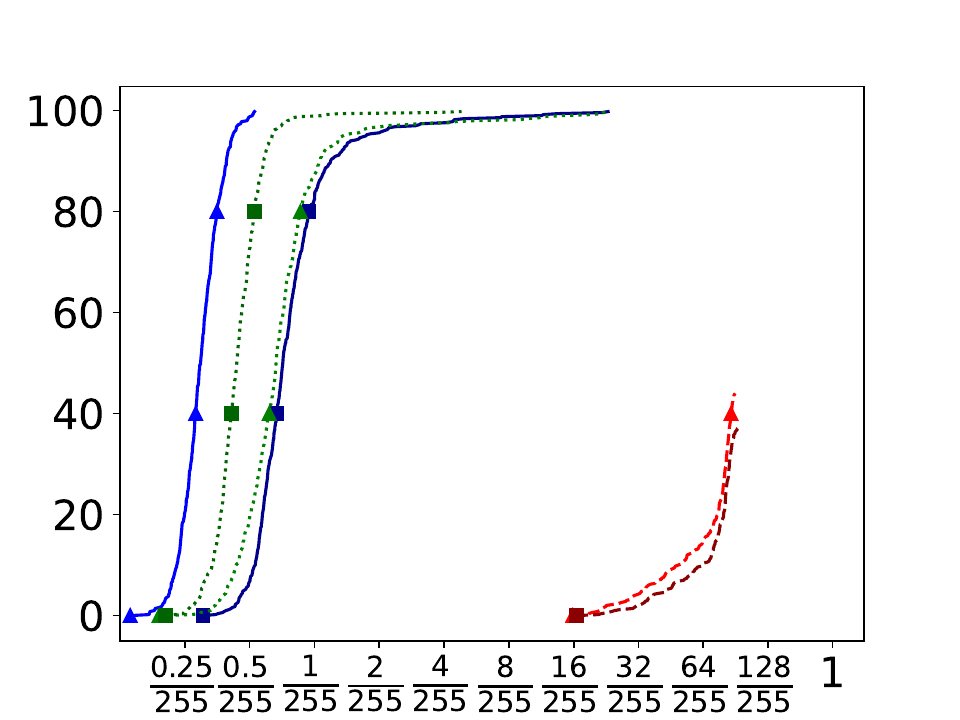} &
\includegraphics[width=0.195\textwidth]{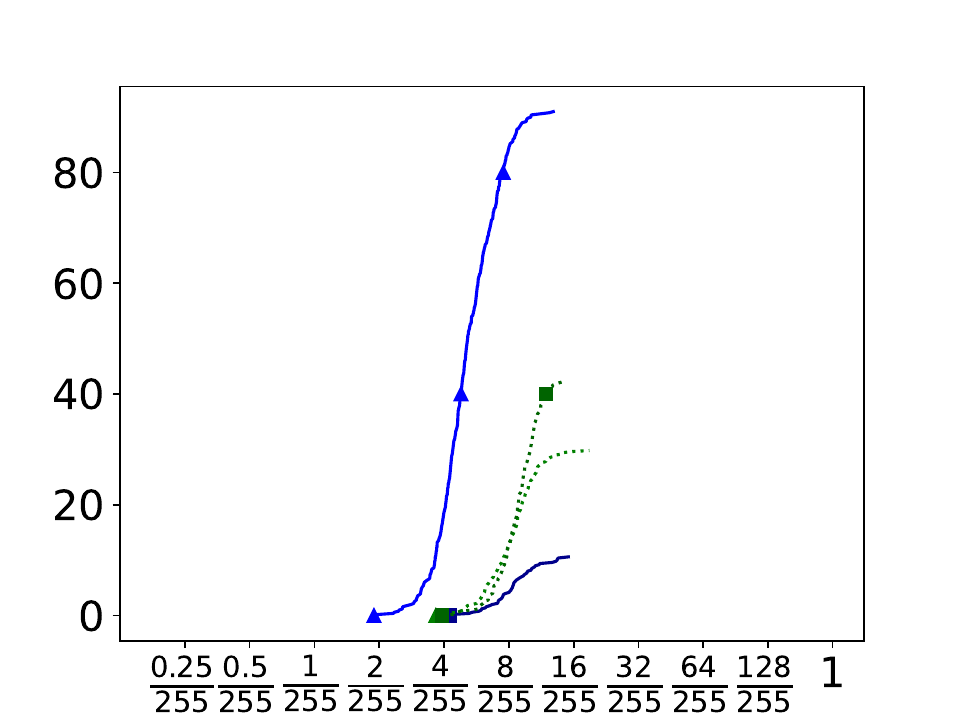} &
\includegraphics[width=0.195\textwidth]{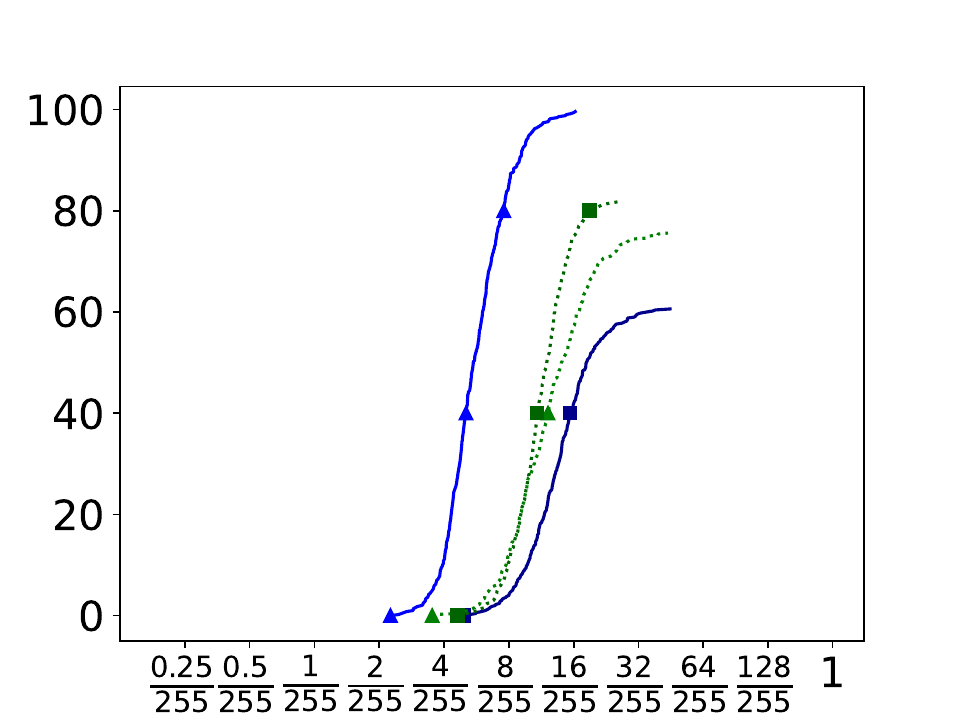} &
\includegraphics[width=0.195\textwidth]{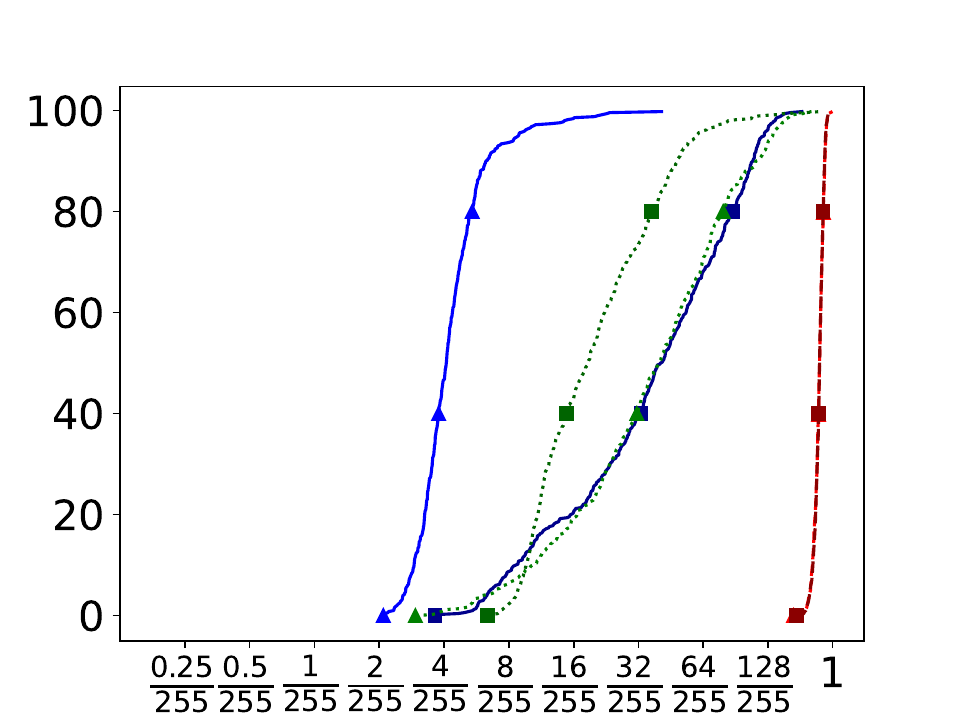} &
\includegraphics[width=0.195\textwidth]{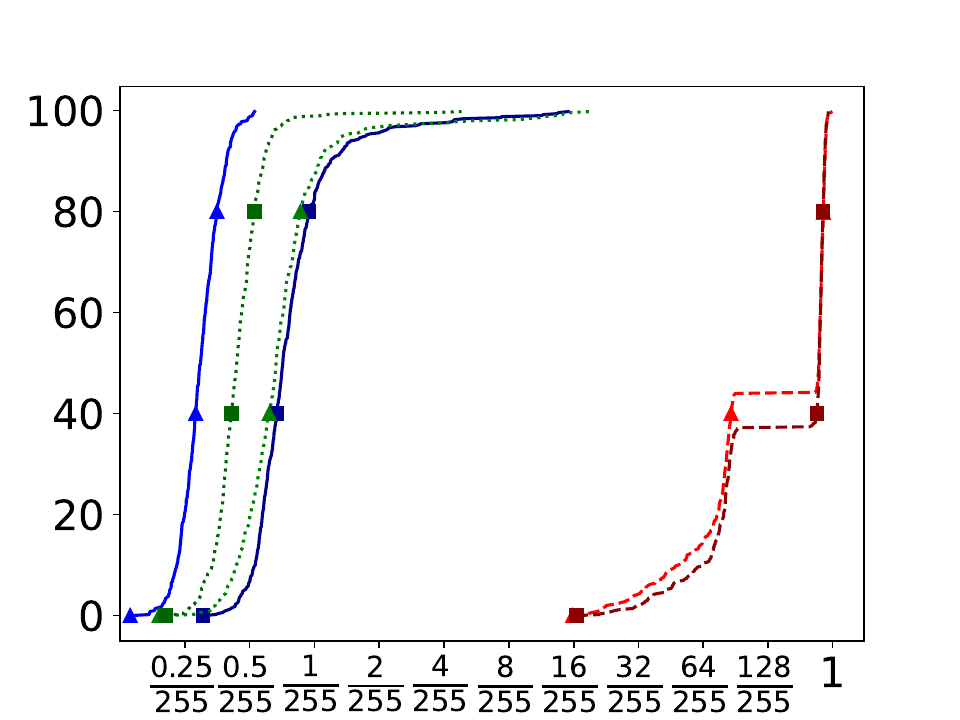} \\
\rotatebox[origin=l]{90}{\hspace{0mm}PSPNet, 90\%} & 
\includegraphics[width=0.195\textwidth]{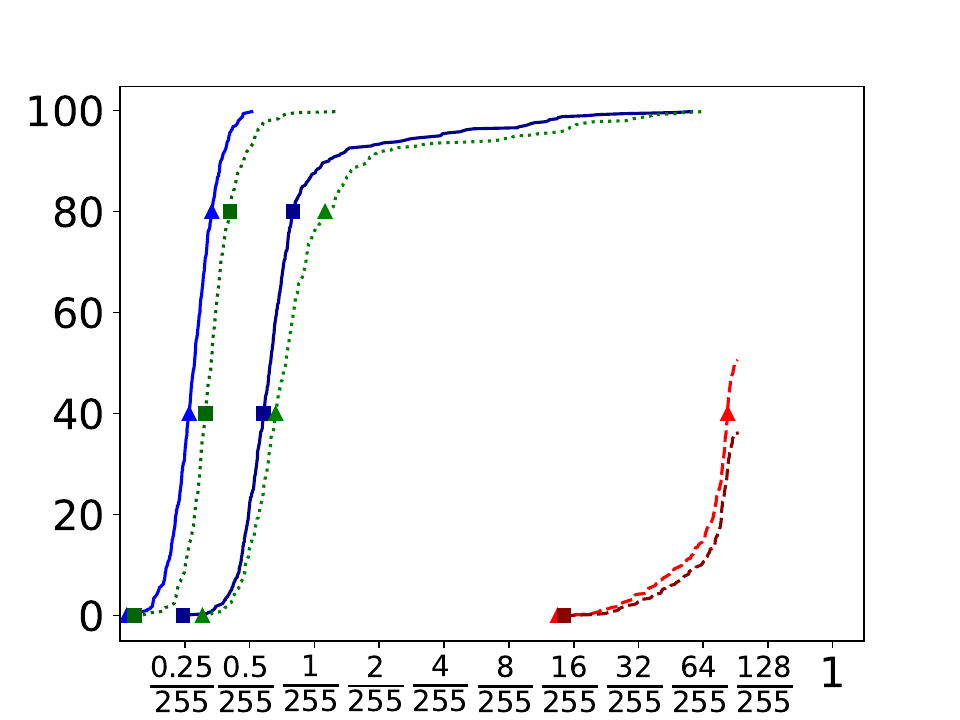} &
\includegraphics[width=0.195\textwidth]{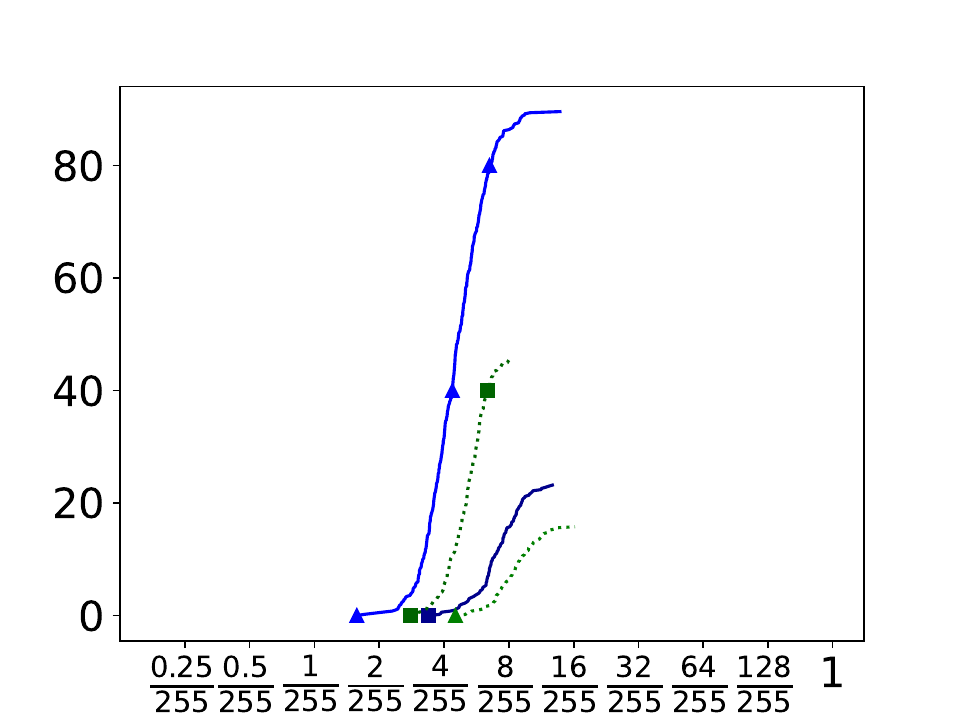} &
\includegraphics[width=0.195\textwidth]{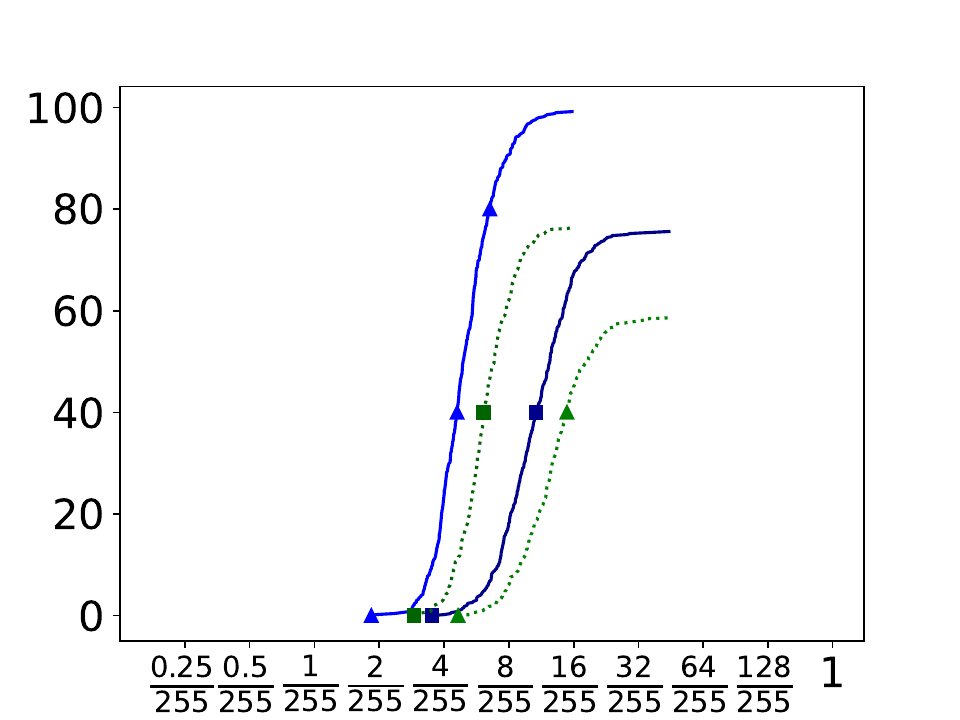} &
\includegraphics[width=0.195\textwidth]{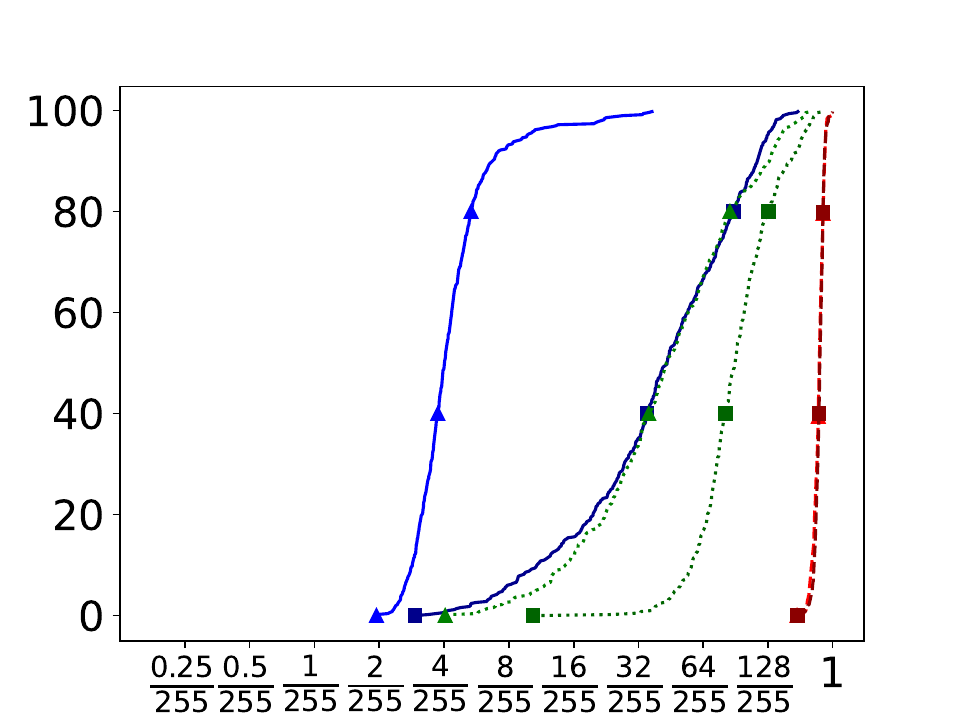} &
\includegraphics[width=0.195\textwidth]{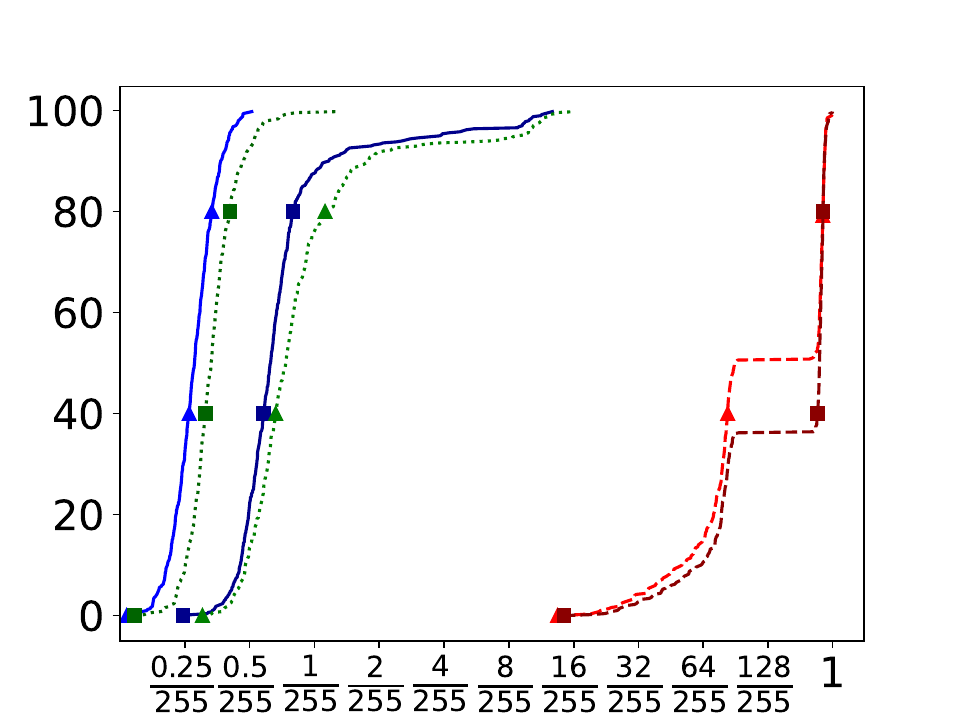} \\
\end{tabular}
\caption{CDF of minimum perturbation to achieve a 99\% or 90\% pixel error obtained using the minimum perturbation attacks on Cityscapes.
The constant 0 and constant 1 segments are not shown. Note the logarithmic scale.}
\label{fig:supp-cs-minpert}
\end{figure*}

\begin{figure*}
\setlength{\tabcolsep}{0pt}
\centering
\tiny
\begin{tabular}{cccccc}
& ALMAProx & DAG-0.001 & DAG-0.003 & PDPGD & Aggregated\\
\rotatebox[origin=l]{90}{\hspace{0mm}DeepLabv3, 99\%} & 
\includegraphics[width=0.195\textwidth]{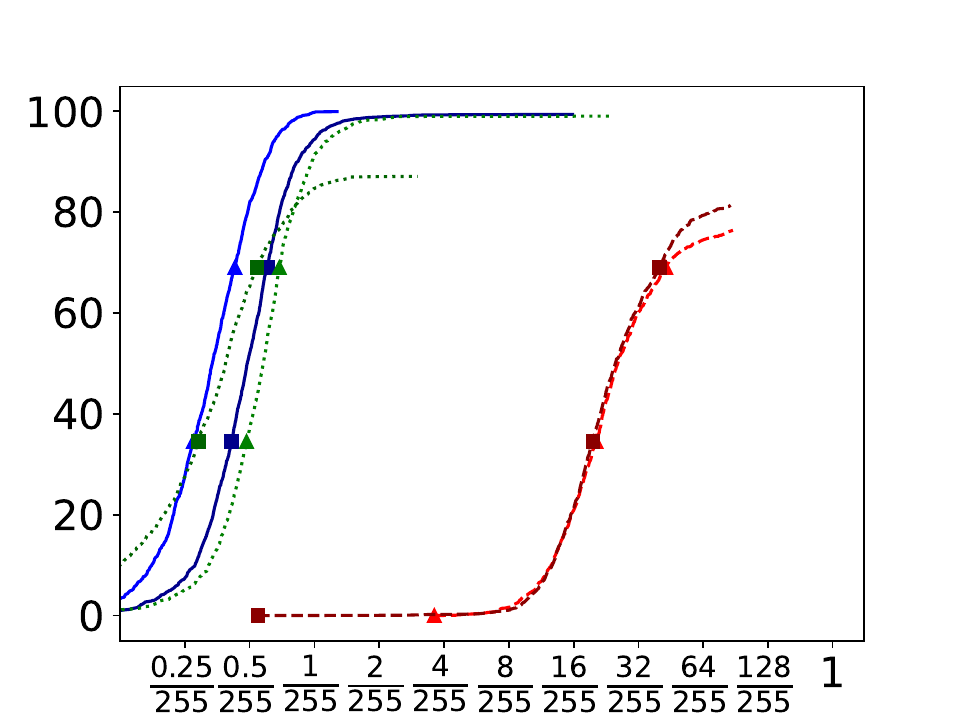} &
\includegraphics[width=0.195\textwidth]{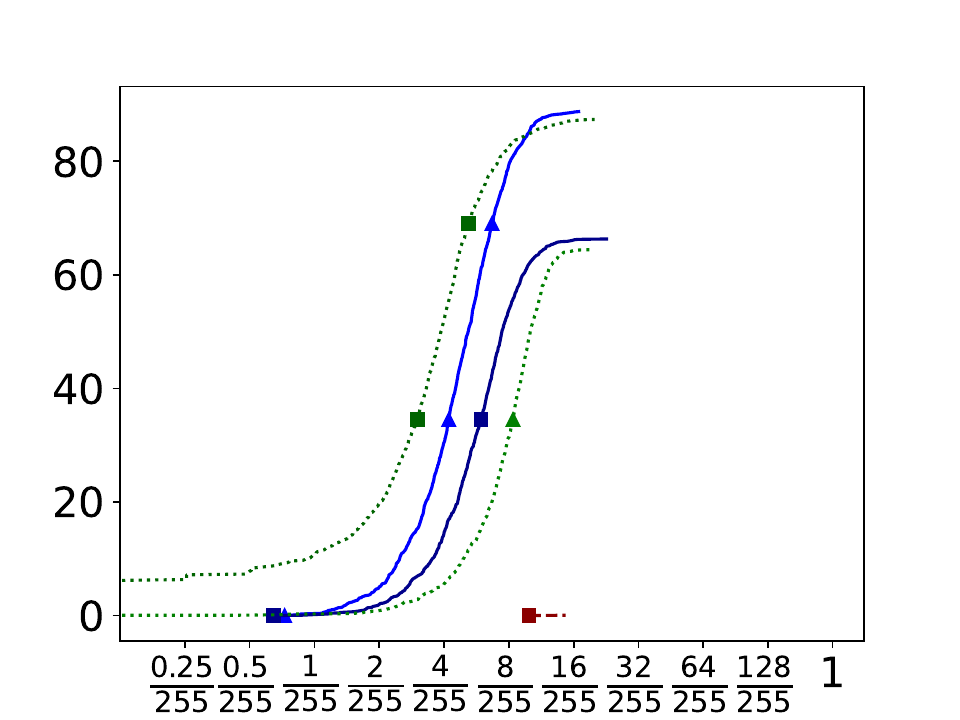} &
\includegraphics[width=0.195\textwidth]{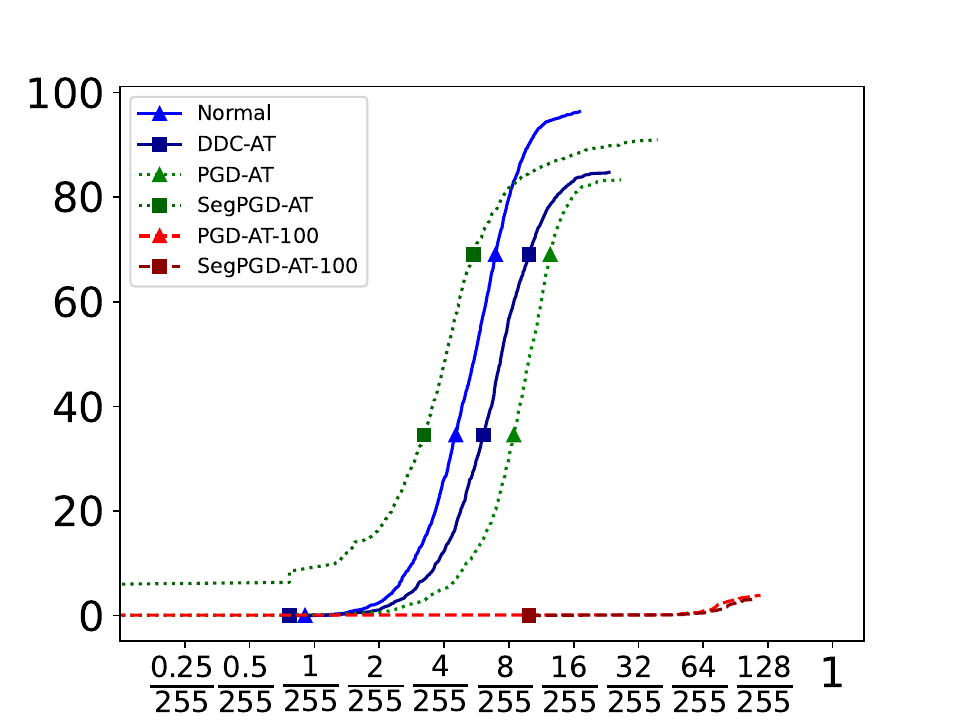} &
\includegraphics[width=0.195\textwidth]{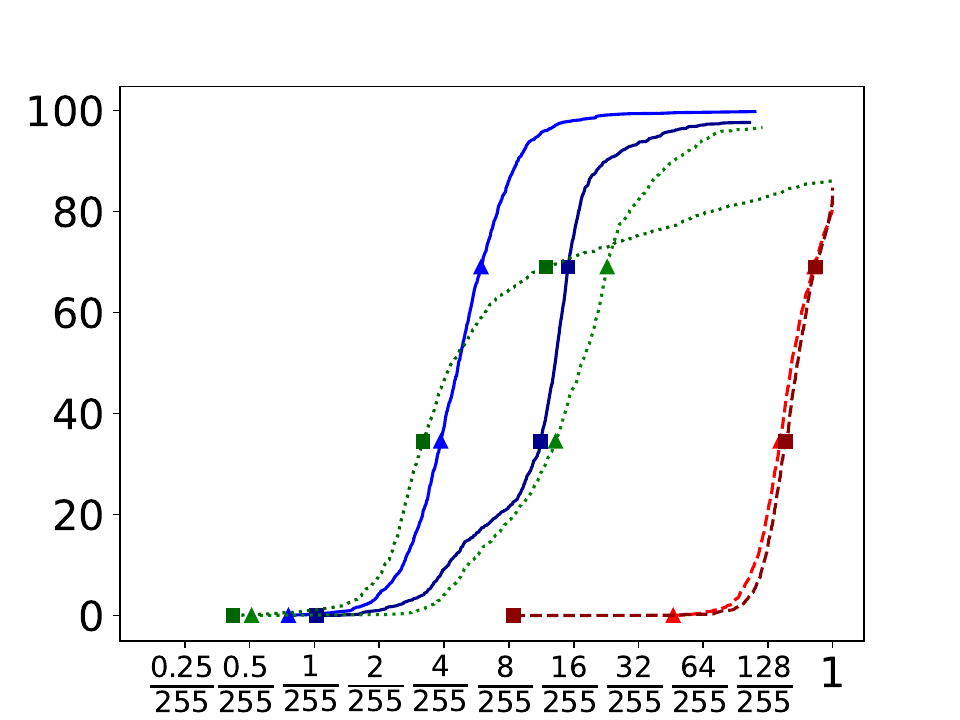} &
\includegraphics[width=0.195\textwidth]{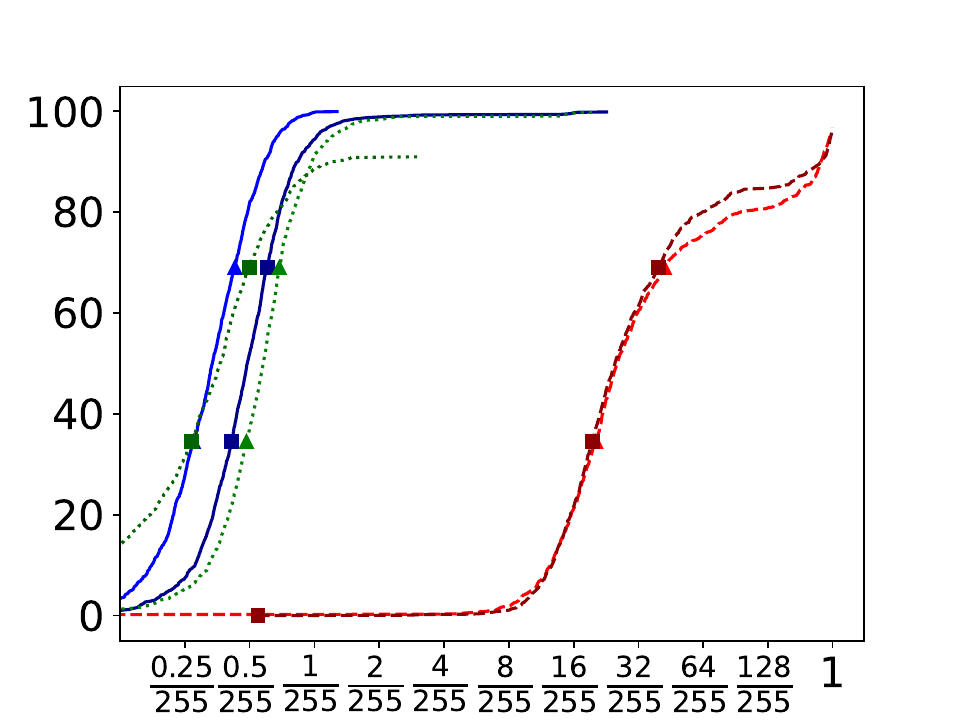} \\
\rotatebox[origin=l]{90}{\hspace{2mm}PSPNet, 99\%} & 
\includegraphics[width=0.195\textwidth]{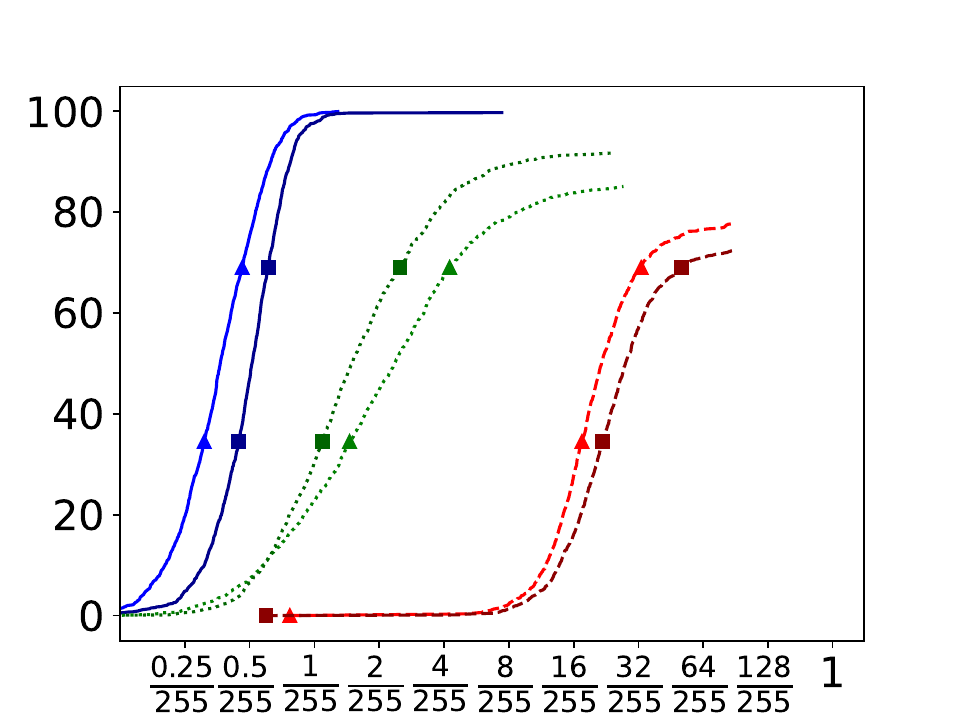} &
\includegraphics[width=0.195\textwidth]{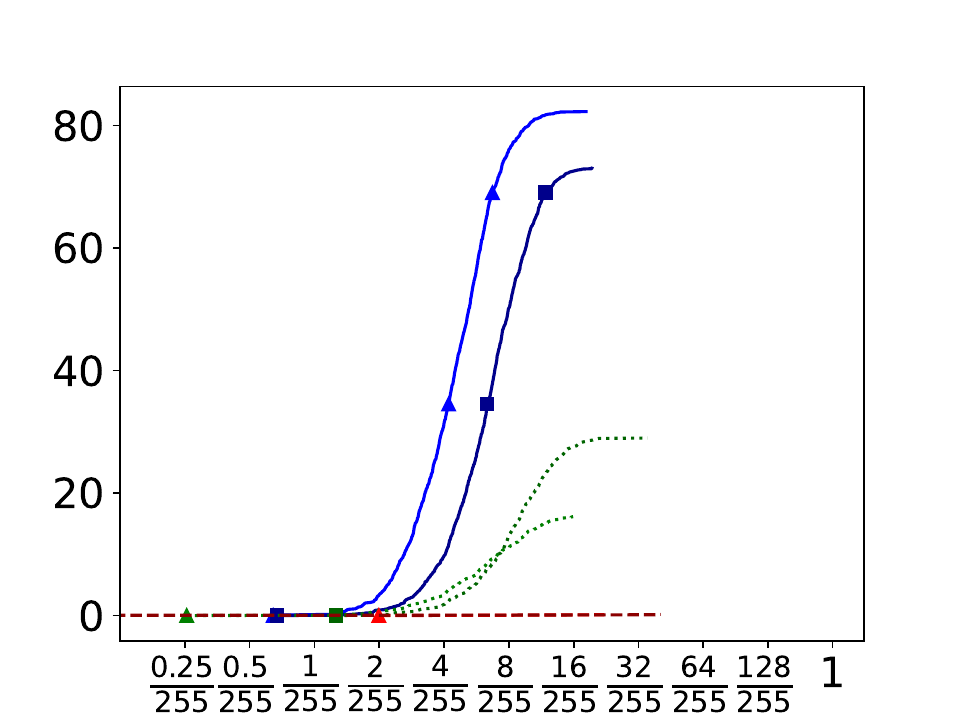} &
\includegraphics[width=0.195\textwidth]{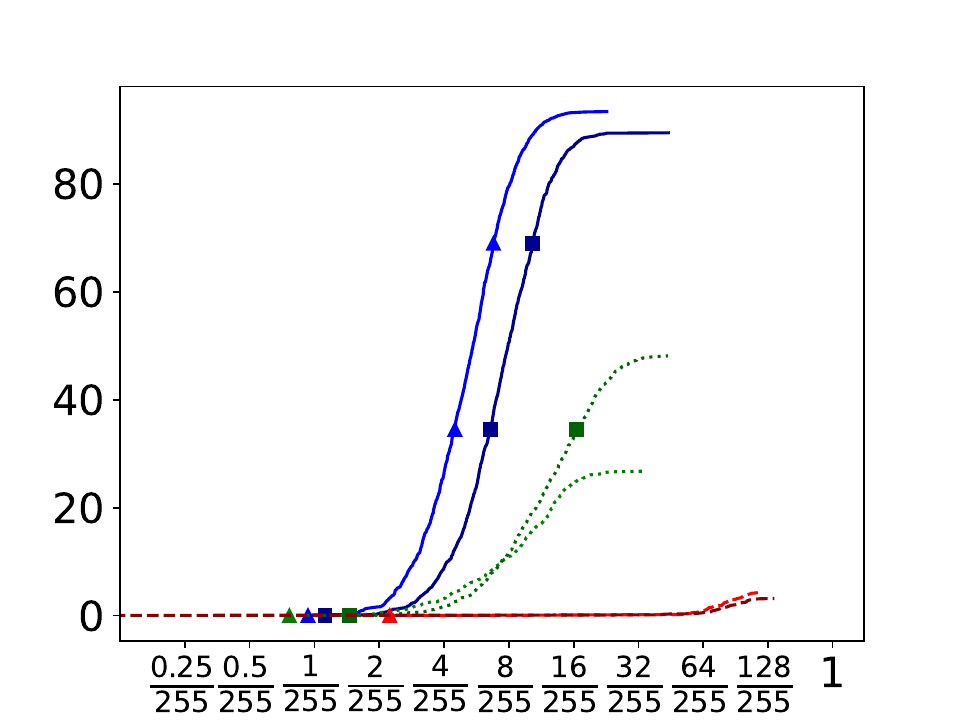} &
\includegraphics[width=0.195\textwidth]{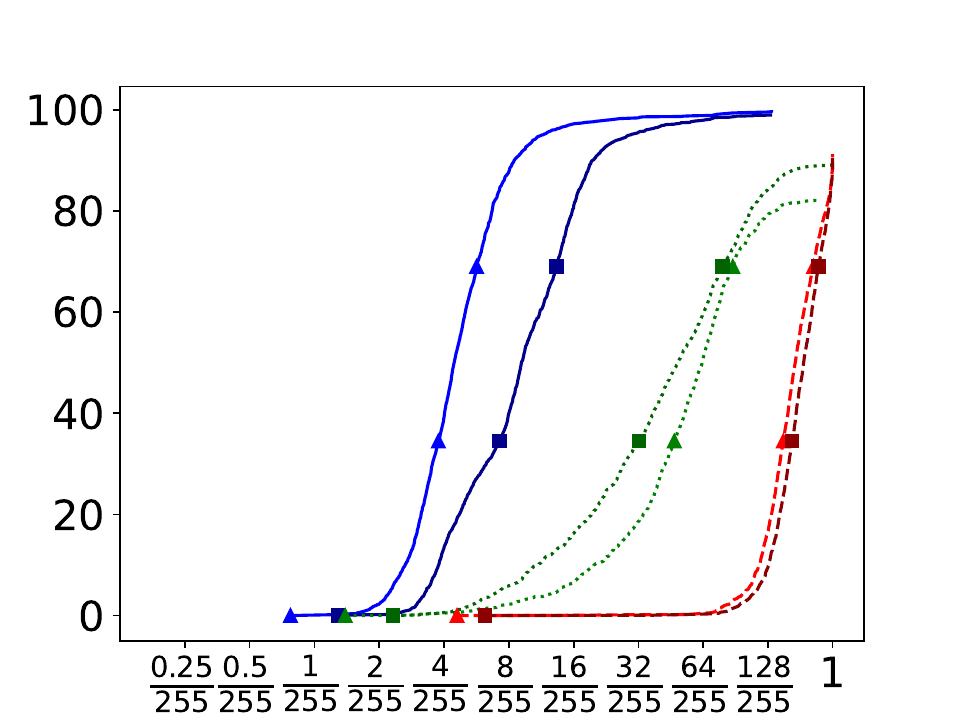} &
\includegraphics[width=0.195\textwidth]{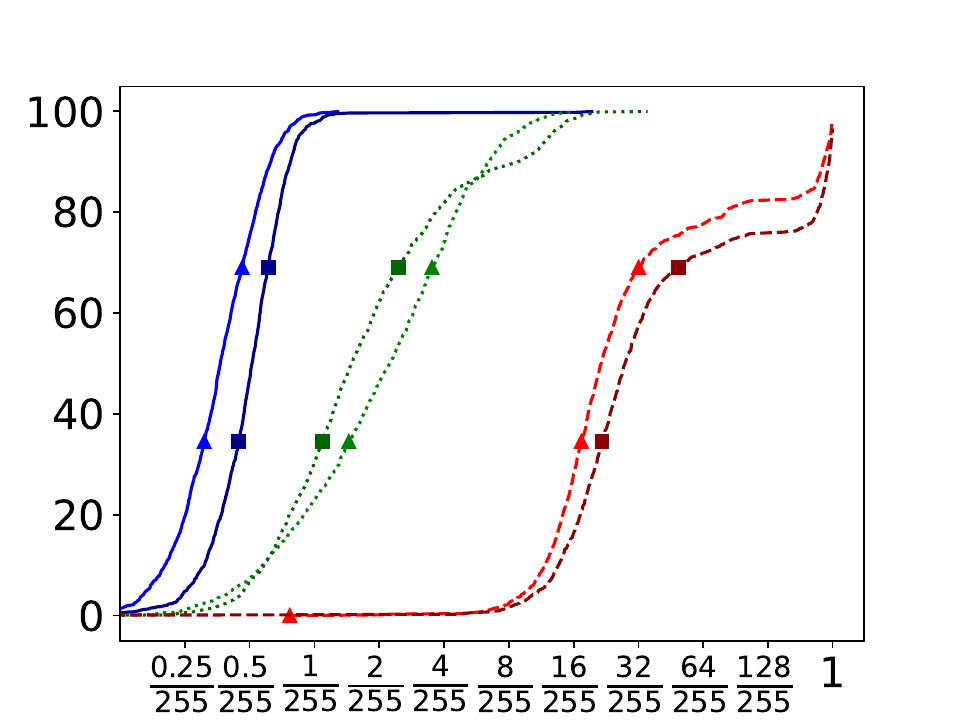} \\
\rotatebox[origin=l]{90}{\hspace{0mm}DeepLabv3, 90\%} & 
\includegraphics[width=0.195\textwidth]{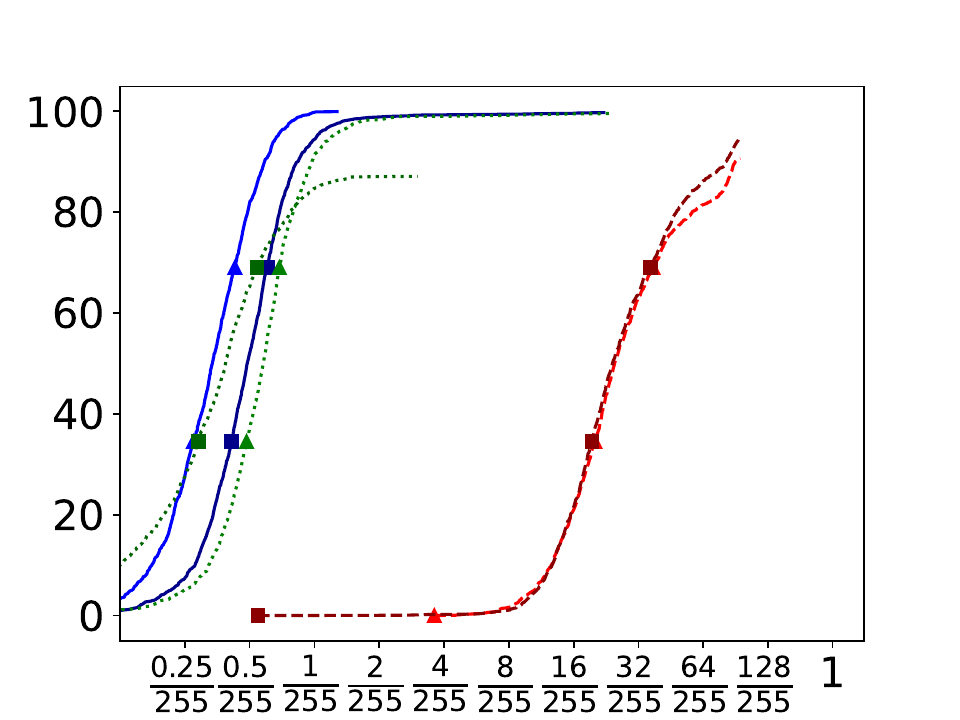} &
\includegraphics[width=0.195\textwidth]{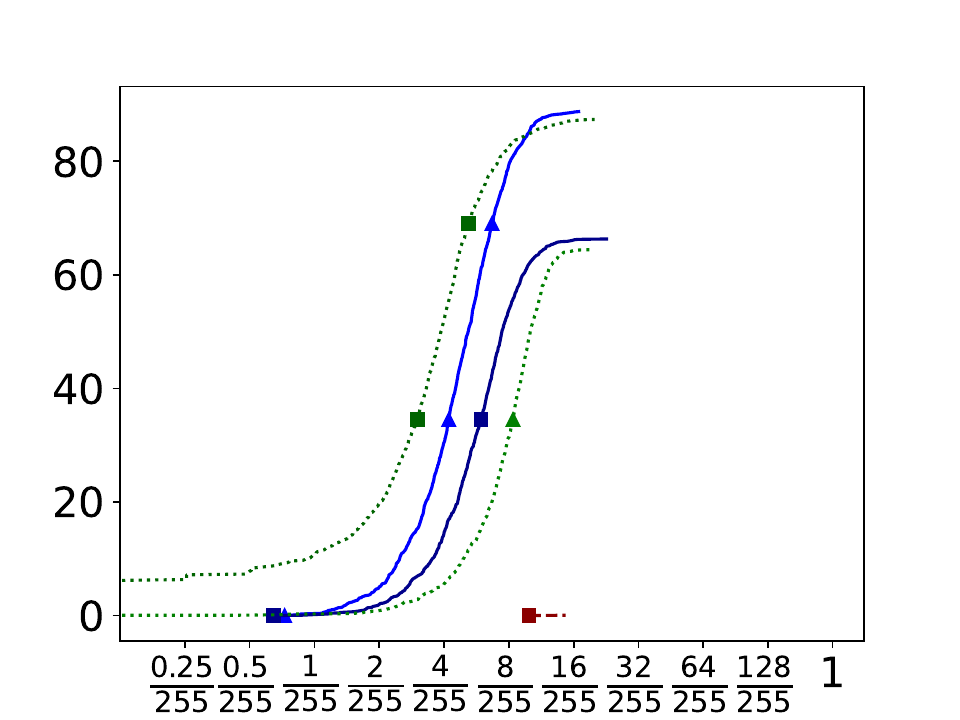} &
\includegraphics[width=0.195\textwidth]{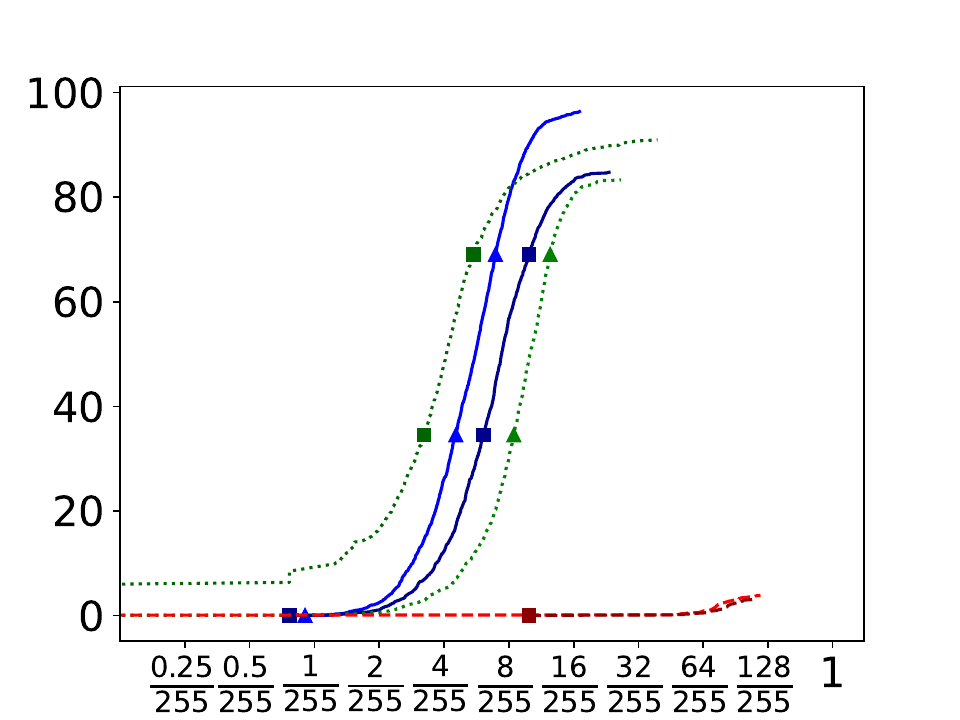} &
\includegraphics[width=0.195\textwidth]{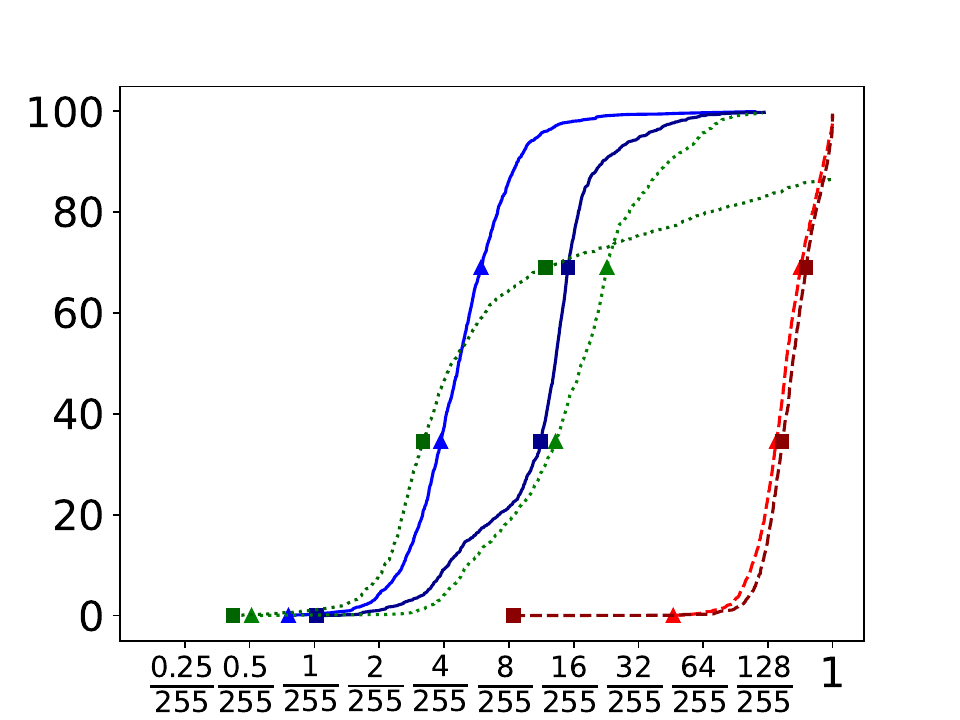} &
\includegraphics[width=0.195\textwidth]{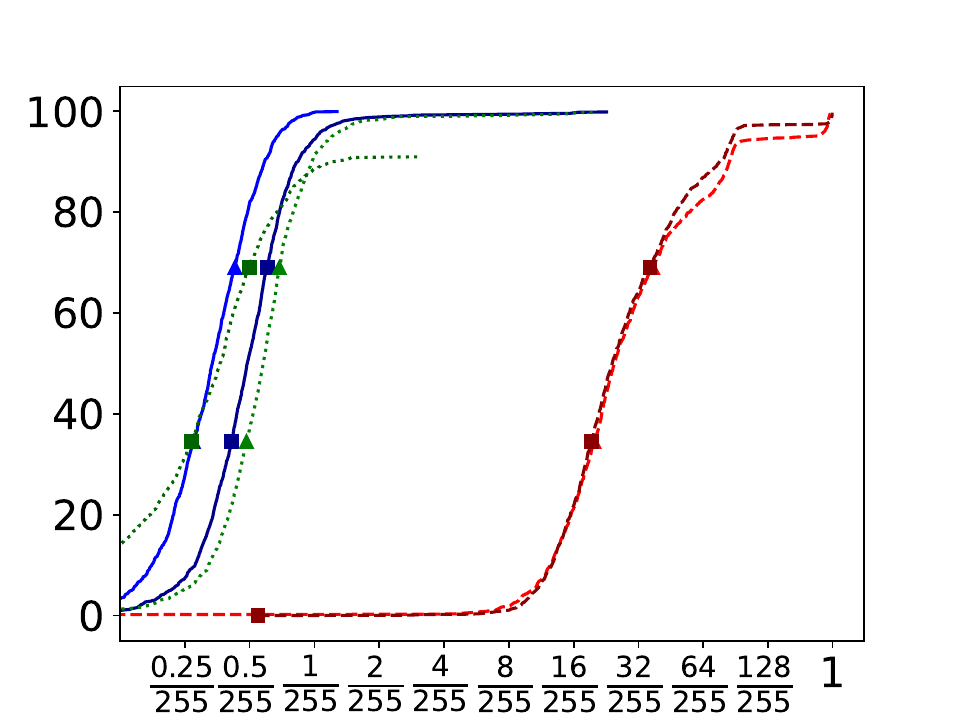} \\
\rotatebox[origin=l]{90}{\hspace{0mm}PSPNet, 90\%} & 
\includegraphics[width=0.195\textwidth]{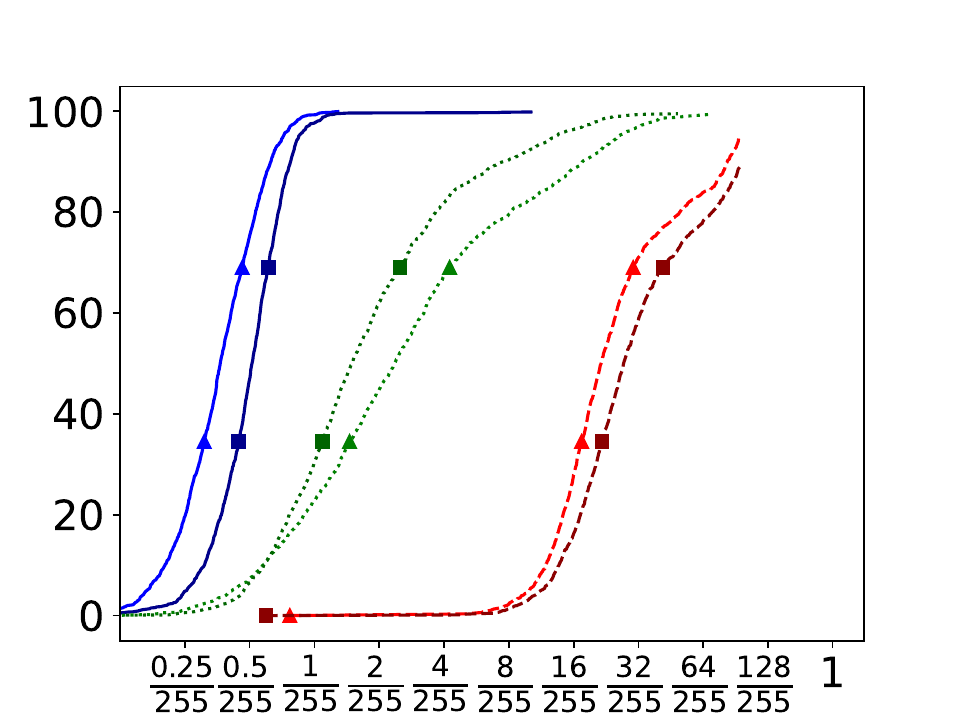} &
\includegraphics[width=0.195\textwidth]{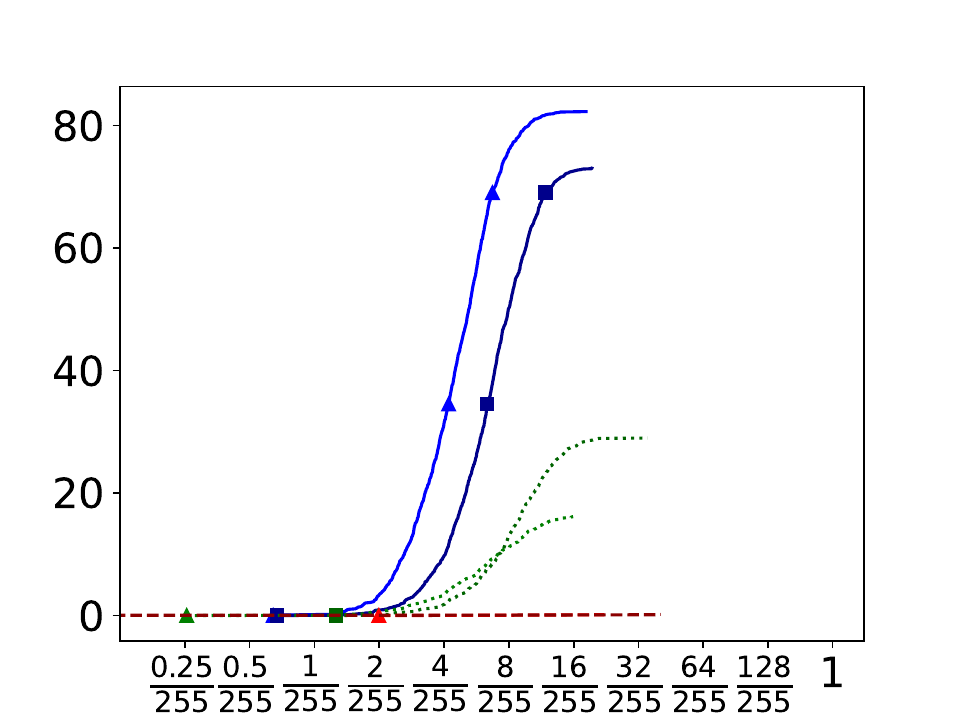} &
\includegraphics[width=0.195\textwidth]{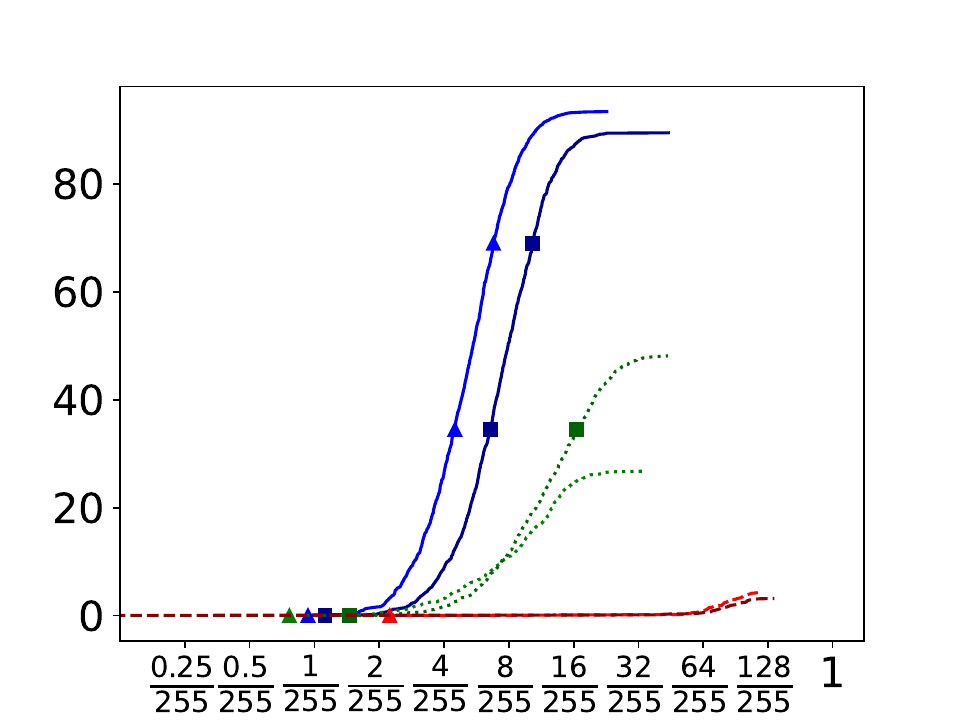} &
\includegraphics[width=0.195\textwidth]{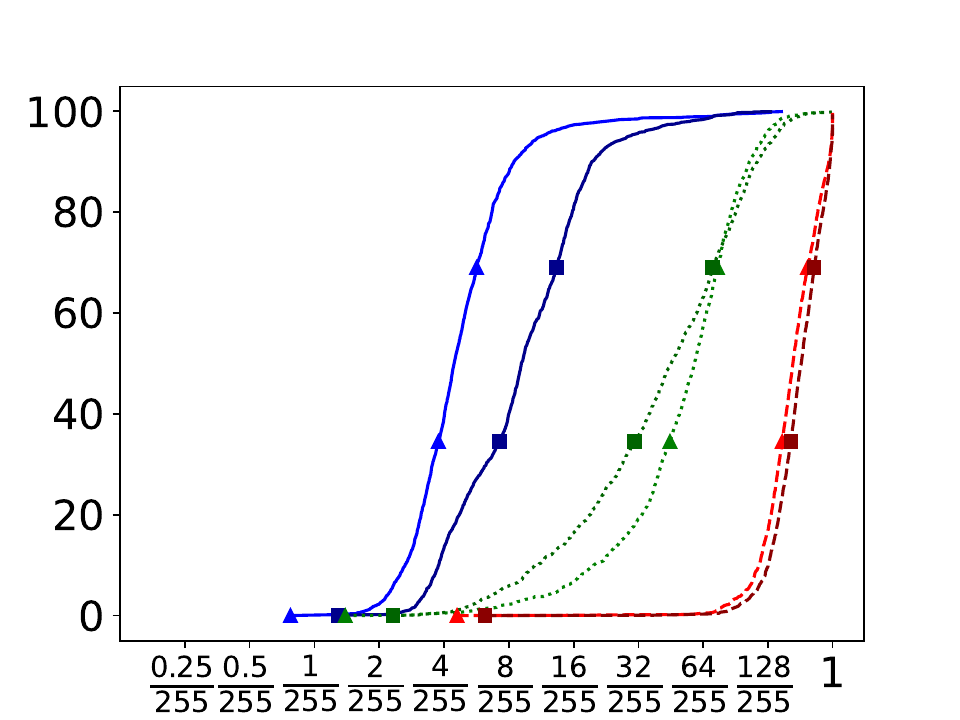} &
\includegraphics[width=0.195\textwidth]{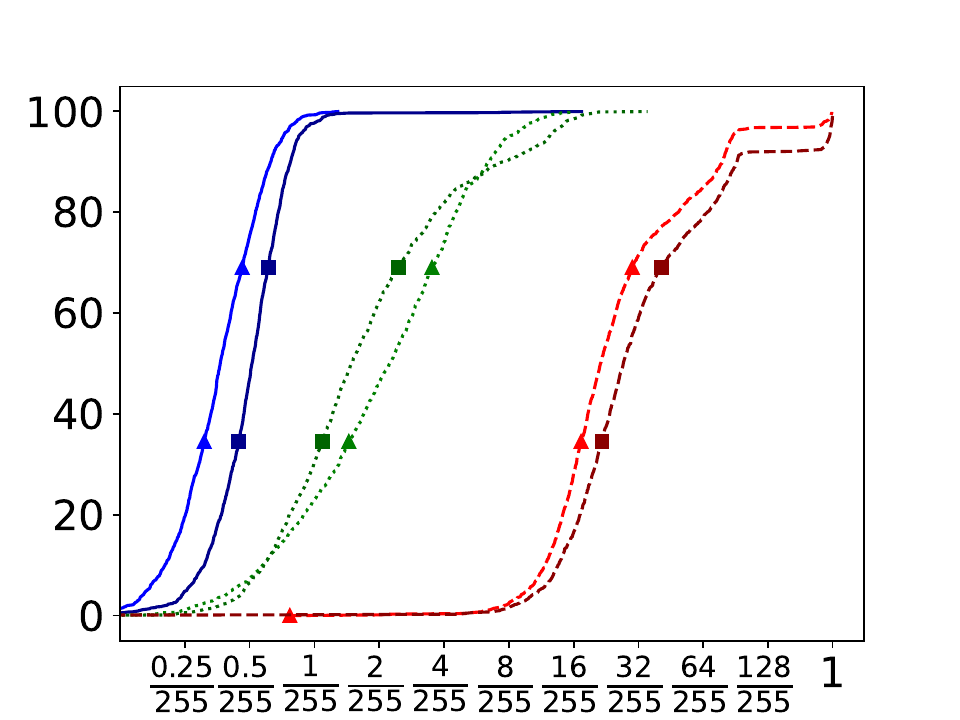} \\
\end{tabular}
\caption{CDF of minimum perturbation to achieve a 99\% or 90\% pixel error obtained using the minimum perturbation attacks on PASCAL VOC 2012.
The constant 0 and constant 1 segments are not shown. Note the logarithmic scale.}
\label{fig:supp-p-minpert}
\end{figure*}

For reference, we include the CDFs of the minimum perturbation over the evaluation set for every individual attack
we performed in \cref{fig:supp-cs-minpert,fig:supp-p-minpert}.
Note that in this work the minimum perturbation attacks were used in a clipped version (see \cref{sec:minpertattacks}).
Here, we present the results before clipping, that is, the plots are based on the raw output of the attacks.

These results add further support to our observation that the models that use only 50\% adversarial samples during adversarial training
have essentially the same (lack of) robustness as the normally trained models.
It is evident that these models are extremely vulnerable to very small perturbations (under one color level), just like normal models
(note the logarithmic scale of the horizontal axis).
Indeed, DDC-AT and SegPGD-AT are robust to $\epsilon < 0.5/255$
while the normal model is robust to $\epsilon < 0.25/255$. At
the same time, other models retain some robustness up to
$\epsilon \approx 16/255$. So, although DDC-AT and SegPGD-AT are
not identical to the normal model, the difference is practically insignificant, because the robustness threshold is
below the resolution of the color representation.

What is also notable is that ALMAProx is clearly the best choice when relatively small perturbations are sufficient to
reach the target error rate, however, in the large perturbation regime the other attacks are able to improve the
aggregated CDFs.

\section{Illustration of Attacks}
\label{sec:supp-examples}

\begin{figure*}[b]
\centering
\begin{tabular}{cc}
clean image & ground truth label \\
\includegraphics[width=.40\textwidth]{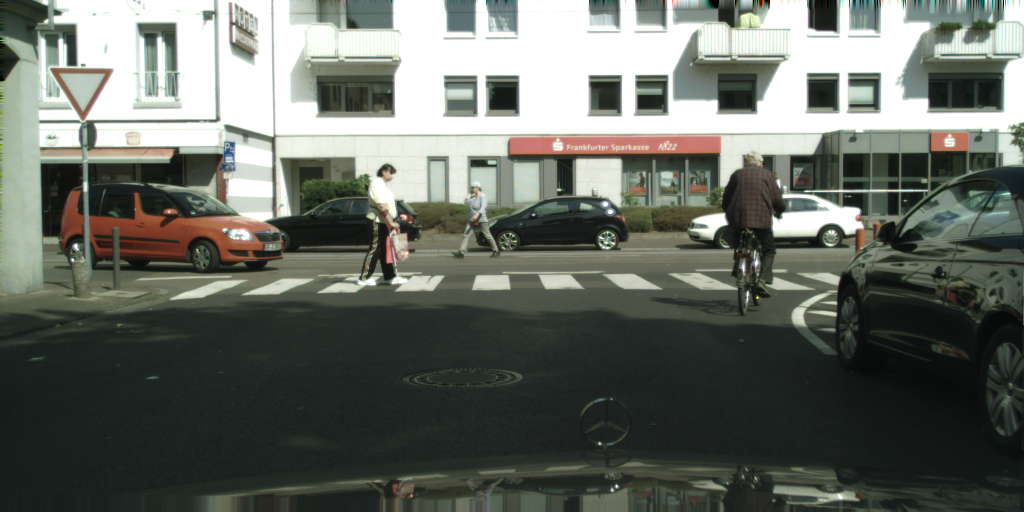} &
\includegraphics[width=.40\textwidth]{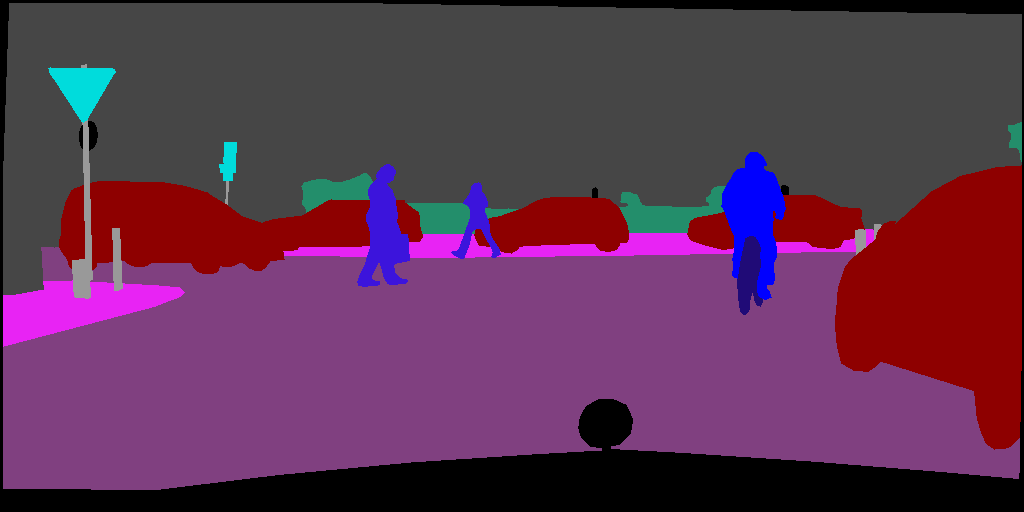}\\
\includegraphics[width=.40\textwidth]{images_adv/saved_images1/original} &
\includegraphics[width=.40\textwidth]{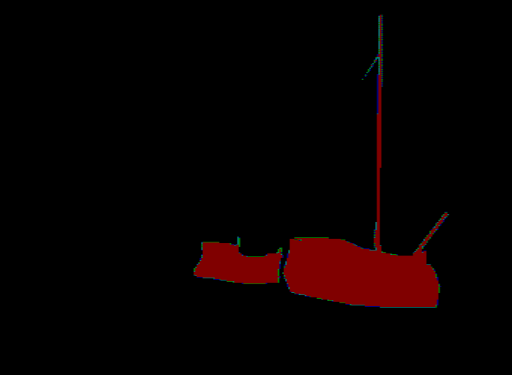} \\
\end{tabular}
\caption{Example images with ground truth labels from Cityscapes (top) and PASCAL VOC (bottom).}
\label{fig:ground-truth}
\end{figure*}

We illustrate our set of attacks in our scenarios using the examples from the Cityscapes and PASCAL VOC datasets shown in
\cref{fig:ground-truth}.

\begin{figure*}[t]
\centering
\setlength{\tabcolsep}{.2pt}
\scriptsize
\begin{tabular}{b{9mm}@{\hspace{2mm}}cccccc}
& Normal & DDC-AT & PGD-AT & SegPGD-AT & PGD-AT-100 & SegPGD-100 \\
No \newline attack &
\includegraphics[width=.15\textwidth]{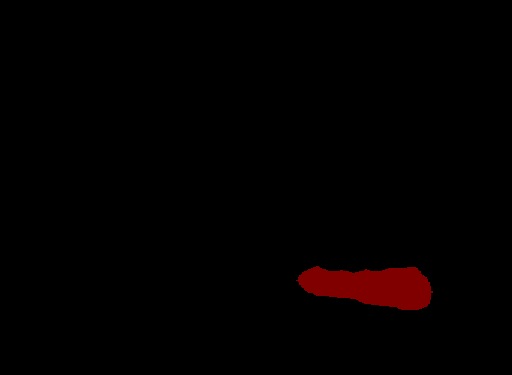} &
\includegraphics[width=.15\textwidth]{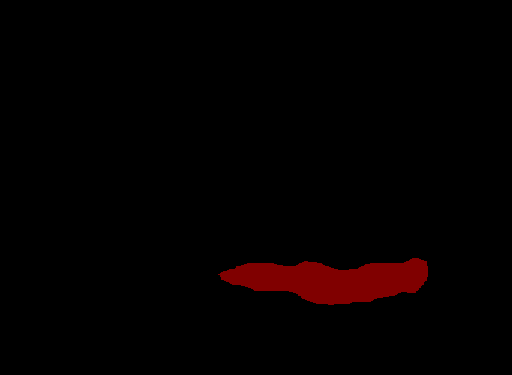} &
\includegraphics[width=.15\textwidth]{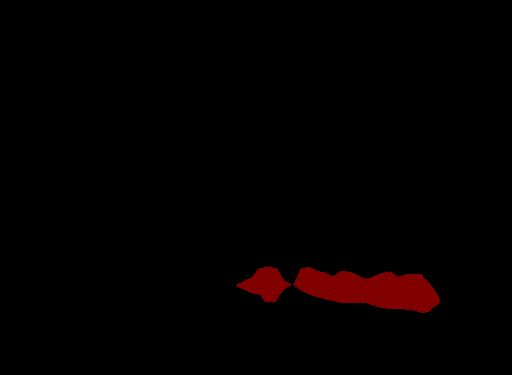} &
\includegraphics[width=.15\textwidth]{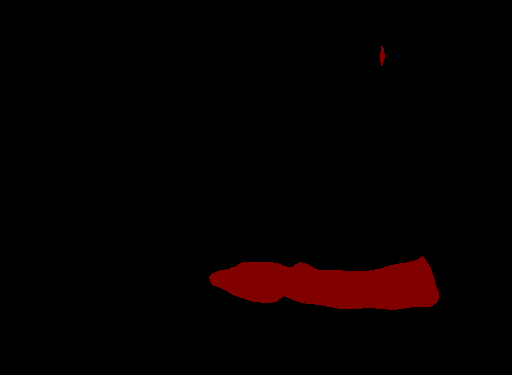} &
\includegraphics[width=.15\textwidth]{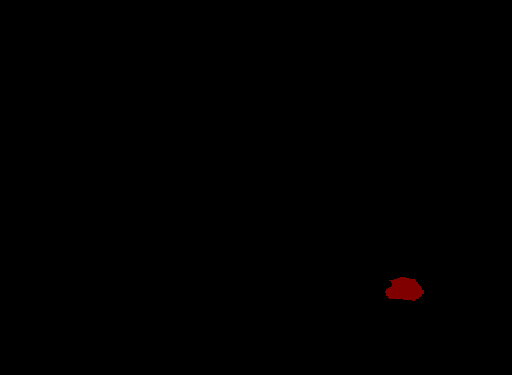} &
\includegraphics[width=.15\textwidth]{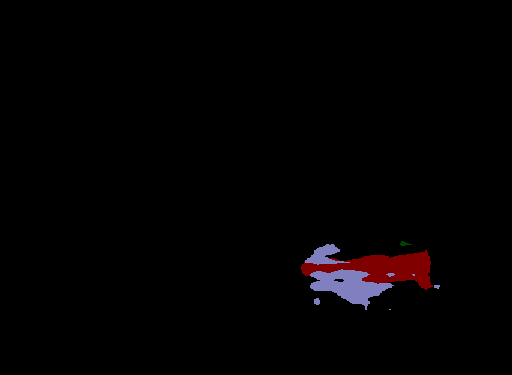} \\
PAdam-CE &
\includegraphics[width=.15\textwidth]{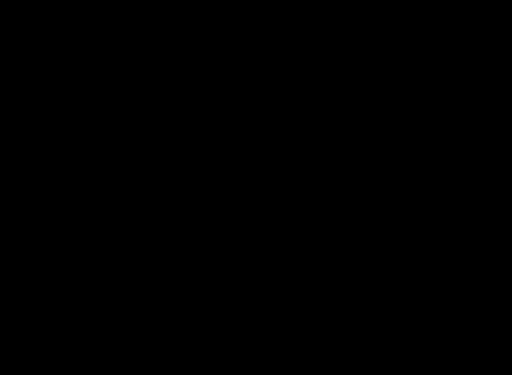} &
\includegraphics[width=.15\textwidth]{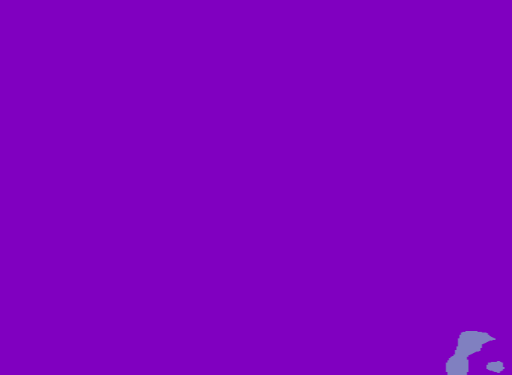} &
\includegraphics[width=.15\textwidth]{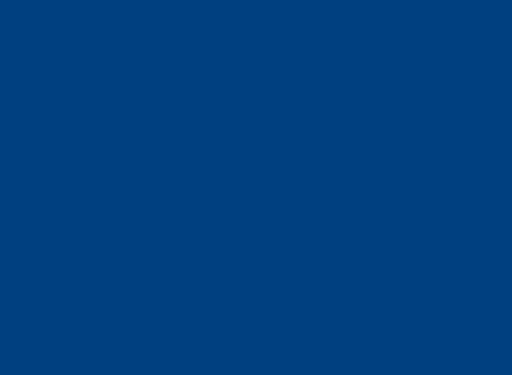} &
\includegraphics[width=.15\textwidth]{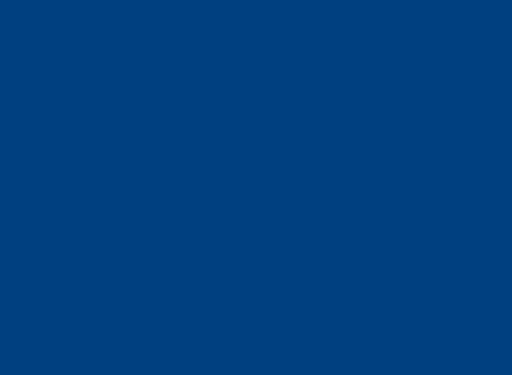} &
\includegraphics[width=.15\textwidth]{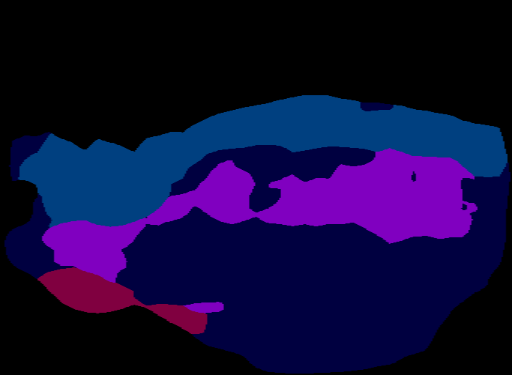} &
\includegraphics[width=.15\textwidth]{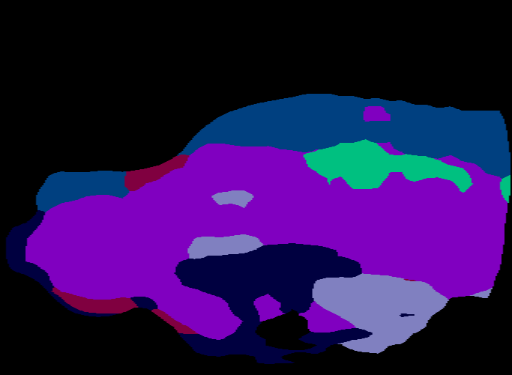} \\
PAdam-Cos &
\includegraphics[width=.15\textwidth]{images/saved_images1/attack_cos-padam-normal_pspnet-2-pred} &
\includegraphics[width=.15\textwidth]{images/saved_images1/attack_cos-padam-ddcat_pspnet-2-pred} &
\includegraphics[width=.15\textwidth]{images/saved_images1/attack_cos-padam-sat_pspnet-2-pred} &
\includegraphics[width=.15\textwidth]{images/saved_images1/attack_cos-padam-seg_pgd_pspnet-2-pred} &
\includegraphics[width=.15\textwidth]{images/saved_images1/attack_cos-padam-sat_100_pspnet-2-pred} &
\includegraphics[width=.15\textwidth]{images/saved_images1/attack_cos-padam-seg_pgd-100_pspnet-2-pred} \\
\end{tabular}
\caption{Result of attacks on an example PASCAL VOC image (see \cref{fig:ground-truth}) for DeepLabv3 models. Only the predicted masks on
the attacked images are shown due to file size limitations, but the perturbations are within the defined range of $8/255$ in
$\ell_\infty$ norm.
Please see \cref{fig:examples-attacks,fig:examples-models} for examples of attacked images.}
\label{fig:p-pspnet-masks}
\end{figure*}

\begin{figure*}[t]
\centering
\setlength{\tabcolsep}{.2pt}
\scriptsize
\begin{tabular}{b{9mm}@{\hspace{2mm}}cccccc}
& Normal & DDC-AT & PGD-AT & SegPGD-AT & PGD-AT-100 & SegPGD-100 \\
No \newline attack &
\includegraphics[width=.15\textwidth]{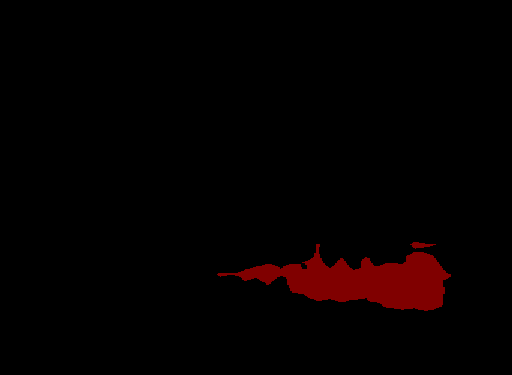} &
\includegraphics[width=.15\textwidth]{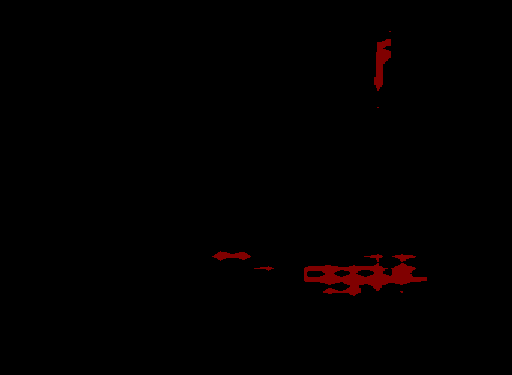} &
\includegraphics[width=.15\textwidth]{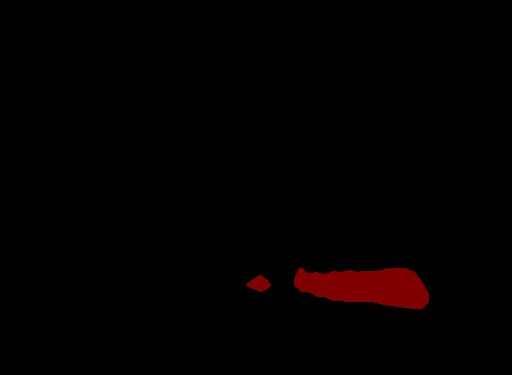} &
\includegraphics[width=.15\textwidth]{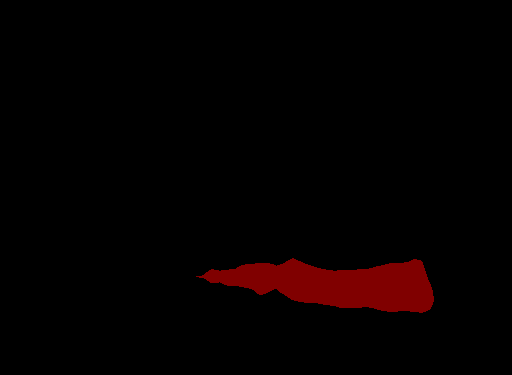} &
\includegraphics[width=.15\textwidth]{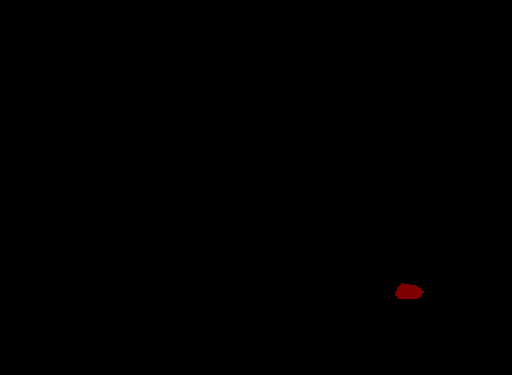} &
\includegraphics[width=.15\textwidth]{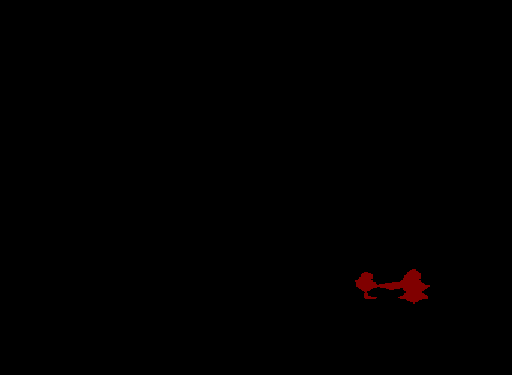} \\
PAdam-CE &
\includegraphics[width=.15\textwidth]{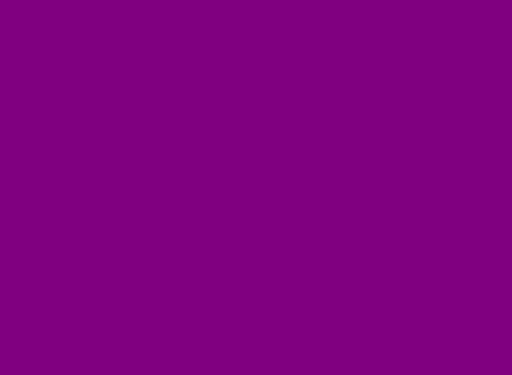} &
\includegraphics[width=.15\textwidth]{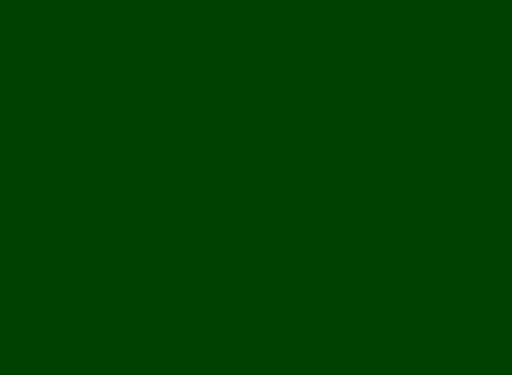} &
\includegraphics[width=.15\textwidth]{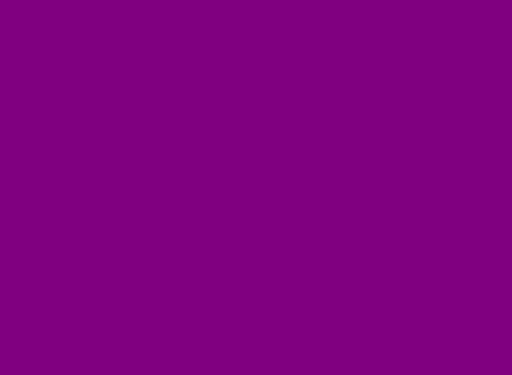} &
\includegraphics[width=.15\textwidth]{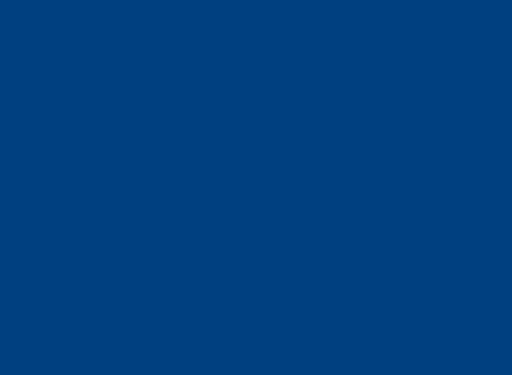} &
\includegraphics[width=.15\textwidth]{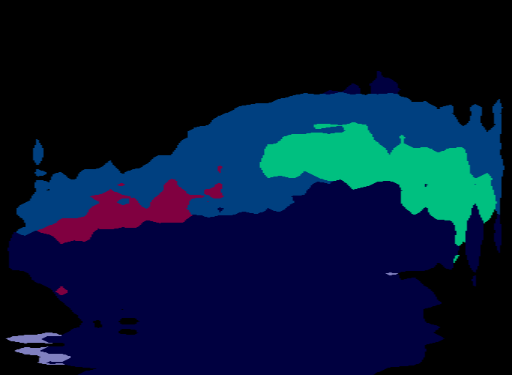} &
\includegraphics[width=.15\textwidth]{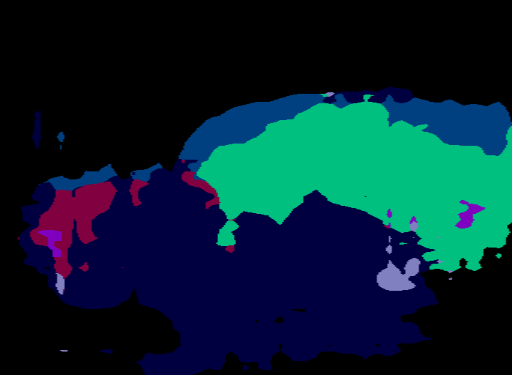} \\
PAdam-Cos &
\includegraphics[width=.15\textwidth]{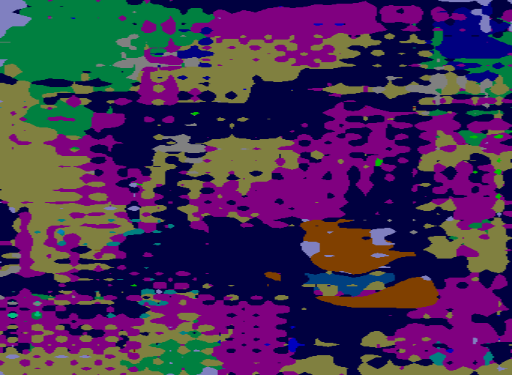} &
\includegraphics[width=.15\textwidth]{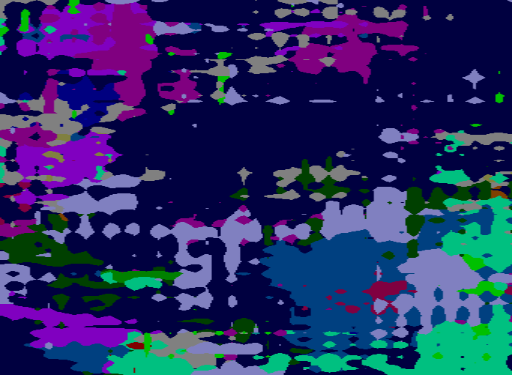} &
\includegraphics[width=.15\textwidth]{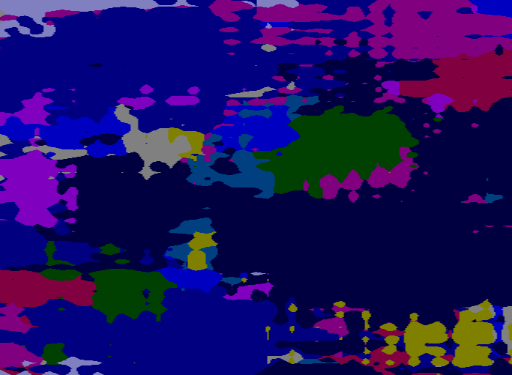} &
\includegraphics[width=.15\textwidth]{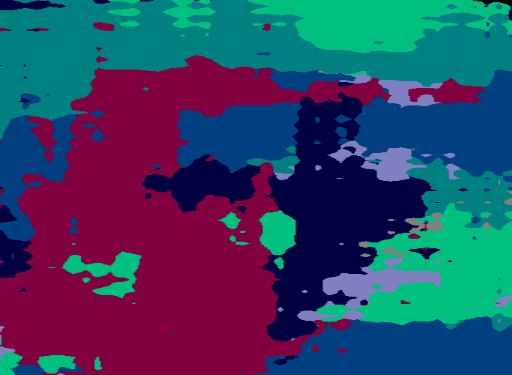} &
\includegraphics[width=.15\textwidth]{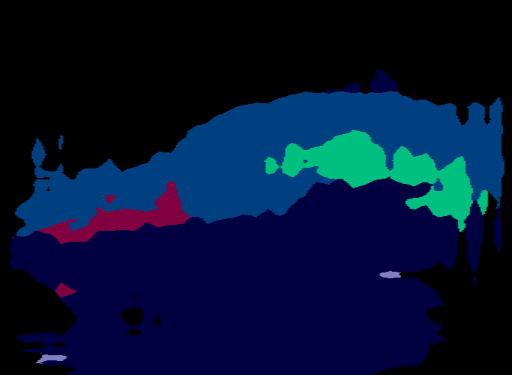} &
\includegraphics[width=.15\textwidth]{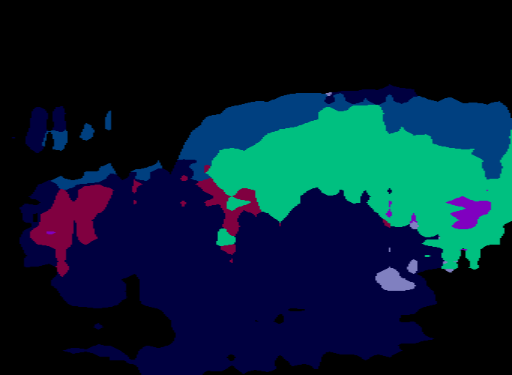} \\
\end{tabular}
\caption{Result of attacks on an example PASCAL VOC image (see \cref{fig:ground-truth}) for PSPNet models. Only the predicted masks on
the attacked images are shown due to file size limitations, but the perturbations are within the defined range of $8/255$ in
$\ell_\infty$ norm.
Please see \cref{fig:examples-attacks,fig:examples-models} for examples of attacked images.}
\label{fig:p-deeplab-masks}
\end{figure*}

\begin{figure*}[t]
\centering
\setlength{\tabcolsep}{.2pt}
\scriptsize
\begin{tabular}{b{9mm}@{\hspace{2mm}}cccccc}
& SEA-AT-Tiny & SEA-AT-Small \\
No \newline attack &
\includegraphics[width=.15\textwidth]{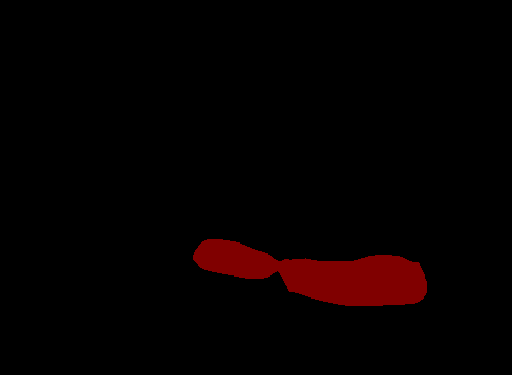} &
\includegraphics[width=.15\textwidth]{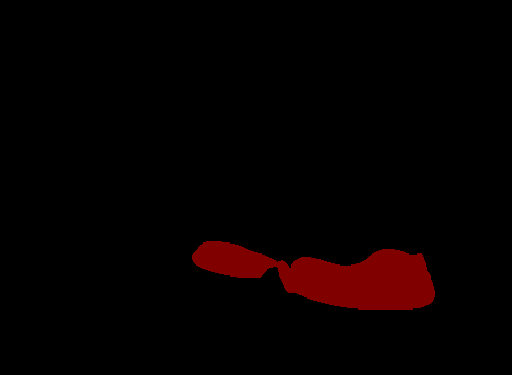} \\
PAdam-CE &
\includegraphics[width=.15\textwidth]{images/saved_images1/attack_ce-padam-ConvNeXt-T_CVST_ROB-2-pred} &
\includegraphics[width=.15\textwidth]{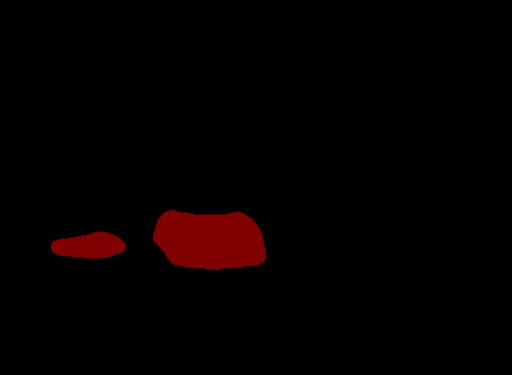} \\
PAdam-Cos &
\includegraphics[width=.15\textwidth]{images/saved_images1/attack_cos-padam-ConvNeXt-T_CVST_ROB-2-pred} &
\includegraphics[width=.15\textwidth]{images/saved_images1/attack_cos-padam-ConvNeXt-S_CVST_ROB-2-pred} \\
SEA-JSD &
\includegraphics[width=.15\textwidth]{images/saved_images1/js-avg-ConvNeXt-T_CVST_ROB-2-pred} &
\includegraphics[width=.15\textwidth]{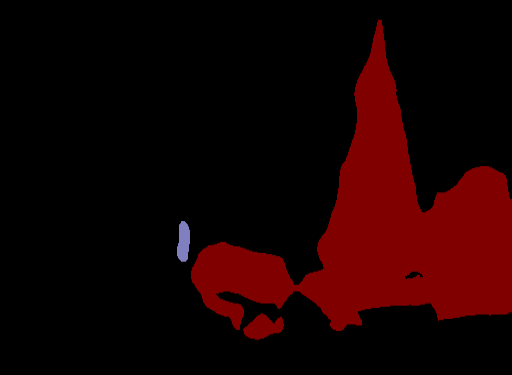} \\
SEA-MCE &
\includegraphics[width=.15\textwidth]{images/saved_images1/mask-ce-avg-ConvNeXt-T_CVST_ROB-2-pred} &
\includegraphics[width=.15\textwidth]{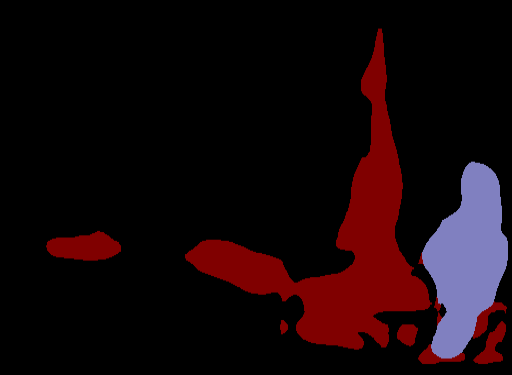} \\
SEA-MSL &
\includegraphics[width=.15\textwidth]{images/saved_images1/mask-norm-corrlog-avg-ConvNeXt-T_CVST_ROB-2-pred} &
\includegraphics[width=.15\textwidth]{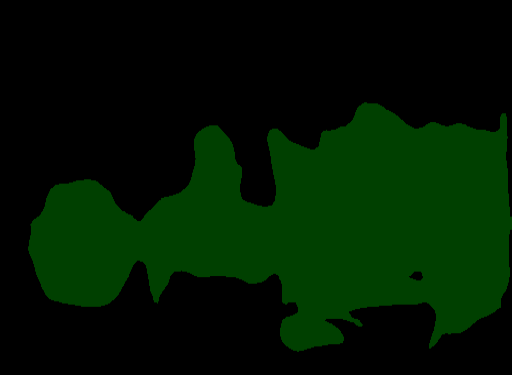} \\
SEA-BCE &
\includegraphics[width=.15\textwidth]{images/saved_images1/segpgd-loss-ConvNeXt-T_CVST_ROB-2-pred} &
\includegraphics[width=.15\textwidth]{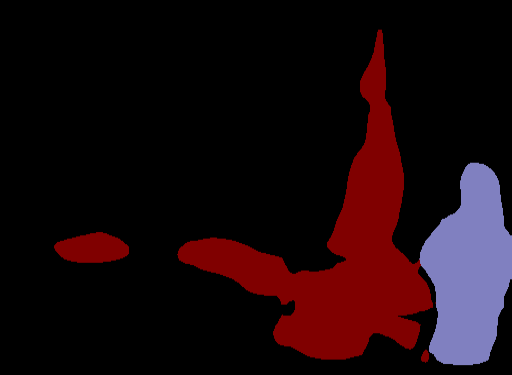} \\
\end{tabular}
\caption{Result of attacks on an example PASCAL VOC image (see \cref{fig:ground-truth}) for the SEA-AT models. Only the predicted masks on
the attacked images are shown due to file size limitations, but the perturbations are within the defined range of $8/255$ in
$\ell_\infty$ norm.
Please see \cref{fig:examples-attacks,fig:examples-models} for examples of attacked images.}
\label{fig:p-seaat-masks}
\end{figure*}

\cref{fig:cs-pspnet-masks,fig:cs-deeplab-masks,fig:p-pspnet-masks,fig:p-deeplab-masks,fig:p-seaat-masks} illustrate our attacks using only
the predicted masks over the attacked inputs.
This is due to the size limit on the supplementary material.
However, we also include perturbed images for a limited number of settings in \cref{fig:examples-attacks,fig:examples-models} in the main paper and in
\cref{fig:cs-examples-images} in the supplementary material.
These images illustrate that the perturbations generated by the attacks within the $\ell_\infty$ ball of radius $8/255$ are indeed hardly noticeable.

The predicted masks over the Cityscapes example shown in \cref{fig:cs-pspnet-masks,fig:cs-deeplab-masks} suggest that the models trained with
100\% adversarial samples are somewhat more robust to all the attacks, while the other models are much more sensitive.
This is consistent with the findings we discussed previously, noting that the price for this (limited) robustness is a large decrease in the
clean performance, hence these models are examples for a strong robustness-accuracy tradeoff.

The same training methodology provides even less robustness over the PASCAL VOC dataset, as evidenced by
\cref{fig:p-pspnet-masks,fig:p-deeplab-masks,fig:p-seaat-masks}.
Here, interestingly, the background seems to be somewhat protected, as we noted previously.
At the same time, the foreground pixels are completely misclassified for some of the attacks, even for the SEA-AT models (note that
it is enough if one of the attacks is successful because the evaluation always takes the best attack from the attack set).
Besides, the example also illustrates our previous observation that using 100\% adversarial training samples might reduce the clean accuracy
as well.

\begin{figure*}
\centering
\setlength{\tabcolsep}{.2pt}
\begin{tabular}{cccccc}
Normal & DDC-AT & PGD-AT \\
\includegraphics[width=.3\textwidth]{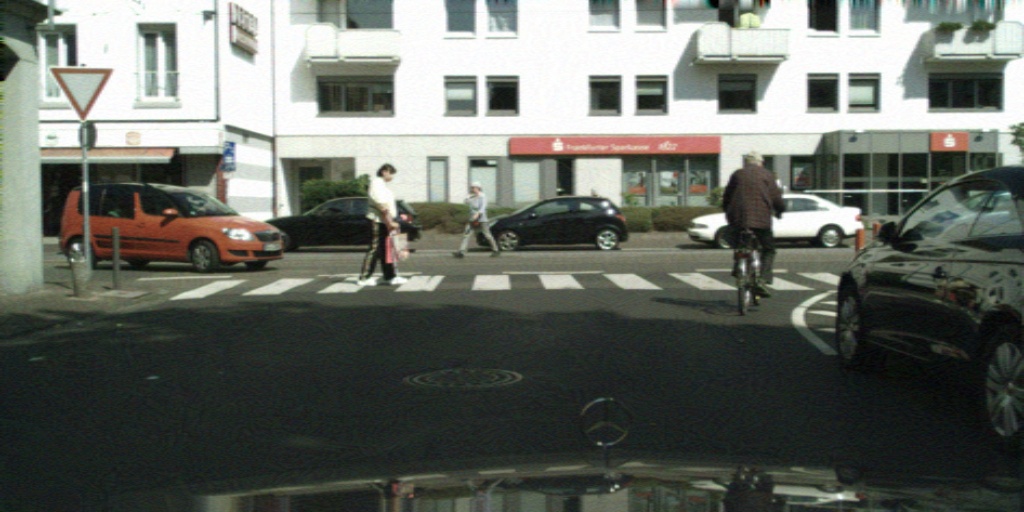} &
\includegraphics[width=.3\textwidth]{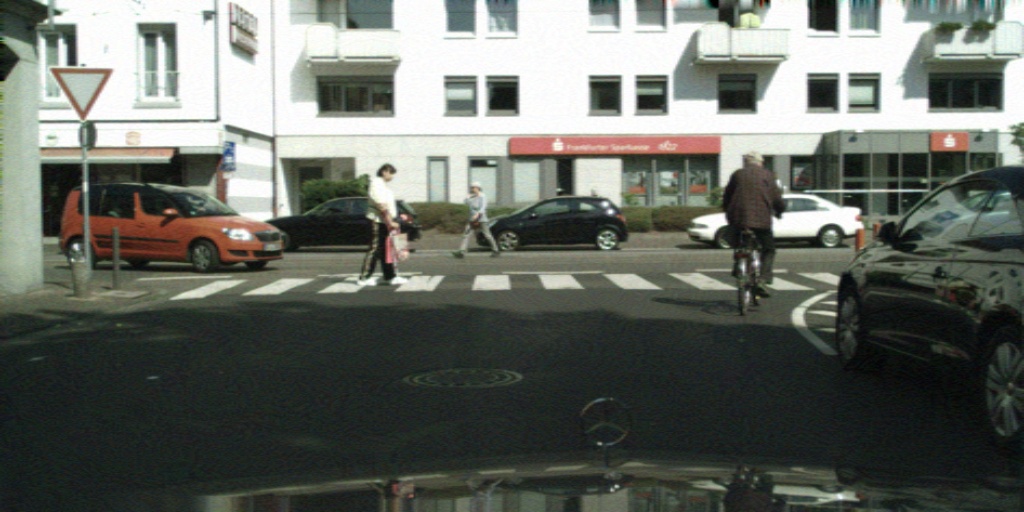} &
\includegraphics[width=.3\textwidth]{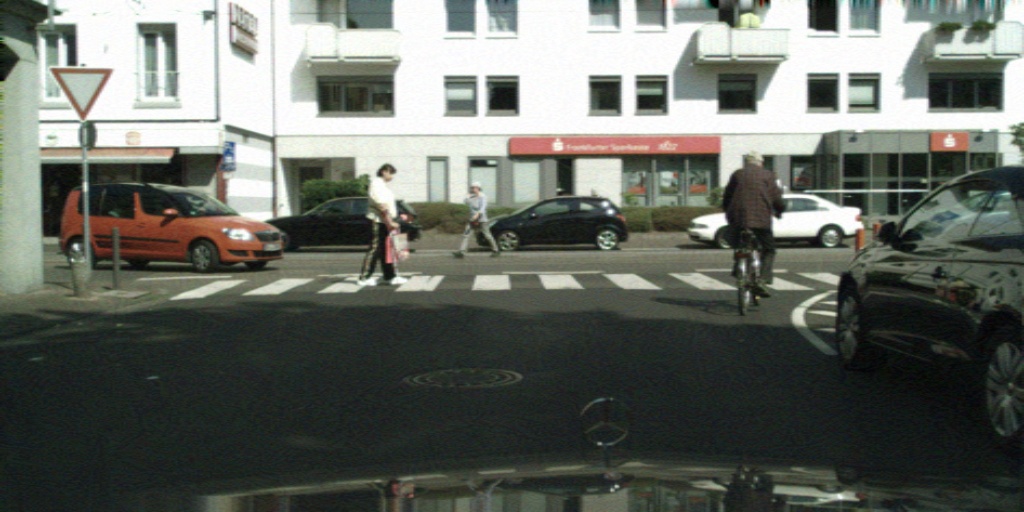} \\[-.5mm]
\includegraphics[width=.3\textwidth]{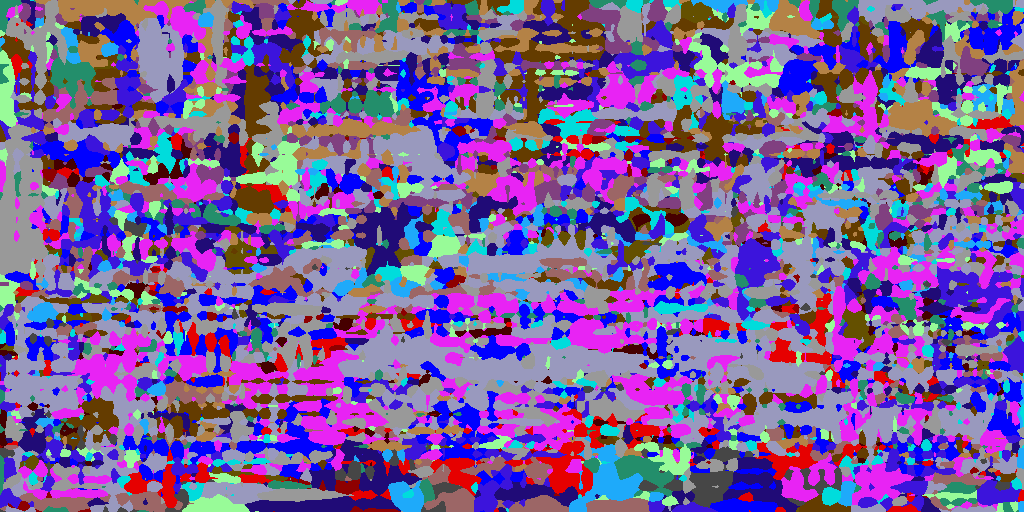} &
\includegraphics[width=.3\textwidth]{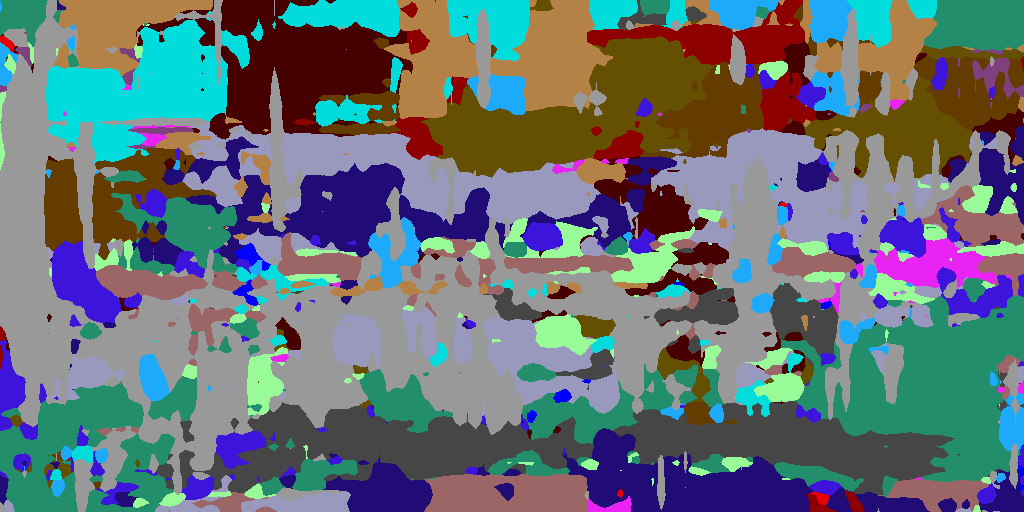} &
\includegraphics[width=.3\textwidth]{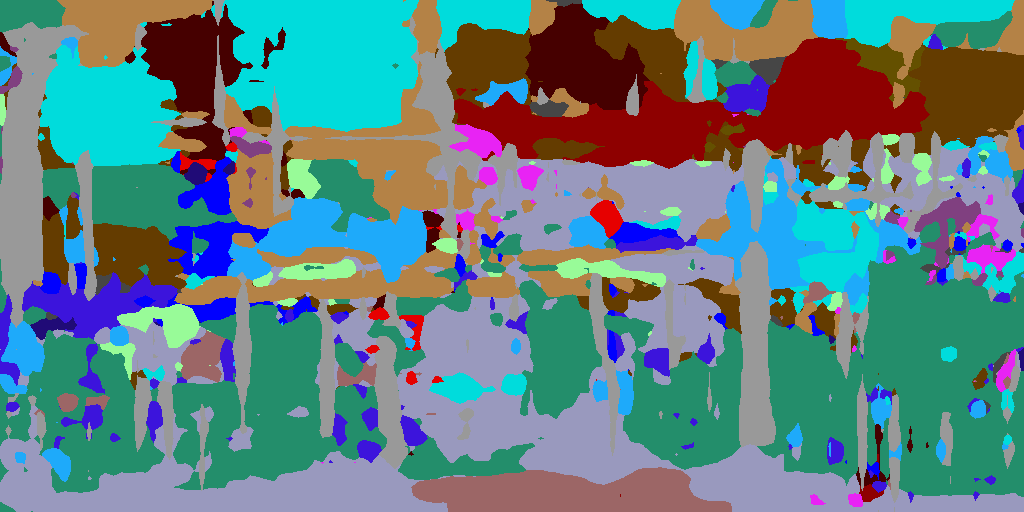} \\[2mm]
SegPGD-AT & PGD-AT-100 & SegPGD-100 \\
\includegraphics[width=.3\textwidth]{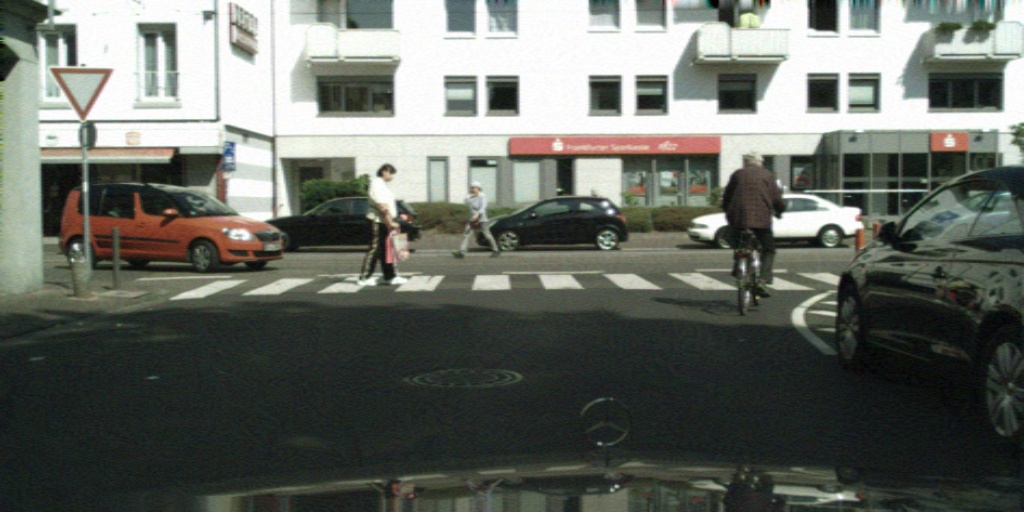} &
\includegraphics[width=.3\textwidth]{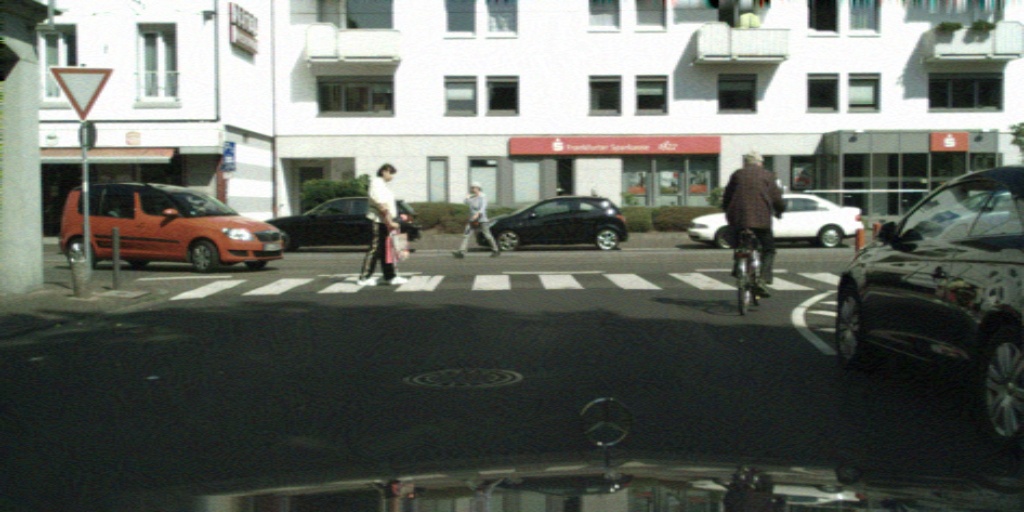} &
\includegraphics[width=.3\textwidth]{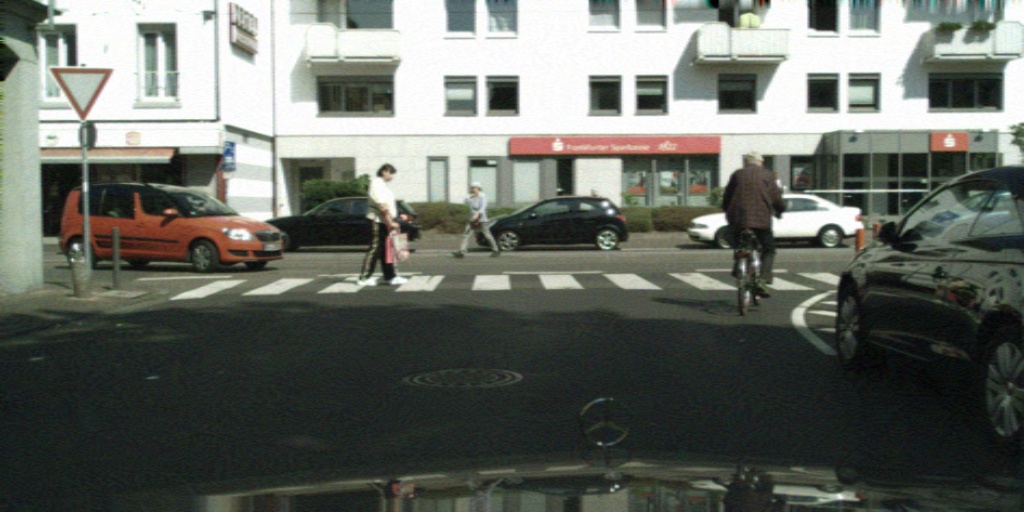} \\[-.5mm]
\includegraphics[width=.3\textwidth]{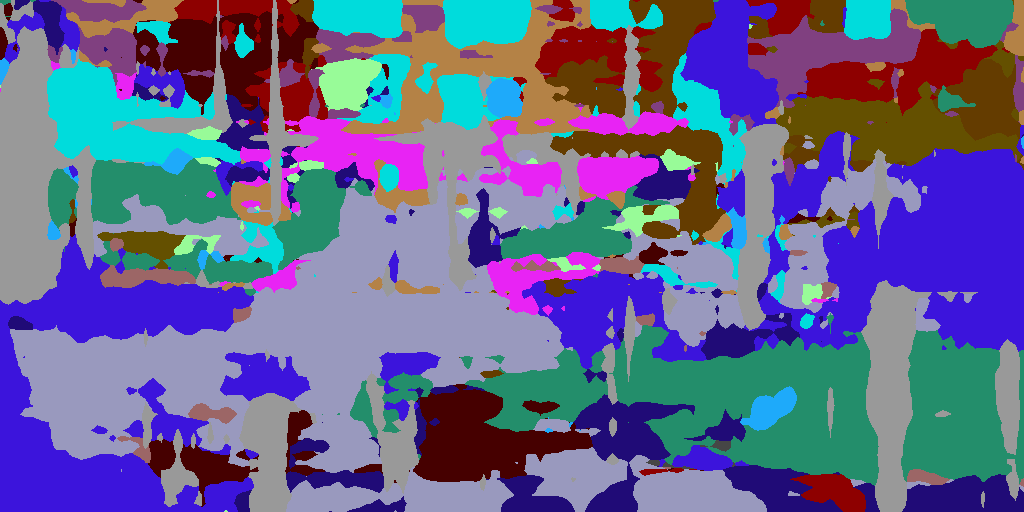} &
\includegraphics[width=.3\textwidth]{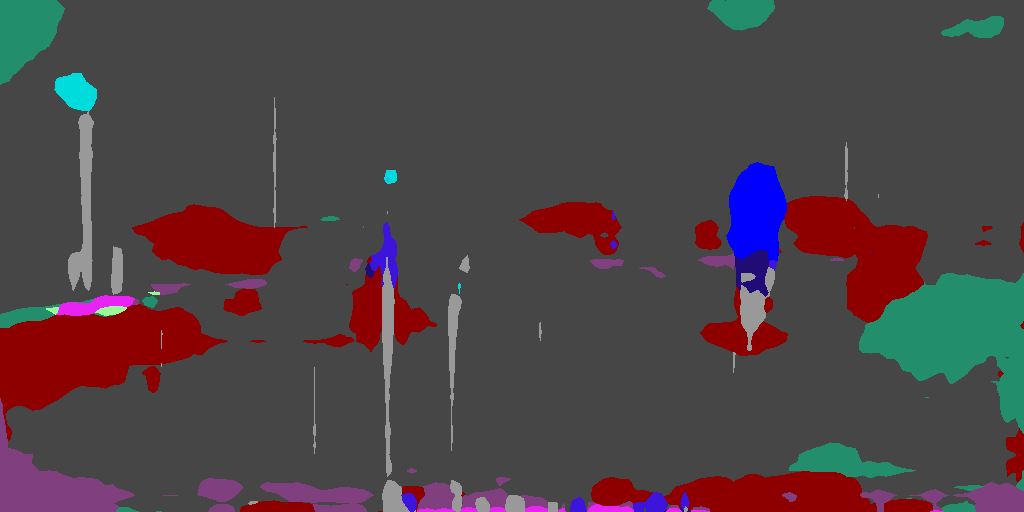} &
\includegraphics[width=.3\textwidth]{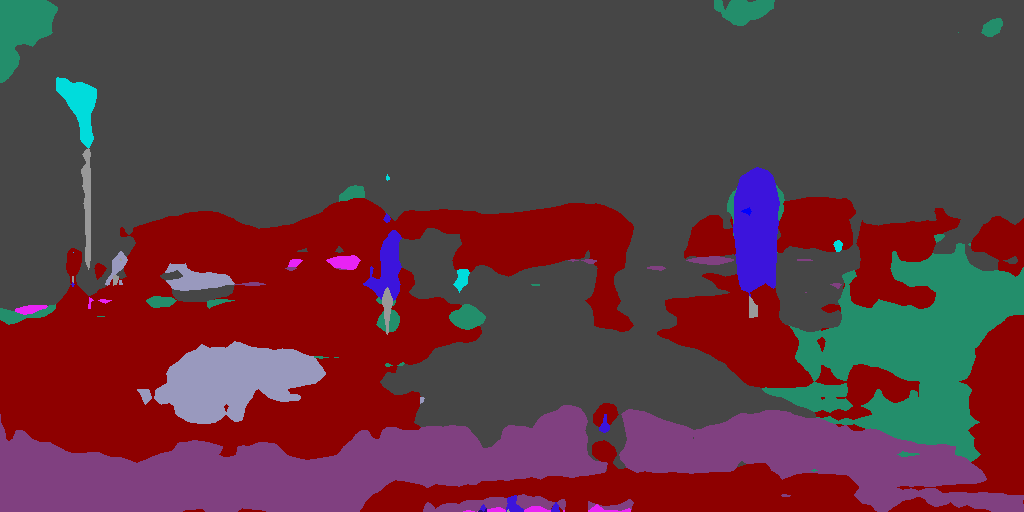} 
\end{tabular}
\caption{Result of PAdam-Cos attack on an example Cityscapes image for various PSPNet models.
Top row: perturbed images; bottom row: predicted mask on the perturbed image.}
\label{fig:cs-examples-images}
\end{figure*}

\begin{figure*}[t]
\centering
\setlength{\tabcolsep}{.2pt}
\scriptsize
\begin{tabular}{b{9mm}@{\hspace{2mm}}cccccc}
& Normal & DDC-AT & PGD-AT & SegPGD-AT & PGD-AT-100 & SegPGD-100 \\
No \newline attack &
\includegraphics[width=.15\textwidth]{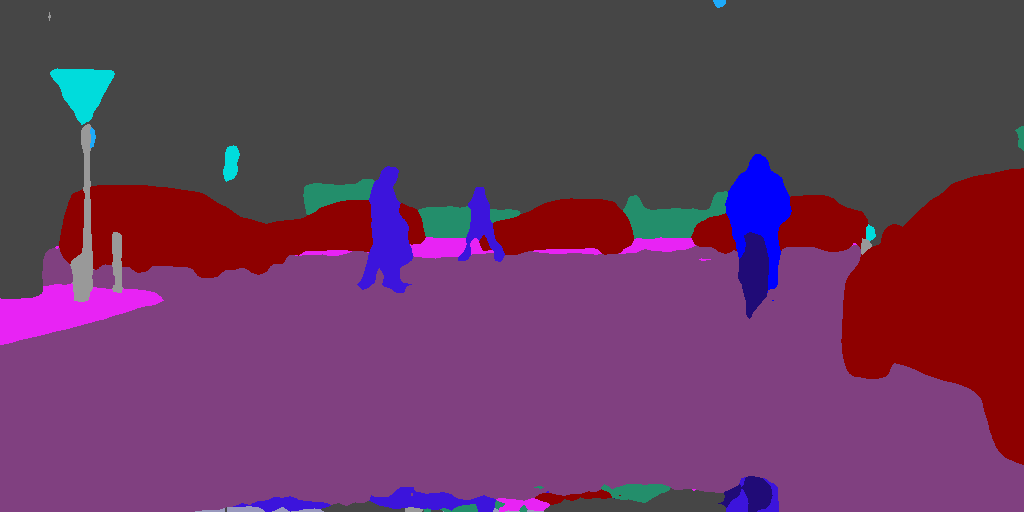} &
\includegraphics[width=.15\textwidth]{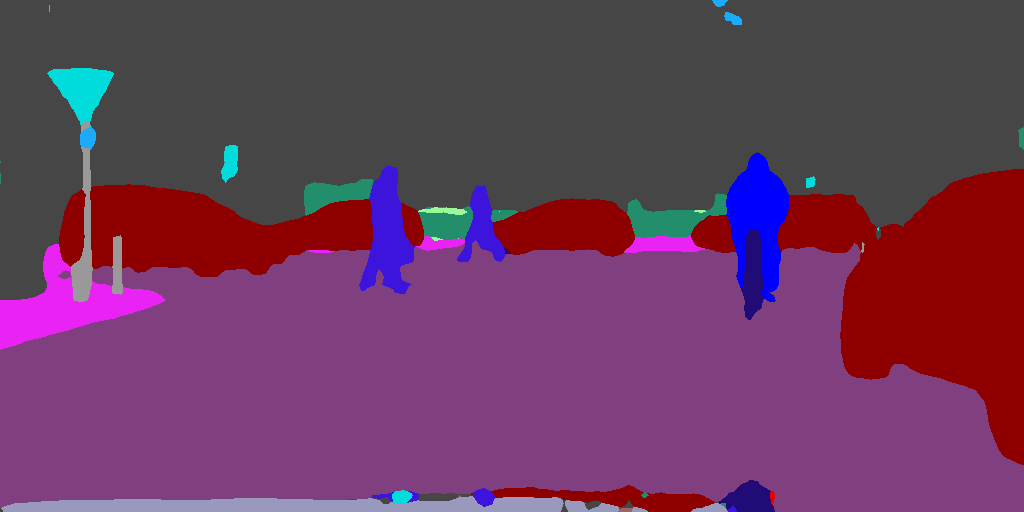} &
\includegraphics[width=.15\textwidth]{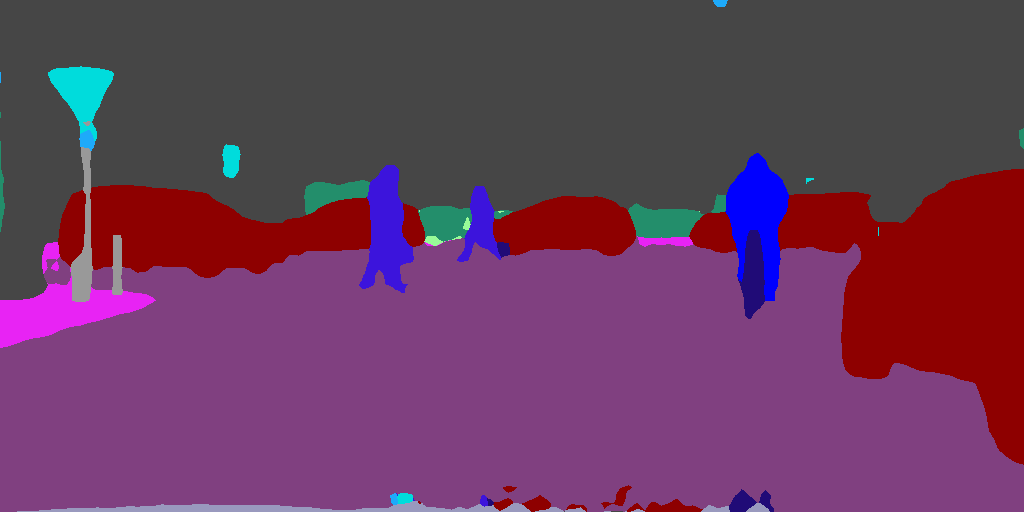} &
\includegraphics[width=.15\textwidth]{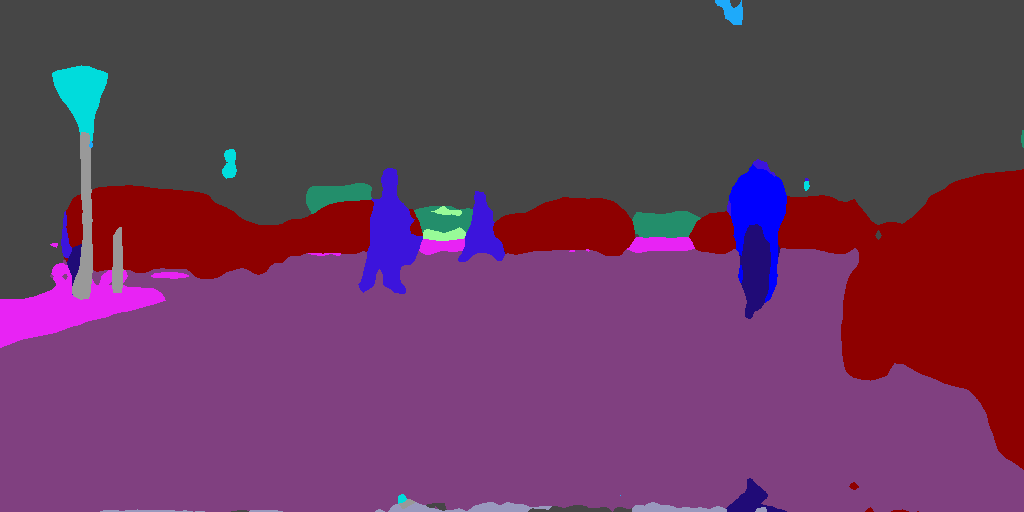} &
\includegraphics[width=.15\textwidth]{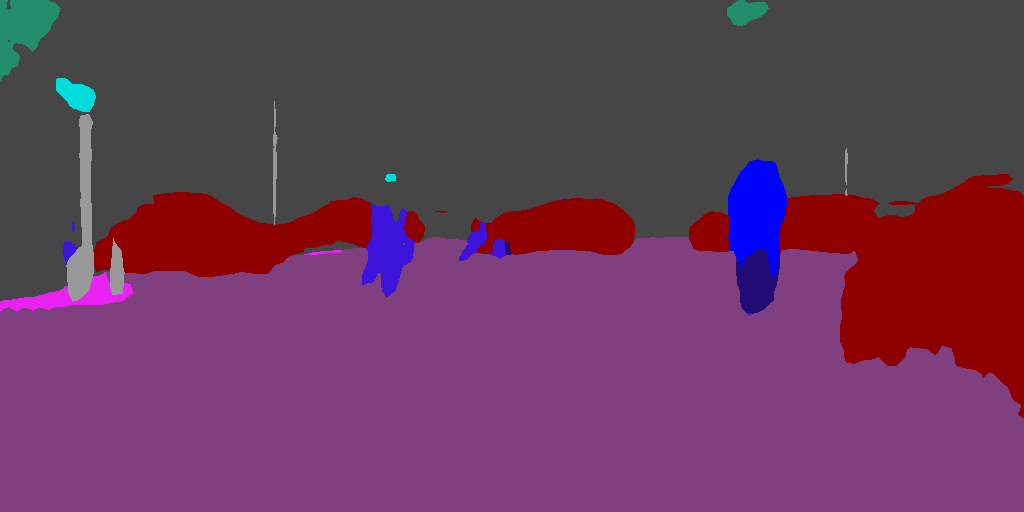} &
\includegraphics[width=.15\textwidth]{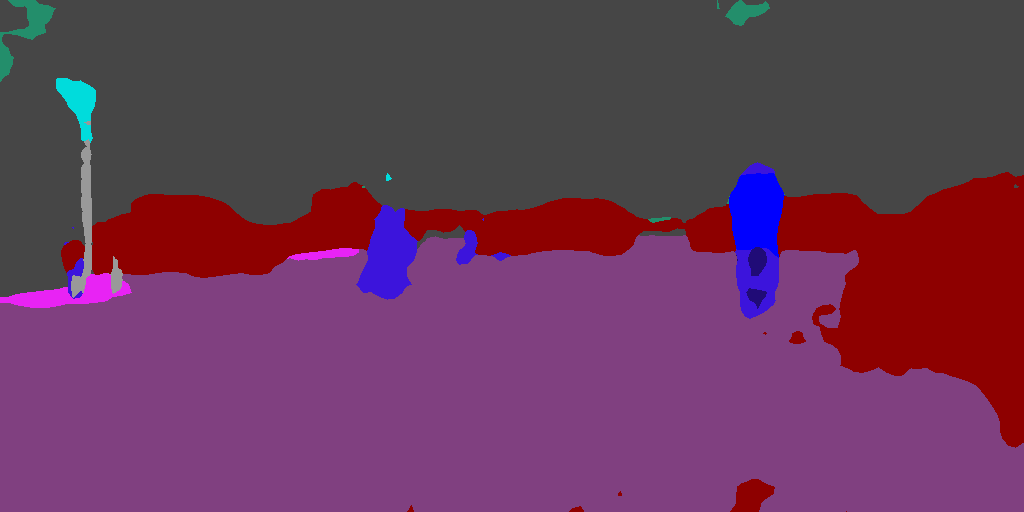} \\
PAdam-CE &
\includegraphics[width=.15\textwidth]{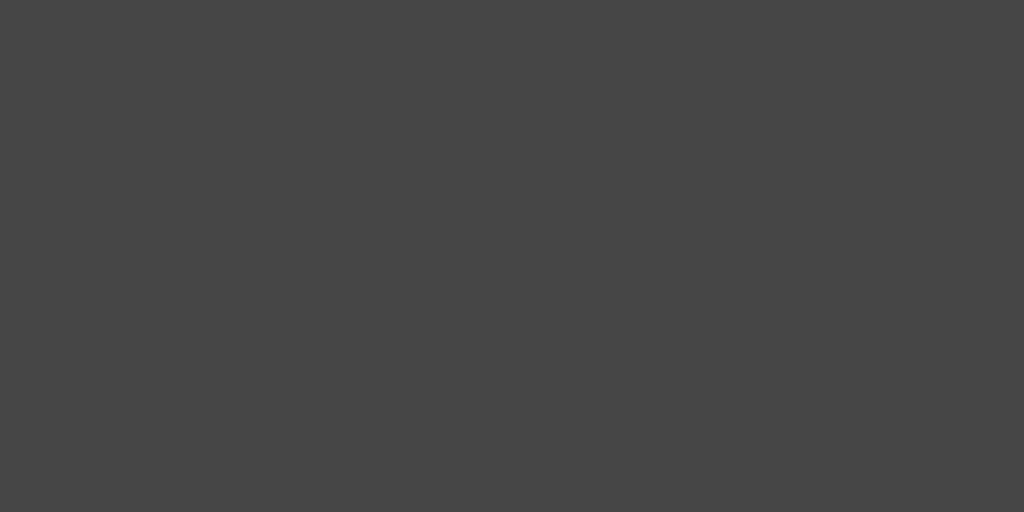} &
\includegraphics[width=.15\textwidth]{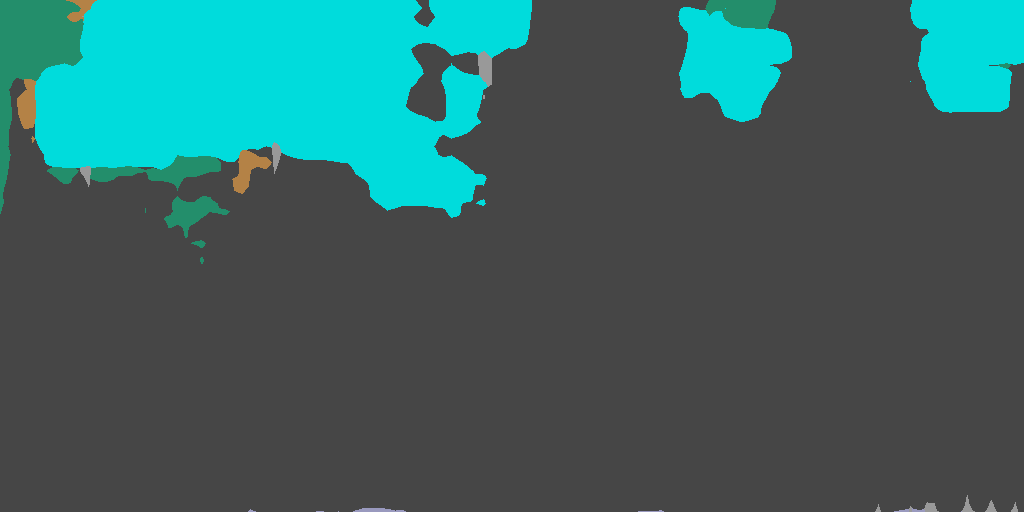} &
\includegraphics[width=.15\textwidth]{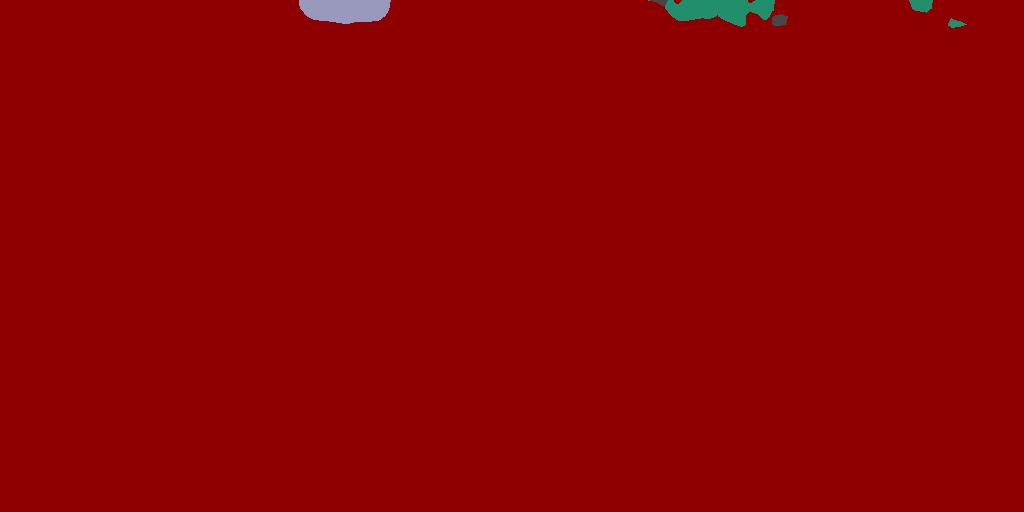} &
\includegraphics[width=.15\textwidth]{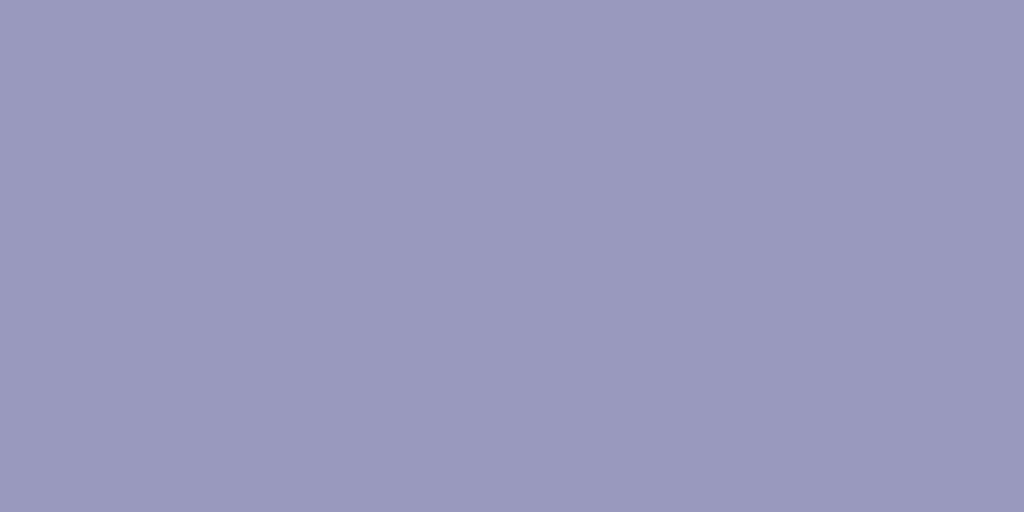} &
\includegraphics[width=.15\textwidth]{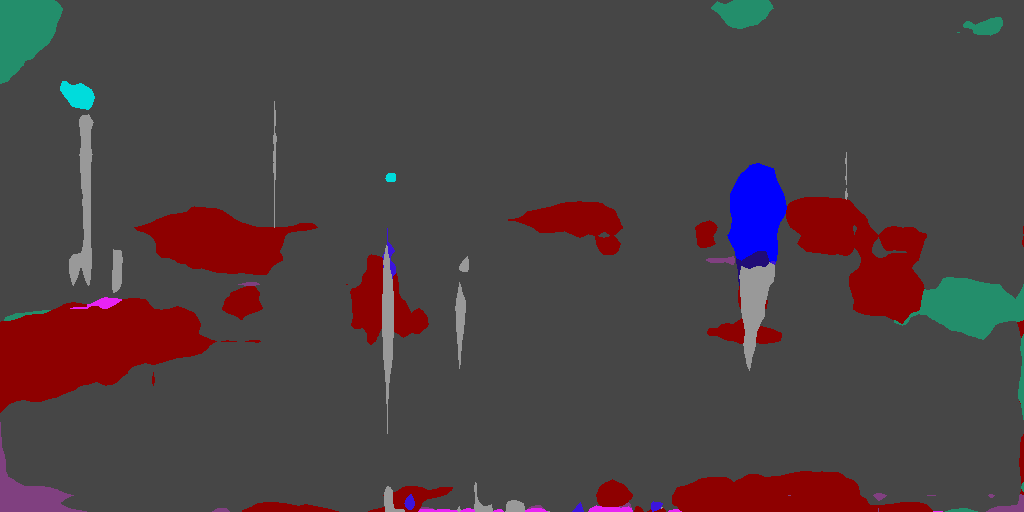} &
\includegraphics[width=.15\textwidth]{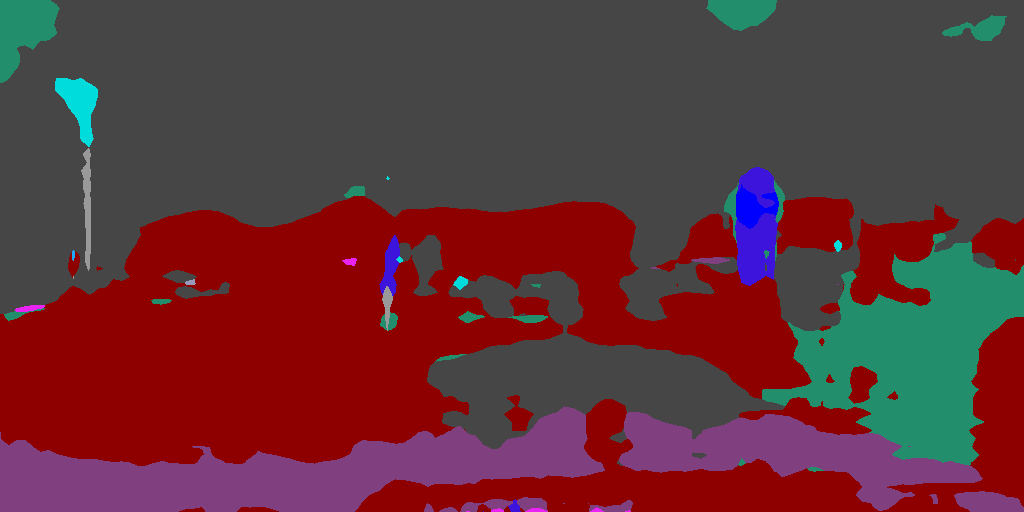} \\
\end{tabular}
\caption{Result of attacks on an example Cityscapes image (see \cref{fig:ground-truth}) for PSPNet models. Only the predicted masks on
the attacked images are shown due to file size limitations, but the perturbations are within the defined range of $8/255$ in
$\ell_\infty$ norm.
Please see \cref{fig:cs-examples-images} for examples of attacked images.}
\label{fig:cs-pspnet-masks}
\end{figure*}

\begin{figure*}[t]
\centering
\setlength{\tabcolsep}{.2pt}
\scriptsize
\begin{tabular}{b{9mm}@{\hspace{2mm}}cccccc}
& Normal & DDC-AT & PGD-AT & SegPGD-AT & PGD-AT-100 & SegPGD-100 \\
No \newline attack &
\includegraphics[width=.15\textwidth]{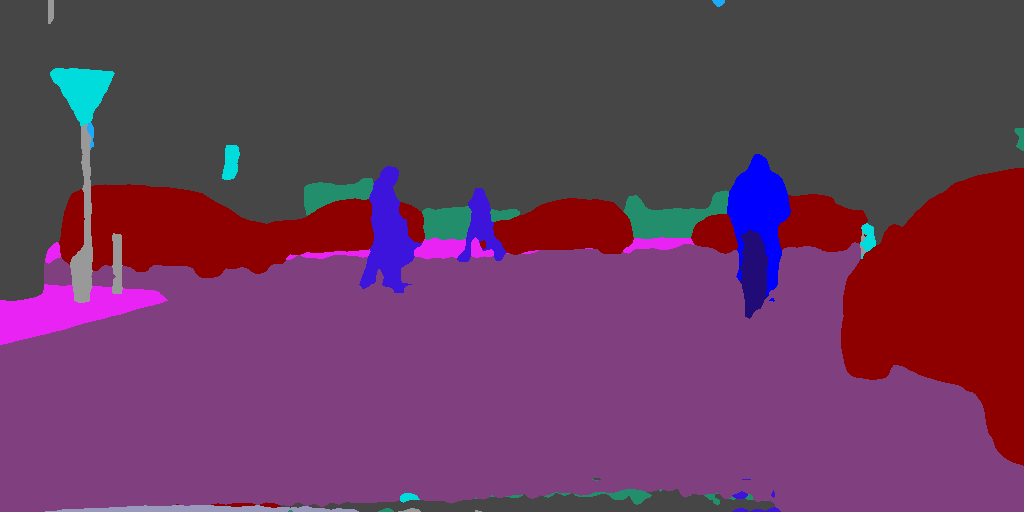} &
\includegraphics[width=.15\textwidth]{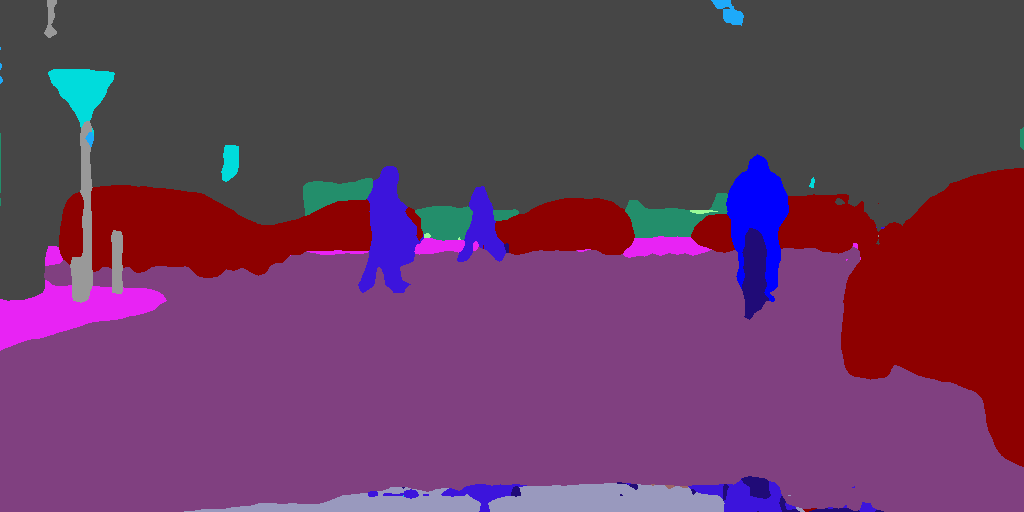} &
\includegraphics[width=.15\textwidth]{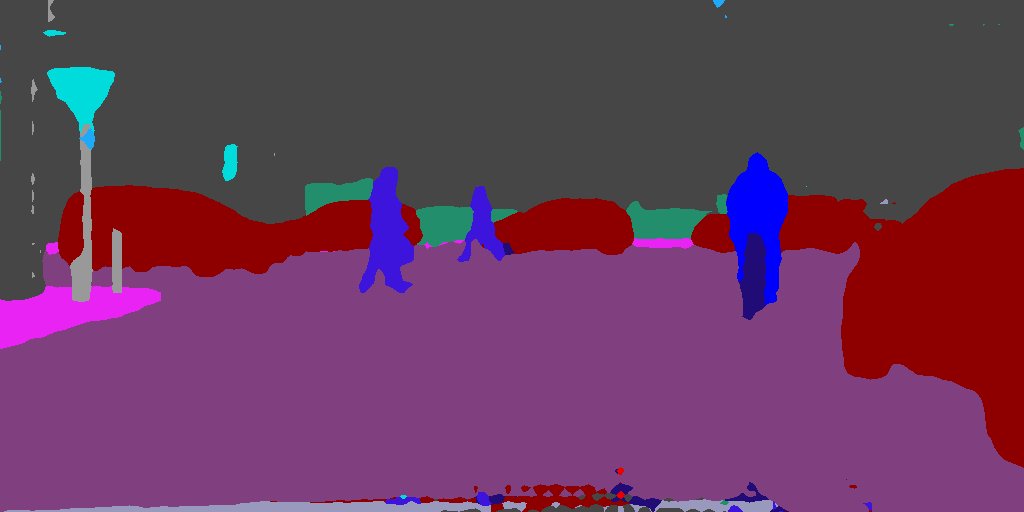} &
\includegraphics[width=.15\textwidth]{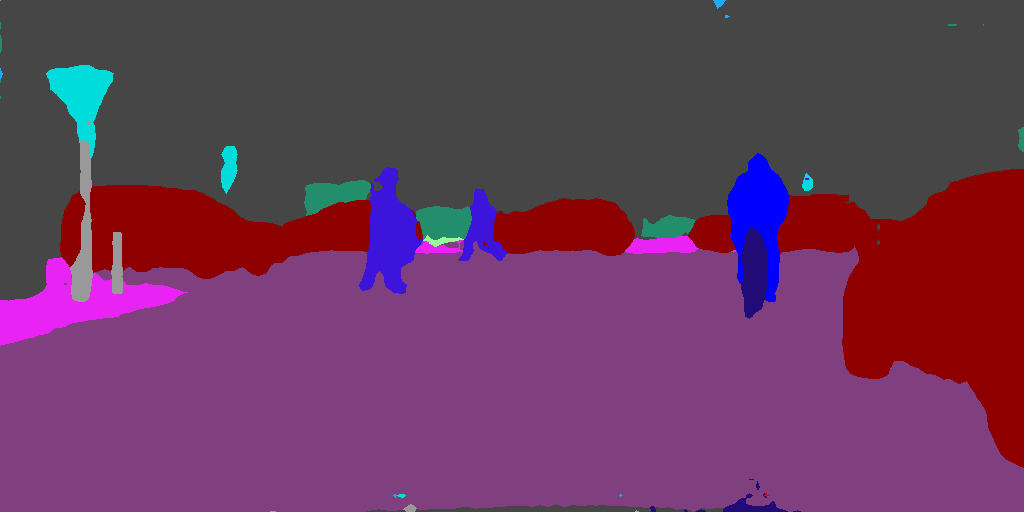} &
\includegraphics[width=.15\textwidth]{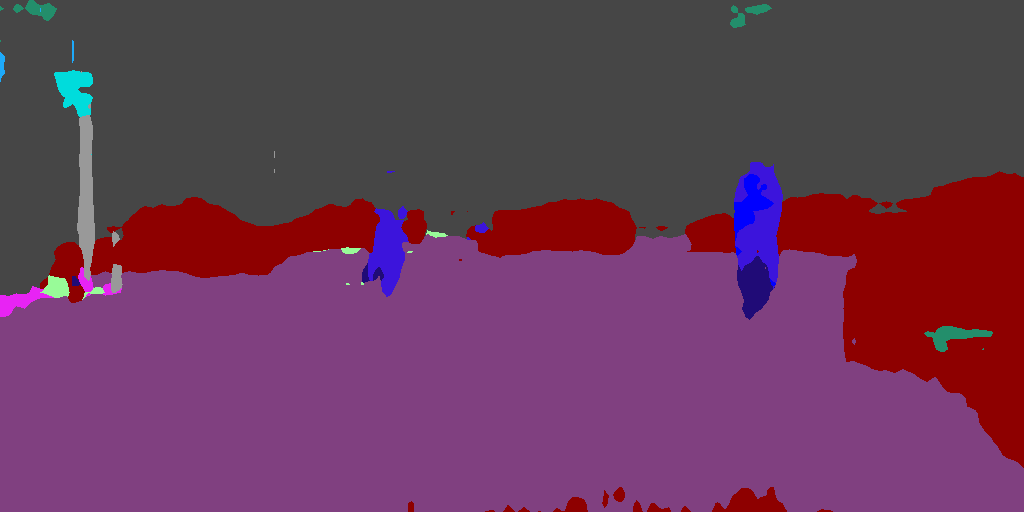} &
\includegraphics[width=.15\textwidth]{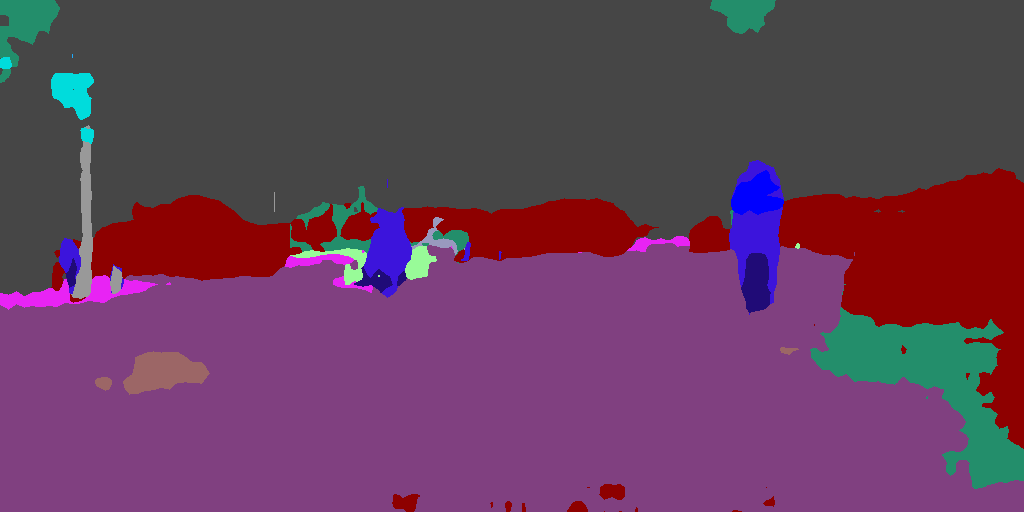} \\
PAdam-CE &
\includegraphics[width=.15\textwidth]{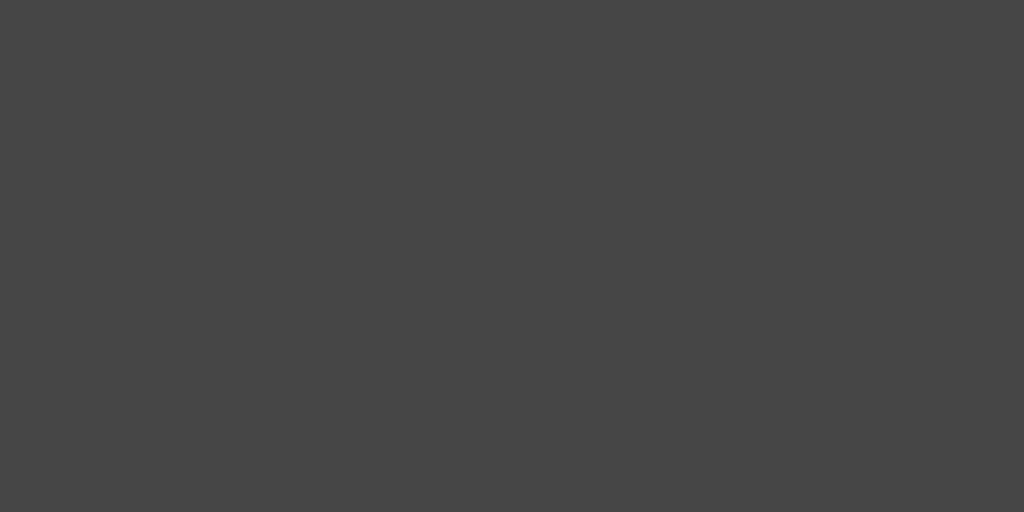} &
\includegraphics[width=.15\textwidth]{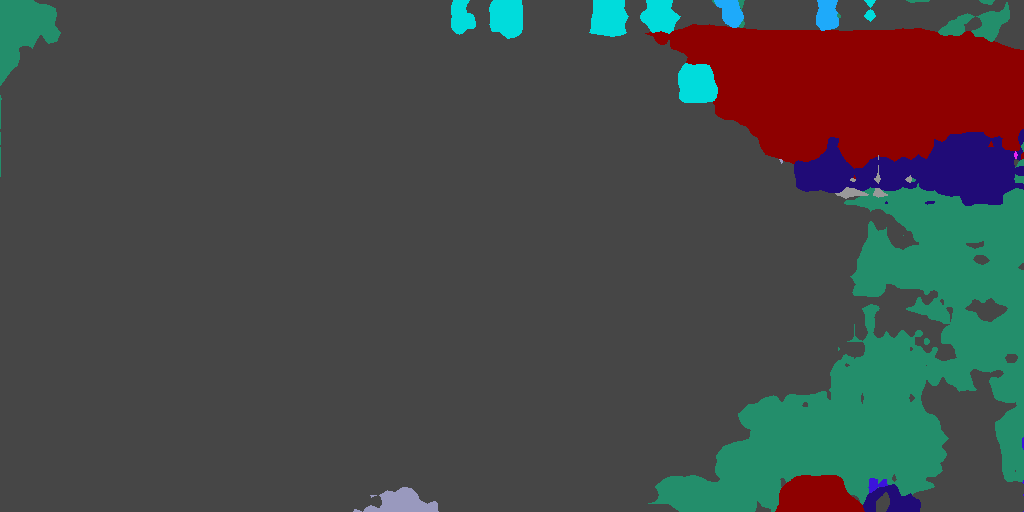} &
\includegraphics[width=.15\textwidth]{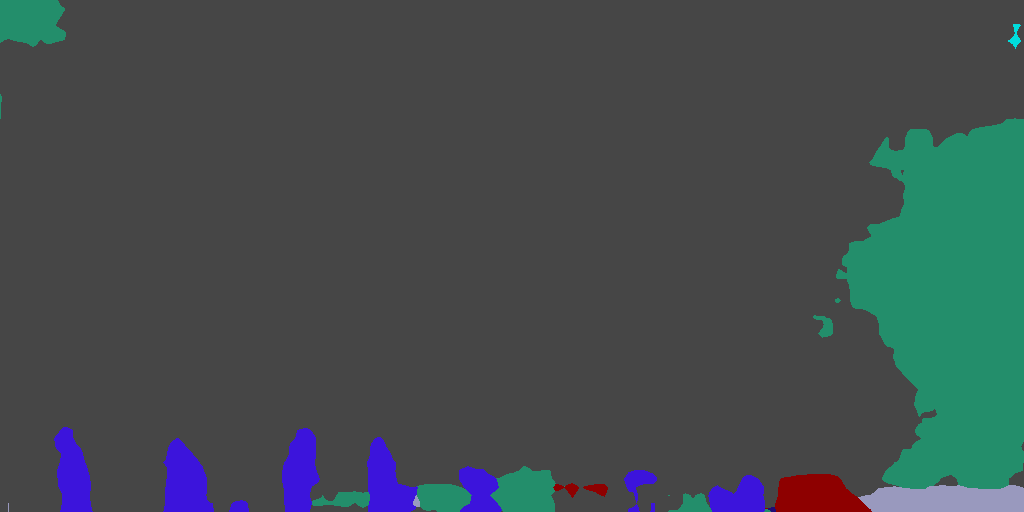} &
\includegraphics[width=.15\textwidth]{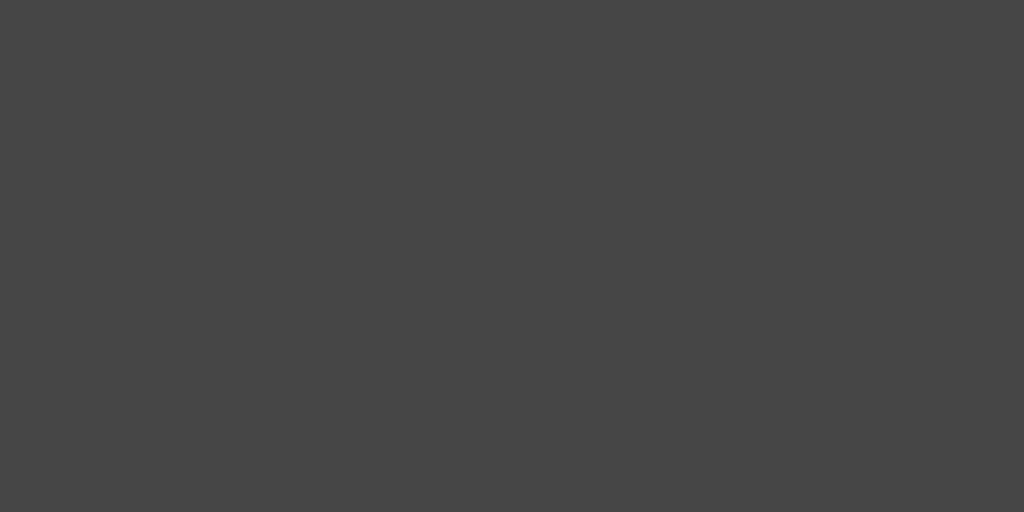} &
\includegraphics[width=.15\textwidth]{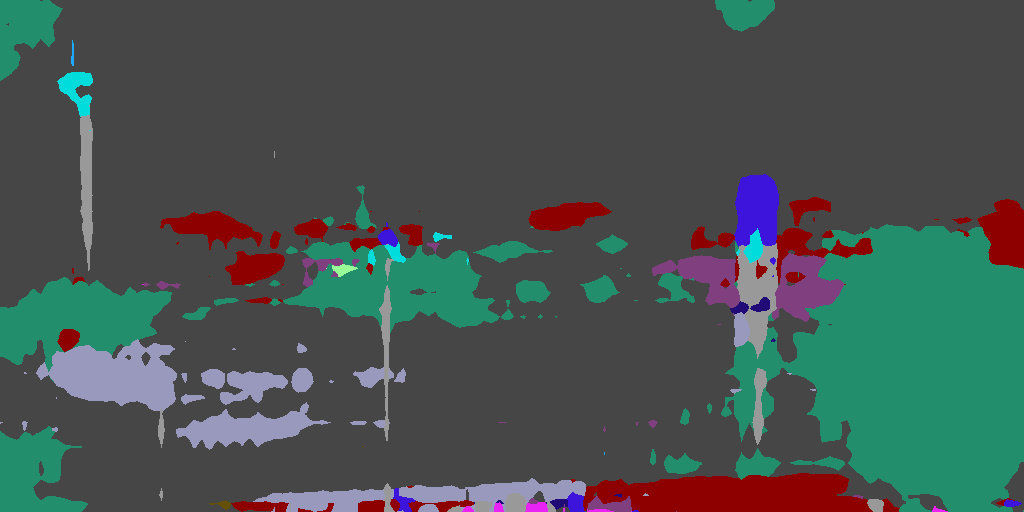} &
\includegraphics[width=.15\textwidth]{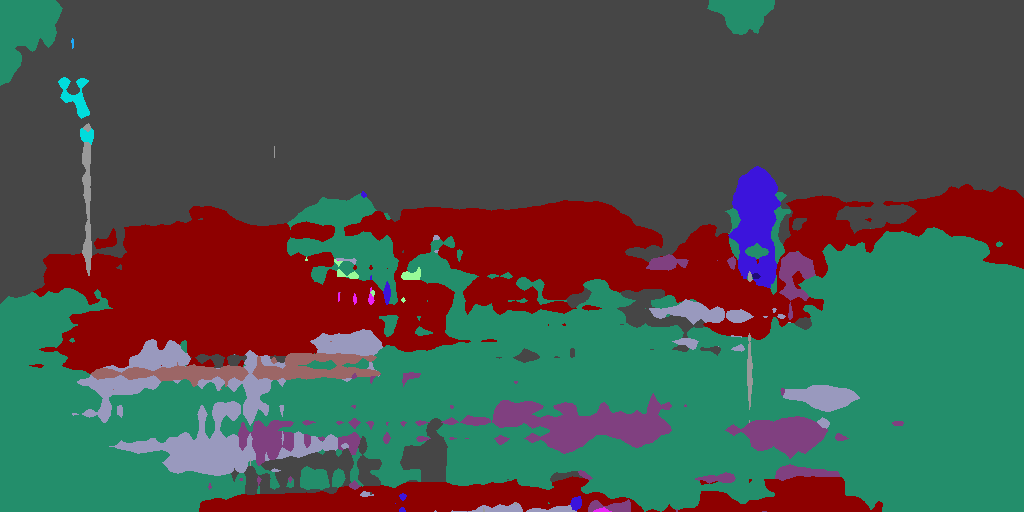} \\
\end{tabular}
\caption{Result of attacks on an example Cityscapes image (see \cref{fig:ground-truth}) for DeepLabv3 models. Only the predicted masks on
the attacked images are shown due to file size limitations, but the perturbations are within the defined range of $8/255$ in
$\ell_\infty$ norm.
Please see \cref{fig:cs-examples-images} for examples of attacked images.}
\label{fig:cs-deeplab-masks}
\end{figure*}

\end{document}